\PassOptionsToPackage{shortlabels,inline}{enumitem}

\documentclass[PhD,with-ethics-statement,beforeDefense]{misc/iitthesis}

\authorEnglish{Yonatan Sverdlov}
\authorHebrew{יונתן סברדלוב}

\titleEnglish{Improving fine-grained expressivity of graph neural networks}
\titleHebrew{משפרים אקספרסיביות עדינה לרשתות למידה על גרפים}

\disciplineEnglish{Mathematics}
\disciplineHebrew{מתמטיקה}

\supervisionEnglish{%
The Research Thesis Was Done in the Interdisciplinary Program in Applied Mathematics Under the Supervision of Prof.~Nadav Dym.%
}
\supervisionHebrew{%
המחקר נעשה בתכנית הבין יחידתית למתמטיקה שימושית בהנחיית פרופ' נדב דים.%
}

\GregorianDateEnglish{September 2026}
\GregorianDateHebrew{ספטמבר 2026}
\JewishDateEnglish{Tishrei 5787}
\JewishDateHebrew{תשרי תשפ"ז}

\financialAcknowledgementEnglish{The generous financial help of the Technion and the Israeli Science Foundation grant no. 272/23 is gratefully acknowledged.}
\financialAcknowledgementHebrew{אני מודה לטכניון ולקרן הלאומית למדע, מענק מס' 272/23, על התמיכה הכספית הנדיבה בהשתלמותי.
}

\publicationInfoEnglish{%
Parts of the research presented in this thesis have appeared in the publications listed below. For each publication, its status and the author's contribution are stated.

\butcheredbibliography{front/pubinfo.bib}
}

\publicationInfoHebrew{%
חלקים מן המחקר המוצג בחיבור זה הופיעו בפרסומים המפורטים להלן. לכל פרסום מצוינים הסטטוס ותרומת המחבר.

\begin{otherlanguage}{english}
\butcheredbibliography{front/pubinfo.bib}
\end{otherlanguage}
}

\ethicsStatementEnglish{%
The author of this thesis states that the research, including the collection, processing and presentation of data, addressing and comparing to previous research, etc., was done entirely in an honest way, as expected from scientific research that is conducted according to the ethical standards of the academic world. Also, reporting the research and its results in this thesis was done in an honest and complete manner, according to the same standards.%
}

\ethicsStatementHebrew{מחבר חיבור זה מצהיר כי המחקר, כולל איסוף הנתונים, עיבודם והצגתם, התייחסות והשוואה למחקרים קודמים וכו', נעשה כולו בצורה ישרה, כמצופה ממחקר מדעי המבוצע לפי אמות המידה האתיות של העולם האקדמי. כמו כן, הדיווח על המחקר ותוצאותיו בחיבור זה נעשה בצורה ישרה ומלאה, לפי אותן אמות המידה.

}

\personalAcknowledgementEnglish{%
[Personal acknowledgements -- include only in the post-defense final version.]
}

\personalAcknowledgementHebrew{%
[תודות אישיות -- להוסיף רק לגרסה הסופית שלאחר הבחינה.]
}

\abstractEnglish{%
Many important learning problems involve structured data rather than vector-valued data. Graphs arise naturally in domains such as molecular modeling, social and communication networks, and relational data, while point clouds and geometric graphs are fundamental representations of three-dimensional objects, physical systems, and spatial data. In such settings, the representation of an object is typically not unique: relabeling the vertices of a graph or rotating a geometric object should not change its semantic meaning. Neural architectures for structured data should therefore respect the appropriate invariances and equivariances. This thesis studies the expressive power of such architectures, with an emphasis on understanding what information invariant representations can preserve and how this depends on the architecture and the underlying structure of the data.

The thesis consists of four works. First, we study the expressive power of message-passing neural networks on sparse geometric graphs. Previous analyses of geometric networks largely focused on complete graphs or on sufficient architectural conditions for expressivity. We provide, to the best of our knowledge, the first generic characterization of expressive power for sparse geometric graphs. We show that rotation-equivariant architectures generically separate inputs whenever the underlying graph is connected, whereas invariant architectures require generic global rigidity. We further introduce EGenNet, a simple equivariant architecture attaining the corresponding expressive guarantees.

Second, we study the construction of equivariant linear layers. A common approach derives such layers by explicitly identifying parameter-sharing constraints imposed by the symmetry group, a procedure that can become cumbersome and error-prone for more complicated group actions. Instead, we use representation theory, and in particular irreducible representations and Schur's lemma, to obtain a systematic characterization of equivariant linear maps. The resulting framework recovers several known architectures and provides a complete characterization in multi-permutation settings, including wreath-product symmetries, where it reveals interactions that are missed by standard Siamese constructions.

Third, we consider representations of sets. Classical architectures such as DeepSets are primarily concerned with permutation invariance and with representations sufficiently expressive to distinguish sets or approximate invariant functions. We ask for a stronger structural property: the representation should preserve set containment through coordinatewise order. We characterize the embedding dimension required for such containment-preserving representations, establish impossibility results for infinite ground sets, and introduce MASNet, a stable relaxed construction. This leads to neural representations that capture not only whether two sets are different, but also the inclusion relation between them.

Finally, we return to message-passing graph neural networks and study quantitative notions of expressivity. Standard expressivity results typically ask whether two graphs can be distinguished, without controlling how far apart their representations remain. We introduce FSW-GNN, an MPNN that combines strong theoretical guarantees---including Weisfeiler--Leman-level expressivity and bi-Lipschitz stability with respect to suitable graph metrics---with strong empirical performance. In particular, the architecture performs very well on long-range graph-learning tasks designed to expose oversquashing, demonstrating that the theoretical preservation of information is reflected in practical performance.

Together, these works develop a broader view of expressive power in structured learning. Beyond asking whether invariant neural networks can distinguish inputs, we study when they can faithfully preserve the geometric, algebraic, order-theoretic, and metric structure that is relevant to the learning problem.
}

\abstractHebrew{בעיות למידה חשובות רבות עוסקות במידע מובנה ולא בנתונים וקטוריים. גרפים מופיעים באופן טבעי בתחומים כגון מידול מולקולרי, רשתות חברתיות ורשתות תקשורת, ונתונים יחסיים, בעוד שענני נקודות וגרפים גאומטריים מהווים ייצוגים בסיסיים של עצמים תלת־ממדיים, מערכות פיזיקליות ונתונים מרחביים. במסגרות כאלה, הייצוג של עצם בדרך כלל אינו יחיד: שינוי תיוג הקודקודים של גרף או סיבוב של עצם גאומטרי אינם אמורים לשנות את משמעותו הסמנטית. לפיכך, ארכיטקטורות נוירוניות עבור מידע מובנה צריכות לכבד את האינווריאנטיות והאקווריאנטיות המתאימות. תזה זו חוקרת את כוח ההפרדה של ארכיטקטורות כאלה, תוך שימת דגש על הבנת סוגי המידע שייצוגים אינווריאנטיים יכולים לשמר, ועל האופן שבו יכולת זו תלויה בארכיטקטורה ובמבנה הבסיסי של הנתונים.

התזה מורכבת מארבע עבודות. ראשית, אנו חוקרים את כוח ההפרדה של רשתות נוירונים מבוססות העברת מסרים על גרפים גאומטריים דלילים. מחקרים קודמים על רשתות גאומטריות התמקדו במידה רבה בגרפים מלאים או בתנאים ארכיטקטוניים מספיקים לכוח הפרדה. למיטב ידיעתנו, אנו מספקים את האפיון הגנרי הראשון של כוח ההפרדה עבור גרפים גאומטריים דלילים. אנו מראים כי ארכיטקטורות אקווריאנטיות לסיבובים מפרידות באופן גנרי בין קלטים כאשר הגרף שבבסיסם קשיר, בעוד שארכיטקטורות אינווריאנטיות דורשות קשיחות גלובלית גנרית. בנוסף, אנו מציגים את EGenNet, ארכיטקטורה אקווריאנטית פשוטה המשיגה את הבטחות כוח ההפרדה המתאימות.

שנית, אנו חוקרים את בנייתן של שכבות לינאריות אקווריאנטיות. גישה מקובלת מפיקה שכבות כאלה באמצעות זיהוי מפורש של אילוצי שיתוף פרמטרים הנכפים על ידי חבורת הסימטריה, תהליך שעלול להפוך למסורבל ומועד לשגיאות עבור פעולות חבורה מורכבות יותר. במקום זאת, אנו משתמשים בתורת ההצגות, ובפרט בהצגות אי־פריקות ובלמה של שור, כדי לקבל אפיון שיטתי של העתקות לינאריות אקווריאנטיות. המסגרת המתקבלת משחזרת מספר ארכיטקטורות מוכרות ומספקת אפיון מלא במצבים הכוללים מספר פרמוטציות, ובכלל זה סימטריות של מכפלת זֵר, שבהן היא חושפת אינטראקציות שאינן נתפסות על ידי ארכיטקטורות סיאמיות סטנדרטיות.

שלישית, אנו עוסקים בייצוגים של קבוצות. ארכיטקטורות קלאסיות כגון DeepSets עוסקות בעיקר באינווריאנטיות לפרמוטציות ובייצוגים בעלי כוח הפרדה מספיק כדי להבחין בין קבוצות או לקרב פונקציות אינווריאנטיות. אנו דורשים תכונה מבנית חזקה יותר: על הייצוג לשמר הכלה בין קבוצות באמצעות סדר לפי רכיבים. אנו מאפיינים את ממד השיכון הנדרש עבור ייצוגים המשמרים הכלה, מוכיחים תוצאות אי־אפשרות עבור קבוצות בסיס אינסופיות, ומציגים את MASNet, בנייה יציבה המבוססת על הרפיה של הדרישה. כך מתקבלים ייצוגים נוירוניים הלוכדים לא רק את השאלה האם שתי קבוצות שונות זו מזו, אלא גם את יחס ההכלה ביניהן.

לבסוף, אנו חוזרים לרשתות נוירונים על גרפים המבוססות על העברת מסרים וחוקרים מושגים כמותיים של כוח הפרדה. תוצאות סטנדרטיות בתחום כוח ההפרדה שואלות בדרך כלל האם ניתן להבחין בין שני גרפים, מבלי לשלוט במרחק הנשמר בין הייצוגים שלהם. אנו מציגים את FSW-GNN, רשת נוירונים מבוססת העברת מסרים המשלבת הבטחות תאורטיות חזקות — ובהן כוח הפרדה ברמת Weisfeiler–Leman ויציבות בי־ליפשיצית ביחס למטריקות מתאימות על גרפים — עם ביצועים אמפיריים חזקים. בפרט, הארכיטקטורה משיגה ביצועים טובים מאוד במשימות למידה על גרפים לטווח ארוך, שנועדו לחשוף את תופעת ה־oversquashing, ובכך מדגימה כי השימור התאורטי של מידע בא לידי ביטוי גם בביצועים מעשיים.

יחד, עבודות אלה מפתחות נקודת מבט רחבה יותר על כוח ההפרדה בלמידה על מידע מובנה. מעבר לשאלה האם רשתות נוירונים אינווריאנטיות יכולות להבחין בין קלטים, אנו חוקרים מתי הן יכולות לשמר באופן נאמן את המבנה הגאומטרי, האלגברי, הסדרי והמטרי הרלוונטי לבעיית הלמידה.

}

\usepackage{misc/iitthesis-extra}

\newglossaryentry{graph}{
    type=notation,
    name={\ensuremath{G=(V,E,X)}},
    sort={01 graph},
    description={A vertex-featured graph, with vertex set \(V\), edge set \(E\), and node-feature matrix \(X\). In the geometric-graph chapter, the equivalent notation \(G=(A,X)\) is also used}
}

\newglossaryentry{vertices}{
    type=notation,
    name={\ensuremath{V}},
    sort={02 vertices},
    description={The vertex set of a graph. In the set-function chapter, \(V\) instead denotes the ground set from which multiset elements are drawn}
}

\newglossaryentry{edges}{
    type=notation,
    name={\ensuremath{E}},
    sort={03 edges},
    description={The edge set of a graph}
}

\newglossaryentry{numvertices}{
    type=notation,
    name={\ensuremath{n=|V|}},
    sort={04 number vertices},
    description={Number of vertices in a graph}
}

\newglossaryentry{maxvertices}{
    type=notation,
    name={\ensuremath{N}},
    sort={05 maximum vertices},
    description={An upper bound on the number of vertices in a graph, when a bounded-size graph space is considered}
}

\newglossaryentry{adjacency}{
    type=notation,
    name={\ensuremath{A}},
    sort={06 adjacency},
    description={Adjacency matrix of a graph}
}

\newglossaryentry{nodefeature}{
    type=notation,
    name={\ensuremath{x_v\in\mathbb{R}^d}},
    sort={07 node feature},
    description={Feature vector associated with node \(v\); in geometric graphs this may be the position of the node}
}

\newglossaryentry{featurematrix}{
    type=notation,
    name={\ensuremath{X}},
    sort={08 feature matrix},
    description={Matrix collecting node features or geometric node positions}
}

\newglossaryentry{featuredomain}{
    type=notation,
    name={\ensuremath{\Omega\subseteq\mathbb{R}^d}},
    sort={09 feature domain},
    description={Domain from which node features are drawn}
}

\newglossaryentry{neighborhood}{
    type=notation,
    name={\ensuremath{\mathcal{N}(v)}},
    sort={10 neighborhood},
    description={Set of neighbors of node \(v\)}
}

\newglossaryentry{hiddenfeature}{
    type=notation,
    name={\ensuremath{h_v^{(t)}}},
    sort={11 hidden feature},
    description={Hidden representation of node \(v\) after \(t\) message-passing iterations}
}

\newglossaryentry{indexset}{
    type=notation,
    name={\ensuremath{[n]=\{1,\ldots,n\}}},
    sort={12 index set},
    description={The set of the first \(n\) positive integers}
}

\newglossaryentry{multiset}{
    type=notation,
    name={\ensuremath{\{\!\{x_1,\ldots,x_n\}\!\}}},
    sort={13 multiset},
    description={A multiset: an unordered collection in which repeated elements are allowed}
}

\newglossaryentry{finitemultisets}{
    type=notation,
    name={\ensuremath{\mathcal{P}_{<\infty}(V)}},
    sort={14 finite multisets},
    description={The collection of all finite multisets whose elements belong to the ground set \(V\)}
}

\newglossaryentry{sets}{
    type=notation,
    name={\ensuremath{S,T}},
    sort={15 sets},
    description={Sets or multisets under comparison}
}

\newglossaryentry{multiplicity}{
    type=notation,
    name={\ensuremath{c_S(v)}},
    sort={16 multiplicity},
    description={Multiplicity of the element \(v\) in the multiset \(S\)}
}

\newglossaryentry{subset}{
    type=notation,
    name={\ensuremath{S\subseteq T}},
    sort={17 subset},
    description={Set or multiset containment; for multisets, \(c_S(v)\leq c_T(v)\) for every ground-set element \(v\)}
}

\newglossaryentry{coordinateorder}{
    type=notation,
    name={\ensuremath{u\leq v}},
    sort={18 coordinatewise order},
    description={Coordinatewise partial order on vectors: \(u_i\leq v_i\) for every coordinate \(i\)}
}

\newglossaryentry{reald}{
    type=notation,
    name={\ensuremath{\mathbb{R}^d}},
    sort={19 real space},
    description={\(d\)-dimensional Euclidean space}
}

\newglossaryentry{symmetrygroup}{
    type=notation,
    name={\ensuremath{\mathcal{G}}},
    sort={20 symmetry group},
    description={A symmetry group acting on a feature space in the equivariance chapters. In the FSW-GNN paper, \(\mathcal{G}\) is locally reused for a graph space}
}

\newglossaryentry{groupaction}{
    type=notation,
    name={\ensuremath{g\cdot x}},
    sort={21 group action},
    description={Action of a group element \(g\in\mathcal{G}\) on an element \(x\) of the representation space}
}

\newglossaryentry{representationspaces}{
    type=notation,
    name={\ensuremath{\mathcal{U},\mathcal{V},\mathcal{W}}},
    sort={22 representation spaces},
    description={Finite-dimensional vector spaces carrying group representations}
}

\newglossaryentry{fixedsubspace}{
    type=notation,
    name={\ensuremath{\mathcal{V}^{\mathcal{G}}}},
    sort={23 fixed subspace},
    description={Fixed-point subspace of a representation: vectors \(v\in\mathcal{V}\) satisfying \(g\cdot v=v\) for all \(g\in\mathcal{G}\)}
}

\newglossaryentry{symmetricgroup}{
    type=notation,
    name={\ensuremath{S_n}},
    sort={24 symmetric group},
    description={Symmetric group of permutations of \(n\) elements}
}

\newglossaryentry{wreathproduct}{
    type=notation,
    name={\ensuremath{\mathcal{G}\wr S_k}},
    sort={25 wreath product},
    description={Wreath-product symmetry for a set of \(k\) unaligned objects, each carrying a \(\mathcal{G}\)-symmetry}
}

\newglossaryentry{permutationmatrix}{
    type=notation,
    name={\ensuremath{P}},
    sort={26 permutation matrix},
    description={Permutation matrix representing a relabeling of elements or graph vertices}
}

\newglossaryentry{orthogonalgroup}{
    type=notation,
    name={\ensuremath{O(d)}},
    sort={27 orthogonal group},
    description={Group of orthogonal transformations of \(\mathbb{R}^d\), including rotations and reflections}
}

\newglossaryentry{rigidmotion}{
    type=notation,
    name={\ensuremath{x\mapsto Qx+t}},
    sort={28 rigid motion},
    description={Rigid transformation with \(Q\in O(d)\) and translation \(t\in\mathbb{R}^d\)}
}

\newglossaryentry{geometricisomorphism}{
    type=notation,
    name={\ensuremath{G\simeq H}},
    sort={29 geometric isomorphism},
    description={Geometric graphs \(G\) and \(H\) differ only by vertex relabeling and a rigid transformation}
}

\newglossaryentry{graphembedding}{
    type=notation,
    name={\ensuremath{F(G)}},
    sort={30 graph embedding},
    description={Invariant finite-dimensional representation of a graph \(G\)}
}

\newglossaryentry{metric}{
    type=notation,
    name={\ensuremath{\rho(G,H)}},
    sort={31 graph metric},
    description={Metric or pseudometric measuring the distance between two input graphs}
}

\newglossaryentry{bilipschitz}{
    type=notation,
    name={\ensuremath{c,C}},
    sort={32 bilipschitz},
    description={Bi-Lipschitz constants satisfying \(0<c\leq C<\infty\) and \(c\,\rho(G,H)\leq\|F(G)-F(H)\|\leq C\,\rho(G,H)\)}
}

\newglossaryentry{graphspace}{
    type=notation,
    name={\ensuremath{\mathcal{G}_{\leq N}(\Omega)}},
    sort={33 graph space},
    description={Space of vertex-featured graphs with at most \(N\) vertices and node features in \(\Omega\); used in the FSW-GNN chapter}
}

\newglossaryentry{wlequivalence}{
    type=notation,
    name={\ensuremath{G\equiv_{\mathrm{WL}} H}},
    sort={34 wl equivalence},
    description={Graphs \(G\) and \(H\) are not distinguished by the corresponding Weisfeiler--Leman procedure}
}

\newabbreviation[type=notation]{gnn}
{GNN}
{graph neural network}

\newabbreviation[type=notation]{mpnn}
{MPNN}
{message passing neural network}

\newabbreviation[type=notation]{wl}
{WL}
{Weisfeiler--Leman; unless otherwise stated, the graph chapter primarily uses 1-WL}

\newabbreviation[type=notation]{mlp}
{MLP}
{multilayer perceptron}

\newabbreviation[type=notation]{eggnn}
{E-GGNN}
{equivariant geometric graph neural network}

\newabbreviation[type=notation]{iggnn}
{I-GGNN}
{invariant geometric graph neural network}

\newabbreviation[type=notation]{ign}
{2-IGN}
{second-order invariant graph network}

\newabbreviation[type=notation]{dws}
{DWS}
{Deep Weight Space}

\newabbreviation[type=notation]{fsw}
{FSW}
{Fourier Sliced-Wasserstein}

\newabbreviation[type=notation]{fswgnn}
{FSW-GNN}
{graph neural network using Fourier Sliced-Wasserstein aggregation}

\newabbreviation[type=notation]{tmd}
{TMD}
{Tree Mover's Distance}

\newabbreviation[type=notation]{ds}
{DS}
{doubly stochastic graph metric}

\newabbreviation[type=notation]{mas}
{MAS}
{monotone and separable}

\newabbreviation[type=notation]{masnet}
{MASNet}
{neural architecture for weakly MAS set representations}

\newabbreviation[type=notation]{egennet}
{EGenNet}
{equivariant geometric message-passing architecture introduced for generic separation}

\keepUnusedAbbreviations

\usepackage[capitalize,noabbrev]{cleveref}
\usepackage{caption}
\usepackage{subfigure} 
\usepackage{footmisc}   
\usepackage{amsthm}
\usepackage{siunitx}
\usepackage{graphicx}
\usepackage{multirow}
\usepackage{colortbl}
\usepackage{xcolor}
\usepackage{bm}
\usepackage{float}
\usepackage{microtype}
\usepackage{csquotes}
\usepackage{verbatim}
\usepackage[capitalize,noabbrev]{cleveref}

\theoremstyle{plain}
\newtheorem{theorem}{Theorem}[section]
\newtheorem{proposition}[theorem]{Proposition}
\newtheorem{corollary}[theorem]{Corollary}
\newtheorem{lemma}[theorem]{Lemma}

\theoremstyle{definition}
\newtheorem{definition}[theorem]{Definition}

\theoremstyle{remark}

\definecolor{snsblue}{RGB}{31,119,180}
\definecolor{snsorange}{RGB}{255,127,14}
\definecolor{snsgreen}{RGB}{44,160,44}
\definecolor{snsred}{RGB}{214,39,40}
\definecolor{snspurple}{RGB}{148,103,189}
\newtheorem{claim}[theorem]{Claim}
\usepackage{appendix}
\usepackage{xspace}
\usepackage{adjustbox}
\usepackage{wrapfig}
\usepackage[thinc]{esdiff}
\usepackage[shortlabels,inline]{enumitem}
\newtheorem{example}[theorem]{Example}
\usepackage{booktabs}
\usepackage{etoolbox}
\usepackage{extarrows}
\usepackage{pgfplots}
\usepackage{booktabs}
\usepackage{multirow}
\usepackage{placeins}
\defbibenvironment{pubinfo}
  {\begin{enumerate}
     \setlength{\itemsep}{0.8em}
     \setlength{\parsep}{0pt}}
  {\end{enumerate}}
  {\item}

\usepackage{xcolor}
\usepackage{tikz}
\usepackage{pgfplots}

\pgfplotsset{compat=1.18}
\usepackage{thm-restate}

\preparebutcheredbibliography{front/pubinfo.bib}
\renewenvironment{abstract}
{%
    \small
    \begin{center}
        {\bfseries \abstractname\vspace{-.5em}}
    \end{center}
    \quotation
}
{%
    \endquotation
}

\begin{document}

\makefrontmatter

\chapter{Preamble}
\section{Background: Symmetries, Sets, and Graphs}
\label{sec:background}

In this section, we introduce the main notions used throughout the thesis for learning on structured data. We first formalize symmetry through group actions and define invariant and equivariant functions. We then review permutation-invariant set representations, message-passing neural networks, and their connection to the Weisfeiler--Leman test. Finally, we discuss stronger notions of expressivity based on separation and quantitative preservation of information.

\subsection{Learning with Symmetries}
\label{subsec:symmetries}

Let $\mathcal{G}$ be a group acting on a set $\mathcal{X}$. A
\emph{group action} is a map
\[
    \mathcal{G}\times\mathcal{X}\to\mathcal{X},
    \qquad
    (g,x)\mapsto g\cdot x,
\]
satisfying
\[
    e\cdot x=x,
    \qquad
    g_1\cdot(g_2\cdot x)=(g_1g_2)\cdot x,
\]
for every $x\in\mathcal{X}$ and $g_1,g_2\in\mathcal{G}$.

The orbit of $x$ is
\[
    \mathcal{O}(x)
    =
    \{g\cdot x:g\in\mathcal{G}\}.
\]
In many learning problems, elements in the same orbit correspond to different
representations of the same underlying object.

Suppose that $\mathcal{G}$ acts on both an input space $\mathcal{X}$ and an
output space $\mathcal{Y}$. A function
$f:\mathcal{X}\to\mathcal{Y}$ is called
\emph{$\mathcal{G}$-equivariant} if
\[
    f(g\cdot x)=g\cdot f(x)
\]
for every $g\in\mathcal{G}$ and $x\in\mathcal{X}$.

It is called \emph{$\mathcal{G}$-invariant} if 
\[
    f(g\cdot x)=f(x).
\]
for every $g\in\mathcal{G}$ and $x\in\mathcal{X}$.
Thus, an invariant function gives the same output for all equivalent
representations of an object, whereas an equivariant function transforms its
output consistently with the input. Building these constraints directly into
an architecture is a standard way to encode known structure and reduce the
need to learn symmetry from data \cite{cohen2016group,ravanbakhsh2017equivariance}.

A typical example is graph learning. Relabeling the nodes of a graph should
not affect a graph-level prediction, such as graph classification, and hence
such a prediction should be invariant. On the other hand, node-level
predictions should be permuted together with the nodes, and are therefore
equivariant.

Symmetry, however, specifies only which inputs \emph{must} be treated as
equivalent. It does not by itself determine whether an architecture can
separate inputs that are not related by the symmetry. This distinction between
symmetry and expressive power will be central throughout the thesis.

\subsection{Sets, Multisets, and Permutation-Invariant Embeddings}
\label{subsec:sets}

Consider a collection of vectors
\[
    X=(x_1,\ldots,x_n),
    \qquad
    x_i\in\Omega\subseteq\mathbb{R}^d.
\]
The symmetric group $S_n$ acts by permuting the elements:
\[
    \pi\cdot X
    =
    (x_{\pi^{-1}(1)},\ldots,x_{\pi^{-1}(n)}).
\]
If the ordering carries no information, a function of $X$ should be invariant
to this action.

We use
\[
    S=\{\!\{x_1,\ldots,x_n\}\!\}
\]
to denote a \emph{multiset}: an unordered collection in which repetitions are
allowed. Multisets arise naturally throughout this thesis, particularly when
aggregating the features of neighboring nodes in a graph.
Let \(\Omega\) be a ground set, and denote by \(\mathcal{M}(\Omega)\) the collection of all finite multisets with elements in \(\Omega\). 

A multiset embedding is a map
$$
\Phi:\mathcal{M}(\Omega)\to\mathbb{R}^m.
$$
Equivalently, multiset functions correspond to permutation-invariant functions on multi-sets.

A basic and widely used construction is the
DeepSets from \cite{zaheer2017deep}
\[
    \Phi(S)
    =
    \rho\left(
        \sum_{x\in S}\phi(x)
    \right),
\]
where $\phi$ and $\rho$ are learned functions. More generally, one can use
any permutation-invariant aggregation operator,
\[
    \Phi(S)
    =
    \rho\left(
        \operatorname{AGG}
        \{\!\{\phi(x):x\in S\}\!\}
    \right).
\]

\subsection{Graphs and Message Passing Neural Networks}
\label{subsec:graphs-mpnn}

A vertex-featured graph is represented as
\[
    G=(V,E,X),
\]
where
\[
    V=\{v_1,\ldots,v_n\}
\]
is the vertex set,
\[
    E\subseteq
    \big\{\{v_i,v_j\}:i,j\in[n]\big\}
\]
is the edge set, and
\[
    X=(x_1,\ldots,x_n),
    \qquad
    x_i\in\mathbb{R}^d,
\]
contains the node features.

We denote the neighborhood of a node $v$ by
\[
    \mathcal{N}(v)
    =
    \{u\in V:\{u,v\}\in E\}.
\]

Equivalently, a graph may be represented by an adjacency matrix
$A\in\{0,1\}^{n\times n}$ and a node-feature matrix $X$.
Given a permutation matrix \(P\), a relabeling of the vertices transforms the graph representation according to

$$
    (A,X)\mapsto (PAP^\top,PX).
$$

Two graphs are isomorphic when their representations are related by such a
relabeling. A graph-level function should therefore satisfy
\[
    F(PAP^\top,PX)=F(A,X),
\]
whereas a node-level function should satisfy
\[
    F(PAP^\top,PX)=PF(A,X).
\]

\paragraph{Message passing.}
Message Passing Neural Networks (MPNNs) construct node representations by
iteratively aggregating information from neighboring nodes
\cite{gilmer2017neural}. Starting from
\[
    h_v^{(0)}=x_v,
\]
a general message-passing layer has the form
\[
    m_v^{(t)}
    =
    \operatorname{AGG}_t
    \left(
        \left\{\!\left\{
        M_t\left(
            h_v^{(t-1)},h_u^{(t-1)}
        \right)
        :
        u\in\mathcal{N}(v)
        \right\}\!\right\}
    \right),
\]
followed by
\[
    h_v^{(t)}
    =
    U_t\left(
        h_v^{(t-1)},m_v^{(t)}
    \right).
\]

The aggregation operator is permutation invariant, ensuring that the overall
network is equivariant to node relabeling.

For graph-level tasks, the final node representations are combined using an
invariant readout,
\[
    h_G
    =
    \operatorname{READOUT}
    \left(
        \{\!\{h_v^{(T)}:v\in V\}\!\}
    \right).
\]

After $T$ iterations, the representation of a node can depend only on its
$T$-hop neighborhood.

\paragraph{Examples of MPNN architectures.}
Many popular graph neural networks are special cases of the message-passing
framework.

\begin{itemize}

\item \textbf{Graph Convolutional Network (GCN).}
A GCN layer performs a normalized average over neighboring features
\cite{kipf2017semi}:
\[
h_v^{(t)}
=
\sigma\left(
W_t
\sum_{u\in\mathcal{N}(v)\cup\{v\}}
\frac{1}{\sqrt{\tilde d_v\tilde d_u}}
h_u^{(t-1)}
\right),
\]
where $\tilde d_v$ is the degree after adding self-loops.

\item \textbf{GraphSAGE.}
GraphSAGE aggregates neighboring features and combines them with the current
node representation \cite{hamilton2017inductive}:
\[
h_v^{(t)}
=
\sigma\left(
W_t
\left[
h_v^{(t-1)}
\,\Vert\,
\operatorname{AGG}
\{\!\{h_u^{(t-1)}:u\in\mathcal N(v)\}\!\}
\right]
\right).
\]

\item \textbf{Graph Attention Network (GAT).}
GAT assigns learned attention weights to the neighbors \cite{velickovic2018graph}:
\[
h_v^{(t)}
=
\sigma\left(
\sum_{u\in\mathcal N(v)\cup\{v\}}
\alpha_{vu}^{(t)}
W_t h_u^{(t-1)}
\right),
\]
where $\alpha_{vu}^{(t)}$ is computed from the features of $v$ and $u$.

\item \textbf{Graph Isomorphism Network (GIN).}
GIN uses sum aggregation followed by an MLP \cite{xu2019powerful}:
\[
h_v^{(t)}
=
\operatorname{MLP}_t
\left(
(1+\epsilon_t)h_v^{(t-1)}
+
\sum_{u\in\mathcal N(v)}
h_u^{(t-1)}
\right).
\]
GIN was designed to match the distinguishing power of the
one-dimensional Weisfeiler--Leman test.
\end{itemize}

\paragraph{The Weisfeiler--Leman test.}
The expressive power of MPNNs is closely related to the one-dimensional
Weisfeiler--Leman (1-WL) graph isomorphism test
\cite{weisfeiler1968reduction,xu2019powerful,morris2019weisfeiler}.

The 1-WL algorithm assigns a color $c_v^{(t)}$ to every node and updates it
according to
\[
    c_v^{(t)}
    =
    \operatorname{HASH}
    \left(
        c_v^{(t-1)},
        \{\!\{
        c_u^{(t-1)}
        :
        u\in\mathcal N(v)
        \}\!\}
    \right),
\]
where the hash function is assumed to be injective.

Two graphs are distinguished if, at some iteration, the multisets of node
colors differ. We write
\[
    G\equiv_{\mathrm{WL}} H
\]
when 1-WL does not distinguish $G$ and $H$.

The similarity with message passing is direct: both procedures repeatedly
combine the current representation of a node with the multiset of
representations in its neighborhood. Standard MPNNs cannot distinguish
graphs that 1-WL fails to distinguish. Conversely, MPNNs with sufficiently
expressive and injective aggregation and update functions can attain the
distinguishing power of 1-WL \cite{xu2019powerful,morris2019weisfeiler}.

This connection provides a natural benchmark for the expressive power of
message-passing architectures.

\subsection{From Symmetry to Expressivity}
\label{subsec:expressivity}

Invariance specifies which inputs a representation should identify, but does
not guarantee that it preserves all other distinctions. This is important for
approximation: if a representation maps two nonequivalent inputs to the same
point, then no subsequent function of that representation can distinguish
between them.

Let $\sim$ denote the equivalence relation induced by the relevant symmetry,
and let

$$
    F:\mathcal X\to\mathbb R^m
$$

be an invariant representation. We say that $F$ \emph{separates} inputs up to
the symmetry if

$$
    F(x)=F(y)
    \quad\Longrightarrow\quad
    x\sim y.
$$

Equivalently,

$$
    x\not\sim y
    \quad\Longrightarrow\quad
    F(x)\neq F(y).
$$

Thus, separation requires the representation to identify only inputs that are
equivalent under the prescribed symmetry.

For graphs, the relevant equivalence relation may be graph isomorphism or a
coarser relation such as 1-WL equivalence. Accordingly, the expressive power
of a graph architecture is commonly measured by which pairs of nonequivalent
graphs it is able to distinguish.

\subsection{Geometric and Multi-Permutation Symmetries}
\label{subsec:structured-symmetries}

The symmetry framework above also extends beyond ordinary vertex relabeling.
Two examples are particularly relevant to this thesis.

\paragraph{Geometric graphs and point clouds.}
For geometric data with positions
$x_i\in\mathbb{R}^d$, one often requires invariance or equivariance not only
to permutations but also to rigid transformations,
\[
    x_i\mapsto Qx_i+t,
    \qquad
    Q\in O(d),\quad t\in\mathbb{R}^d.
\]
This is the basic symmetry principle behind many geometric and Euclidean
equivariant neural networks \cite{thomas2018tensor,satorras2021n}.
Accordingly, two geometric graphs should be regarded as equivalent when they
differ only by a relabeling of their vertices and a rigid motion of their
coordinates. Chapter~\ref{chap:paper1} studies when sparse geometric
message-passing architectures can separate such objects, with particular
attention to generic completeness and to the role of the underlying graph
structure.

\subsection{Multiple permutation symmetries.}
Some structured inputs contain several unordered axes. For example, for
\[
    X\in\mathbb{R}^{n\times m\times d},
\]
the product group
\[
    S_n\times S_m
\]
may act independently on the first two indices:
\[
    ((\pi,\sigma)\cdot X)_{i,j,:}
    =
    X_{\pi^{-1}(i),\sigma^{-1}(j),:}.
\]
Linear layers equivariant to such actions can be characterized through
parameter-sharing arguments \cite{ravanbakhsh2017equivariance,hartford2018deep}.
Chapter~\ref{chap:paper2} develops a complementary representation-theoretic
approach based on irreducible decompositions and Schur's lemma, and applies it
to several settings with multiple permutation symmetries.

\subsection{Thesis Overview}
\label{subsec:thesis-overview}

The main body of this thesis is organized around four research works, presented in Chapters~\ref{chap:paper1}--\ref{chap:paper4}. Each of these chapters contains the corresponding paper and develops one of the main contributions of the thesis in full detail. Chapter~\ref{chap:methods} provides a more concise and self-contained account of the main ideas and methodology underlying these works, and may therefore serve as an overview before turning to the individual papers.

Chapter~\ref{chap:paper1} studies the expressive power of message-passing neural networks on sparse geometric graphs \cite{thesisEGenNet2025}. Chapter~\ref{chap:paper2} develops a representation-theoretic approach to the characterization of equivariant linear layers \cite{thesisSchurNet2025}. Chapter~\ref{chap:paper3} studies set representations that preserve set inclusion \cite{thesisMonotone2025}, while Chapter~\ref{chap:paper4} introduces FSW-GNN, a WL-equivalent MPNN with strong theoretical guarantees and empirical performance on long-range graph-learning tasks \cite{thesisFSWGNN2025}.

Taken together, these works study how neural representations can respect the symmetries of structured data while preserving progressively richer forms of information.

\let\citep\parencite
\let\citet\textcite
\chapter{Research Methods}
\label{chap:methods}

This chapter provides a concise overview of the four papers comprising the thesis. Each section summarizes the main problem, methodology, and principal results of one paper. Specifically, \cref{sec:methods-geometric} summarizes the first paper presented in \cref{chap:paper1}, \cref{sec:methods-equivariant} summarizes the second paper presented in \cref{chap:paper2}, \cref{sec:methods-sets} summarizes the third paper presented in \cref{chap:paper3}, and \cref{sec:methods-fsw} summarizes the fourth paper presented in \cref{chap:paper4}.

\section{Geometric Graphs and Generic Separation}
\label{sec:methods-geometric}

This section summarizes the first paper of the thesis, presented in
\cref{chap:paper1}. The main question is simple: when can a geometric
message-passing network distinguish two geometric graphs that are not related
by the natural symmetries of the problem?

\subsection{Geometric Graphs}

A \emph{geometric graph} is a graph whose vertices are embedded in Euclidean
space. We write

$$
    G=(A,X),
$$

where \(A\in\{0,1\}^{n\times n}\) is the adjacency matrix and

$$
    X=(x_1,\ldots,x_n),\qquad x_i\in\mathbb{R}^d,
$$

contains the vertex positions.

Such graphs admit several natural symmetries. Relabeling the vertices by a
permutation matrix \(P\) transforms the representation as

$$
    (A,X)\mapsto(PAP^\top,PX),
$$

while a rigid transformation acts on the coordinates as

$$
    x_i\mapsto Qx_i+t,
$$

where \(Q\in O(d)\) and \(t\in\mathbb{R}^d\).
Two geometric graphs are considered equivalent if one can be obtained from
the other by these transformations.

Our results are stated for \emph{generic} geometric graphs. Informally,
genericity rules out special degenerate configurations and holds for almost
every choice of vertex positions.

Another important notion is \emph{global rigidity}. A geometric graph is
globally rigid if its edge lengths determine the entire realization, up to a
rigid transformation. Thus, if a graph is globally rigid, knowing the lengths
of its edges is enough to reconstruct its geometry up to the symmetries that
we wish to ignore.

\subsection{Message Passing on Geometric Graphs}

We study two families of geometric message-passing networks.

An \emph{invariant geometric GNN} (I-GGNN) propagates only invariant scalar
features. Geometric information enters the network through edge lengths:

$$
    s_i^{(t+1)}
    =
    f_t\!\left(
        \{\!\{
        (s_j^{(t)},\|x_i-x_j\|)
        :j\in\mathcal N(i)
        \}\!\}
    \right).
$$

As a result, an I-GGNN does not retain directional information during message
passing.

An \emph{equivariant geometric GNN} (E-GGNN), in contrast, may propagate
vector-valued features. Its messages can depend on relative positions
\(x_i-x_j\), allowing directional information to be preserved:

$$
    v_i^{(t+1)}
    =
    f_t\!\left(
        \{\!\{
        (v_j^{(t)},x_i-x_j)
        :j\in\mathcal N(i)
        \}\!\}
    \right).
$$

The message-passing layers are required to be permutation equivariant and
rotation equivariant, while the final graph-level output is invariant to
vertex permutations and rigid transformations.

To study the limitations of the message-passing framework itself, rather than
those of a particular choice of neural networks, we consider
\emph{maximally expressive} I-GGNNs and E-GGNNs. Roughly speaking, all
operations are assumed to preserve as much information as their required
symmetries allow. In particular:

\begin{itemize}
\item the neighborhood aggregation does not identify two different
multisets unless this is forced by the relevant symmetry;
\item equivariant functions do not identify two inputs unless they are
related by the corresponding group action;
\item invariantization identifies only inputs related by the symmetry;
and
\item the final readout is injective on multisets of node features.
\end{itemize}

These assumptions give the strongest possible expressive power within each
message-passing framework and allow us to ask which limitations are inherent
to invariant or equivariant message passing itself.

\subsection{Main Expressivity Results}

Our main theoretical result gives a complete generic characterization for
both classes of networks.

For I-GGNNs, the relevant graph property is \emph{generic global rigidity}.
The reason is intuitive: an I-GGNN observes the geometry only through the
lengths of graph edges. If two different realizations have exactly the same
edge lengths, then the network receives the same geometric information from
both and cannot distinguish them.

We prove that a maximally expressive I-GGNN can generically identify a
geometric graph precisely when the underlying graph is generically globally
rigid:

$$
    \boxed{
    \text{I-GGNN generic identification}
    \iff
    \text{generic global rigidity}.
    }
$$

For E-GGNNs, the situation is different. Since vector features can preserve
directional information, the network is not limited to edge lengths. We show
that global rigidity is no longer required: the only necessary graph property
is connectivity.

Indeed, if the graph is disconnected, its connected components can be moved
independently and message passing cannot communicate between them. On the
other hand, for a connected graph, a maximally expressive E-GGNN with
sufficient depth can generically identify the entire geometric graph. Hence,

$$
    \boxed{
    \text{E-GGNN generic identification}
    \iff
    \text{connectivity}.
    }
$$

This gives a sharp distinction between the two frameworks. Invariant message
passing requires the considerably stronger condition of global rigidity,
whereas equivariant message passing requires only connectivity.

\subsection{EGenNet}

The previous results describe what a \emph{maximally expressive} E-GGNN
could achieve. We next ask whether these requirements can actually be
realized by a concrete and practical neural architecture.

To this end, we introduce \emph{EGenNet}. The architecture is designed as an
explicit implementation of the E-GGNN framework described above. Its
message-passing layers are permutation and rotation equivariant, translation
is handled through relative coordinates, and its final representation is
invariant to the symmetries of the geometric graph.

More importantly, EGenNet satisfies the maximal-expressivity requirements:
its aggregation, equivariant transformations, invariantization, and final
readout can all be chosen so that they do not lose information beyond what is
required by the corresponding symmetries.

We prove that EGenNet therefore attains the generic expressive power predicted
by our characterization of E-GGNNs. In particular, with sufficient depth it
generically distinguishes connected geometric graphs exactly up to geometric
isomorphism.

Thus, EGenNet connects the abstract expressivity result with an explicit
neural architecture: the theoretical advantage of equivariant message passing
is not merely an idealized property of the framework, but can be realized by
a concrete model. We further validate the architecture empirically on
synthetic separation tasks and molecular learning benchmarks.

\section{Characterizing Equivariant Linear Layers}
\label{sec:methods-equivariant}

A recurring problem in equivariant deep learning is the following.
Suppose that a symmetry group acts linearly on a vector space
$\mathcal{V}$. We would like to characterize all linear maps
\[
    T:\mathcal{V}\to\mathcal{V}
\]
that respect the symmetry, namely,
\[
    T(g\cdot x)=g\cdot T(x)
    \qquad
    \forall g\in\mathcal{G},\ x\in\mathcal{V}.
\]
Once these linear maps are characterized, they can be used as the basic
linear layers of an equivariant neural network.

\paragraph{The traditional approach.}
For permutation symmetries, this problem has traditionally been addressed
through \emph{parameter sharing}. One begins with a general linear map and
uses the equivariance constraints to determine which entries of its parameter
tensor must be equal.

Equivalently, one can study the orbits induced by the group on the input and
output indices: parameters associated with indices that belong to the same
orbit must be shared. This approach leads, for example, to the classical
characterizations of DeepSets and higher-order invariant graph networks.

The parameter-sharing viewpoint is effective, but the required bookkeeping
becomes increasingly involved when several permutation groups act
simultaneously, or when the group acts on several interacting tensors.

\paragraph{Our approach.}
In Chapter~\ref{chap:paper2}, we study the same characterization problem from
a different perspective. Instead of working directly with individual entries
of the parameter tensor, we decompose the underlying vector space into
minimal subspaces that are preserved by the symmetry group.

Let $\mathcal{G}$ be a finite group acting linearly on a finite-dimensional
real vector space $\mathcal{V}$. A subspace
\[
    \mathcal{U}\subseteq\mathcal{V}
\]
is called \emph{invariant} if
\[
    g\cdot u\in\mathcal{U}
    \qquad
    \forall g\in\mathcal{G},\ u\in\mathcal{U}.
\]
An invariant subspace is called \emph{irreducible} if it contains no
nonzero proper invariant subspace.

For the finite groups considered here, the space can be decomposed as a
direct sum
\[
    \mathcal{V}
    =
    \mathcal{V}_1\oplus\cdots\oplus\mathcal{V}_s,
\]
where each $\mathcal{V}_i$ is irreducible.

The main idea is that after this decomposition has been found, the possible
linear equivariant maps between the different components are extremely
restricted.

\subsection{A Simple Example}

Consider the action on $\mathbb{R}^2$ that exchanges the two coordinates,
\[
    (x_1,x_2)\mapsto(x_2,x_1).
\]
There are two natural invariant subspaces:
\[
    \mathcal{U}_{+}
    =
    \operatorname{span}\{(1,1)\},
    \qquad
    \mathcal{U}_{-}
    =
    \operatorname{span}\{(1,-1)\}.
\]
Hence
\[
    \mathbb{R}^2
    =
    \mathcal{U}_{+}\oplus\mathcal{U}_{-}.
\]

The symmetry acts trivially on $\mathcal{U}_{+}$, while it acts by a sign
change on $\mathcal{U}_{-}$. These are therefore two different symmetry
types.

An equivariant linear map cannot arbitrarily mix the two components. It can
only scale each component independently:
\[
    T(x_{+}+x_{-})
    =
    a x_{+}+b x_{-}.
\]

This elementary example already illustrates the basic principle of our
approach: first decompose the space according to how the symmetry acts, and
then determine which interactions between the resulting components are
possible.

\subsection{Schur's Lemma}

This principle is formalized by Schur's lemma.

Let $\mathcal{U}$ and $\mathcal{W}$ be irreducible invariant subspaces.
We say that they are of the same symmetry type if there exists an equivariant
linear bijection
\[
    L:\mathcal{U}\to\mathcal{W}.
\]

Schur's lemma states that an equivariant linear map
\[
    T:\mathcal{U}\to\mathcal{W}
\]
satisfies the following:

\begin{itemize}
    \item if $\mathcal{U}$ and $\mathcal{W}$ are not of the same symmetry
    type, then
    \[
        T=0;
    \]

    \item if they are of the same symmetry type, then, in the settings
    considered in this work,
    \[
        T=\lambda L
    \]
    for some scalar $\lambda\in\mathbb{R}$ and a fixed equivariant
    isomorphism $L$.
\end{itemize}

Technically, the second statement over the real numbers requires the
irreducible spaces to be \emph{absolutely irreducible}; all irreducible
components used in the constructions below satisfy this property.

Consequently, suppose that
\[
    \mathcal{V}
    =
    \bigoplus_{i=1}^{s}\mathcal{V}_i
\]
is an irreducible decomposition. A linear equivariant map is completely
determined by the pairs of components that are of the same symmetry type.

If the distinct symmetry types occur with multiplicities
\[
    \alpha_1,\ldots,\alpha_k,
\]
then the dimension of the space of equivariant linear maps is
\[
    \sum_{i=1}^{k}\alpha_i^2.
\]

Thus, the characterization of an equivariant layer reduces to two main
steps:

\begin{enumerate}
    \item decompose the feature space into irreducible invariant subspaces;
    \item determine which of these subspaces are of the same symmetry type.
\end{enumerate}

Schur's lemma then provides the form of all possible linear equivariant
mappings automatically.

\subsection{The Classical Permutation Example: DeepSets}

We first apply this approach to the classical action of the symmetric group
$S_n$ on
\[
    \mathcal{V}=\mathbb{R}^n
\]
by permutation of coordinates.

There are two irreducible invariant subspaces:
\[
    \mathbf{S}
    =
    \operatorname{span}\{\mathbf{1_n}\},
\]
and
\[
    \mathbf{V}(n)
    =
    \left\{
        x\in\mathbb{R}^n:
        \sum_{i=1}^{n}x_i=0
    \right\}.
\]
Every vector can be uniquely decomposed as
\[
    x
    =
    \bar{x}\mathbf{1_n}
    +
    \left(x-\bar{x}\mathbf{1_n}\right),
    \qquad
    \bar{x}
    =
    \frac{1}{n}\sum_{i=1}^{n}x_i.
\]

The two subspaces are not of the same symmetry type. Therefore Schur's lemma
implies that a linear permutation-equivariant map has two free parameters:
one acting on the constant component and one on the zero-sum component.

Equivalently, every such map has the form
\[
    T(x)
    =
    a x
    +
    b
    \left(
        \sum_{i=1}^{n}x_i
    \right)\mathbf{1}.
\]

This recovers the classical linear equivariant layer underlying DeepSets.
The usual two-parameter characterization, commonly derived through parameter
sharing, therefore follows immediately from the decomposition
\[
    \mathbb{R}^n
    =
    \mathbf{S}\oplus\mathbf{V}(n).
\]

\subsection{Second-Order Graph Equivariance: 2-IGN}

A more involved example is the second-order graph setting underlying
2-IGN. Here
\[
    \mathcal{V}
    =
    \mathbb{R}^{n\times n},
\]
and a permutation $\tau\in S_n$ acts simultaneously on the two indices:
\[
    (\tau\cdot X)_{ij}
    =
    X_{\tau^{-1}(i),\tau^{-1}(j)}.
\]
In matrix notation,
\[
    X\mapsto PXP^\top,
\]
where $P$ is the permutation matrix associated with $\tau$.

For $n\geq4$, we show that
$\mathbb{R}^{n\times n}$ decomposes into seven irreducible invariant
subspaces.

Two of these components are of the same symmetry type, three further
components are of a second common type, and the remaining two components
each occur once. The multiplicities are therefore
\[
    2,\quad 3,\quad 1,\quad 1.
\]

Schur's lemma immediately gives
\[
    2^2+3^2+1^2+1^2=15
\]
free parameters.

Thus the space of second-order permutation-equivariant linear layers is
15-dimensional, recovering exactly the classical 2-IGN characterization.

The significance here is methodological. The fifteen parameters are usually
obtained by studying fifteen different parameter-sharing patterns in a
four-index tensor. Our derivation instead follows from the structure of the
seven invariant subspaces and their isomorphism relations.
\subsection{Independent permutations of rows and columns.}
Consider the action of \(S_n\times S_m\) on
\(\mathbb{R}^{n\times m}\), where \(S_n\) permutes the rows and
\(S_m\) permutes the columns:
\[
(P,Q)\cdot X = PXQ^\top .
\]
Under this action, the matrix space decomposes into four natural
invariant subspaces. The first is the one-dimensional space of
constant matrices,
\[
U_{1}
=
\left\{
\begin{bmatrix}
c & c & \cdots & c\\
c & c & \cdots & c\\
\vdots & \vdots & \ddots & \vdots\\
c & c & \cdots & c
\end{bmatrix}
:\; c\in\mathbb{R}
\right\}.
\]
The second consists of matrices whose rows are all identical,
with the common row having zero sum,
\[
U_{2}
=
\left\{
\begin{bmatrix}
-\,x^\top\,-\\
-\,x^\top\,-\\
\vdots\\
-\,x^\top\,-
\end{bmatrix}
:\;
x\in\mathbb{R}^m,\;
\sum^{m}_{i=1} x_i=0
\right\},
\]
and therefore has dimension \(m-1\). The third consists of matrices
whose columns are all identical,
\[
U_{3}
=
\left\{
\begin{bmatrix}
| & | &        & |\\
y & y & \cdots & y\\
| & | &        & |
\end{bmatrix}
:\;
y\in\mathbb{R}^n,\;
\sum^{n}_{i=1} y_i=0
\right\},
\]
and has dimension \(n-1\). Finally,
\[
U_{4}
=
\left\{
Z\in\mathbb{R}^{n\times m}:
Z\mathbf{1}_m=0,\;
\mathbf{1}_n^\top Z=0
\right\}
\]
is the space of matrices whose row sums and column sums are all zero,
and has dimension \((n-1)(m-1)\). Hence
\[
\mathbb{R}^{n\times m}
=
U_{1}\oplus U_{2}\oplus U_{3}\oplus U_{4}.
\]
These four subspaces are irreducible and pairwise non-isomorphic
\(S_n\times S_m\)-representations. Therefore, by Schur's lemma, every
linear \(S_n\times S_m\)-equivariant layer acts by an independent
scalar on each component: if
\[
X=X_{00}+X_{01}+X_{10}+X_{11},
\qquad X_{ab}\in U_{ab},
\]
then
\[
T(X)
=
\lambda_{00}X_{00}
+\lambda_{01}X_{01}
+\lambda_{10}X_{10}
+\lambda_{11}X_{11}.
\]
Equivalently, such a layer is determined by four parameters and can
be written entrywise as
\[
(TX)_{ij}
=
aX_{ij}
+b\sum_{j'=1}^m X_{ij'}
+c\sum_{i'=1}^n X_{i'j}
+d\sum_{i'=1}^n\sum_{j'=1}^m X_{i'j'}.
\]

\subsection{Wreath-Product Equivariance}
\label{subsec:methods-wreath}

The direct-product example above assumes that the same collection of row and
column permutations acts on a single object. A different symmetry appears
when the input itself is a set of \emph{unaligned symmetric objects}. Suppose
that each object belongs to a representation space $\mathcal V$ of a finite
group $\mathcal G$, and consider a $k$-tuple
\[
    (v_1,\ldots,v_k)\in\mathcal V^k.
\]
Each object may be transformed independently by an element of $\mathcal G$,
while the $k$ objects themselves may also be permuted. The resulting symmetry
group is the wreath product
\[
    \mathcal G\wr S_k
    =
    \mathcal G^k\rtimes S_k,
\]
where $S_k$ acts on $\mathcal G^k$ by permuting its $k$ factors
\citep{wreath,maron2020learning}. With the standard convention, its action on
$\mathcal V^k$ can be written as
\[
    ((g_1,\ldots,g_k),\tau)\cdot(v_1,\ldots,v_k)
    =
    \bigl(
        g_1\cdot v_{\tau^{-1}(1)},\ldots,
        g_k\cdot v_{\tau^{-1}(k)}
    \bigr).
\]
Thus the symmetry contains two distinct operations: independent
``alignments'' inside each object and permutation of the objects themselves.
This setting appears, for example, in collections of graphs, collections of
neural-network weight spaces, hierarchical data, and alignment problems.

A simple way to obtain a wreath-equivariant layer is to apply the same
$\mathcal G$-equivariant map independently to every element,
\[
    L_{\mathrm{Siam}}(v_1,\ldots,v_k)
    =
    \bigl(\widehat L(v_1),\ldots,\widehat L(v_k)\bigr).
\]
We refer to such a layer as \emph{Siamese}. The main question studied in
Chapter~\ref{chap:paper2} is whether this exhausts all linear
$\mathcal G\wr S_k$-equivariant maps. In general, it does not.

The characterization is controlled by the fixed-point subspace
\[
    \mathcal V_{\mathrm{fixed}}
    =
    \{v\in\mathcal V:
      g\cdot v=v\ \text{for every }g\in\mathcal G\}.
\]
Let
\[
    s=\dim \mathcal V_{\mathrm{fixed}},
\]
and choose a basis $e_1,\ldots,e_s$ of this subspace. Since $\mathcal G$ is
finite, $\mathcal V$ admits a $\mathcal G$-invariant inner product; it can be
obtained by averaging any inner product over the group
\citep{fulton2013representation}.

Our characterization shows that every linear
$\mathcal G\wr S_k$-equivariant map $L:\mathcal V^k\to\mathcal V^k$ has the
form
\[
\begin{split}
    L(v_1,\ldots,v_k)
    ={}&
    \bigl(
       \widehat L(v_1),\ldots,\widehat L(v_k)
    \bigr)
    \\
    &+
    \sum_{i,j=1}^{s} a_{ij}
    \left(
      \sum_{\ell=1}^{k}\langle v_\ell,e_i\rangle e_j,
      \ldots,
      \sum_{\ell=1}^{k}\langle v_\ell,e_i\rangle e_j
    \right),
\end{split}
\]
where $\widehat L:\mathcal V\to\mathcal V$ is an arbitrary linear
$\mathcal G$-equivariant map and $a_{ij}\in\mathbb R$. The first term is the
Siamese part. The second term couples different elements of the set: it
extracts invariant scalar information from all $k$ elements, aggregates that
information globally, and broadcasts it back to every element in a fixed
direction. These are precisely the additional non-Siamese interactions.

The representation-theoretic reason for this form is particularly simple.
Every nontrivial irreducible component of $\mathcal V$ remains irreducible
when lifted to $\mathcal V^k$ under the wreath-product action. In contrast,
each one-dimensional trivial component produces the usual two irreducible
$S_k$ components: a constant component and a zero-sum component. Schur's
lemma therefore creates an additional family of interactions only through
the trivial part of $\mathcal V$.

Consequently, beyond the Siamese equivariant maps there are exactly $ s^2$.
independent non-Siamese linear maps. This recovers the previously studied
transitive case, where $s=1$ and there is only one additional interaction,
but gives substantially richer layers when the group action has several
independent invariant directions. For example, for graph matrices
$\mathcal V=\mathbb R^{n\times n}$ under simultaneous node permutations,
$s=2$, giving four non-Siamese maps. For deep weight spaces, the dimension
of the fixed-point subspace can be much larger, yielding a correspondingly
large family of additional equivariant interactions.

\section{Monotone and Separable Set Representations}
\label{sec:methods-sets}

This section summarizes the third paper of the thesis, presented in
\cref{chap:paper3}. A central goal in learning on sets is to construct
permutation-invariant representations that preserve the information contained
in the input set. DeepSets \citep{zaheer2017deep} provides a fundamental
framework for this purpose, using representations of the form

$$
    F(S)=\rho\left(\sum_{x\in S}\phi(x)\right).
$$

Much of the classical theory asks whether such representations are expressive
enough to distinguish different sets. Our work asks for a stronger property:
can the representation preserve not only whether two sets are different, but
also whether one is contained in the other?

\subsection{Monotone and Separable Set Functions}

Let \(V\) be a ground set. We seek a representation

$$
    F(S)\in\mathbb{R}^m
$$

for finite multisets \(S\subseteq V\), such that multiset containment is
represented exactly by the coordinatewise order:

$$
    S\subseteq T
    \quad\Longleftrightarrow\quad
    F(S)\leq F(T).
$$

We call \(F\) \emph{monotone} if

$$
    S\subseteq T
    \quad\Longrightarrow\quad
    F(S)\leq F(T),
$$

and \emph{separable} if

$$
    F(S)\leq F(T)
    \quad\Longrightarrow\quad
    S\subseteq T.
$$

A function satisfying both properties is called a
\emph{monotone and separable} (MAS) function.

This requirement is stronger than ordinary injectivity. An injective
representation only distinguishes different sets; a MAS representation
preserves the entire partial order induced by containment.

\subsection{Finite Ground Sets}

We first assume that the ground set is finite,

$$
    |V|=n,
$$

while placing no bound on the cardinality of the input multisets.

In this setting we obtain an exact result. The smallest dimension in which a
MAS representation can exist is precisely

$$
    \boxed{
    m^*(V,\infty)=n.
    }
$$

The upper bound is simple: use one coordinate for each element of \(V\), and
let \(F(S)\) record its multiplicity in \(S\). Coordinatewise comparison then
coincides exactly with multiset containment. Our lower bound shows that no
representation into fewer than \(n\) coordinates can satisfy both
monotonicity and separability for multisets of arbitrary size.

\subsection{Infinite Ground Sets}

The situation changes fundamentally when the ground set is infinite, as is
typically the case in machine learning, for example when
\(V=\mathbb{R}^d\).

We prove that for every \(k\geq2\), there is no finite-dimensional MAS
representation on multisets of cardinality at most \(k\):

$$
    \boxed{
    |V|=\infty,\ k\geq2
    \quad\Longrightarrow\quad
    \text{no finite-dimensional MAS representation exists}.
    }
$$

Thus, exact preservation of containment is too strong to be achieved in
finite dimension on general continuous domains.

\subsection{Bounded Multiset Cardinality}

For a finite ground set, the dimension \(n\) may still be prohibitively large.
We therefore ask whether the situation improves if every input multiset
contains at most \(k\) elements.

It does. We prove the upper bound

$$
    m^*(V,k)
    \leq
    \min\left\{
        n,\,
        (k+2)^{k+2}\log n
    \right\}.
$$

Thus, for fixed \(k\), the required dimension can grow only logarithmically
with the size of the ground set rather than linearly.

We complement this with lower bounds. For \(k\geq2\),

$$
    m^*(V,k)
    \geq
    \log_2(\log_3 n),
$$

and, when \(k\leq (n-1)/2\),

$$
    m^*(V,k)\geq2k.
$$

Hence, in this regime,

$$
    m^*(V,k)
    \geq
    \max\left\{
        2k,\,
        \log_2(\log_3 n)
    \right\}.
$$

These results show that bounding the input size dramatically reduces the
dimension required for containment-preserving representations, while some
dependence on both \(k\) and \(n\) remains unavoidable.

\subsection{Approximate Containment}

Exact containment is a Boolean notion, whereas in learning problems the set
elements are often noisy or continuous. We therefore also introduce an
asymmetric distance that measures how far a set \(S\) is from being contained
in another set \(T\).

For \(V\subseteq\mathbb{R}^d\) and \(|S|\leq|T|\), we define

$$
    d_{\subseteq}(S,T)
    =
    \min_{\substack{T'\subseteq T\\|T'|=|S|}}
    \operatorname{EMD}(S,T'),
$$

where \(\operatorname{EMD}\) denotes the Earth Mover Distance.

The asymmetry is intentional. We are not asking how similar \(S\) and \(T\)
are, but rather how close \(S\) is to \emph{some subset} of \(T\). In
particular,

$$
    d_{\subseteq}(S,T)=0
    \quad\Longleftrightarrow\quad
    S\subseteq T.
$$

If \(S\) is not exactly contained in \(T\), but each of its elements can be
matched closely to elements of \(T\), then \(d_{\subseteq}(S,T)\) is small.

This allows us to strengthen the purely Boolean notion of separation. We
introduce a lower H"older separability condition, under which the amount by
which the embeddings violate the coordinatewise order is quantitatively
controlled from below by \(d_{\subseteq}(S,T)\). Thus, sets that are farther
from satisfying containment are also easier to separate in the embedding
space. This extends ideas of H"older stability for multiset representations
\citep{davidson2024h} to the asymmetric
set-containment setting.

\subsection{MASNet}

The impossibility result for infinite ground sets motivates relaxing exact
MAS while preserving its most useful structural property. We introduce a
parametric notion called \emph{weakly MAS}. A family \(F(S;w)\) is weakly MAS
if

\begin{enumerate}
\item for every parameter choice \(w\), \(F(\cdot;w)\) is monotone; and
\item whenever \(S\not\subseteq T\), there exists a parameter choice \(w\)
for which
\(        F(S;w)\not\leq F(T;w).
   \)
\end{enumerate}

Thus, monotonicity is guaranteed for every model in the family, while
different parameter choices may separate different pairs of non-contained
sets.

Based on this relaxation, we introduce \emph{MASNet}, a DeepSets-style
architecture of the form

$$
    \operatorname{MASNet}(S)
    =
    M_2\left(
        \sum_{x\in S}
        \sigma\!\left(M_1(x)\right)
    \right).
$$

The construction is deliberately simple. The element-wise network \(M_1\)
maps each input element to a latent representation, the non-negative
activation \(\sigma\) produces contributions that are summed over the set,
and \(M_2\) is chosen to be a monotone vector-to-vector map. Consequently,
adding elements to the input can only increase the resulting representation,
so monotonicity holds by construction.

The choice of \(\sigma\) determines the separation properties. Using
hat-shaped activations yields a weakly MAS model and, under the assumptions
developed in the paper, also provides lower H"older separability with respect
to \(d_{\subseteq}\). Alternatively, a two-layer ReLU realization of \(M_1\)
can also produce a weakly MAS model.

Hence, MASNet retains exact monotonicity while replacing the impossible
requirement of a single finite-dimensional MAS embedding on continuous
domains by a parametric separation guarantee. The resulting inductive bias is
particularly suited to exact and approximate set-containment problems, where
our experiments show improvements over standard set architectures such as
DeepSets and Set Transformers.

\newpage
\section{Bi-Lipschitz Message Passing and FSW-GNN}
\label{sec:methods-fsw}

The previous section focused on whether a neural architecture respects a
given symmetry. For graph neural networks, however, respecting permutation
symmetry is only part of the problem. Even an expressive MPNN may gradually
lose information as messages are repeatedly aggregated and propagated across the graph.

This issue becomes particularly apparent in \emph{long-range} graph learning
tasks, where solving the problem requires many message-passing iterations.

\subsection{Long-Range Information and Oversquashing}

Recall that after $t$ message-passing iterations, the feature of a node can
depend on information from nodes at distance at most $t$. Therefore, a task
whose relevant information is $r$ hops away requires at least $r$
message-passing iterations.

In principle, increasing the depth should allow information to travel farther.
In practice, however, deep MPNNs often become difficult to train and fail on long-range tasks.

One important explanation for the difficulty of long-range message passing is
\emph{oversquashing}. As the receptive field of a node grows with depth,
information from an increasingly large number of distant nodes must be
compressed into a fixed-dimensional representation and propagated through
relatively few intermediate nodes. Consequently, information originating far
away may have only a weak influence on the final representation, even when
the network is sufficiently deep for that information to reach the target
node.

\begin{figure}[t]
    \centering

    \begin{minipage}[c]{0.55\linewidth}
        \centering
        \includegraphics[width=\linewidth]
        {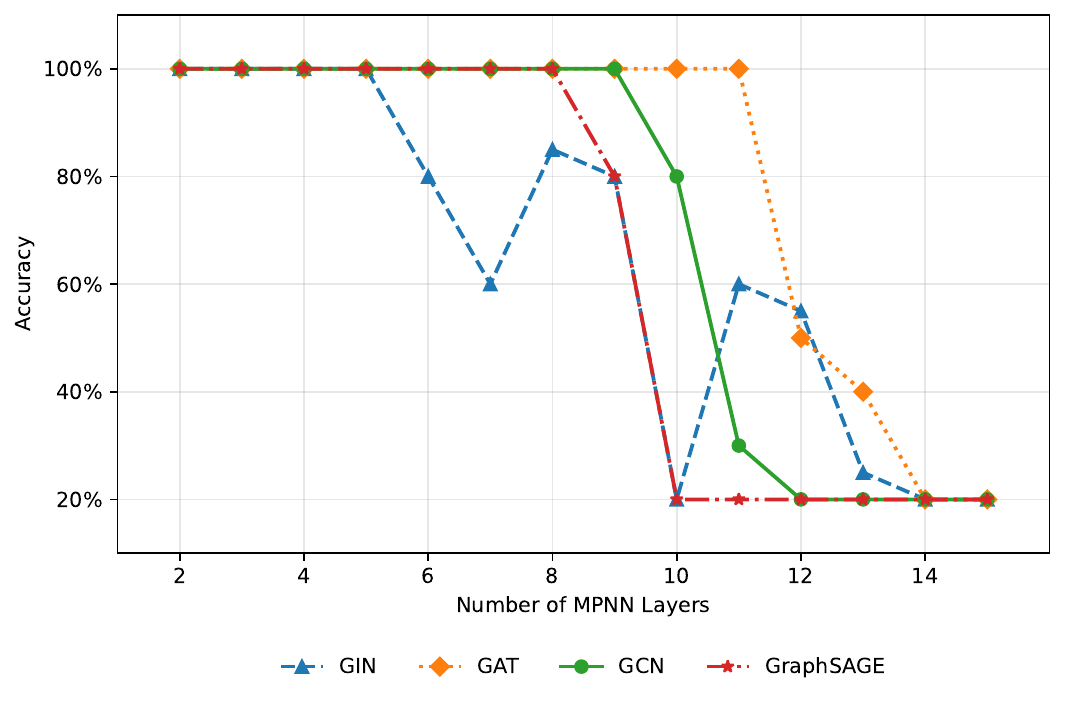}
    \end{minipage}
    \hfill
    \begin{minipage}[c]{0.40\linewidth}
        \centering
        \includegraphics[width=\linewidth]
        {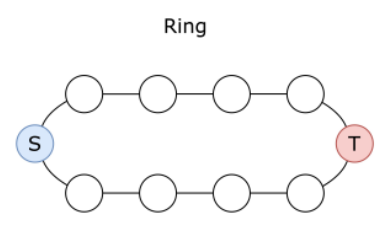} 
    \end{minipage}

    \caption{
        Performance of standard MPNNs on the Ring long-range task.
        As the required message-passing depth increases, the models
        gradually fail to propagate information across the graph.
        The Ring topology is illustrated on the right.
    }
    \label{fig:methods-standard-ring}
\end{figure}

A useful example is the \emph{Ring} benchmark. Each graph is a cycle with a designated source node and target node at graph distance \(r\), as illustrated in \cref{fig:methods-standard-ring}. The target is assigned one of five possible labels, and the task is to predict this label at the source node. Thus, solving the task requires information to travel over \(r\) message-passing steps. As \(r\) increases, this becomes increasingly difficult for standard MPNNs, as shown in \cref{fig:methods-standard-ring}. Since there are five possible labels, random guessing achieves \(20\%\) accuracy. The comparison is shown in \cref{fig:methods-standard-ring}.

This behavior is particularly interesting because it cannot be explained
solely through the usual binary notion of expressive power. Some standard
MPNNs, such as GIN, are theoretically as expressive as the 1-WL test.
Nevertheless, being able to separate two inputs in principle does not imply
that their learned representations remain numerically well separated.

This motivates a quantitative question:
\[
    \textit{How much of the distance between different inputs is preserved
    by the network?}
\]

\subsection{WL-Compatible Metrics on Graphs}
\label{subsec:methods-wl-metrics}

Binary expressivity alone does not quantify how much information is retained
by an embedding. To formulate a quantitative notion, we first need a distance
on the graph domain. Let $\mathcal G_{\leq N}(\Omega)$
denote vertex-featured graphs with at most $N$ vertices and features in
$\Omega\subseteq\mathbb R^d$.
The natural equivalence relation for an MPNN is not graph isomorphism but
1-WL equivalence: no MPNN can distinguish a pair of graphs that 1-WL fails
to distinguish \citep{xu2019powerful,morris2019weisfeiler}. We therefore use
metrics whose zero sets agree exactly with this limitation.

A \emph{WL metric} is a function
\[
    \rho:
    \mathcal G_{\leq N}(\Omega)
    \times
    \mathcal G_{\leq N}(\Omega)
    \longrightarrow \mathbb R_{\geq 0}
\]
that is symmetric, satisfies the triangle inequality, and obeys
\[
    \rho(G,H)=0
    \quad\Longleftrightarrow\quad
    G\equiv_{\mathrm{WL}}H.
\]
Strictly speaking, this is a pseudometric on the space of labeled graph
representations, since distinct graphs may have zero distance. Equivalently,
it is an ordinary metric on the quotient space obtained by identifying all
WL-equivalent graphs. The point is that a WL metric refines the binary WL
test: it still regards exactly the same pairs as indistinguishable, but it
also assigns a magnitude to the difference between distinguishable pairs.

We use two such metrics, which capture graph difference from complementary
viewpoints.

\subsubsection{The Doubly Stochastic Metric}

The Doubly Stochastic (DS) metric originates from a convex relaxation of
graph isomorphism \citep{grohe,boker2021graph}. For two featureless graphs
with the same number $n$ of vertices and adjacency matrices $A$ and
$\widetilde A$, isomorphism requires a permutation matrix $P$ satisfying
\[
    AP=P\widetilde A.
\]
Replacing the discrete set of permutation matrices by its convex hull, the
set $\mathcal D_n$ of doubly stochastic matrices, leads to the discrepancy
\[
    \min_{S\in\mathcal D_n}
    \|AS-S\widetilde A\|.
\]
A classical characterization of fractional graph isomorphism implies that
this quantity vanishes exactly when the two graphs are 1-WL equivalent
\citep{grohe2021logic}. Thus the relaxation is closely aligned with the
expressive power of message passing rather than with exact graph
isomorphism.

For the setting of Chapter~\ref{chap:paper4}, graphs may have different
numbers of vertices and continuous vertex features. We therefore use an
extended DS metric. Let
\[
    G=(A,X),\qquad
    \widetilde G=(\widetilde A,\widetilde X),
\]
with $n$ and $\widetilde n$ vertices, respectively, and define
$\Pi(n,\widetilde n)$ as the set of nonnegative
$n\times\widetilde n$ matrices whose row sums are $1/n$ and whose column
sums are $1/\widetilde n$. We set
\[
\begin{split}
    \rho_{\mathrm{DS}}(G,\widetilde G)
    ={}& |n-\widetilde n| \\
    &+ \min_{S\in\Pi(n,\widetilde n)}
       \left(
       \|AS-S\widetilde A\|_1
       +
       \sum_{i=1}^{n}\sum_{j=1}^{\widetilde n}
       S_{ij}\|x_i-\widetilde x_j\|_1
       \right).
\end{split}
\]
The first term penalizes different graph cardinalities, the matrix term
measures disagreement between the two adjacency structures under a soft
vertex correspondence, and the final term measures feature mismatch under
the same correspondence. We prove that this extension is again a WL metric.

\subsubsection{Tree Mover's Distance}

The second metric is the Tree Mover's Distance (TMD) of
\cite{TMD}. Its construction follows the computation-tree interpretation of
message passing. After $t$ rounds, the representation at a vertex depends on
the rooted depth-$t$ computation tree obtained by recursively unfolding its
neighborhood. TMD defines a distance between two such rooted trees by
comparing the root features and then optimally matching their child subtrees;
the matching cost is itself defined recursively. A graph-level distance is
obtained by an optimal-transport matching of the rooted computation trees of
all vertices.

This construction mirrors the WL recursion. In particular, the tree distance
between two rooted computation trees is zero precisely when WL assigns the
corresponding roots the same color after the relevant number of iterations.
Under the standard assumptions on the feature domain, TMD therefore induces
a WL metric at graph level \citep{TMD}. Whereas the DS metric is based on a
soft global correspondence between adjacency matrices, TMD measures graph
difference through the local computation trees actually seen by message
passing.

\subsection{Bi-Lipschitz Embeddings and Distortion}
\label{subsec:methods-bilipschitz}

Once a WL metric $\rho$ is fixed, we can ask whether a graph embedding
\[
    F:\mathcal G_{\leq N}(\Omega)\to\mathbb R^m
\]
preserves that geometry. We say that $F$ is \emph{bi-Lipschitz} with respect
to $\rho$ if there exist constants
\[
    0<c\leq C<\infty
\]
such that
\[
    c\,\rho(G,H)
    \leq
    \|F(G)-F(H)\|
    \leq
    C\,\rho(G,H)
\]
for every pair $G,H$ in the considered domain.

The upper inequality provides stability: nearby inputs cannot be amplified
arbitrarily. The lower inequality provides quantitative separation: inputs
that are far apart according to the WL geometry cannot collapse to nearly
the same representation. The ratio $\frac{C}{c}$
upper-bounds the metric distortion of the embedding.

This is strictly stronger than WL-equivalent separation. A WL-equivalent
embedding only guarantees
\[
    F(G)=F(H)
    \quad\Longleftrightarrow\quad
    G\equiv_{\mathrm{WL}}H;
\]
it places no lower bound on the distance between the embeddings of two
non-equivalent graphs. Consequently, a sequence of distinguishable graph
pairs can be mapped arbitrarily close together while the model remains fully
WL-expressive. Bi-Lipschitzness rules out precisely this form of numerical
collapse.

\subsection{The Aggregation Bottleneck}

A central source of oversquashing lies in the aggregation step itself.

At every message-passing iteration, a node compresses a multiset of
neighboring features,
\[
    \{\!\{
        h_u^{(t-1)}
        :
        u\in\mathcal{N}(v)
    \}\!\},
\]
into a single finite-dimensional vector.

Standard MPNNs commonly use simple aggregation operations such as mean,
maximum, or summation. Mean and maximum aggregation are not injective on
general multisets and therefore necessarily discard information. Sum
aggregation can be sufficiently expressive to obtain WL-equivalent MPNNs \cite{amir2023neural},
but it's provably not a bi-Lipschitz embedding of multiset space.

Consequently, even when an aggregation rule is expressive enough to
distinguish the required multisets theoretically, it may strongly distort
their distances. Repeating such aggregation over many layers provides a
possible mechanism by which information becomes increasingly compressed in
deep MPNNs.

This observation motivates replacing standard aggregation by a multiset
embedding that not only distinguishes multisets but also preserves their
geometry.

\subsection{FSW Aggregation}

Our construction is based on the
\emph{Fourier Sliced-Wasserstein (FSW) embedding}, a permutation-invariant
embedding of multisets into Euclidean space introduced by
\citet{amir2024injective}.

At a high level, FSW first projects the elements of a multiset onto several
one-dimensional directions. The projected values are sorted and interpreted
through their quantile functions, from which Fourier-type features are
extracted. The resulting finite-dimensional representation is independent of
the ordering of the input elements and naturally accommodates multisets of
different cardinalities.

Most importantly for our purpose, the FSW embedding is bi-Lipschitz with
respect to a natural Wasserstein metric on multisets.

We therefore use FSW as the neighborhood aggregation mechanism of an MPNN.
At iteration $t$, we compute
\[
    q_v^{(t)}
    =
    \operatorname{FSW}_t
    \left(
        \{\!\{
            h_u^{(t-1)}
            :
            u\in\mathcal{N}(v)
        \}\!\}
    \right),
\]
and update the node representation by first concatenating the previous
node state with the aggregated message, and then projecting the result
back to the hidden dimension:
\[
    h_v^{(t)}
    =
    \Phi_t
    \left(
        \left[
        h_v^{(t-1)} \,\|\, q_v^{(t)}
        \right]
    \right),
\]
where \(\|\) denotes concatenation and \(\Phi_t\) is a learned projection
back to the hidden dimension.

A final FSW aggregation over all node representations is used for graph-level
prediction. We refer to the resulting architecture as
\emph{FSW-GNN}.

\subsection{A Bi-Lipschitz WL-Equivalent MPNN}
\label{subsec:methods-fsw-theory}

We show that FSW-GNN can attain the full distinguishing power available to standard message-passing networks.
For graphs with at most $N$ vertices and $d$-dimensional input features, an
FSW-GNN with $T=N$ iterations and sufficiently large hidden dimension is, for
almost every choice of its parameters, equivalent in expressive power to
1-WL. The proof uses the Finite Witness Theorem for multiset embeddings
\citep{amir2023neural}.

The stronger result concerns metric preservation. Let $\Omega\subset
\mathbb R^d$ be compact. Under the same assumptions that guarantee
WL-equivalence, FSW-GNN is bi-Lipschitz with respect to
$\rho_{\mathrm{DS}}$ on $\mathcal G_{\leq N}(\Omega)$. If, additionally,
$0\notin\Omega$, it is also bi-Lipschitz with respect to TMD. 

\begin{figure}[t]
    \centering

    \begin{minipage}[t]{0.48\textwidth}
        \centering
        \includegraphics[width=\linewidth]{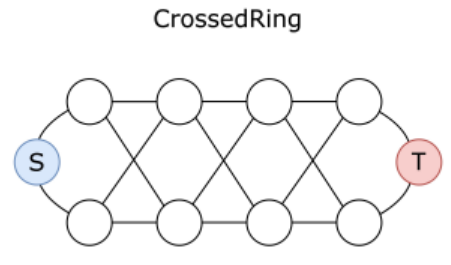}
        \small (a) First topology
    \end{minipage}
    \hfill
    \begin{minipage}[t]{0.48\textwidth}
        \centering
        \includegraphics[width=\linewidth]{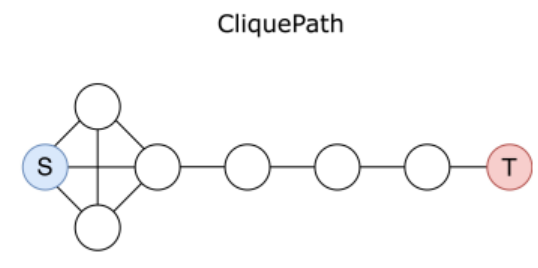}
        \small (b) Second topology
    \end{minipage}

    \caption{Additional graph-transfer topologies used in the long-range experiments.}
    \label{fig:methods-transfer-topologies}
\end{figure}

\subsection{Long-Range Behavior}

The motivation for bi-Lipschitz message passing is reflected particularly
clearly in long-range experiments.

On the Ring task, we vary the problem radius from

$$
    2\leq r\leq 15.
$$

\begin{figure}[t]
\centering
\includegraphics[width=0.75\linewidth]{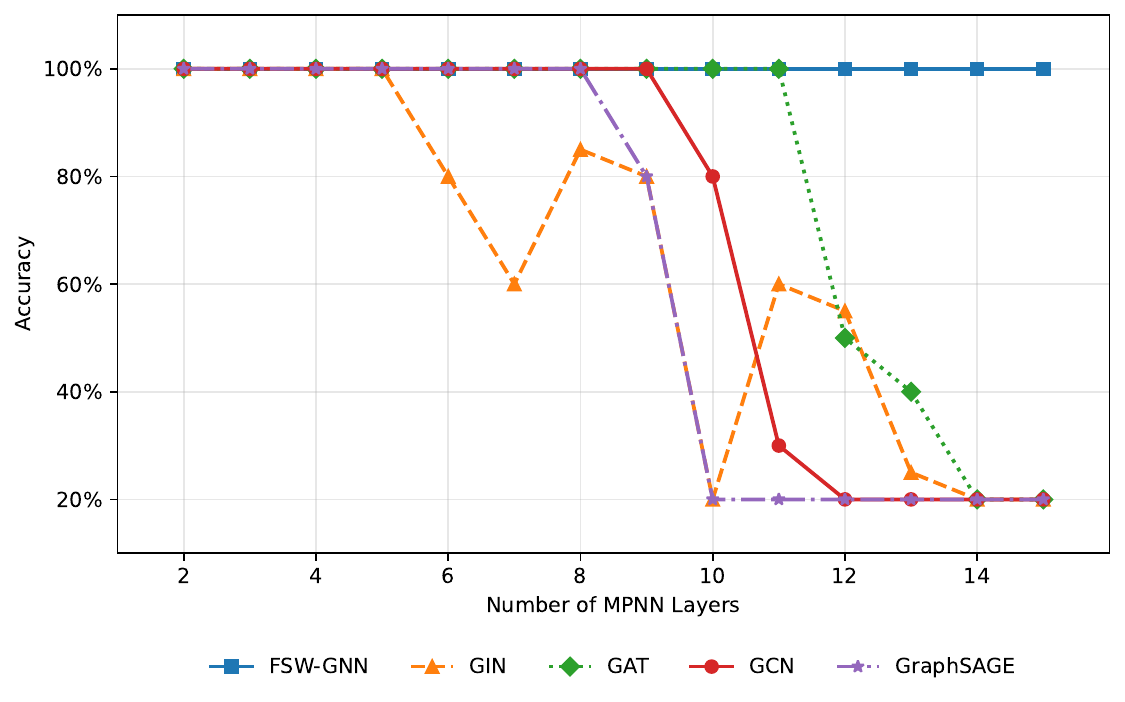}
\caption{Performance on the Ring long-range task, including FSW-GNN.}
\label{fig:methods-fsw-ring}
\end{figure}

FSW-GNN achieves perfect accuracy throughout the entire tested range, as shown
in \cref{fig:methods-fsw-ring}.

We observe the same qualitative behavior on additional graph-transfer tasks.
The different graph topologies considered in these experiments---CliquePath,
Ring, and CrossRing---are illustrated in
\cref{fig:methods-transfer-topologies}. In each case, information must be
propagated from a source node to a target node located \(r\) steps away.

\begin{figure}[t]
    \centering

    \begin{minipage}[t]{0.48\textwidth}
        \centering
        \includegraphics[width=\linewidth]{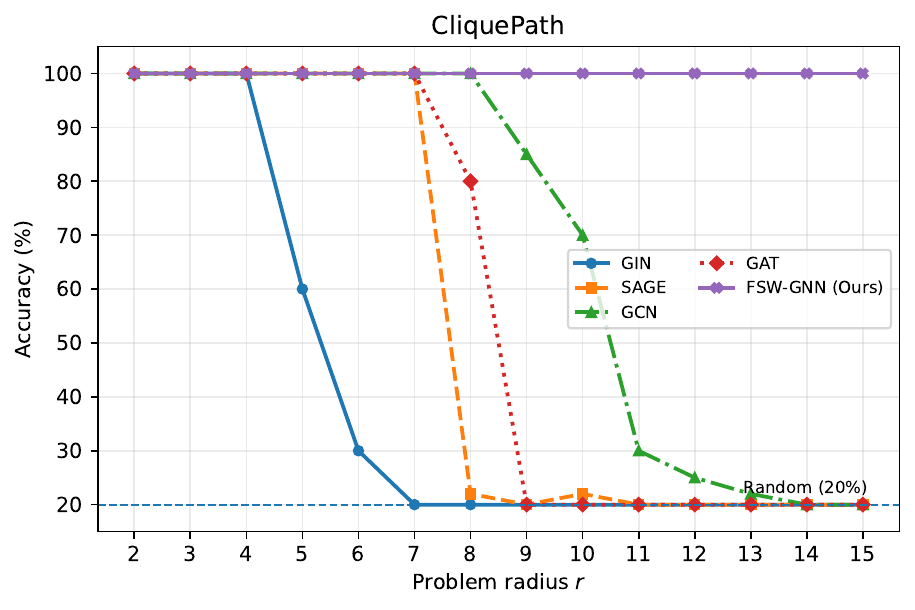}
        {\small (a) CliquePath\par}
    \end{minipage}
    \hfill
    \begin{minipage}[t]{0.48\textwidth}
        \centering
        \includegraphics[width=\linewidth]{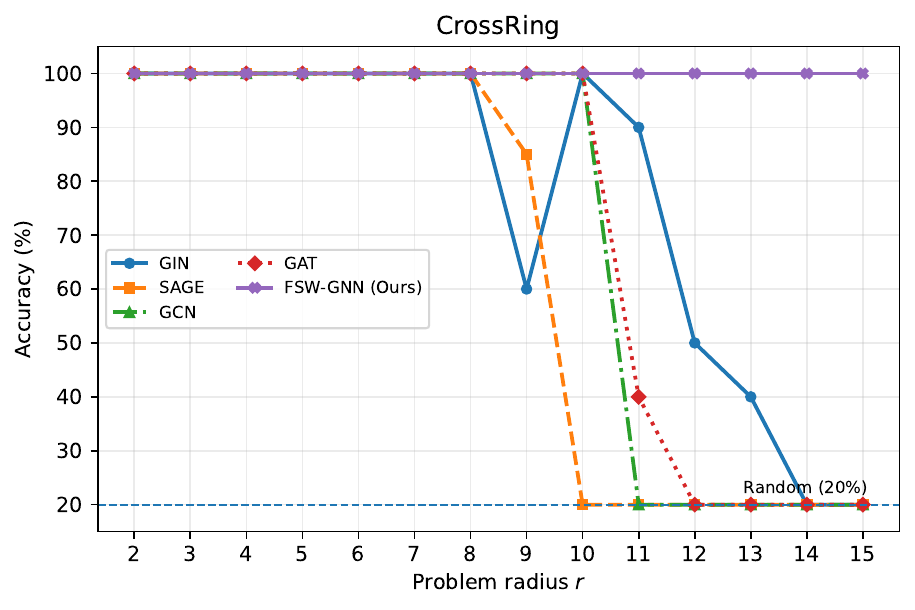}
        {\small (b) CrossRing\par}
    \end{minipage}

    \caption{Performance on the CliquePath and CrossRing long-range graph-transfer tasks. FSW-GNN maintains $100\%$ accuracy for all tested radii up to $r=15$.}
    \label{fig:methods-transfer-results}
\end{figure}

On all three benchmarks, FSW-GNN is the only tested method to maintain
\(100\%\) accuracy for all tested radii up to \(r=15\), as shown in
\cref{fig:methods-transfer-results}. 

The competing MPNNs begin to fail
considerably earlier, with the precise failure point depending on the graph
topology.

FSW-GNN is not completely immune to the difficulties of very deep message
passing. Its performance eventually deteriorates when the required radius
becomes sufficiently large; in our experiments this occurs only at
substantially larger radii, beyond approximately \(r=20\).

Thus, our claim is not that bi-Lipschitz aggregation eliminates
oversquashing altogether. Rather, both the theoretical guarantees and the
experiments indicate that FSW-GNN preserves useful information through many
more message-passing iterations, and consequently fails substantially later
than standard MPNNs on long-range problems.

\chapter{On the Expressive Power of Sparse Geometric MPNNs}
\label{chap:paper1}

\noindent
\textbf{Yonatan Sverdlov and Nadav Dym.}\\
International Conference on Learning Representations (ICLR), 2025.

\bigskip


\begin{refsection}[papers/paper1/references.bib]

\begingroup

\graphicspath{{papers/paper1/}}

%

\def\lms{ \{\!\!\{ }
\def\rms{ \}\!\!\} }

\def\sglobal{s^{\mathrm{global}}}
\def\sglobalhat{\hat{s}^{\mathrm{global}}}

\def\gennet{\textsc{EGenNet}}
\def\us{\gennet}

\def\RR{\mathbb{R}}

\def\G{\mathcal{G}}
\def\H{\mathcal{H}}
\def\M{\mathbb{M}}
\def\N{\mathcal{N}}
\def\O{\mathcal{O}}
\def\W{\mathbb{W}}

\def\bz{\boldsymbol{z}}
\def\btheta{\boldsymbol{\theta}}
\def\bth{\btheta}

\def\vs{{\bm{s}}}
\def\vt{{\bm{t}}}
\def\vv{{\bm{v}}}
\def\vx{{\bm{x}}}

\def\mA{{\bm{A}}}
\def\mE{{\bm{E}}}
\def\mF{{\bm{F}}}
\def\mH{{\bm{H}}}
\def\mP{{\bm{P}}}
\def\mQ{{\bm{Q}}}
\def\mV{{\bm{V}}}
\def\mX{{\bm{X}}}

\def\gG{{\mathcal{G}}}
\def\gN{{\mathcal{N}}}

\def\Gd{\mathbb{G}(d)}
\def\GdN{\mathbb{G}_{N}(d)}
\def\GdNdistinct{\mathbb{G}_{N}^{\mathrm{distinct}}(d)}
\def\GdNgeneric{\mathbb{G}_{\mathrm{generic}}(d,N)}

\def\vvv{{\vv}}
\def\finv{f_{\mathrm{inv}}}

\def\gray#1{\textcolor{gray}{#1}}

\def\eqref#1{equation~\ref{#1}}

\let\citep\parencite
\let\citet\textcite

\begin{abstract}
Motivated by applications in chemistry and other sciences, we study the expressive power of message-passing neural networks for geometric graphs, whose node features correspond to 3-dimensional positions. Recent work has shown that such models can separate \emph{generic} pairs of non-isomorphic geometric graphs, though they may fail to separate some rare and complicated instances. However, these results assume a fully connected graph, where each node possesses complete knowledge of all other nodes. In contrast, often, in applications, every node only possesses knowledge of a small number of nearest neighbors.

This paper shows that generic pairs of non-isomorphic geometric graphs can be separated by message-passing networks with rotation equivariant features as long as the underlying graph is connected. When only invariant intermediate features are allowed, generic separation is guaranteed for generically globally rigid graphs. We introduce a simple architecture, $\us$, which achieves our theoretical guarantees and compares favorably with alternative architectures on synthetic and chemical benchmarks. Our code is available at
\href{https://github.com/yonatansverdlov/E-GenNet}{GitHub}.
\end{abstract}

\section{Introduction}
Geometric graphs are graphs whose nodes are a 'position' vector in $\RR^d$ (typically $d=3$) and whose symmetries include the permutation symmetries of combinatorial graphs and translation and rotation of the node positions. Geometric graphs arise naturally in learning applications for chemistry, physical dynamics, and computer vision as natural models for molecules, particle systems, and 3D point clouds. These applications motivated the introduction of many learning models for geometric graphs, which were often inspired by 'standard' \emph{graph neural networks} (GNNs) for combinatorial graphs \citep{egnn,han2024survey,gemnet}. Subsequently, several theoretical works aimed primarily at understanding the expressive power and limitations of GNNs for geometric graphs. 

GNNs for geometric graphs produce global graph features invariant to geometric graph symmetries. The expressive power of GNNs is typically assessed primarily by their ability to assign different global features to pairs of geometric graphs that are not geometrically isomorphic. 

Recent research on this problem has uncovered several interesting results, mainly assuming that the graphs are full (an edge connects each pair of nodes). Under the full graph assumptions, several models for geometric GNNs are \emph{complete}, that is, capable of separating \emph{all} pairs of non-isomorphic graphs.
However, this typically comes at a relatively high computational price. Examples of complete models include $(d-1)$-WL-based GNNs \citep{three_iterations,gramnet,disgnn}, GNNs based on sub-graph aggregation \citep{Invariant_models}, and message passing models based on arbitrarily high-dimensional irreducible representations \citep{TFN,gemnet,dym2020universality,finkelshtein2022simple}. 
This paper focuses on more efficient geometric GNNs: message-passing neural networks (MPNN) based on simple invariant and equivariant features. In \citep{gcnn,disgnn}, it was shown that, even under the full graph assumption, invariant MPNN-based networks are not complete. In contrast, \citet{Invariant_models} and \citet{gramnet} proved that these models are 'generically complete,' which means that, under the full graph assumption, they are capable of separating most geometric graphs and only fail on a Lebesgue measure zero set of corner cases, which are difficult to separate. 

In practical applications, the full graph assumption is often not fulfilled: to improve computational efficiency, geometric GNNs often operate on sparse graphs, where each node is only connected to a few nearest neighbors.
Accordingly, the first question we address in this paper is:

\textbf{Question 1.} For which graphs, other than the full graphs, are geometric message-passing neural networks generically complete?

Our answer to this question is inspired by the results of \cite{on_the_expressive_power}. They divided Message-passing-based networks for geometric graphs into \emph{Equivariant Geometric Graph Neural Networks} (E-GGNNs), which use rotation \emph{equivariant} hidden features, and \emph{Invariant Geometric Graph Neural Networks} (I-GGNNs), which use rotation \emph{invariant} hidden features, and showed that E-GGNNs have substantial advantages over I-GGNNs in terms of their expressive power. 
    
Our answer to Question 1 is also strongly related to this dichotomy. We prove in \autoref{thm:ggnn} that E-GGNNs are generically complete for a given graph if and only if the graph is connected. This is visualized in Figure \ref{fig:teaser}. E-GGNNs will not be able to differentiate between the pair of geometric graphs in Figure \ref{fig:teaser}(A) as the underlying graph is disconnected but will be able to differentiate between the pairs in (B) and (C). 
\begin{figure}[t]
    \centering
     \includegraphics[width=13.0cm]{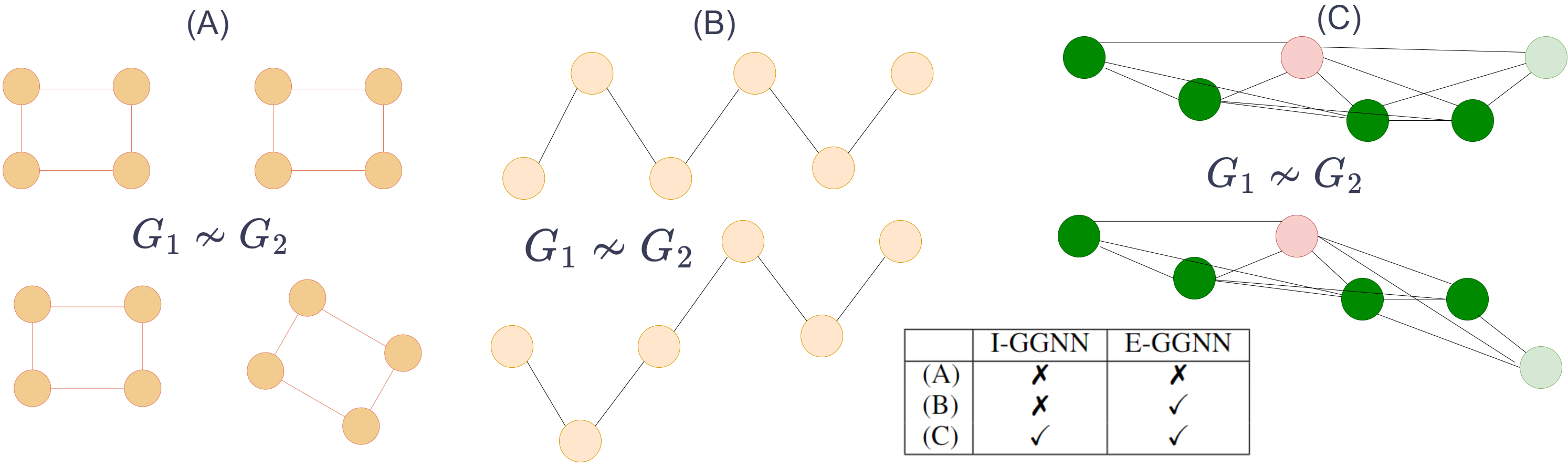}
    \caption{(A) and (B) are pairs of non-isomorphic geometric graphs with the same distances along edges. I-GGNN cannot distinguish such pairs, while E-GGNN can (generically) if the graph is connected, as in (B). Subplot (C) depicts an example of globally rigid graphs, where non-isomorphic geometric graphs do not share the same distances across edges (in the figure, the edge between pink and light green). Such examples can be separated by both I-GGNN and E-GGNN.}
    \centering
    \label{fig:teaser}
\end{figure}
In contrast, I-GGNNs will not be able to differentiate between the non-isomorphic pairs in Figure \ref{fig:teaser}(A)-(B), as the distances across all graph edges are the same. On the other hand, in (C), the distances along graph edges determine the positions of all nodes up to rotation and translation. In rigidity theory, such graphs are called 'generically globally rigid.' In \autoref{thm:iggnn}, we prove that I-GGNNs are generically complete if and only if the underlying graph is generically globally rigid. These two theorems give a complete answer to Question 1. 

To guarantee generic separation, our theorems make the standard assumption \citep{on_the_expressive_power} that the I-GGNN and E-GGNN we consider are maximally expressive, meaning that all aggregation and readout functions used throughout the networks are injective. Recent work has shown that relatively standard and simple I-GGNN can fulfill this assumption \citep{Neural_Injective}. However, for E-GGNN, it is unclear how this assumption can be fulfilled. Accordingly, our second question is 

\textbf{Question 2.} Can we devise a simple E-GGNN architecture that is maximally expressive hence, generically complete on connected graphs?

We answer this question affirmatively by devising a simple architecture, which we name $\us$, and proving in \autoref{thm:maximally} that it is maximally expressive (for generic graphs). This architecture is based on a simple message-passing architecture that resembles EGNN \citep{egnn} but uses multiple equivariant channels as in \citep{levy2023using}.

We experimentally find that $\us$ is highly successful in synthetic separation tasks and on several learning tasks involving medium-sized molecules, with a performance comparable to, or better than, state-of-the-art methods. These results indicate that simple, generically complete methods like $\us$  are sufficient for the tasks we considered. However, we acknowledge that there may be other tasks, perhaps involving molecules with non-trivial symmetry groups, where more complex geometric graph neural networks that offer full separation could be beneficial. 
\paragraph{Contributions} To summarize, our main results are 
\begin{itemize}
    \item Showing generic separation of E-GGNNs depends on graph connectivity.
    \item Showing generic separation of I-GGNNs depends on graph rigidity.
    \item Proposing a simple E-GGNN architecture that is provably (generically) maximally expressive and performs well in practical tasks.
\end{itemize}

\paragraph{Appendix} See the appendix for all full proofs and related work.

\section{Setup: Geometric Graph Neural Networks and Rigidity Theory}
\textbf{Geometric Graphs. }
In our discussion of geometric graphs and GNNs, we loosely follow the definitions of \cite{on_the_expressive_power}. We define a geometric graph as a pair $\G = (\mA, \mX)$. The matrix $\mA \in \{0,1 \}^{n \times n}$ is an adjacency matrix, which fully defines a combinatorial graph. The neighborhood of a node $i$ is denoted by $\N_{i}$. Accordingly, we sometimes use statements like 'the graph $\mA$.' We assume for simplicity that the graph nodes are the integers $1,\ldots,n$. The matrix $\mX=(\vx_1,\ldots,\vx_n)\in \RR^{n \times d}$ denotes node positions.
 We denote the set of all geometric graphs with coordinates in $\mathbb{R}^{d}$ by $\Gd$, and the subset of graphs with $\leq N$ nodes by $\GdN$. In Appendix \ref{app:features}, we discuss incorporating additional node and edge features into our model and theoretical results.

The symmetry group of geometric graphs is the product of the permutation group $S_n$ and the group of rigid motions $\mathcal{E}(d)$. The action of a permutation matrix $\mP$ on a geometric graph is given by $\mP \gG := (\mP\mA\mP^\top,  \mP\mX)$. A rigid motion in $\mathcal{E}(d)$ is a rotation $\mQ \in \O(d)$ and a translation vector $\vt \in \RR^d$. They act on $\mX$ by applying the rotation and translation to all its coordinates, $\vx_i\mapsto \mQ \vx_i+\vt$. 

Two geometric graphs $\G$ and $\H$ are \emph{geometrically isomorphic} if there exists a transformation in $S_n \times \mathcal{E}(d) $  which can be applied to $\mathcal{G}$ to obtain $\mathcal{H}$. We will say that a function $F$ defined on $\Gd$ is \emph{invariant} if $F(\G)=F(\H) $ for all $\G$ and $\H$ which are geometrically isomorphic.

The expressive power of an invariant model $F$ is closely related to its ability to separate pairs $\G,\H$ that are not geometrically isomorphic: 
clearly, if $F$ cannot separate $\G$ from $\H$, then 
 $F$ will not be a good approximation for functions $f$ for which $f(\G)\neq f(\H)$. Conversely, if $F$ can separate all pairs of geometric graphs, then it can be used to approximate all continuous invariant \citep{gortler_and_dym,gramnet} and equivariant \citep{villar2021scalars,hordan2} functions. In Appendix \ref{sec:approx}, we discuss the connection between approximation and separation in the context of our results.
 
Constructing an \( F \) which can separate all geometric graphs is at least as challenging as distinguishing all combinatorial graphs (by setting all node features to zero). Without any assumptions on the point cloud, achieving this is infeasible. To address this, we employ the concept of generic geometric graphs which originates from rigidity theory.
\paragraph{Generic separation and Rigidity Theory}
As discussed in the introduction, our paper focuses on generic separation. The results we present in this section are primarily influenced by rigidity theory, so we also use this field's common notion of 'generic'.
We will say that $\mX\in \RR^{d\times n}$ is \emph{generic} if it is not the root of any non-zero multivariate polynomial $p:\RR^{d\times n} \to \RR$ with rational coefficients. The genericity assumption guarantees that $\mX$ avoids many degeneracies, making separation easier. For example, for generic $\mX$ all pairs of points have different norms, and every $d$ tuple of points $\vx_{i_1},\ldots,\vx_{i_d}$ is full rank because $\|\vx_i\|^2-\|\vx_j\|^2 $ and $\det[\vx_{i_1},\ldots,\vx_{i_d}] $ are polynomials with natural coefficients. 
We note that Lebesgue almost all point clouds in $\RR^{d\times n}$  are generic \citep{mityagin2020zero}. 

A globally rigid graph is a central topic in rigidity theory:
\begin{definition}[Globally rigid]( \cite{reconstructible}) A geometric graph $\mathcal{G} = (\mA,\mX)$ is \emph{globally rigid} in $\RR^d$, if the only $\mX' \in \mathbb{R}^{n \times d}$ which satisfy that $\| \vx_i-\vx_j\|=\|\vx_i'-\vx_j'\|$ for all edges $(i,j)$ in $\mA$, are those that are related to $\mX$ by a rigid motion.\\
A combinatorial graph $\mA$ is \emph{generically globally rigid} in $\RR^d$, if the geometric graph $\mathcal{G} = (\mA,\mX)$ is \emph{globally rigid} in $\RR^d$ for every generic $\mX\in \mathbb{R}^{n \times d}$.
\end{definition}

The geometric graphs in \autoref{fig:teaser} (A)-(B) are not globally rigid, as they have the same distances along edges but are not related by a rigid motion. The underlying combinatorial graph, disconnected in (A) or line graph in (B), is not generically globally rigid. 
In contrast, the underlying combinatorial graph in (C) is generically globally rigid. Another simple example of a generically globally rigid graph is the full graph. Additional non-trivial examples can be found in \citep{power-graph,power-graph2}.
A necessary condition for generic global rigidity of a graph with $n \geq d + 2$ nodes in $\mathbb{R}^{d}$ is to be $d+1$ connected \footnotemark, which explains why the graphs in (A)-(B) are not generically globally rigid.
\footnotetext{We say a graph $G$ is $k$ connected if the graph remains connected after removing any $k-1$ vertices.}

Note that global rigidity definitions focus on reconstructing $\mX$ when $\mA$ is known. In the study of GGNNs, we are interested in identifying both $\mX$ and $\mA$:

\begin{definition}
Let $d$ be a natural number. Following \cite{Invariant_models}, we will say that an invariant function $F$ defined on $\Gd$, \emph{identifies} a geometric graph $\G$, if for every $\hat \G \in \Gd$, we have that $F(\G)=F(\hat{\G})$ if and only if $\G$ and $\hat \G$ are geometrically isomorphic.\\
We say that $F$ \emph{generically identifies} a \emph{combinatorial graph} $\mA$ in $\mathbb{R}^{d}$, if $F$ can identify $\G=(\mA,\mX)$ for every generic $\mX\in \mathbb{R}^{n \times d}$. \\
We say that $F$ \emph{generically fails to identify} a \emph{combinatorial graph} $\mA$ in $\mathbb{R}^{d}$, if $F$ does not identify $\G=(\mA,\mX)$ for every generic $\mX \in \mathbb{R}^{n \times d}$.
\end{definition}
Note that the fact that $F$ does not generically identify $\mA$ only guarantees that there exists a generic $\mX$ so that $F$ fails to identify $(\mA,\mX)$. The statement ‘$F$ generically fails to identify' is, a priori, significantly stronger. Also note that invariant $F$ separates all non-isomorphic graphs (is \emph{complete}) if and only if it identifies all graphs.
As discussed in the introduction, our goal will be to classify which graphs are generically identifiable by standard message-passing-based networks for geometric graphs, which can be classified as either I-GGNN or E-GGNN. We will now define these concepts.

\paragraph{E-GGNN}
In the context of combinatorial graphs (with no geometric information), \citet{gilmer2017neural,jogl2023expressivitypreserving} and  \citet{velivckovic2022message} showed that many famous graph neural networks are instantiations of MPNNs. The goal of \cite{on_the_expressive_power} was to show, analogously, that many neural networks for \emph{geometric} graphs follow a generalized MPNN framework. Loosely following \cite{on_the_expressive_power}, we define \emph{Equivariant Geometric Graph Neural Networks (E-GGNN)} to be functions defined by a sequence of layers that propagates vector features from iteration $t$ to $t+1$ via a learnable function $f_{(t)}$ (at initialization we set $v_j^{(0)}=0$).
\begin{equation}
\label{eq:GGNN} 
\vv_i^{(t+1)}=f_{(t)} \left( \lms  \vv_j^{(t)}, \vx_i-\vx_j , \mid j \in \mathcal{N}_i \rms \right).
\end{equation}
To ensure the equivariance of the construction, the function $f_{(t)}$ is required to be rotation equivariant. Translation invariance is implicitly guaranteed by the translation invariance of $\vx_{i}-\vx_j$, and the requirement that $f_{(t)}$ is defined on multi-sets implicitly enforces permutation equivariance. We will often refer to $f_{(t)}$ as an \emph{aggregation function}.

We note that for the sake of simplicity, we don't require a \emph{combine} step that incorporates $\vv_i^{(t)}$ in the process of computing $\vv_i^{(t+1)} $ since we find that this step has little influence on our theoretical and empirical results.
We also note that the vector $\vv_i^{(t)}$ is not necessarily $d$ dimensional, and the action of $O(d)$ on $\vv_i^{(t)}$ is allowed to change from layer to layer. For example, in the $\us$ architecture we will introduce later on, each  $\vv_i^{(t)}$ will be a $d\times m$ matrix, and the action of $O(d)$ will be multiplication on each of the $m$ coordinates.

After $T$ iterations of E-GGNNs, a permutation, rotation, and translation \emph{invariant} feature vector can be obtained via a two-step process involving a rotation-invariant function $\finv$ and a multi-set \textsc{ReadOut} function:
\begin{equation} 
    \vs_i =\finv(\vvv_i^{(T)}), \quad  
    \sglobal = \textsc{ReadOut}\lms\vs_{1},..,\vs_{n}\rms
    \label{eq:readout}
\end{equation}

As shown by \cite{on_the_expressive_power}, many popular geometric graph neural networks can be seen as instances of E-GGNNs with a specific choice of functions $f_{(t)}$ and readout functions. Examples include EGNN \citep{egnn}, TFN \citep{TFN}, GVP \citep{gvp-egnn}, and MACE \citep{mace}.
In Section \ref{architecture}, we present $\us$, our instantiation of this framework. 

\paragraph{I-GGNNs}
An important subset of E-GGNNs are \emph{invariant} GGNNs (I-GGNNs). These networks only maintain scalar features $\vs_i^{(t)}$ and replace $\vx_i-\vx_j$ with its norm to obtain
\begin{equation}\label{eq:IGGNN} \vs_i^{(t+1)}=f_{(t)} \left( \lms  \vs_j^{(t)}, \|\vx_i-\vx_j\| , \mid j \in \mathcal{N}_i \rms \right)
 \end{equation}
A global invariant feature is obtained via a permutation invariant readout function as in \eqref{eq:readout}. 
An advantage of I-GGNNs is that there are no rotation equivariance constraints on $f_{(t)}$. Therefore, any standard message-passing neural network that can process continuous edge weights can be employed as an I-GGNN by setting the edge weights to $\|{\vx}_{ij}\|$. Examples of I-GGNNs in the literature include a variant of EGNN used for invariant tasks \citep{egnn}, SchNet \citep{SchNet}, DimeNet \citep{DimeNet} and SphereNet \citep{SphereNet}\footnotemark.
Our definition of I-GGNN follows the definition from \citep{gcnn,disgnn}. 
In Appendix \ref{I-GGNN types}, we explain that the definition of I-GGNN \cite{on_the_expressive_power} has a seemingly minor difference that renders it significantly stronger.
\footnotetext{\label{invariant models} DimeNet \citep{DimeNet} and SphereNet \citep{SphereNet} have access to the power graph $\G^2$ (defined later in the paper) and thus are invariant models on $\G^{2}$ and have potential expressive power stronger than invariant I-GGNN on $\G$.}

\paragraph{Maximally expressive E-GGNN and I-GGNNs}
The separation power of E-GGNN and I-GGNN architectures depends on the specific instance used. For example, if we choose the final readout function as the zero function, all geometric graphs will be assigned the same value. To address this, it is customary to assume that all functions $f_{(t)}$, as well as $\finv$ and the readout function, are injective, up to the ambiguities these functions have by construction. Following \cite{on_the_expressive_power}, we will call E-GGNN (respectively I-GGNN) architectures, which are composed of injective functions, maximally expressive E-GGNN (respectively I-GGNN) architectures.
We note that the existence of the injective functions necessary for maximal expressivity follows from set-theoretic considerations, similar to those used by \cite{wang2024rethinking}. The question of whether practical differentiable aggregation functions are injective is discussed in Section \ref{sec:maximally}.

\section{Expressive power of I-GGNN}
\label{sec:inv}
Our next goal is to analyze the separation abilities of I-GGNN.
\begin{restatable}{theorem}{IGGNN}
    [expressive power of I-GGNN]
\label{thm:iggnn}
Let $d$ be a natural number. Let $F$ be an I-GGNN.  Let $\mA$ be a graph not generically globally rigid on $\RR^d$. Then, $F$ generically fails to identify $\mA$. \\ Conversely, if $\mA$ is generically globally rigid on $\RR^d$ and $F$ is a maximally expressive I-GGNN with depth $T=1$, then $F$ generically identifies $\mA$. 
\end{restatable}

\begin{proof}[Proof idea]
if $\mA$ is not globally rigid, then according to \cite{gortler2010characterizing}, for \emph{every} generic $\mX$ there exists $\mX'$ for which distances along edges are preserved, but $\mX$, $\mX'$ are not related by a rigid motion.
It's easy to see any I-GGNN, which is based only on distances along edges, can't separate such graphs.

On the other hand,  generic geometric graphs
 have distinct pairwise distances. The full proof (see Appendix \ref{Proof of I-GGNN}) shows that this can be used to reconstruct the combinatorial graph $\mA$. By global rigidity, we can then reconstruct $\mX$, up to rigid motion.
\end{proof}
Note that \autoref{thm:iggnn} is a generalization of \cite{gramnet} that showed maximally expressive I-GGNN generically identify full graphs. We now showed this is true for all generically globally rigid graphs.
\paragraph{I-GGNN generic separation by power graph preprocessing} 
A natural strategy to overcome the lack of generic separation for graphs that are not generically globally rigid, as proven in Theorem \ref{thm:iggnn}, is to replace these graphs with graphs that \emph{are} generically globally rigid. This replacement can be done using power graphs:
For a given graph $\mathcal{G}$ and natural number $k$, the power graph $\mathcal{G}^{k}$ is defined to have the same nodes as the original graph $\mathcal{G}$. A pair of nodes are connected by an edge in the power graph if and only if there is a path between them in the original graph of length $\leq k$. 

In \citep{power-graph, power-graph2}, it was shown that if the original graph $\G$ is connected, then the power graph $\G^{d+1}$ is generically globally rigid in $\mathbb{R}^{d}$. Accordingly, when the input graph $\G$ is connected but not generically globally rigid, our theory suggests that applying I-GGNN not to the original graph but to its $d+1$ power may benefit separation. An experiment illustrating this idea is presented in \autoref{tab:power_graph}.

\section{Expressive power of E-GGNN}
When using E-GGNN rather than I-GGNN,  we will have generic separation if and only if the graph is connected. 
\begin{restatable}{theorem}{GGNN}
[expressive power of E-GGNN]\label{thm:ggnn}
Let $d$ be a natural number. Let $F$ be an E-GGNN.  Let $\mA$ be a disconnected graph. Then, $F$ generically fails to identify $\mA$.\\ Conversely, if $\mA$ is connected and $F$ is a maximally expressive E-GGNN with depth $T\geq d+1$, then $F$ generically identifies $\mA$.
\end{restatable}

\begin{proof}[Proof idea]
Suppose a geometric graph $\mathcal{G}$ isn't connected. We can apply two distinct rigid motions to two connected components and obtain a new geometric graph, $\mathcal{G}'$, which cannot be separated from $\mathcal{G}$ using E-GGNN. This is illustrated in \autoref{fig:teaser}(A).
Conversely, in the appendix \autoref{Identifiable}, we show that the features $s_i$ obtained after $d+1$ iterations of a maximally expressive E-GGNN can encode the multi-set of all distances $\|x_i-x_j\| $ for all $j$ which are $d+1$-hop neighbors of $i$ (a similar claim was shown by \cite{on_the_expressive_power}). Thus, $d+1$ iterations of maximally expressive E-GGNN can simulate the application of a single iteration of an I-GGNN on the $d+1$ power graph. The proof then follows from the fact that the $(d+1)$ power graph is generically globally rigid \citep{power-graph, power-graph2}. 
\end{proof}

An immediate result of this theorem is that generic graph separation is a local problem with a problem radius of at most $d+1$  \citep{alon2020bottleneck} .\footnote{A task's problem radius denoted by $r$ is the minimal integer where the task becomes solvable upon each node observing its neighborhood within a $r$-hop radius and aggregating all node's information.} We note that the generic assumption allows us to reconstruct the coordinates of the nodes and the combinatorial graph. Without the generic assumption, even models based on $2$-WL that are complete on full graphs cannot reconstruct the combinatorial graph. Setting all positions to zero nullifies the geometric information, and $2$-WL is insufficient to distinguish all combinatorial graphs.

\section{Building a generically maximally expressive E-GGNN}\label{sec:maximally}
To obtain generic separation guarantees as in Theorems \ref{thm:iggnn}, \ref{thm:ggnn} for a practical instantiation of E-GGNN (or I-GGNN), we will need to  
 show that the instantiation is fully expressive. This type of question was studied rather extensively for combinatorial graphs \citep{Neural_Injective,aamand2022exponentially}, and these results can be used to show that some simple realizations of I-GGNN are maximally expressive. We discuss this in more detail in Appendix \ref{building i-ggnn}. 
 
For E-GGNN, the requirement that $f_{(t)}$ is simultaneously injective (up to permutation) and equivariant to rotations seems to be challenging to attain with a simple E-GGNN construction, and indeed, \cite{on_the_expressive_power} who suggested the notions of maximally expressive E-GGNN, did not give any indication of how they could be constructed in practice. Moreover, in Appendix \ref{maximally_expressive_egnn}, we show that a maximally expressive E-GGNN can separate all pairs of full graphs and separate geometrically all pairs of connected graphs\footnote{We define the geometric separation of a geometric graph if we can separate the point clouds up to group action, without necessarily reconstructing the combinatorial graph.}, indicating that constructing such a network may be computationally expensive, as currently, such guarantees are only achieved by networks of relatively high complexity based on $2$-WL subgraph aggregation techniques. 

In our setting, however, the problem is less severe since, from the outset, we are only interested in analyzing network behavior on generic graphs. Accordingly, we would like to require our E-GGNN to be maximally expressive only for generic graphs. It turns out that this goal can be achieved with a straightforward E-GGNN architecture, which we name $\us$. Our next step will be to present this architecture.

\subsection{The $\us$ architecture}
\label{architecture}
Our goal in this section is to build a maximally expressive EGNN. For this, we need to implement injective readout and aggregation functions $f_{(t)},\finv, \textsc{ReadOut}$ (see equations \ref{eq:GGNN}-\ref{eq:readout}). 
The $\us$ architecture we suggest resembles an EGNN architecture \citep{egnn} with multiple equivariant channels as in \citep{levy2023using}.
The $\us$ architecture depends on two hyper-parameters: depth $T$ (the term ‘depth’ refers to the number of MPNN iterations) and channel number $C$. The input to the architecture is a geometric graph $(\mA,\mX)$.
\paragraph{Aggregation}
The architectures then maintains, at each iteration $t$ and node $i$, an 'equivariant' node  feature with $C$ channels
\begin{equation}
    \vv^{(t)}_i=(v^{(t)}_{i,1},\ldots,v^{(t)}_{i,C}), \quad v^{(t)}_{i,c} \in \RR^{d}, c=1,\ldots,C
\end{equation}
These feature values are initialized to zero and are updated via the rotation and permutation equivariant aggregation formula
\begin{equation}
    v^{(t+1)}_{i,q}=\sum_{j\in \gN_i}\left( \phi^{(t,q,0)}(\|\vx_i-\vx_j\|,\|\vv_j\|)(\vx_i-\vx_j)+\sum_{c=1}^C \phi^{(t,q,c)}(\|\vx_i-\vx_j\|,\|\vv_j\|)v_{jc}^{(t)}  \right) 
\end{equation}
where $\|\vv_j\|$ is the $C$ dimensional vector containing the norms of the vectors $v_{jc}, c=1,\ldots,C$. 
\paragraph{Construction of $\finv$ and \textsc{ReadOut}}
After $T$ iterations, we obtain rotation and permutation invariant node features $\vs_i\in \RR^C$ and global feature $\sglobal \in \RR^C $ via
\begin{align}
s_{i,q}=\|\sum_{c=1}^C \theta_{c,q}v^{(T)}_{i,c}\|^2 , \quad
\sglobal_q=\sum_{i=1}^n \phi_{\mathrm{global}}^{(q)}(\vs_i), \quad q=1,\ldots,C
\end{align}
Where $\theta$ is a learnable matrix, and 
the functions $\phi_{\mathrm{global}}^{(q)},\phi^{(t,q,c)}$ are
all fully connected neural networks with a single layer, an output dimension of $1$, and an analytic non-polynomial activation (in our implementation, TanH activation). 
The complexity of $\us$ is linear in $N\cdot C$, as we discuss in Appendix \ref{complexity}. 

The next theorem shows that $\us$ is generically maximally expressive: \begin{restatable}{theorem}{express}
\label{thm:maximally}
Let $d, N, T$ be natural numbers. Let $F$ be a maximally expressive E-GGNN of depth $T$, and let $F_\theta$ denote the $\us$ architecture with depth $T$ and $C=2Nd+1$ channels. Then, for Lebesgue almost every choice of network parameters $\theta$, we have that for all generic $\G,\hat \G \in \GdN$, 
$$ F(\G)=F(\hat \G) \iff F_\theta(\G)=F_\theta(\hat \G) $$
\end{restatable}

\begin{proof}[Proof idea]
 The proof relies on the methodology developed by \citet{gortler_and_dym, Neural_Injective} and \citet{gramnet}. We note that each step in $\us$ employs $C$ identical analytic functions indexed by $q=1,\ldots, C$, with independent parameters. \citet{gortler_and_dym} and \citet{Neural_Injective} showed that in this setting, if every pair of non-isomorphic inputs can be separated by some parameter of the original function, then $C=2Nd+1$ copies of this function with random parameters will suffice to separate \emph{all pairs} uniformly. The proof is then reduced to showing that any pair of non-isomorphic inputs can be separated by some parameter vector, which we prove in the appendix \ref{sub:maximal}.
\end{proof}
Joining this theorem with our theorem on maximally expressive E-GGNN, we deduce 
\begin{corollary}\label{cor:us}
Let $d, N$ be natural numbers. 
let $F_\theta$ denote the $\us$ architecture with depth $T=d+1$ and $C=2Nd+1$ channels. Then for Lebesgue almost every $\theta$, for every pair of graphs $\G, \G'\in \GdN $ which are generic and connected,
$$F_{\theta}(\G)=F_{\theta}(\hat \G) \iff \G, \hat \G \text{ are geometrically isomorphic}$$
\end{corollary}
Note that by \cite{on_the_expressive_power}, theorems 20, 22, the dimension of any injective embedding must be at least $nd-\frac{d(d-1)}{2}$, showing our embedding dimension $2nd+1$ is optimal up to factor of $2$ (when $n\gg d$).
\section{Experiments}
In the experiments section, we have three main objectives. First, we aim to demonstrate that invariant models are indeed strictly less expressive than equivariant models and that power graphs help to mitigate this gap. Second, we seek to show that our model can distinguish challenging graph examples, including those requiring multiple layers for separation, thereby avoiding the bottleneck phenomenon encountered in other geometric models. Finally, we aim to demonstrate the improved performance of our model on real-world benchmarks.
\subsection{Separation experiments}
\label{sepearion exps}
\paragraph{Enhancing Model Expressiveness: Power Graphs in Invariant vs. Equivariant Comparisons}
We begin with experiments to corroborate our theoretical findings. We consider examples where the graph $\mA$ is a line graph, which is connected but not generically globally rigid. We choose two pairs of geometric line graphs, Pair a and Pair b, depicted in Figure \ref{fig:Two pairs for power Graphs}. These pairs have identical distances along graph edges but are not isomorphic. Therefore, they cannot be distinguished by any I-GGNN but (at least when generic) can be distinguished by E-GGNN and our $\us$ architectures. 
Following the protocol from \cite{on_the_expressive_power}, we construct binary classification problems from Pair a and Pair b and apply our E-GGNN architecture $\us$, as well as three popular I-GGNN architectures: SchNet \citep{SchNet}, DimeNet \citep{DimeNet}, and SphereNet \citep{SphereNet}. 

The results are shown in the first column in Table \ref{tab:power_graph}. As expected, $\us$ perfectly separates pairs a and b, while SchNet fails to separate both. Note that DimeNet and SphereNet separate Pair a but not Pair b. This is because DimeNet and SphereNet compute angles at edges, which utilize information not contained in the graph $\G$ but only in the power graph $\G^{2}$. Indeed, distances between two-hop neighbors are sufficient for discriminating between the two geometric graphs in Pair a but not in Pair b, where the distance between the three-hop red and orange nodes is needed for separation. 
\begin{figure}[t]
    \centering
    \includegraphics[width=1.0\textwidth]{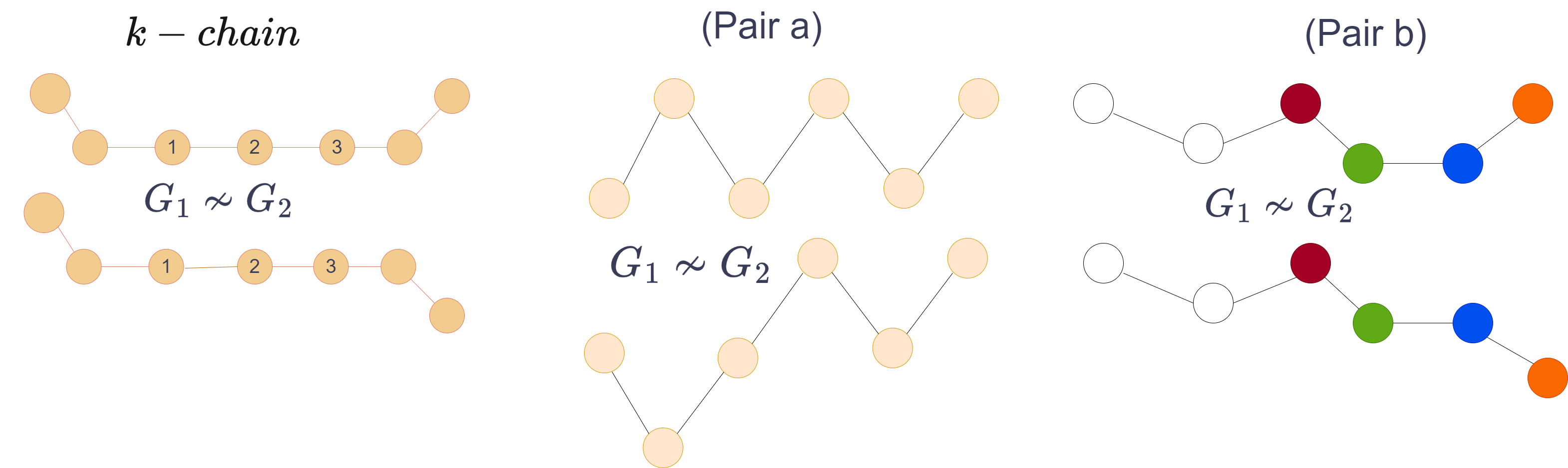}
    \caption{From Left to Right: $k$-chain pair graphs that are non-generic and require $\sim k/2$ blocks for separation, Pair a in which $2$-power graph is enough for I-GGNN to distinguish and Pair b in which $3$-power graph is necessary and sufficient for separation.}
    \label{fig:Two pairs for power Graphs}
\end{figure}
Next, as we suggested in Section \ref{sec:inv}, we apply the same algorithms to the powers of the original graph. As shown in Table \ref{tab:power_graph}, this improves the separation capabilities of the I-GGNN models. All invariant models succeed when the third power is taken, as expected since a connected graph's $d+1$ power (here $d=2$) is generically globally rigid. 
\begin{table}
\begin{center}
     \caption{E-GGNN models like $\us$ can separate the two pairs of connected graph in Figure \ref{fig:Two pairs for power Graphs}. I-GGNN models cannot separate unless power graphs are considered.}       
\begin{tabular}{clcccc}
            \toprule
            & \multicolumn{5}{c}{\textbf{Power graph}} \\
            & \textbf{GNN Layer} & $1$ & $2$ & $3$ &  \\
            \midrule                                              
            \multirow{4}{*}{\rotatebox[origin=c]{90}{Pair a}} &
            \gray{Maximal I-GGNN} & \gray{50\%} & \gray{100\%} & \gray{100\%} \\
            & SchNet \cite{SchNet} & \cellcolor{red!10} 50.0 {\scriptsize {\scriptsize ± 0.0}} & \cellcolor{green!10} 100.0 {\scriptsize {\scriptsize ± 0.0}} & \cellcolor{green!10} 100.0 {\scriptsize {\scriptsize ± 0.0}} \\
            & DimeNet \cite{DimeNet}  & \cellcolor{green!10} 100.0 {\scriptsize {\scriptsize ± 0.0}} & \cellcolor{green!10} 100.0 {\scriptsize {\scriptsize ± 0.0}} & \cellcolor{green!10} 100.0 {\scriptsize {\scriptsize ± 0.0}} \\
            & SphereNet \cite{SphereNet}  & \cellcolor{green!10} 100.0 {\scriptsize {\scriptsize ± 0.0}} & \cellcolor{green!10} 100.0 {\scriptsize {\scriptsize ± 0.0}} & \cellcolor{green!10} 100.0 {\scriptsize {\scriptsize ± 0.0}} \\
            \bottomrule
            & SchNet $_{\text{full graph}}$ \footref{schnet models}  & \cellcolor{green!10} 100.0 {\scriptsize ± 0.0} & \cellcolor{green!10} 100.0 {\scriptsize ± 0.0} & \cellcolor{green!10} 100.0 {\scriptsize ± 0.0} \\
            & SchNet$_{\text{global feat}}$ \footref{schnet models} & \cellcolor{green!10} 100.0 {\scriptsize ± 0.0} & \cellcolor{green!10} 100.0 {\scriptsize ± 0.0} & \cellcolor{green!10} 100.0 {\scriptsize ± 0.0} \\
            \bottomrule
             & $\us$(us) & \cellcolor{green!10} 100.0 {\scriptsize ± 0.0} & \cellcolor{green!10} 100.0 {\scriptsize ± 0.0} & \cellcolor{green!10} 100.0 {\scriptsize ± 0.0} \\
            \bottomrule
            \multirow{4}{*}{\rotatebox[origin=c]{90}{Pair b}} &
            \gray{Maximal I-GGNN} & \gray{50\%} & \gray{50\%} & \gray{100\%} \\
            & SchNet \cite{SchNet}  & \cellcolor{red!10} 50.0 {\scriptsize {\scriptsize ± 0.0}} & \cellcolor{red!10} 50.0 {\scriptsize {\scriptsize ± 0.0}} & \cellcolor{green!10} 100.0 {\scriptsize {\scriptsize ± 0.0}} \\
            & DimeNet \cite{DimeNet} & \cellcolor{red!10} 50.0 {\scriptsize {\scriptsize ± 0.0}} & \cellcolor{green!10} 100.0 {\scriptsize {\scriptsize ± 0.0}} & \cellcolor{green!10} 100.0 {\scriptsize {\scriptsize ± 0.0}} \\
            & SphereNet \cite{SphereNet}  & \cellcolor{red!10} 50.0 {\scriptsize {\scriptsize ± 0.0}} & \cellcolor{green!10} 100.0 {\scriptsize {\scriptsize ± 0.0}} & \cellcolor{green!10} 100.0 {\scriptsize {\scriptsize ± 0.0}} \\
            \bottomrule
            & SchNet $_{\text{full graph}}$ \footref{schnet models}  & \cellcolor{green!10} 100.0 {\scriptsize ± 0.0} & \cellcolor{green!10} 100.0 {\scriptsize ± 0.0} & \cellcolor{green!10} 100.0 {\scriptsize ± 0.0} \\
            & SchNet$_{\text{global feat}}$ \footref{schnet models} & \cellcolor{green!10} 100.0 {\scriptsize ± 0.0} & \cellcolor{green!10} 100.0 {\scriptsize ± 0.0} & \cellcolor{green!10} 100.0 {\scriptsize ± 0.0} \\
            \bottomrule
             & $\us$ (us) & \cellcolor{green!10} 100.0 {\scriptsize ± 0.0} & \cellcolor{green!10} 100.0 {\scriptsize ± 0.0} & \cellcolor{green!10} 100.0 {\scriptsize ± 0.0} \\
           \bottomrule
            \end{tabular}

            \label{tab:power_graph}
    \end{center} 
    \end{table}

\paragraph{Unpacking the Bottleneck: Layer Requirements for Complex Graph Separation}
\paragraph{k-chains}
Next, we consider the $k$ chain pairs suggested in \cite{on_the_expressive_power} and illustrated in the left of Figure \ref{fig:Two pairs for power Graphs}. By design, this is a pair of graphs with $k$ nodes, which can only be separated by $k-1$ hop distances. In \cite{on_the_expressive_power}, it was shown that a depth of at least $\lfloor k/2 \rfloor+1 $ is necessary to achieve separation by E-GGNN in this setting. Note that as most nodes in the chain graphs are on the same line, this point cloud is not generic, so the fact that for large $k$, more than $d+1$ iterations are needed does not contradict our theory.
In our experiments, we include two variations of SchNet \citep{SchNet}: full graph and global feature. Those variations use all pairwise distances, thus showing better results than other invariant models. 
In the first experiment, we set $k = 4$.
Theoretically, $3$ blocks are needed for propagating the information, and $2$ blocks are insufficient.
As shown in Appendix \autoref{tab:4chains}, we succeed with a probability of $1$ over $10$ trials with $3$ blocks.
For checking long-term dependencies, we choose $k=12$ again; $7$ blocks are needed, and $6$ is insufficient. We examined all models with $7,8,9,10,11$ blocks, each $10$ times. As shown in \autoref{tab:12chains}, we are the only ones succeeding with $7$ blocks, and all other models need a much higher number of blocks. This may suggest our model is less influenced by over-squashing \cite{alon2020bottleneck}. In the appendix, \autoref{tab:time_com}, we add a time comparison table for all models using the $128$ dimensional feature.


\footnotetext{\label{schnet models} Here, SchNet global and full are invariant models utilizing the full graph structure (unlike other models utilizing the given combinatorial structure).}

\paragraph{Hard examples}
In \autoref{tab:hard_pc_results} in the appendix, we show that our model can also perfectly separate challenging examples from \citep{PhysRevLett.125.166001,gcnn}, using the experiment setup taken from \cite{gramnet}. These geometric graph pairs cannot be separated by I-GGNN, even on full graphs, and are also difficult to separate using equivariant methods like EGNN. 

\paragraph{Chemical property experiments}
Finally, we check our model's ability to learn regression-invariant real-world chemical properties, a critical task in computational chemistry and material science. Learning chemical properties accurately is essential for predicting the behaviors of compounds, enabling advancements in drug discovery, material design, and environmental science. By effectively capturing these properties, the model demonstrates its potential to assist in solving complex chemical problems, where precise property prediction can reduce experimental costs and guide targeted research.
As in \citep{baselines}, we compare to the following models: Random Forest \citep{Random_forest}, LSTM \citep{LSTM}, Transformer \citep{Transformer}, GIN \citep{GIN}, GIN+VN \citep{GIN-VN}, ChemProp \citep{ChemProp}, GraphGPS \citep{GraphGPS}, SchNet \citep{SchNet}, DimeNet++ \citep{DimeNet++}, GemNet \citep{gemnet}, PaiNN \citep{painn}, ClofNet \citep{clofnet}, LEFTNet \citep{leftnet}, and Marcel \citep{baselines}.
Our task is learning chemical property prediction on three datasets: Drugs \citep{Drugs}, Kraken \citep{Kraken}, and BDE \citep{BDE}. Drugs is the largest and most challenging dataset, containing 560K samples; BDE is a medium-sized dataset with 80K samples; and Kraken is the smallest, with 22K samples.
The baseline model performance, data split, and training procedure are taken from \cite{baselines}. In particular, we ran each experiment three times and reported the model's test error with the best validation performance. 
In Table \ref{mean} in the appendix, we report the mean and standard deviation. 
\begin{table}[t]
    \centering
       \caption{Performance of 1D, 2D, and 3D baseline MRL models and the best results from ensemble learning strategies on 3D GNNs. The metric used is the Mean Absolute Error (MAE, \(\downarrow\)). The \textbf{bold} indicates the best-performing model, while \underline{underlined} denotes the second-best.}
    \small  
    \setlength{\tabcolsep}{2pt}  
    \renewcommand{\arraystretch}{1.2}  
    \begin{tabular}{@{}lccccccccccc@{}}
        \toprule
        \rowcolor{gray!25}
        & & \multicolumn{3}{c}{\textbf{Drugs}} & \multicolumn{4}{c}{\textbf{Kraken}} & \multicolumn{2}{c}{\textbf{BDE}} \\
        \cmidrule(lr){3-5} \cmidrule(lr){6-9} \cmidrule(lr){10-11}
        \textbf{Category} & \textbf{Model} & \textbf{IP} & \textbf{EA} & $\boldsymbol{\chi}$ & $\mathbf{B_5}$ & \textbf{L} & $\mathbf{BurB_5}$ & $\mathbf{BurL}$ & \textbf{BE} \\
        \midrule
        \multirow{3}{*}{1D} 
        & Random Forest  & 0.4987 & 0.4747 & 0.2732 & 0.4760 & 0.4303 & 0.2758 & 0.1521 & 3.0335 \\
        & LSTM           & 0.4788 & 0.4648 & 0.2505 & 0.4879 & 0.5142 & 0.2813 & 0.1924 & 2.8279 \\
        & Transformer    & 0.6617 & 0.5850 & 0.4073 & 0.9611 & 0.8389 & 0.4929 & 0.2781 & 10.0771 \\
        \midrule
        \multirow{4}{*}{2D} 
        & GIN            & 0.4354 & 0.4169 & 0.2260 & 0.3128 & 0.4003 & 0.1719 & 0.1200 & 2.6368 \\
        & GIN+VN         & 0.4361 & 0.4169 & 0.2267 & 0.3567 & 0.4344 & 0.2422 & 0.1741 & 2.7417 \\
        & ChemProp       & 0.4595 & 0.4417 & 0.2441 & 0.4850 & 0.5452 & 0.3002 & 0.1948 & 2.6616 \\
        & GraphGPS       & 0.4351 & 0.4085 & 0.2212 & 0.3450 & 0.4363 & 0.2066 & 0.1500 & 2.4827 \\ 
        \midrule
        \multirow{8}{*}{3D} 
        & SchNet         & 0.4394 & 0.4207 & 0.2243 & 0.3293 & 0.5458 & 0.2295 & 0.1861 & 2.5488 \\
        & DimeNet++      & 0.4441 & 0.4233 & 0.2436 & 0.3510 & 0.4174 & 0.2097 & 0.1526 & \underline{1.4503} \\
        & GemNet         & 0.4069 & 0.3922 & \textbf{0.1970} & 0.2789 & 0.3754 & 0.1782 & 0.1635 & 1.6530 \\
        & PaiNN          & 0.4505 & 0.4495 & 0.2324 & 0.3443 & 0.4471 & 0.2395 & 0.1673 & 2.1261 \\
        & ClofNet        & 0.4393 & 0.4251 & 0.2378 & 0.4873 & 0.6417 & 0.2884 & 0.2529 & 2.6057 \\
        & LEFTNet        & 0.4174 & 0.3964 & 0.2083 & 0.3072 & 0.4493 & 0.2176 & 0.1486 & 1.5328 \\
        & Marcel         & \underline{0.4066} & \underline{0.3910} & \underline{0.2027} & \underline{0.2225} & \underline{0.3386} & \underline{0.1589} & \underline{0.0947} & 1.4741 \\
        & $\us$(Ours)      & \textbf{0.3965} & \textbf{0.3686} & 0.2068 & \textbf{0.1878} & \textbf{0.1738} & \textbf{0.1530} & \textbf{0.0626} & \textbf{0.3170} \\
        \bottomrule
    \end{tabular}
    \label{cpe}
\end{table}

As we see in \autoref{cpe}, we show comparable or improved results in all tasks. For example, in BDE, we have an improvement of $78\%$, in Kraken property \emph{burL} $33\%$, and in the property \emph{L} $48\%$. 
Remarkably, our method outperforms competing more complex methods like GemNet, which employs high-order irreducible representations. This can be explained by the generic completeness of our method, coupled with results we will show in the appendix \autoref{tab:norms_repeat}, which show that these datasets have a small number of norm-repetitions, indicating  that they are  ‘close to generic’ .

\section{Limitation and Future Work}
In this work, we characterized the generic expressive power of I-GGNN and E-GGNN, showed that $\us$ is a generically maximally expressive E-GGNN, and showed the effectiveness of this architecture on chemical regression tasks. A limitation of our work is that it may struggle with symmetric and non-generic input. Future work could consider assessing the expressive power of non-MPNN models like geometric 2-WL on sparse graphs to address such input efficiently.

\newpage
\section*{Appendix}
\phantomsection
\addcontentsline{toc}{section}{Appendix}
\addtocontents{toc}{\protect\setcounter{tocdepth}{0}}
\setcounter{section}{0}
\renewcommand{\thesection}{\Alph{section}}
\renewcommand{\theHsection}{appendix.\thechapter.\Alph{section}}
\section{Related Work}
This section discusses related work that is not discussed in the main text. 
\paragraph{Equivariant models}
In the context of learning on geometric graphs, models should be equivariant to the joint action of permutations, rotations, and translations. Besides the models discussed in the paper, some prominent models with this equivariant structure include Cormorant \citep{cormorant}, Vector Neurons \citep{deng2021vector}, and SE(3) transformers \citep{se3transformer}.

\paragraph{Completeness, universality and generic completeness}
The main text discusses related works on completeness focused on methods coming from the geometric learning community. Other complete methods, generally using similar ideas, appeared independently in the more chemistry-oriented community\citep{kurlin2023polynomial,widdowson2023recognizing,shapeev,nigam2020recursive}. Completeness in $2D$ was achieved in \cite{zz}, and completeness with respect to rigid motions (without permutations) was achieved by Comenet \citep{comenet}. Comenet also uses some genericity assumptions (e.g., that there is a well-defined notion of nearest neighbors).  

The main text discussed how generic completeness can be achieved using I-GGNN when the graph is globally rigid. A similar result can be obtained when considering the list of all distances \citep{reconstructible}. Generic completeness for full graphs can also be obtained using more complex distance-based features \citep{widdowson2022generic} or shape principal axes \citep{puny2021frame} (under the generic assumption that the principal eigenvalues are distinct).

\paragraph{Comparison with \cite{new}}  
Generic completeness without the full graph assumption, employing only the assumption that local neighborhoods are not symmetric, is discussed by \cite{new}. However, there are notable differences in key aspects of the results. Specifically, Theorem 4.2 in \cite{new} demonstrates that their proposed equivariant model can identify generic molecules using \( L \) message-passing iterations, where \( L \) is the graph's diameter. In contrast, our Theorem \ref{thm:ggnn} requires only \( d+1 \) iterations (where \( d = 3 \) in typical applications, resulting in a total of just four iterations).  

Moreover, while Theorem 4.2 in \cite{new} focuses on recovering the 3D structure, it does not address the combinatorial structure. Our Theorem \ref{thm:ggnn}, however, guarantees the recovery of both combinatorial and geometric structures in the generic case.  

Additionally, our Theorem \ref{thm:iggnn} precisely describes the conditions under which invariant models are, and are not, generically complete—a result not covered in \cite{new}.

\section{Implementation details}
Here, we detail the hyper-parameter choice.
For all experiments, we use AdamW optimizer \cite{AdamW}. We use a varying weight decay of $1e^{-4},1e^{-5}$ and learning rate of $1e^{-4}$. Reduce On Plateau scheduler is used with a rate decrease of 0.75 with the patience of 15 epochs.
We use two blocks with $40$ channels in the challenging point cloud separation experiments.
For \emph{k-chain} experiments, we repeat each experiment $10$ times, with $10$ different seeds, and set the number of channels to be $60$ and use a varying number of blocks, as detailed in the table.
For power graph experiments, we repeat each experiment $10$ times, with $10$ different seeds, and set the number of channels to be $60$ and $3$ blocks.
For chemical property experiments, the entire procedure is taken from \cite{baselines}: each dataset is partitioned randomly into three subsets: 70\% for training, 10\% for validation, and
20\% for test. We set the number of channels to be $256$ and used $6$ blocks. Each model is trained over $1,500$ epochs. Experiments are repeated three times for all eight regression targets, and the results reported correspond to the model that performs best on the validation set in terms of Mean Absolute Error (MAE). In contrast to the experiments in \cite{baselines}, we didn't consider the EE dataset as it's currently not public on GitHub.
We utilize PyTorch and PyTorch-Geometric to implement all deep learning models. All
experiments are conducted on servers with NVIDIA A$40$ GPUs, each with 48GB of memory. All dataset details, including statistics, can be found in \cite{baselines}.

\section{Building perfect I-GGNN}
\label{building i-ggnn}
Here, we describe in detail how to construct a perfect I-GGNN. We will assume that we consider geometric graphs with up to $N$ nodes. The main ingredients needed are aggregation functions and \textsc{ReadOut} function that are injective on the space of all multisets of size $\leq N$. \cite{Neural_Injective} proved that the sum of projections and analytic activation can be used to construct such layers. Formally, let $\M$ be a semi-analytic set of dimension $D$, and $N$ vectors in $\mathbb{R}^{d}$ denoted by $\mX$. let $A\in \mathbb{R}^{(2D+1) \times d}, b\in \mathbb{R}^{2D+1}$, $\sigma:\mathbb{R}\rightarrow \mathbb{R}$ analytic non-polynomial activation, then for Lebesgue almost every $A,b$ $$f_{A,b}(\mX) = \sum^{N}_{i=1} \sigma(A\cdot \mX_{i}+b)$$ is injective up to permutation.
Thus, the composition of such blocks and such \textsc{ReadOut} function with dimension $D = d\cdot N$ yields a maximally expressive model on all input domain. Here, $D$ is the intrinsic dimension of all features, ultimately produced by the $D$ dimensional input space. 
\section{Separation Power of Fully Maximally expressive E-GGNN }
\label{maximally_expressive_egnn}
In the main text, we proposed $\us$ a simple \emph{generically} maximal expressive E-GGNN. In this section, we show that if we have a maximally expressive E-GGNN on all input domain, we can reconstruct a full graph up to group action, and if the graph is connected, we can reconstruct its point cloud. This indicates that fully maximally expressive E-GGNN is as challenging as achieving completeness, which generally requires complex methods like $2$-WL-based GNNs.
\begin{theorem}
    \label{reconstruction}
    Let $\G = (\mA,\mX)$ be a geometric graph, where $\mA$ is the full graph on $n$ nodes, and $\mX$ is an arbitrary $d\times n$ matrix, then one iteration of maximally expressive E-GGNN can reconstruct $\mX$.
\end{theorem}
\begin{proof}
    Assume we run one iteration of E-GGNN on our geometric graph, then for each node $i$, we know up to some rotation $Q_{i}$
    \begin{align*}
        \lms Q_{i}\cdot (x_{i} - x_{j}) | j \in [n] \rms 
    \end{align*}
  Choosing $i=1$, and $x_{i}=0$, we have all other positions $Q_{i} \cdot x_{j},j\neq i$, thus up to translation (as we set $x_{i}$), rotation $Q_{i}$ and node permutation, we can reconstruct our point cloud. 
\end{proof}
We now show that using $n$ iterations of E-GGNN can reconstruct the point cloud of each connected graph. 
\begin{theorem}
        Assuming a connected geometric graph $\G = (\mA,\mX)$ with $n$ nodes, then $n$ iteration of maximally expressive E-GGNN can reconstruct the point cloud.
\end{theorem}
\begin{proof}
    Assume we run $n$ iterations of a maximally expressive E-GGNN on our connected graph.
    By the injectivity of the aggregation functions, we can recursively reconstruct from
    $\vv_i^{(t)}$, up to a rotation $Q_i$, all relative displacement vectors appearing in
    the $t$-hop computation tree rooted at node $i$. In particular, for every node $j$
    whose graph distance from $i$ is at most $t$, we can reconstruct
    $Q_i(\vx_i-\vx_j)$ by summing the relative displacement vectors along a path from
    $i$ to $j$. Since the graph is connected and its diameter is at most $n-1$, after
    $n$ iterations this allows us to reconstruct all node positions relative to $\vx_i$,
    up to a common rotation $Q_i$. Hence, up to translation, rotation, and node
    permutation, we can reconstruct the point cloud.
\end{proof}
\section{I-GGNN+}
\label{I-GGNN types}
Here, we present the definition of I-GGNN presented by \cite{on_the_expressive_power}, denoted by I-GGNN+, and show that it can separate all geometric full graphs. After, we explain why this model is strictly stronger than our I-GGNN.

\cite{on_the_expressive_power} defines (neglecting the update step also used there)
\label{I-GGNN+ types}
\begin{equation}
    \vv_i^{(t+1)}=f^{t}\left( \lms  \vv_j^{(t)}, \vx_i-\vx_j , \mid j \in \mathcal{N}_i \rms \right).
\end{equation}
Such that $f^{t}$ holds that for all $Q\in O(d)$ 
\begin{align*}
    f^{t}\left( \lms  \vv_j^{(t)}, \vx_i-\vx_j , \mid j \in \mathcal{N}_i \rms \right)= f^{t}\left( \lms  \vv_j^{(t)},Q \cdot(\vx_i-\vx_j) , \mid j \in \mathcal{N}_i \rms \right)
\end{align*}
I-GGNN+ is maximally expressive if $f^{t}$ holds:
\begin{align*}
    f^{t}\left( \lms  \vv_j^{(t)}, \vx_i-\vx_j , \mid j \in \mathcal{N}_i \rms \right)= f^{t}\left( \lms  \vv_j^{(t)}, \hat{\vx_i}-\hat{\vx_j} , \mid j \in \mathcal{N}_i \rms \right) \iff \\
    \exists Q \in O(3):  \lms  \vv_j^{(t)}, \vx_i-\vx_j , \mid j \in \mathcal{N}_i \rms = \lms  \vv_j^{(t)}, Q \cdot(\hat{\vx_i}-\hat{\vx_j}) , \mid j \in \mathcal{N}_i \rms
\end{align*}
We now prove that if we run one iteration of maximally expressive I-GGNN+ on a full graph, we can reconstruct it.
\begin{theorem}
      Let $\G = (\mA,\mX)$ be a geometric graph, where $\mA$ is the full graph on $n$ nodes, and $\mX$ is an arbitrary $d\times n$ matrix, then one iteration of maximally expressive I-GGNN+ can reconstruct $\mX$.
\end{theorem}
\begin{proof}
    Assume we run one iteration of I-GGNN+. Then, for each node $i$, we have $$f^{t}\left( \lms  \vv_j^{(t)}, \vx_i-\vx_j , \mid j \in [n] \rms \right)$$ then up to some rotation $Q_{i}\in O(3)$, we know $$\lms  Q_{i}\cdot(\vx_i-\vx_j) , \mid j \in [n] \rms$$ Then, we can reconstruct the point cloud by \autoref{reconstruction}. 
\end{proof}
It's easy to see that I-GGNN+ can simulate our I-GGNN as the norm function is invariant under multiplication by $Q\in O(3)$. But \citep{PhysRevLett.125.166001,gcnn} gave examples of full-graphs each I-GGNN can't separate, but by what we proved above, maximally expressive I-GGNN+ can separate them. Thus, I-GGNN+ is strictly more powerful than I-GGNN in its expressive power.
\section{Approximation and separation}\label{sec:approx}
It is known that a tight connection exists between the approximation of invariant functions and complete invariant models \cite{gramnet}. 

In the context of our paper, we discussed generic completeness. In particular, we showed in Corollary \ref{cor:us} that $\us$ with random parameters is complete on $\GdNgeneric$, which denotes the set of all geometric graphs $\G=(\mA,\mX)$ with exactly $N$ nodes, $\mA$ connected, and $\mX\in \RR^{d\times n}$ generic. Accordingly, we can obtain the following theorem:
\begin{theorem}
Let $K \subseteq \GdNgeneric$ be a compact set, $f^{sep}:K \rightarrow R^{m}$ a continuous function invariant to rotations, translations, and permutations. Then $f^{sep}$ is injective on $K$ (up to symmetries) if and only if every continuous invariant $f:K\rightarrow R^{M}$ can be approximated uniformly on $K$ using functions of the form $\N(f^{sep})$, where $\N$ is a fully connected neural network.
\end{theorem}
\begin{proof}
    Assume, on the one hand,  $f^{sep}$ is injective on $K$ (up to symmetries). Given a continuous $f: K \rightarrow R^{M}$ we want to approximate, then, by \cite{gramnet}, $\forall \epsilon > 0$ $\exists$ neural network $\N$: \begin{align*}
        sup_{\G \in K} |f(\G) - \N(f^{sep}(\G))| < \epsilon 
    \end{align*}
On the other hand, assume by contradiction there exists $\G_{1} = (\mA_{1},\mX_{1})$ and $\G_{2}=(\mA_{2},\mX_{2})$ which are not related by translation, rotation or permutation, such that $f^{sep}(\G_{1}) = f^{sep}(\G_{2})$. 

Define $f(\G) = Inf_{g \in G} \|(\mA,\mX)- g \cdot (\mA_{1},\mX_{1})\|$, where $\G=(\mA,\mX$), and $G$ is the group of permutations, rotations, and translation. Note that this minimum is obtained as the product group of rotation and permutation is compact, and the translation can be bounded by the sum of the norms of the two graphs. Then, according to the Tychonoff's theorem, our group is compact, and we take a minimum continuous function over a compact set. Note that $f$ is an invariant, continuous function such that $f(\G_{1})=0$, and $f(\G_{2})> 0$, but for each neural network $\N$, $\N(f^{sep}(\G_{1})) = \N(f^{sep}(\G_{2}))$, and we can't approximate our desired function $f$ with $\epsilon = \frac{f(\G_{2})}{2}$ accuracy, yielding a contradiction.
\end{proof}
\section{Proofs}
In this section, we restate and prove theorems that have not been fully proved in the main text.
\IGGNN*
\begin{proof}
\label{Proof of I-GGNN}
Let $\mathcal{G}=(\mA,\mX) $ be a geometric graph, and assume that $\mX$ is generic. 
First, we assume that $\mA$ is generically globally rigid. Assume that there is some geometric graph $\mathcal{G}'$ which is assigned the same global feature as $\mathcal{G}$ after a single iteration. We need to show the two graphs are geometrically isomorphic.
The multisets of $s_i$ and $s_i'$ features in \eqref{eq:readout} are equal, and in particular, both graphs have the same number of nodes $n$, and by relabeling if necessary, we can assume that $s_i=s_i'$ for all $i=1,\ldots,n$. This equation, in turn, implies that for every node $i$, 
$$\lms \|\vx_{ij}\|, j \in \N_i \rms=\lms \|\vx_{ij}'\|, j \in \N_i' \rms$$
In particular, the $i$'th node in $\G$ and $\G'$ have the same degree, so both graphs have the same number of edges.
For any fixed $(i,j)$ which is an edge of $\G$, the distance $\|x_{ij}\|'$ will appear exactly once in the multiset corresponding to $s_i'$ and $s_j'$ and will not appear in the other multisets. This observation implies that $(i,j)$ is also an edge in the graph $\G'$. Since the number of edges in both graphs is the same, we deduce that $\mA=\mA'$.
By global rigidity, we deduce that all pairwise distances are the same, and by \cite{egnn}, the two graphs are geometrically isomorphic.

We mention a theorem proved in \cite{gortler2010characterizing} for the other direction. \cite{gortler2010characterizing} proved $\mA$ is globally rigid if exists generic $\mX$ such that $(\mA,\mX)$
is globally rigid. Thus if $\mA$ is not globally rigid, for \emph{every} generic $\mX$, there exists $\mX'$ for whom distances along edges are preserved, but $\mX$ and $\mX'$ aren't geometrically isomorphic. By immediate induction on the number of layers, it can be shown that any I-GGNN cannot separate any pair of geometric graphs whose distances across edges are preserved. This concludes the proof.
\end{proof}
\GGNN*
\begin{proof}
\label{Identifiable}
Assume $\mA$ is not connected. Let $\mX \in \RR^{n\times d}$. Let $\mV_{1}$ be a connected component of the graph, and let $\G'$ be the geometric graph whose adjacency matrix is $\mA'=\mA$, and where $\mX'$ is obtained from $\mX$ by applying a non-trivial translation $\vt \in \RR^d$ to all nodes in $\mV_1$: $\hat{\vx}_i = \vx_i + \vt$ for all $i \in V_1$, and setting $\hat{\vx}_j=\vx_j$ for all $j\notin V_1$. We choose $\vt$ with sufficiently large norm so that the two graphs are not geometrically isomorphic. However, since there are no edges between $\mV_1$ and the other connected components, for every edge $(i,j)$ we have $\hat{\vx}_i-\hat{\vx}_j=\vx_i-\vx_j$. Therefore, for any E-GGNN, we can easily see recursively that all features $\vv_i^{(t)}$ constructed from $\mX$ and $\mX'$, respectively, will be the same, and consequently, the E-GGNN will assign the same global features to $\mathcal{G}$ and $\mathcal{G}'$.

Now assume $\mA$ is connected, and we are given two geometric graphs $\G$ and $\hat{\G}$ in $(\GdNgeneric,\GdN)$. By assumption, $\mX$ is generic. Assume that the E-GGNN assigns the same value of $\G$ and $\hat{\G}$. We need to show that the two graphs are geometrically isomorphic.
First, we note that since the two graphs have the same global features, they have the same number of nodes $n$, and  by applying relabeling if necessary, we can assume that 
\begin{align*}
    s_i=\hat{s}_i, \quad \forall i=1,\ldots,n
\end{align*}
which implies that for every node $i$ there is some rotation $R_i\in O(d)$ such that $\vv_i^{(T)}=R_i\hat{\vv}_i^{(T)}$. By rotation equivariance, we know that $R_i\hat{\vv}_i^{(T)}$ is the feature that we would have obtained by applying the maximally expressive E-GGNN to $R_i \cdot \hat\mX $. We deduce that
$$\lms \vv_j^{(T-1)}, \vx_i-\vx_j|j \in \gN_i \rms=\lms R_i\hat{\vv}_j^{(T-1)}, R_i(\hat{\vx}_i-\hat{\vx}_j)| j \in \hat{\gN}_i \rms$$
The equality of multisets above implies that the node $i$ has the same degree in $\mA$ and $\hat{\mA}$. Since this is true for all nodes, $\mA$ and $\hat{\mA}$ have the same number of edges.
Moreover 
$$\lms \|\vx_i-\vx_j\|, j \in \gN_i \rms=\lms \|\hat{\vx_i}-\hat{\vx_j}\|, j \in \hat{\gN}_i \rms $$
Since $\mX$ is generic, for every edge $(i,j)\in \mA$ the distance $||\vx_i-\vx_j|| $ will appear only in the 1-hop multi-sets corresponding to the $i$ and $j$ nodes. Thus, they will appear only once in the 1-hop multisets corresponding to the second graph's $i$ and $j$ nodes. This implies that $(i,j)$ is an edge in $\hat{\mA}$ too, and 
$$ R_i(\hat{\vx}_i-\hat{\vx}_j)=\vx_i-\vx_j, \forall j \in \gN_i.$$
 Since the two graphs have the same number of edges, it follows that $\mA=\hat{\mA} $.
We now show recursively that for all $1\leq t \leq T$, we have $\vv_j^{T-t}=R_i \hat{\vv}_j^{T-t} $ and $\vx_i-\vx_j=R_i(\hat{\vx}_i-\hat{\vx}_j)$ for all $j$ that is a $t$-hop neighbor of $i$ (we say $j$ is an $t$-hop neighbor of $i$, if there is a path of length at most $t$ from $i$ to $j$). 

We have already proved the claim for $t=1$. Now we assume correctness for $t$ and show for $t+1$. Since by the induction hypothesis we know that $\vv_j^{T-t}=R_i \hat{\vv}_j^{T-t} $ for every $j$ which is $t$ hop neighbor of $i$, we deduce the multi-set equality
$$\lms \vv_k^{T-(t+1)}, \vx_j-\vx_k | k \in \gN_j\rms=\lms R_i\hat{\vv}_k^{T-(t+1)}, R_i(\hat{\vx}_j-\hat{\vx}_k)|k \in \gN_j \rms  $$
Since we know that the pairwise norms of $\mX$ are all distinct due to the genericity of $\mX$, and also that for every edge $(j,k)$ we have the norm equality $\|\vx_j-\vx_k\|=\|\hat \vx_j-\hat \vx_k\| $, we see that necessarily
$$\vv_k^{T-(t+1)}= R_i \hat{\vv}_k^{T-(t+1)}, \text {and  }  \vx_j-\vx_k=R_i(\hat{\vx}_j-\hat{\vx}_k).$$

Now,  every $k$, an $t+1$ hop neighbor of $i$, is connected by an edge to a node $j$, an $t$ hop neighbor of $i$. 
By the induction assumption we know that $\vx_i-\vx_j=R_i(\hat{\vx}_i-\hat{\vx}_j) $. We deduce that 
$$\vx_i-\vx_k=\vx_i-\vx_j+\vx_j-\vx_k=R_i(\hat{\vx}_i-\hat{\vx}_j+\hat{\vx}_j-\hat{\vx}_k)=R_i (\hat{\vx}_i-\hat{\vx}_k) $$
for all $t+1$ hop neighbors of $i$. Continuing with this argument inductively, we see that for all $d+1$ hop neighbors $k$ of $i$, we obtain
$$\vx_i-\vx_k=R_i (\hat{\vx}_i-\hat{\vx}_k) $$
and in particular
$$\|\vx_i-\vx_k\|=\|\hat{\vx}_i-\hat{\vx}_k\| $$
for all $d+1$ hop neighbors. By global generic rigidity of the $d+1$ power graph of a connected graph \citep{power-graph,power-graph2}, it follows that the two graphs are geometrically isomorphic.
Thus, as previously, we reconstructed the combinatorial graph, and we now reconstructed the geometric graphs; we are done.
\end{proof}
\subsection{Proof of Maximal Expressivity of $\us$}
\label{sub:maximal}
This subsection is devoted to proving Theorem \ref{thm:maximally} on the generic maximal expressivity of $\us$. This proof will require some background on the finite witness theorem from \cite{Neural_Injective}:

\paragraph{Background on the finite witness theorem} Before we prove the theorem, we review some definitions and results from \cite{Neural_Injective}, which we will use for the proof. 
The main tool we require from \cite{Neural_Injective} is Theorem A.2 from that paper, the finite witness theorem. We state it here in the special case where the parameter space is Euclidean, and the function $F$ is analytic in the parameters:
\begin{theorem}[Finite Witness Theorem, \cite{Neural_Injective}]\label{thm_fwt_full}
Let $\M \subseteq \RR^p$ be a $\sigma$-sub-analytic set of dimension $D$. Let $F:\M \times \mathbb{R}^{q} \to \mathbb{R}$ be a $\sigma$-sub-analytic function, and assume that $F(\bz;\bth)$ is analytic as a function of $\bth$ for all fixed $\bz \in \M$. Define the set
$$\N=\{\bz\in \M| \quad F(\bz;\bth)=0,\ \forall \bth\in \RR^q \}.$$
Then for Lebesgue almost every \footnotemark $\left( \bth^{(1)},\ldots,\bth^{(D+1)}\right) \in \RR^{q\times (D+1)}$,
\begin{equation}\label{eq:Nequals}
\N=\{\bz\in \M| \quad F(\bz;\bth^{(i)})=0,\ \forall i=1,\ldots D+1\} .
\end{equation}
\end{theorem}
\footnotetext{The original statement in \cite{Neural_Injective} discussed \emph{generic} $\left( \bth^{(1)},\ldots,\bth^{(D+1)}\right) \in \W^{D+1}$, in the sense defined in \cite{Neural_Injective}. In particular, this implies the "Lebesgue almost every" statement we used here. }
\textbf{Remark:} We note that this theorem also holds when the output of $F$ is a vector in $\RR^d$, which is what we will use in our proofs below. 
To see the theorem holds for a function $\hat F:\M \times \mathbb{R}^q \to \mathbb{R}^{d}$, we can define a new parametric function $F$ with additional parameters $a\in \RR^d$,
$$F(\bz;\bth,a)=\langle \hat F(\bz;\bth),a \rangle $$
and use the theorem for $F$, which has scalar output, to deduce the theorem for $\hat F$.

The finite witness theorem assumes the domain is a $\sigma$ sub-analytic set. The full definition of $\sigma$-sub-analytic sets is rather involved, and we refer the reader to Appendix A in \cite{Neural_Injective} for a full definition. Rather than giving a full definition here, we will recall the properties of these sets, which we will need. 

The class of $\sigma$ subanalytic sets in $\RR^d$ is rather large. In particular, it contains all open sets in $\RR^d$, all semi-algebraic sets in $\RR^d$ (sets defined by polynomial equalities and inequalities, like spheres and linear sub-spaces), as well as sets defined by analytic equality and inequality constraints. The class of $\sigma$ subanalytic sets is closed under countable unions and finite intersections. 

$\sigma$-sub-analytic sets are always a countable union of smooth manifolds. The dimension of a $\sigma$ sub-analytic set is the maximal dimension of the manifolds composing it. 

A $\sigma$-sub-analytic function is a function whose graph is a $\sigma$ sub-analytic set. Analytic functions are, in particular, $\sigma$-sub-analytic. When a $\sigma$ sub-analytic function is applied to a $\sigma$ sub-analytic set, the image is a $\sigma$ sub-analytic set whose dimension is no larger than the dimension of the $\sigma$ sub-analytic domain \cite{Neural_Injective}. 

We can now restate and prove Theorem \ref{thm:maximally}:
\express*
\begin{proof}
In the proof, we consider the set of all generic geometric graphs in $\GdN$. We will prove the theorem for an even larger set: the set $\GdNdistinct$ of all geometric graphs in $\GdN$, which additionally satisfies that for any distinct edges $(i,j)$ and $(s,t)$ in the graph, 
$0<\|\vx_i-\vx_j\|\neq \|\vx_s-\vx_t\|$. Since polynomial inequalities with integer coefficients define it, we note that $\GdNdistinct$ contains \emph{all generic} graphs with $\leq N$ nodes. The advantage of considering $\GdNdistinct$ is that the set of all geometric graphs in $\GdNdistinct$ of the same cardinality $n \leq N$ can be  
thought of as a subset of $\RR^{n\times n}\oplus \RR^{n \times d} $. Since a constant number of polynomial inequalities defines it, it is a semi-algebraic set, and hence, in particular, $\sigma$-subanalytic, so the finite witness theorem can be applied. In contrast, it is unclear that the set of all generic points of fixed dimension is a $\sigma$ subanalytic set. 

Our goal is to show that for the networks described in the theorem, for almost every choice of parameter values, all functions described in the procedure are injective. To be more accurate, let us introduce some notation given an arbitrary pair 
of geometric graphs 
$$\G=(\mA,\mX),  \quad \hat \G=(\hat \mA,\hat \mX) $$
in our domain $\GdNdistinct$, we denote the features corresponding to the first graph by $\vv_{i}^{(t)},\vs_i$ and $\sglobal$ as in the main text, and the features corresponding to the second graph by $\hat{\vv}_{i}^{(t)},\hat{\vs}_i$ and $\sglobalhat$. The neighborhood of a node $k$ in $\hat \G$ is denoted by $\hat \gN_k $. The number of nodes in the two graphs is denoted by $n$ and $m$, respectively.

We need to show that for all $t=0,\ldots,T-1$ and all nodes $i$ of $\G$ and $k$ of $\hat \G$,
\begin{align}
\vv_{i}^{(t+1)}&=\hat{\vv}_{k}^{(t+1)} \text{ if and only if } \lms (\vx_i-\vx_j,\vv_{j}^{(t)}), j \in \gN_i \rms=\lms (\hat{\vx}_k-\hat{\vx}_j,\hat{\vv}_{j}^{(t)})| j \in \hat{\gN}_k \rms \label{eq:first}\\
\vs_{i}&=\hat{\vs}_k \text{ if and only if } \exists \mQ \in O(d), \mQ\vv_{i}^{(T)}=\hat{\vv}_{k}^{(T)} \label{eq:middle}\\
\sglobal&=\sglobalhat \text{ if and only if } \lms \vs_i| i=1,\ldots,n \rms= \lms \hat \vs_i| i=1,\ldots,m \rms \label{eq:last}
\end{align}
Our first step for reaching this result is using the finite witness theorem, similar to what was done in \cite{gramnet}. We note that the  features we are interested in are composed of $C=2Nd+1$ coordinates, denoted by $q=1,\ldots, C$, which  are of the general form
\begin{align}
v_{i,q}^{(t+1)}-\hat{v}_{i,q}^{(t+1)}&=F_t(\mX,\mV^{(t)},\hat \mX, \hat{\mV}^{(t)};\alpha_{q,t}), \quad t=0,\ldots,T-1 \label{eq:bla}\\
s_{i,q}^{(t+1)}-\hat{s}_{i,q}^{(t+1)}&=F_{T}(\mV^{(T)}, \hat{\mV}^{(t)};\alpha_{q,T})\\
\sglobal_q-\sglobalhat_g&=F_{T+1}(\vs_1,\ldots,\vs_n,\hat{\vs}_1,\ldots,\hat {\vs}_{m}; \alpha_{q,T+1})
\end{align}
where $\alpha_{q,t}$ denote the parameters of the appropriate function, $\mV^{(t)}$ is the concatenation of all $n$ features $\vv_i^{(t)}$, and $\hat \mV^{(t)}$ is the concatenation of all $m$ features $\hat{\vv}_k^{(t)}$. In essence, we claim that by the finite witness theorem, for almost every choice of network parameters, we will have the following equalities for all geometric graphs $\G,\hat \G $ satisfying the theorem's assumptions: 
\begin{align}
\vv_{i}^{(t+1)}&-\hat{\vv}_{k}^{(t+1)}=0 \text{ if and only if } F_t(\mX,\mV^{(t)},\hat \mX, \hat{\mV}^{(t)};\alpha_{t})=0, \forall \alpha \label{eq:first2}\\
\vs_{i}&-\hat{\vs}_k=0 \text{ if and only if } F_{T}(\mV^{(T)}, \hat{\mV}^{(t)};\alpha), \forall \alpha \label{eq:middle2}\\
\sglobal&-\sglobalhat=0 \text{ if and only if } F_{T+1}(\vs_1,\ldots,\vs_n,\hat{\vs}_1,\ldots,\hat \vs_{m}; \alpha)=0, \quad \forall \alpha \label{eq:last2}
\end{align}
In other words, the finite witness theorem enables us to reduce the problem to the problem of showing that some choice of network parameter exists, which enables separation. 

Some explanation is needed as to the details of how the finite witness theorem is applied. It is convenient to think of the network parameters being chosen recursively. For $t=0$, the finite witness theorem allows us to choose the parameters $\alpha_{q,0}, q=1,\ldots, C $ in \eqref{eq:bla} so that the theorem applies. Formally, to do this, we need to split into a finite number of cases: once the number of nodes $n$ and $m$ in the two graphs are specified, as well as the combinatorial graphs $\mA, \hat \mA$ and the nodes $i,k$, we can think of $F_0$ as a function which is analytic in the parameters, and $\sigma$-subanalytic on its domain (the norms used in the network are semi-algebraic functions, and hence $\sigma$-subanalytic functions. All other functions used in the network are analytic). The function $F_0$ operates on the domain of pairs $\mX \in \RR^{d\times n}, \hat \mX \in \RR^{d\times m}$ which satisfy a finite number of polynomial inequalities. This domain is semi-algebraic and has dimensions $d(n+m)\leq 2Nd$. Therefore, we can apply the finite witness theorem with $C=2Nd+1$.  
Next, once the parameters $\alpha_{q,t}$ are chosen for all $t<t'$, we can choose the parameters $\alpha_{q,t'}$ again using the finite witness theorem. Central to the usage here is that, once the previous parameters were fixed, the features in the $t'$ step are an image of a $\sigma$-subanalytic function applied to a pair $(\mX,\hat \mX) $. Therefore, the intrinsic dimension of the $t'$ generation features is no larger than $2Nd=C-1$ (for more on this argument, see \cite{gramnet}). This concludes the recursive definition of parameter choice. By Fubini's theorem, one can verify that, in fact, for Lebesgue almost every parameter will be 'good.'
It remains to show that
\begin{align}
    \text{The right hand side of \eqref{eq:first} holds if and only if the right hand side of \eqref{eq:first2} holds} \label{eq:first_claim}\\
    \text{The right hand side of \eqref{eq:middle} holds if and only if the right hand side of \eqref{eq:middle2} holds} \label{eq:middle_claim}\\
    \text{The right hand side of \eqref{eq:last} holds if and only if the right hand side of \eqref{eq:last2} holds} \label{eq:last_claim}
\end{align}

We now reformulate these three claims as three lemmas.

We use some fixed analytic non-polynomial function $\sigma$ in the following lemmas. We say that $\phi$ is a \emph{basic analytic network} if it is of the form  
$$\phi(\vs;a,b)=\sigma(a \cdot \vs+b)$$
for some $a\in \RR^d, b\in \RR$. 

Our first lemma will show \ref{eq:first_claim}, where  $z_j$ will assume the role of $\vx_i-\vx_j$. 
\begin{lemma}
\label{eq:hardest}
Let $z_1,\ldots,z_n$ and $\hat{z}_1,\ldots,\hat{z}_m$ be non-zero vectors in $\RR^d$. Additionally assume that $\|z_1\|<\|z_2\|<\ldots <\|z_n\| $ and  $\|\hat{z}_1\|<\|\hat{z}_2\|<\ldots <\|\hat{z}_m\| $.

Let $\vv_1,\ldots,\vv_n $ and $\hat{\vv}_1,\ldots,\hat{\vv}_m $ be matrices in $\RR^{d \times C}$, and denote 
$\vv_i=(v_{i,1},\ldots,v_{i,C})$ and  $\hat \vv=(\hat{v}_{i,1},\ldots,\hat{v}_{i,C})$. If for all basic analytic neural networks $\phi^{(0)},\ldots,\phi^{(C)} $ we have that 
\begin{align}\sum_{j=1}^n\left( \phi^{(0)}(\|z_j\|)(z_j)+\sum_{c=1}^C \phi^{(c)}(\|z_j\|)v_{jc}  \right)-
\sum_{j=1}^m\left( \phi^{(0)}(\|\hat{z}_j\|)(\hat{z}_j)+\sum_{c=1}^C \phi^{(c)}(\|\hat{z}_j\|)\hat{v}_{jc}  \right)=0 \label{eq:big}
\end{align}
then
$$\lms (z_1,\vv_1), \ldots(z_n,\vv_n)  \rms
=\lms (\hat{z}_1,\hat{\vv}_1), \ldots(\hat{z}_m,\hat{\vv}_m)  \rms$$
\end{lemma}
We note that this lemma is slightly stronger than we need since, in contrast with the original MPNN, we do not use the norms of $v_{i,c}$ as features.

Our second lemma will show \ref{eq:middle_claim}
\begin{lemma}\label{eq:rot}
Let $\vv=(v_1,\ldots,v_C)\in \RR^{d \times C} $ and  $\hat \vv=(\hat{v}_1,\ldots,\hat{v}_C)\in \RR^{d \times C} $. If for all $\theta \in \RR^C$ we have that 
$$\|\sum_{c=1}^C \theta_cv_c\|=\|\sum_{c=1}^C \theta_c\hat{v}_c\| $$
then there exists an $\mQ \in O(d)$  such that $\mQ v_c=\hat v_c $ for all $c=1,\ldots,C$.
\end{lemma}
For proof of this lemma, see Section 3.2 in  \cite{gortler_and_dym}.

Our third lemma will show \ref{eq:last_claim}
\begin{lemma}
Let $\vs_1,\ldots,\vs_n $ and $\hat {\vs}_1, \ldots ,\hat {\vs}_m$ be vectors in $\RR^d$. 

If for all basic analytic $\phi$, we have that 
$\sum_{i=1}^n \phi(\vs_i)=\sum_{i=1}^m \phi(\hat \vs_i) $
then
$\lms \vs_1, \ldots,\vs_n \rms=\lms \hat{\vs}_1,\ldots, \hat{\vs}_m \rms $.

\end{lemma}
For proof of this lemma, see Proposition 3.2 in \cite{Neural_Injective}.

It remains to prove Lemma \ref{eq:hardest}. We note that if \eqref{eq:big} holds for all basic analytic networks $\phi^{(c)}, c=0,\ldots, C$, this equality is also true if each $\phi^{(c)}$ is a linear combination of a finite number of shallow neural networks. Next, since we are only considering the values of $\phi^{(c)}$ on a fixed finite collection of inputs, and since linear combinations of shallow neural networks with analytic (or even continuous) non-polynomial functions can approximate any continuous function uniformly on a compact set (and in particular on a finite set) \cite{approximations}, it follows that \eqref{eq:hardest} holds for all \emph{continuous} $\phi^{(c)}$.

Next, we consider the norm multisets: 
$$S=\lms \|z_1\|, \ldots,\|z_n\| \rms, \quad \hat{S} =\|\hat{z}_1\|, \ldots,\|\hat{z}_m\| .$$
We claim that $S=\hat{S}$. We first show that $S\subseteq \hat S$. Fix  some $s$ in $\{1,\ldots,n\}$, we need to show  that $\|z_s\| \in \hat S$. We define a continuous function $f^s$ such that $f^s(\|z_s\|)=1$ and $f^s(\|z_i\|)=0=f^s(\|\hat z_j\|)  $ for all $i\neq s$, and for all $j=1,\ldots,m$ such that the norms $\|\hat z_j\|$ and $\|z_s\|$ are not equal. We claim that there exists $j$ for which the norms \emph{are} equal. Otherwise, we  choose the continuous functions $\phi^{(c)}, c=1,\ldots,C $ to be the zero functions, and choose $\phi^{(0)}$ to be $f^s$.  This will lead to the equation $z_s=0$  in \ref{eq:hardest}, which contradicts the assumption that $z_s\neq 0$, so we see that $S \subseteq \hat S $. A symmetric argument shows that  $\hat S \subseteq  S $ and so $S=\hat S$. We deduce that $n=m$ and $\|z_i\|=\| \hat z_i \| $ for all $i=1,\ldots,n$. 
We can now conclude the proof: fix an arbitrary $s$ in $\{1,\ldots,n\}$. Choose $\phi^{(0)}=f^s$ and take the other $\phi^{(c)}$ to be zero. Then \ref{eq:hardest} gives
$$z_s-\hat{z}_s=0 $$
Similarly, for an arbitrary index $q$ in $\{1,\ldots,C\}$, we can choose $\phi^{(q)}=f^s $  and $\phi^{(c)}=0$ for all $c\neq q$ in $\{0,1,\ldots,C\}$. In this case \ref{eq:hardest} gives
$$v_{jq}-\hat{v}_{jq}=0 .$$
This concludes the proof of the lemma and the proof of the theorem.
\end{proof}

\subsection{Complexity}
\label{complexity}
Running $\us$ for $T$ iterations with $C$ channels requires $O(C \cdot d \cdot deg_{i})$ elementary arithmetic operations for every node $i$, at every iteration $t$. Overall, this requires $O(C \cdot d \cdot N \cdot d_{avg} \cdot T)$ to run $\us$, where $d_{avg}$ denotes the average degree in the graph. This is linear in $N$, which dominates the other terms in the complexity bound. If we set $C=2Nd+1$ as required for maximal expressivity in Theorem \ref{thm:maximally}, then the complexity will be quadratic in $N$, which is less by a factor of $n$ than \citep{three_iterations,disgnn}, and by a factor of $n^{2}$ than \cite{gramnet}. 
 We also note that the cardinality in the finite witness theorem, the main tool for our proof, depends on the intrinsic dimension of the input and not the ambient dimension. Thus, under the standard 'manifold assumption' \cite{manifold}, which asserts that data in learning tasks typically resides in a manifold of low dimension $r$,  embedded in a high dimensional Euclidean space, the proof can be adapted to show that the number of channels necessary for maximal expressivity is only $2r+1$. Thus, under the manifold assumption, the complexity of $\us$ is linear in the number of nodes $N$. 
\section{Adding node and edge features}
\label{app:features}
This section explains how we incorporate node and edge features into our architecture and theory. 

A featured geometric graph is a $4$-tuple $\G =(\mA_{1},\mX_{1},\mF_{1},\mE_{1})$ where as before $\mA$ is an $n$ by $n$ adjacency matrix, and $\mX$ is a $n $ by $d$ matrix denoting node positions. $\mF$ is a $n$ by $d_n$ matrix denoting node features (without the action of rigid motions), and $\mE$ is a $n\times n \times d_e$ tensor of edge features. We denote the feature at a node $i$ and edge $(i,j)$ by $f_i$ (a column of $\mF$) and $f_{i,j}$, respectively.

We incorporate edge and node features into $\us$ by slightly modifying the aggregation step to
\begin{align*}
    v^{(t+1)}_{i,q}=\sum_{j\in \gN_{i}}\psi_{n}^{(t,q)}(f_{i})\cdot \psi_{e}^{(t,q)}(f_{i,j}) \cdot (\phi^{(t,q,0)}(\|\vx_{i}-\vx_{j}\|,\|\vv_{j}\|)(\vx_{i}-\vx_{j})+ \\
    \sum_{c=1}^{k} \phi^{(t,q,c)}(\|\vx_{i}-\vx_{j}\|,\|\vv_{j}\|)v_{jc}^{(t)})
\end{align*}

In our analysis we will take $\psi_{n}^{(t,q)}:\RR^{d_{n}} \to \RR$ and $\psi_{e}^{(t,q)}:\RR^{d_{e}} \to \RR$ to be simple linear functions 
$$\psi_{n}^{(t,q)}(f_{i})=a_{n}^{(t,q)} \cdot f_{i}+b_{n}^{(t,q)}, \quad \psi_{e}^{(t,q)}(f_{i,j})=a_{e}^{(t,q)} \cdot f_{i,j}+b_{q}^{(t,q)} $$
In practice, in our code, we use a slightly more complex function involving radial basis functions as in \cite{disgnn}.

To extend our analysis to the case of node and edge features, we first extend our definition of isomorphism of geometric graphs to include edge and node features.

\begin{definition}
    Given two featured geometric graphs 
    \begin{align*}
    \G =(\mA_{1},\mX_{1},\mF_{1},\mE_{1}), \quad  \mH = (\mA_{2},\mX_{2},\mF_{2},\mE_{2})
    \end{align*}
     $\G$ and $\H$ are \emph{featured geometrically isomorphic} if there exists a permutation matrix $\mP$, a rotation matrix $\mQ$, and a translation ${\vt}$ such that:
     \begin{enumerate}
         \item $\mA_{2}$ = $ \mP \cdot \mA_{1} \cdot \mP^{T}$.
         \item $\mX_{2} = \mP \cdot \mX_{1} \cdot \mQ + \vt$
         \item $\mF_{2} = \mP \cdot \mF_{1}$
         \item $\mE_{2}$ = $ \mP \cdot \mE_{1} \cdot \mP^{T}$
     \end{enumerate}
     where $ \mP \cdot \mE_{1} \cdot \mP^{T}$ should be understood as operating element-wise on the $n_e$ matrices of dimension $n \times n$ which compose the tensor $\mE $.
\end{definition}

We can then prove the following theorem: a generalization of \autoref{cor:us} to the featured graph case. 

\begin{theorem}
Let $N,d,d_e,d_n$ be natural numbers. Let $F_\theta$ denote the $\us$ architecture for featured geometric graphs, with depth $T=d+1$ and $C=2Nd+d_{e}+d_{n}+1$ channels. Then for Lebesgue almost every $\theta$, we have for all featured graphs $\G, \G'\in \GdN $ which are generic and connected,
$$F_\theta(\G)=F_\theta(\hat{\G}) \iff \G, \hat{\G} \text{ are featured geometrically isomorphic}$$
\end{theorem}
We note that the assumption that the geometric graph is generic is only on the position coordinates $\mX$. The statement will hold for all node and edge features in this case. 
\begin{proof}
We explain how to modify the proof of Theorem \ref{thm:maximally} to obtain this result. First, we note that the number of channels was adapted to take into account the higher dimensionality of the data now that node and edge features are included. The finite witness theorem can be applied with this adapted dimension, as explained in Theorem \ref{thm:maximally}. Thus, we only need to show that, for given featured graphs $\hat{\G},\G $ satisfying the conditions of the theorem, if for all $\theta$ we have that $F_\theta(\G)=F_\theta(\hat \G) $, then $\G $ and $\hat \G$ are featured geometrically isomorphic.  

Now, let us assume that, for all $\theta$ we have  $F_\theta(\G)=F_\theta(\hat \G)$. We can obtain all the results we had for the case where there were no additional features by choosing the parameters 
$$a_n^{(t,q)}=0, b_n^{(t,q)}=1, a_e^{(t,q)}=0, b_e^{(t,q)}=1   $$
so that $\psi_n^{(t,q)}=1=\psi_e^{(t,q)}$, and the effect of node and edge features on the architectures is 'cancelled out'. In particular, we can deduce that, after an appropriate permutation and rigid motion are applied to the data, the graphs have the same number of nodes $n$, and for all node $i$ and iteration $t$, we have
$$\vv_i^{(t)}=\hat {\vv}_i^{(t)}, \vx_i=\hat{\vx}_i $$
Now,  by setting $\phi^{(t,q,c)}=0$ for $c\geq 1$, and considering  $\phi^{(t,q,0)}$ which depend only on the first coordinate, we can deduce that 
\begin{align*}
    \sum_{j\in \gN_i} \left[(\psi_n^{(t,q)}(f_i)\cdot \psi_e^{(t,q)}(f_{i,j})-\psi_n^{(t,q)}(\hat{f}_i)\cdot \psi_e^{(t,q)}(\hat{f}_{i,j})\right]\cdot \left( \phi^{(t,q,0)}(\|\vx_i-\vx_j\|)(\vx_i-\vx_j) \right)=0
\end{align*}
for all choices of parameters of the functions $\psi_n^{(t,q)},\psi_e^{(t,q)}$ and $\phi^{(t,q,0)}$. As we showed in the proof of \autoref{thm:maximally}, this equation implies that this equation will also hold for any continuous $\phi^{(t,q,0)}$ (and any choice of parameters of the other functions) since the span of all such basic analytic functions is dense. In particular, for fixed $i,j$, we can choose $\phi^{(t,q,0)}$ to be a continuous function with $\phi^{(t,q,0)}(\|\vx_i-\vx_j\|)=1$ and $\phi^{(t,q,0)}(\|\vx_i-\vx_k\|)=0$ if $k \neq j$. Inserting this into the last equality we obtain 
$$\left[\psi_n^{(t,q)}(f_i)\cdot \psi_e^{(t,q)}(f_{i,j})-\psi_n^{(t,q)}(\hat{f}_i)\cdot \psi_e^{(t,q)}(\hat{f}_{i,j})\right] \cdot  (\vx_i-\vx_j) =0$$
by genericity $\vx_i-\vx_j \neq 0$, and so we deduce that 
$$ \psi_n^{(t,q)}(f_i)\cdot \psi_e^{(t,q)}(f_{i,j})-\psi_n^{(t,q)}(\hat{f}_i)\cdot \psi_e^{(t,q)}(\hat{f}_{i,j})=0$$
for every choice of function parameters. Setting 
$a_n^{(t,q)}=0, b_n^{(t,q)}=1, b_e^{(t,q)}=0   $
we obtain that 
$$a_e \cdot (f_{ij}-\hat{f}_{ij})=0, \quad \forall a_e \in \RR^{d_e}$$
and therefore $f_{ij}-\hat{f}_{ij}=0 $. Similarly, setting $a_e^{(t,q)}=0, b_e^{(t,q)}=1, b_n^{(t,q)}=0   $ we obtain that 
$$a_n \cdot (f_{i}-\hat{f}_{i})=0, \quad \forall a_n \in \RR^{d_n}$$
and therefore $f_{i}-\hat{f}_{i}=0 $. This concludes the proof of the theorem.
\end{proof}

\section{Additional Empirical Results}
\paragraph{k-chain results}
Here, we show the results of the $k$-chain experiment for $k=4$. As we can see, only our method $\us$  and GVP \cite{gvp-egnn} succeed using the minimal number of blocks. This experiment illustrates our model's ability to distinguish very sparse graphs. Next, we show results for $k=12$. As can be seen, we are the only ones succeeding with the minimal number of blocks, and other models need a higher number of blocks to succeed in distinguishing.
\begin{table*}[t!]
     \caption{Separation Learning of $4$-chain paths \autoref{fig:Two pairs for power Graphs}. We show the learned accuracy over ten trials for each model, depending on the number of blocks. We show the minimal needed number of blocks (3) is sufficient for separation, and only \cite{gvp-egnn} also succeeded (among equivariant models working on the sparse graph) all $10$ trials using $3$ blocks.}
    \begin{subtable}
        \centering
        \resizebox{\linewidth}{!}{
        \begin{tabular}{clccccc}
            \toprule
            & ($k=\mathbf{4}$-chains) & \multicolumn{5}{c}{\textbf{Number of layers}} \\
            & \textbf{GNN Layer} & $\lfloor \frac{k}{2} \rfloor$ & \cellcolor{gray!10} $\lfloor \frac{k}{2} \rfloor + 1 = \mathbf{3}$ & $\lfloor \frac{k}{2} \rfloor + 2$ & $\lfloor \frac{k}{2} \rfloor + 3$ & $\lfloor \frac{k}{2} \rfloor + 4$ \\
            \midrule
            \multirow{5}{*}{\rotatebox[origin=c]{90}{Equiv.}} &
            \gray{GWL} & \gray{50\%} & \gray{\textbf{100\%}} & \gray{\textbf{100\%}} & \gray{\textbf{100\%}} & \gray{\textbf{100\%}} \\
            & E-GNN \cite{egnn} & 50.0 {\scriptsize ± 0.0} & \cellcolor{red!10} 50.0 {\scriptsize ± 0.0} & \cellcolor{red!10} 50.0 {\scriptsize ± 0.0} & \cellcolor{red!10} 50.0 {\scriptsize ± 0.0} & \cellcolor{green!10} 100.0 {\scriptsize ± 0.0} \\
            & GVP-GNN \cite{gvp-egnn} & 50.0 {\scriptsize ± 0.0} & \cellcolor{green!10} 100.0 {\scriptsize ± 0.0} & \cellcolor{green!10} 100.0 {\scriptsize ± 0.0} & \cellcolor{green!10} 100.0 {\scriptsize ± 0.0} & \cellcolor{green!10} 100.0 {\scriptsize ± 0.0} \\
            & TFN \cite{TFN} & 50.0 {\scriptsize ± 0.0} & \cellcolor{red!10} 50.0 {\scriptsize ± 0.0} & \cellcolor{red!10} 50.0 {\scriptsize ± 0.0} & \cellcolor{blue!10} 80.0 {\scriptsize ± 24.5} & \cellcolor{blue!10} 85.0 {\scriptsize ± 22.9} \\
            & MACE \cite{mace} & 50.0 {\scriptsize ± 0.0} & \cellcolor{blue!10} 90.0 {\scriptsize ± 20.0} & \cellcolor{blue!10} 90.0 {\scriptsize ± 20.0} & \cellcolor{blue!10} 95.0 {\scriptsize ± 15.0} & \cellcolor{blue!10} 95.0 {\scriptsize ± 15.0} \\
             & $\us$(ours) & 50.0 {\scriptsize ± 0.0} & \cellcolor{green!10} 100.0 {\scriptsize ± 0.0} & \cellcolor{green!10} 100.0 {\scriptsize ± 0.0} & \cellcolor{green!10} 100.0 {\scriptsize ± 0.0} & \cellcolor{green!10} 100.0 {\scriptsize ± 0.0} \\
            \midrule
            \multirow{4}{*}{\rotatebox[origin=c]{90}{Inv.}} &
            \gray{IGWL} & \gray{50\%} & \gray{50\%} & \gray{50\%} & \gray{50\%} & \gray{50\%} \\
            & SchNet \cite{SchNet} & 50.0 {\scriptsize {\scriptsize ± 0.0}} & 50.0 {\scriptsize {\scriptsize ± 0.0}} & 50.0 {\scriptsize {\scriptsize ± 0.0}} & 50.0 {\scriptsize {\scriptsize ± 0.0}} & 50.0 {\scriptsize {\scriptsize ± 0.0}} \\
            & DimeNet \cite{DimeNet} & 50.0 {\scriptsize {\scriptsize ± 0.0}} & 50.0 {\scriptsize {\scriptsize ± 0.0}} & 50.0 {\scriptsize {\scriptsize ± 0.0}} & 50.0 {\scriptsize {\scriptsize ± 0.0}} & 50.0 {\scriptsize {\scriptsize ± 0.0}} \\
            & SphereNet \cite{SphereNet} & 50.0 {\scriptsize {\scriptsize ± 0.0}} & 50.0 {\scriptsize {\scriptsize ± 0.0}} & 50.0 {\scriptsize {\scriptsize ± 0.0}} & 50.0 {\scriptsize {\scriptsize ± 0.0}} & 50.0 {\scriptsize {\scriptsize ± 0.0}} \\
            \cmidrule{2-7}
            & SchNet $_{\text{full graph}} \footref{schnet models}$ \cite{SchNet} & \cellcolor{green!10} 100.0 {\scriptsize ± 0.0} & \cellcolor{green!10} 100.0 {\scriptsize ± 0.0} & \cellcolor{green!10} 100.0 {\scriptsize ± 0.0} & \cellcolor{green!10} 100.0 {\scriptsize ± 0.0} & \cellcolor{green!10} 100.0 {\scriptsize ± 0.0} \\
            & SchNet$_{\text{global feat}} \footref{schnet models}$ \cite{SchNet} & \cellcolor{green!10} 100.0 {\scriptsize ± 0.0} & \cellcolor{green!10} 100.0 {\scriptsize ± 0.0} & \cellcolor{green!10} 100.0 {\scriptsize ± 0.0} & \cellcolor{green!10} 100.0 {\scriptsize ± 0.0} & \cellcolor{green!10} 100.0 {\scriptsize ± 0.0} \\
            \bottomrule
        \end{tabular}
        }
    \end{subtable}
    \hfill
    \label{tab:4chains}
\end{table*}

\begin{table*}[b!]
    \hfill
      \caption{Separation Learning of $12$-chain paths \autoref{fig:Two pairs for power Graphs}. We show the learned accuracy over ten trials for each model depending on the number of blocks. As we can see, we are the only ones that succeed with the minimal number of blocks ($7$), and other models need many more layers to show high Performance.}
    \begin{subtable}
        \centering
        \resizebox{\linewidth}{!}{
        \begin{tabular}{clcccccc}
            \toprule
            & ($k=\mathbf{12}$-chains) & \multicolumn{5}{c}{\textbf{Number of layers}} \\
            & \textbf{GNN Layer} & $ \lfloor \frac{k}{2} \rfloor $ & \cellcolor{gray!10} $\lfloor \frac{k}{2} \rfloor +1$ & $ \lfloor \frac{k}{2} \rfloor 2$ & $ \lfloor \frac{k}{2} \rfloor +3$ & $ \lfloor \frac{k}{2} \rfloor +4$\\
            \midrule
            \multirow{5}{*}{\rotatebox[origin=c]{90}{Equiv.}} &
            \gray{Maximally expressive E-GGNN} & \gray{50\%} & \gray{\textbf{100\%}} & \gray{100\%} & \gray{100\%} & \gray{\textbf{100\%}} &  \\
            & E-GNN \cite{egnn} & 50.0 {\scriptsize ± 0.0} & \cellcolor{red!10} 50.0 {\scriptsize ± 0.0} & \cellcolor{red!10} 50.0 {\scriptsize ± 0.0} &  
            \cellcolor{red!10} 50.0 {\scriptsize ± 0.0} &
            \cellcolor{blue!10} 95 {\scriptsize ± 15.0} &    \\
            & GVP-GNN \cite{gvp-egnn}  & 50.0 {\scriptsize ± 0.0} & \cellcolor{red!10} 50.0 {\scriptsize ± 0.0} & \cellcolor{blue!10} 60.0 {\scriptsize ± 20.0} & \cellcolor{red!10} 50.0 {\scriptsize ± 0.0} & \cellcolor{blue!10} 60.0 {\scriptsize ± 20.0} & \\
            & TFN \cite{TFN} & 50.0 {\scriptsize ± 0.0} & \cellcolor{red!10} 50.0 {\scriptsize ± 0.0} & \cellcolor{red!10} 50.0 {\scriptsize ± 0.0} & \cellcolor{red!10} 50.0 {\scriptsize ± 0.0} & \cellcolor{red!10} 50.0 {\scriptsize ± 0.0} & \\
            & MACE \cite{mace} & 50.0 {\scriptsize ± 0.0} & \cellcolor{red!10} 50.0 {\scriptsize ± 0.0} & \cellcolor{red!10} 50.0 {\scriptsize ± 0.0} & \cellcolor{red!10} 50.0 {\scriptsize ± 0.0} & \cellcolor{red!10} 50.0 {\scriptsize ± 0.0} & \\
             & $\us$(ours) & 50.0 {\scriptsize ± 0.0} & \cellcolor{green!10} 100.0 {\scriptsize ± 0.0} & \cellcolor{green!10} 100.0 {\scriptsize ± 0.0} & \cellcolor{green!10} 100.0 {\scriptsize ± 0.0} & \cellcolor{green!10} 100.0 {\scriptsize ± 0.0}& \\
            \midrule
            \multirow{4}{*}{\rotatebox[origin=c]{90}{Inv.}} &
            \gray{Maximally expressive I-GGNN} & \gray{50\%} & \gray{50\%} & \gray{50\%} & \gray{50\%} & \gray{50\%} & \\
            & SchNet \cite{SchNet} & 50.0 {\scriptsize {\scriptsize ± 0.0}} & 50.0 {\scriptsize {\scriptsize ± 0.0}} & 50.0 {\scriptsize {\scriptsize ± 0.0}} & 50.0 {\scriptsize {\scriptsize ± 0.0}} & 50.0 {\scriptsize {\scriptsize ± 0.0}} \\
            & DimeNet \cite{DimeNet} & 50.0 {\scriptsize {\scriptsize ± 0.0}} & 50.0 {\scriptsize {\scriptsize ± 0.0}} & 50.0 {\scriptsize {\scriptsize ± 0.0}} & 50.0 {\scriptsize {\scriptsize ± 0.0}} & 50.0 {\scriptsize {\scriptsize ± 0.0}}& \\
            & SphereNet \cite{SphereNet} & 50.0 {\scriptsize {\scriptsize ± 0.0}} & 50.0 {\scriptsize {\scriptsize ± 0.0}} & 50.0 {\scriptsize {\scriptsize ± 0.0}} & 50.0 {\scriptsize {\scriptsize ± 0.0}} & 50.0 {\scriptsize {\scriptsize ± 0.0}} \\
            \cmidrule{2-7}
            & $\text{SchNet}_{\text{full graph}}$\footref{schnet models} \cite{SchNet} & \cellcolor{green!10} 100.0 {\scriptsize ± 0.0} & \cellcolor{green!10} 100.0 {\scriptsize ± 0.0} & \cellcolor{green!10} 100.0 {\scriptsize ± 0.0} & \cellcolor{green!10} 100.0 {\scriptsize ± 0.0} & \cellcolor{green!10} 100.0 {\scriptsize ± 0.0}&  \\
            & $\text{SchNet}_{\text{global feat}}$\footref{schnet models} \cite{SchNet} & \cellcolor{green!10} 100.0 {\scriptsize ± 0.0} & \cellcolor{green!10} 100.0 {\scriptsize ± 0.0} & \cellcolor{green!10} 100.0 {\scriptsize ± 0.0} & \cellcolor{green!10} 100.0 {\scriptsize ± 0.0} & \cellcolor{green!10} 100.0 {\scriptsize ± 0.0}& \\
            \bottomrule
        \end{tabular}
        }
    \end{subtable}
    \hfill
    \label{tab:12chains}
\end{table*}

\paragraph{Hard examples}
This section presents the results mentioned in Section \ref{sepearion exps}. We use a dataset of four difficult-to-separate pairs of point clouds. Three (Hard1,Hard2,Hard3) are taken from
\cite{PhysRevLett.125.166001}. The last pair, 'Harder,' is taken from \cite{gcnn} and is an example of a non-isomorphic pair that I-GGNN models cannot distinguish. We show we succeeded in separating all tuples with a probability of $1$.
The tested models are $\us$, GramNet \cite{gramnet}, GeoEGNN \cite{gramnet}, EGNN \cite{egnn}, LinearEGNN, MACE \cite{mace}, TFN \cite{TFN}, DimeNet \cite{DimeNet}, GVPGNN \cite{gvp-egnn} and are taken from \cite{gramnet}. 


\begin{table*}[t]
    \centering
    \caption{
        Accuracy on challenging point-cloud benchmarks
        \cite{gcnn,PhysRevLett.125.166001}.
        Higher values are better; \(0.5\) corresponds to chance-level performance.
    }
    \label{tab:hard_pc_results}

    \setlength{\tabcolsep}{12pt}
    \renewcommand{\arraystretch}{1.08}

    \begin{tabular}{@{}lcccc@{}}
        \toprule
        \textbf{Model}
        & \textbf{Hard1}
        & \textbf{Hard2}
        & \textbf{Hard3}
        & \textbf{Harder} \\
        \midrule

        GramNet
        & \textbf{1.000}
        & \textbf{1.000}
        & \textbf{1.000}
        & \textbf{1.000} \\

        GeoEGNN
        & 0.998
        & 0.970
        & 0.850
        & 0.899 \\

        EGNN
        & 0.500
        & 0.500
        & 0.500
        & 0.500 \\

        LinearEGNN
        & \textbf{1.000}
        & \textbf{1.000}
        & \textbf{1.000}
        & 0.500 \\

        MACE
        & \textbf{1.000}
        & \textbf{1.000}
        & \textbf{1.000}
        & \textbf{1.000} \\

        TFN
        & 0.500
        & 0.500
        & 0.550
        & 0.500 \\

        DimeNet
        & \textbf{1.000}
        & \textbf{1.000}
        & \textbf{1.000}
        & \textbf{1.000} \\

        GVPGNN
        & \textbf{1.000}
        & \textbf{1.000}
        & \textbf{1.000}
        & \textbf{1.000} \\

        \midrule
        \rowcolor{gray!10}
        \textbf{\(\us\) (Ours)}
        & \textbf{1.000}
        & \textbf{1.000}
        & \textbf{1.000}
        & \textbf{1.000} \\

        \bottomrule
    \end{tabular}
\end{table*}

\paragraph{Chemical property experiments with full statistics}
the table for the chemical property prediction tasks in the main text, Table \ref{mean}, was produced using the protocol from \cite{baselines}. In particular, each method was run with three different seeds. The seed that did best on the validation set was chosen, and its results on the test set were reported.

Table \ref{mean} presents each task's mean and standard deviation over $3$ seeds. As we can see, in most tasks, the standard deviation is rather low, and the results are qualitatively similar to those obtained in Table \ref{cpe}. In particular, in the tasks in the Kraken and BDE datasets, we outperform the competitors, often by a large margin. In the Drugs 75K, we obtain comparable, but somewhat higher, results than the best method (while in Table \ref{cpe}, in two of three Drugs-75K  cases, our results were the best by a small margin).

\renewcommand{\arraystretch}{2.2} 
\begin{table}[t]
    \centering
      \caption{ Raw performance data (mean ± standard deviation) of 1D, 2D, 3D, and conformer ensemble MRL models all taken from \cite{baselines} in terms of absolute test error. Our model is in the 3D category, named $\us$.}
    \resizebox{\textwidth}{!}{%
    \begin{tabular}{llccccccccc}
        \toprule
        \rowcolor{gray!25} Category & Model & IP & EA & $\chi$ & B\textsubscript{5} & L & BurB\textsubscript{5} & BurL & BDE \\
        \midrule
        \rowcolor{gray!10} 1D & Random Forest & 0.4987±0.0037 & 0.4747±0.0022 & 0.2732±0.0031 & 0.4760±0.0041 & 0.4303±0.0090 & 0.2758±0.0180 & 0.1521±0.0149 & 3.03±0.27 \\
        \rowcolor{gray!10} 1D & LSTM & 0.4788±0.0024 & 0.4648±0.0002 & 0.2505±0.0050 & 0.4879±0.0280 & 0.5142±0.0411 & 0.2813±0.0041 & 0.1924±0.0028 & 2.82±0.07 \\
        \rowcolor{gray!10} 1D & Transformer & 0.6617±0.0023 & 0.5850±0.0031 & 0.4073±0.0006 & 0.9611±0.0813 & 0.8389±0.0431 & 0.4929±0.0369 & 0.2781±0.0207 & 10.08±0.64 \\
        \midrule
        \rowcolor{snsblue!15} 2D & GIN & 0.4354±0.0029 & 0.4169±0.0032 & 0.2260±0.0017 & 0.3128±0.0264 & 0.4003±0.0341 & 0.1719±0.0031 & 0.1200±0.0040 & 2.63±0.22 \\
        \rowcolor{snsblue!15} 2D & GIN-VN & 0.4361±0.0059 & 0.4169±0.0083 & 0.2267±0.0002 & 0.3567±0.0031 & 0.4344±0.0416 & 0.2422±0.0033 & 0.1741±0.0109 & 2.74±0.24 \\
        \rowcolor{snsblue!15} 2D & ChemProp & 0.4595±0.0028 & 0.4417±0.0045 & 0.2441±0.0012 & 0.4850±0.0068 & 0.5452±0.0454 & 0.3002±0.0086 & 0.1948±0.0138 & 2.66±0.14 \\
        \rowcolor{snsblue!15} 2D & GraphGPS & 0.4351±0.0049 & 0.4085±0.0055 & 0.2212±0.0054 & 0.3450±0.0324 & 0.4363±0.0133 & 0.2066±0.0115 & 0.1500±0.0138 & 2.48±0.19 \\
        \midrule
        \rowcolor{snsorange!15} 3D & SchNet & 0.4394±0.0062 & 0.4207±0.0021 & 0.2243±0.0089 & 0.3293±0.0068 & 0.5458±0.0341 & 0.2295±0.0111 & 0.1861±0.0095 & 2.54±0.00 \\
        \rowcolor{snsorange!15} 3D & DimeNet++ & 0.4441±0.0087 & 0.4233±0.0072 & 0.2436±0.0075 & 0.3510±0.0107 & 0.4174±0.0397 & 0.2097±0.0160 & 0.1526±0.0072 & \underline{1.45±0.03} \\
        \rowcolor{snsorange!15} 3D & GemNet & 0.4069±0.0007 & \textbf{0.3922±0.0024} & \textbf{0.1970±0.0039} & 0.2789±0.0125 & 0.3754±0.0086 & 0.1782±0.0099 & 0.1635±0.0063 & 1.65±0.30 \\
        \rowcolor{snsorange!15} 3D & PaiNN & 0.4505±0.0041 & 0.4495±0.0054 & 0.2324±0.0040 & 0.3443±0.0388 & 0.4471±0.0324 & 0.2395±0.0176 & 0.1673±0.0088 & 2.12±0.09 \\
        \rowcolor{snsorange!15} 3D & ClofNet & 0.4393±0.0084 & 0.4251±0.0066 & 0.2378±0.0020 & 0.4873±0.0093 & 0.6417±0.0362 & 0.2884±0.0166 & 0.2529±0.0052 & 2.60±0.02 \\
        \rowcolor{snsorange!15} 3D & LEFTNet & 0.4174±0.0007 & 0.3964±0.0009 & 0.2083±0.0054 & 0.3072±0.0012 & 0.4493±0.0261 & 0.2176±0.0010 & 0.1486±0.0095 & 1.53±0.05 \\
        \rowcolor{snsorange!15} 3D & $\us$(Ours) & 0.4251±0.0463 & \underline{0.3935±0.0361} & 0.2332±0.0248 & \textbf{0.1953±0.0068} & \textbf{0.1956±0.0264} & \textbf{0.1545±0.0019} & \textbf{0.0644±0.0017} & \textbf{0.4455±0.1305} \\
        \midrule
        \rowcolor{snsgreen!15} 3D + Sampling & SchNet & 0.4452±0.0080 & 0.4232±0.0042 & 0.2243±0.0022 & 0.3235±0.0147 & 0.4598±0.0041 & 0.2086±0.0111 & 0.1739±0.0142 & 1.97±0.01 \\
        \rowcolor{snsgreen!15} 3D + Sampling & DimeNet++ & 0.4395±0.0032 & 0.4217±0.0040 & 0.2432±0.0048 & 0.3323±0.0320 & 0.4153±0.0208 & 0.2237±0.0122 & 0.1561±0.0241 & 1.47±0.03 \\
        \rowcolor{snsgreen!15} 3D + Sampling & GemNet & \textbf{0.4066±0.0015} & 0.3910±0.0004 & \underline{0.2027±0.0013} & 0.2694±0.0221 & 0.3488±0.0252 & 0.1796±0.0098 & 0.1184±0.0033 & 1.60±0.10 \\
        \rowcolor{snsgreen!15} 3D + Sampling & PaiNN & 0.4466±0.0087 & 0.4393±0.0045 & 0.2331±0.0037 & 0.3441±0.0161 & 0.4358±0.0343 & 0.2476±0.0070 & 0.1543±0.0022 & 1.92±0.01 \\
        \rowcolor{snsgreen!15} 3D + Sampling & ClofNet & 0.4430±0.0074 & 0.4237±0.0005 & 0.2335±0.0090 & 0.4524±0.0935 & 0.5962±0.0074 & 0.2442±0.0109 & 0.1756±0.0112 & 2.51±0.23 \\
        \rowcolor{snsgreen!15} 3D + Sampling & LEFTNet & \underline{0.4149±0.0019} & 0.3988±0.0048 & 0.2141±0.0084 & 0.2834±0.0068 & 0.4407±0.0531 & 0.2120±0.0097 & 0.1547±0.0101 & 1.52±0.00 \\
        \midrule
        \rowcolor{snspurple!10} SchNet & Mean & 0.4583±0.0019 & 0.4410±0.0018 & 0.2371±0.0098 & 0.3075±0.0151 & 0.4691±0.0234 & 0.2282±0.0206 & 0.1619±0.0062 & 2.53±0.02 \\
        \rowcolor{snspurple!10}        & DeepSet & 0.4537±0.0065 & 0.4396±0.0010 & 0.2385±0.0066 & 0.3105±0.0381 & 0.4322±0.0464 & 0.2249±0.0234 & 0.1535±0.0076 & 2.29±0.22 \\
        \rowcolor{snspurple!10}        & Attention & 0.4556±0.0075 & 0.4382±0.0125 & 0.2380±0.0007 & 0.2704±0.0187 & 0.4517±0.0132 & 0.2024±0.0183 & 0.1443±0.0043 & 2.64±0.00 \\
        \midrule
        \rowcolor{snspurple!10} DimeNet++ & Mean & 0.4488±0.0086 & 0.4340±0.0079 & 0.2425±0.0060 & 0.2630±0.0122 & 0.3828±0.0331 & 0.1960±0.0059 & 0.1268±0.0060 & 1.79±0.12 \\
        \rowcolor{snspurple!10}           & DeepSet & 0.4126±0.0076 & 0.3944±0.0034 & 0.2267±0.0047 & 0.2889±0.0069 & 0.3468±0.0090 & 0.1783±0.0110 & 0.1339±0.0087 & 1.75±0.01 \\
        \rowcolor{snspurple!10}           & Attention & 0.4188±0.0024 & 0.4030±0.0075 & 0.2325±0.0028 & 0.3718±0.0300 & 0.3628±0.0259 & 0.1899±0.0081 & 0.1185±0.0105 & 2.57±0.21 \\
        \midrule
        \rowcolor{snspurple!10} GemNet & Mean & 0.4505±0.0052 & 0.4334±0.0023 & 0.2289±0.0032 & 0.2635±0.0053 & 0.3753±0.0036 & 0.1671±0.0154 & 0.1587±0.0029 & 2.19±0.06 \\
        \rowcolor{snspurple!10}        & DeepSet & 0.4187±0.0022 & 0.4002±0.0012 & 0.2169±0.0036 & 0.2313±0.0026 & \underline{ 0.3386±0.0269} & \underline{ 0.1589±0.0068} & \underline{ 0.0947±0.0012} & 2.25±0.21 \\
        \rowcolor{snspurple!10}        & Attention & 0.4212±0.0017 & 0.4221±0.0097 & 0.2260±0.0056 & 0.2670±0.0026 & 0.3554±0.0147 & 0.1769±0.0153 & 0.1346±0.0075 & 2.68±0.02 \\
        \midrule
        \rowcolor{snspurple!10} PaiNN & Mean & 0.4591±0.0024 & 0.4425±0.0064 & 0.2360±0.0032 & 0.2877±0.0252 & 0.3950±0.0233 & 0.1817±0.0091 & 0.1472±0.0039 & 1.87±0.16 \\
        \rowcolor{snspurple!10}       & DeepSet & 0.4471±0.0071 & 0.4269±0.0033 & 0.2294±0.0065 & \underline{0.2225±0.0218} & 0.3619±0.0192 & 0.1693±0.0111 & 0.1324±0.0091 & 2.20±0.05 \\
        \rowcolor{snspurple!10}       & Attention & 0.4641±0.0016 & 0.4567±0.0094 & 0.2471±0.0049 & 0.3496±0.0140 & 0.4109±0.0167 & 0.2123±0.0005 & 0.1506±0.0029 & 2.23±0.12 \\
        \midrule
        \rowcolor{snspurple!10} ClofNet & Mean & 0.4536±0.0030 & 0.4301±0.0007 & 0.2365±0.0075 & 0.3555±0.0193 & 0.4485±0.0053 & 0.2473±0.0076 & 0.2022±0.0212 & 2.01±0.08 \\
        \rowcolor{snspurple!10}          & DeepSet & 0.4280±0.0056 & 0.4033±0.0024 & 0.2199±0.0073 & 0.3228±0.0020 & 0.4742±0.0161 & 0.2263±0.0249 & 0.1548±0.0039 & 2.35±0.04 \\
        \rowcolor{snspurple!10}          & Attention & 0.4330±0.0071 & 0.4107±0.0048 & 0.2220±0.0084 & 0.3734±0.0267 & 0.4963±0.0286 & 0.2178±0.0186 & 0.1690±0.0281 & 2.66±0.14 \\
        \midrule
        \rowcolor{snspurple!10} LEFTNet & Mean & 0.4402±0.0062 & 0.4267±0.0026 & 0.2183±0.0007 & 0.2949±0.0001 & 0.3643±0.0352 & 0.2098±0.0146 & 0.1386±0.0007 & 2.04±0.00 \\
        \rowcolor{snspurple!10}         & DeepSet & 0.4167±0.0043 & 0.3953±0.0000 & 0.2069±0.0022 & 0.2644±0.0130 & 0.3866±0.0270 & 0.2023±0.0026 & 0.1441±0.0042 & 2.51±0.30 \\
        \rowcolor{snspurple!10}         & Attention & 0.4229±0.0059 & 0.4067±0.0047 & 0.2198±0.0011 & 0.3161±0.0116 & 0.4324±0.0292 & 0.2017±0.0023 & 0.1508±0.0075 & 2.63±0.15 \\
        \midrule
        \bottomrule
    \end{tabular}%
    }

  \label{mean}
\end{table}
\paragraph{Time measurements}
To compare our running time with other models, we measured the execution time for each model on the 12-chain task with all models using 128 hidden feature size. The results \ref{tab:time_com} show that our running time is on par with the competitors.

\paragraph{Norm Accuracies}
To heuristically estimate  ‘how generic’ the molecules in the chemical property datasets \citep{Drugs,Kraken,BDE} we considered are,  we computed all norms of atoms in each (centralized) molecule and counted the number of unique norms relative to the total number of points. We considered a norm unique if its distance from all other norms exceeded a specified threshold $\mu$, and considered various threshold values. The results \ref{tab:norms_repeat} indicate that the norms exhibit considerable variation, suggesting that the molecules are close to being generic. This supports the conjecture stated in the main text that the surprising success of our simple method, compared to more expressive methods, is related to the near-genericity of the molecules in these datasets.


\begin{table}[t]
    \centering
    \caption{
        Runtime comparison on the 12-chain problem.
        Lower values indicate faster execution.
    }
    \label{tab:time_com}

    \setlength{\tabcolsep}{12pt}
    \renewcommand{\arraystretch}{1.05}

    \begin{tabular}{@{}l S[table-format=3.1]@{}}
        \toprule
        \textbf{Method} & {\textbf{Time (s)}} \\
        \midrule
        SchNet \cite{SchNet}             &  53.3 \\
        DimeNet \cite{DimeNet}           & 222.6 \\
        SphereNet \cite{SphereNet}       & 360.9 \\
        EGNN \cite{egnn}                 &  98.8 \\
        GVP \cite{gvp-egnn}              & 222.4 \\
        TFN \cite{TFN}                   & 268.1 \\
        MACE \cite{mace}                 & 440.6 \\
        \addlinespace[2pt]
        \midrule
        \textbf{\(\us\) (Ours)}          & \bfseries 112.0 \\
        \bottomrule
    \end{tabular}
\end{table}

\begin{table}[t]
    \centering
    \caption{
        Fraction of repeated norms at different thresholds across datasets.
    }
    \label{tab:norms_repeat}

    \setlength{\tabcolsep}{10pt}
    \renewcommand{\arraystretch}{1.08}

    \begin{tabular}{
        @{}
        S[table-format=1.0e-1]
        S[table-format=1.4]
        S[table-format=1.4]
        S[table-format=1.4]
        @{}
    }
        \toprule
        {\textbf{Threshold}}
        & {\textbf{Drugs}}
        & {\textbf{Kraken}}
        & {\textbf{BDE}} \\
        \midrule
        0       & 1.0000 & 1.0000 & 1.0000 \\
        1e-4    & 0.9967 & 0.9804 & 0.9950 \\
        1e-3    & 0.9678 & 0.9139 & 0.9573 \\
        1e-2    & 0.7230 & 0.5217 & 0.6698 \\
        \bottomrule
    \end{tabular}
\end{table}

\addtocontents{toc}{\protect\setcounter{tocdepth}{1}}
\printbibliography[
    heading=subbibintoc,
    title={References}
]

\endgroup

\end{refsection}

\chapter{Revisiting Multi-Permutation Equivariance through the Lens of Irreducible Representations}
\label{chap:paper2}

\noindent
\textbf{Yonatan Sverdlov, Ido Springer, and Nadav Dym.}\\
International Conference on Learning Representations (ICLR), 2025.\\
Yonatan Sverdlov and Ido Springer contributed equally.

\bigskip


\begin{refsection}[papers/paper2/iclr2025_conference.bib]

\begingroup


\def\RR{\mathbb{R}}
\def\CC{\mathbb{C}}

\def\X{\textbf{X}}
\def\G{\mathcal{G}}
\def\C{\mathbf{C}}
\def\B{\mathcal{B}}
\def\W{\mathcal{W}}
\def\V{\mathcal{V}}
\def\R{\mathcal{R}}
\def\U{\mathcal{U}}
\def\K{\mathcal{K}}
\def\H{\mathcal{H}}
\def\S{\mathcal{S}}

\def\p{\textbf{p}}

\def\Scalar{\mathbf{S}}
\def\Vector{\mathbf{V}}
\def\Matrix{\mathbf{M}}

\def\Diag{\operatorname{Diag}}
\def\GL{\operatorname{GL}}

\def\eqref#1{equation~\ref{#1}}

\def\ys#1{%
  \textcolor{magenta}{[Yon:] \textit{#1}: }%
}

\let\citep\parencite
\let\citet\textcite

\begin{abstract}
This paper explores the characterization of equivariant linear layers for representations of permutations and related groups.
Unlike traditional approaches, which address these problems using parameter-sharing, we consider an alternative methodology based on irreducible representations and Schur's lemma. Using this methodology, we obtain an alternative derivation for existing models like DeepSets, 2-IGN graph equivariant networks, and Deep Weight Space (DWS) networks. The derivation for DWS networks is significantly simpler than that of previous results.

Next, we extend our approach to unaligned symmetric sets, where equivariance to the wreath product of groups is required. Previous works have addressed this problem in a rather restrictive setting, in which almost all wreath equivariant layers are Siamese. In contrast, we give a full characterization of layers in this case and show that there is a vast number of additional non-Siamese layers in some settings. We also show empirically that these additional non-Siamese layers can improve performance in tasks like graph anomaly detection, weight space alignment, and learning Wasserstein distances.
Our code is available at \href{https://github.com/yonatansverdlov/SchurNet}{GitHub}.
\end{abstract}
\section{Introduction}

Learning with symmetries has recently attracted great attention in machine learning. In this learning setting, a group acts on an input space, and the hypothesis mappings are restricted to be equivariant with respect to the group action.
Motivated by the structure of fully connected neural networks, group equivariant models are often defined by a composition of parametric linear equivariant functions and non-parametric non-linear activation functions \citep{pmlr-v48-cohenc16}. Thus, characterizing all equivariant layers for a given group action is fundamental for understanding and designing equivariant deep neural networks. Indeed, this question has attracted a considerable amount of attention in various scenarios (e.g., \cite{finzi,pmlr-v80-kondor18a,NEURIPS2019_b9cfe8b6,pearce2023brauer,pearce2023jellyfish}).

For permutation groups, the standard strategy for characterizing equivariant layers is through studying parameter-sharing \citep{param2017}. Perhaps the first result in this direction was given by \cite{DeepSets}, who showed that only two parameters are required to characterize all permutation equivariant layers on a set of scalars. Another famous example is \cite{IGN}, who described the specific parameter-sharing scheme for graph equivariant networks and other tensor representations of the symmetric group. Recently, parameter-sharing has been applied in \cite{DWS,zhou2024permutation} to characterize all equivariant layers of weight space neural functionals. These neural networks operate on weights of another neural network, a problem with a complex multi-permutation equivariant structure.

In this paper, we revisit equivariant linear layer characterization for permutation groups from the perspective of irreducible representations.  Given a group $\G$ acting on a linear space $\V$, we explore linear equivariant mappings $f: \V \rightarrow \V$ by decomposing $\V$ to its minimal invariant components, which are irreducible representations of $\G$. Once this decomposition is performed, a basic result in representation theory called Schur's lemma automatically provides us with a full characterization of all equivariant mappings. 

As a first step, we use our method to get an alternative derivation for known equivariant layer characterization at the basis of the DeepSets \citep{DeepSets}, 2-IGN \citep{IGN}, and DWSNets \citep{DWS} architectures. Notably, for the Deep Weight Space problem, our derivation is significantly simpler than the cumbersome derivations previously obtained by parameter sharing \citep{DWS,zhou2024permutation}.   

Next, we consider the problem of sets of \emph{unaligned} symmetric elements, where equivariance to the joint action of a group $\G$ and the permutation group $S_k$ is required (also known as the wreath product $\G \wr S_k $).
 This setting arises naturally when considering alignment, distance prediction, or anomaly detection of sets of unaligned symmetric elements. It was studied by \cite{wreath} under the assumption that $\G$ acts transitively, in which case 
Siamese Networks capture an overwhelming majority of the equivariant layers. Indeed, in practice, Siamese networks are often employed \citep{DWS-align, Samanta}. In contrast, we give a full characterization of the equivariant layers in the general setting and show that in some settings (e.g., alignment of weights spaces), there can be a large number of non-Siamese equivariant layers. Empirically, we show that these additional layers improve performance on a synthetic graph anomaly detection task and the deep weight space alignment task discussed in \cite{DWS-align}. In summary, our contributions are the following:
\begin{itemize}[wide, labelwidth=!, labelindent=5pt]
    \item We give an alternative irreducible-based derivation for the  DeepSets \citep{DeepSets}, 2-IGN \citep{IGN}, and DWSNets \citep{DWS} architectures. The DWSnets derivation is significantly simpler than previous approaches.
    \item We provide a characterization of all linear equivariant functions on  sets of unaligned symmetric elements.    
    \item We empirically validate the importance of our equivariant layers for sets of unaligned symmetric elements by showing their superiority on a synthetic graph anomaly detection task, Wasserstein set distance computation, and deep weight space alignment task.
\end{itemize}

\section{Related work}
\paragraph{Learning Sets of Symmetric Elements.}
When dealing with multiple occurrences of symmetric objects as a set, additional symmetries arise as the set elements can be permuted. In this case, the layers are required to be equivariant under the permutation of elements in addition to the original element group action.
\cite{maron2020learning} suggested a new layer called Deep Sets for Symmetric Elements (DSS) that operates on a set of symmetric objects. They showed that incorporating DSS layers is strictly more expressive than using Siamese networks.  
The DSS framework was also used in sub-graph aggregation networks \citep{bevilacqua2021equivariant}, where the network should be invariant to the order of the sub-graphs.
In the main setting of \cite{maron2020learning}, the same group element $g \in G$ acts uniformly on all set objects. However, the more challenging setting is where we allow different elements $g_i \in G$ to act on the objects. This corresponds to the action of the restricted wreath product $G \wr S_k$ on the set. \cite{maron2020learning} deal with this setting as well, but only when $G$ is a subgroup of $S_n$ and the action of $G$ is transitive. Later,
\cite{wreath} characterized all linear mappings for general wreath products, however they also assumed that the group actions are transitive.

\paragraph{Equivariant Characterization Via Irreducibles.}
Using irreducible representations to characterize equivariant layers is a popular approach for rotation-equivariant networks \citep{tensorFields,cormorant,dymuniversality}. For permutation groups, the primary work which considered this approach, to the best of our knowledge, is \cite{kondor}. They suggest using the known characterization of permutation irreducible representations with Young diagrams \citep{fulton2013representation}  to characterize permutation-equivariant layers and explain how to reconstruct the results from DeepSets and 2-IGN. Our approach for these results is similar but gives a more explicit derivation of the DeepSet and 2-IGN layers. More importantly, our analysis covers deep weight spaces and sets of unaligned symmetric elements, not discussed in this paper. 

\section{preliminaries} 
Let $\V$ be a finite-dimensional vector space over the field $\mathbb{F}=\CC$ or  $\RR $. Let $\G$ be a finite group acting linearly on $\V$. We will sometimes say that $\V$ is a representation of $\G$. We will say $\V$ is a trivial representation of $\G$ if the action of $\G$ on $\V$ is the trivial action $gv=v$ for all $g\in \G$ and $v\in \V$.

A subspace $\U \subseteq \V$ is invariant to $\G$ if $g\cdot u \in \U$ for all $g\in G$ and $u\in \U$. We say an invariant subspace $\U$ is irreducible if it does not strictly contain an invariant subspace except the zero space. By Maschke's theorem, each finite-dimensional space $\V$ can be decomposed into a direct sum of irreducible invariant spaces \citep{fulton2013representation}.

A mapping $T: \V \rightarrow \V$ is equivariant if $T(g\cdot v) = g \cdot T(v)$ for every $v \in V, g \in \G$. Two representations $\V,\U$ of $\G$ are isomorphic if there is an equivariant linear bijection $L_{\V,\U}:\V \rightarrow \U$, which we notate by $\V \cong \U$.

The following lemma, named Schur's lemma, is a key component of our work.
Schur's lemma describes equivariant mappings between irreducible representations:
\paragraph{Schur's lemma.}
Let $\V, \W$ be finite-dimensional irreducible representations of $\G$ over  $\mathbb{F}$, and let $T: \V \to \W$ be a $\G$-equivariant linear map \footnote{The equivariant map is often referred to as a $G$-module homomorphism, or an intertwiner.}. Then:
\begin{itemize}
    \item Either $T$ is an isomorphism or $T=0$.
    \item If $L:\V \to \W$ is an isomorphism, and    $\mathbb{F}=\CC$, then $T=\lambda L$ for $\lambda\in \CC$.
    
\end{itemize}

In this paper, we will only consider representations over $\RR$, which are the common setting in applications. In all cases, we will discuss, our real irreducible representations are \emph{absolutely irreducible}. This means that their natural extension to complex vector spaces is irreducible. In this case, if $L:\V \to \W$ is an isomorphism,    $\mathbb{F}=\RR$,  and $\V$ is \emph{absolutely irreducible}, then $T=\lambda L$ for $\lambda \in \RR$ \citep{note}. In Appendix \ref{app:absolutely} we elaborate more on this topic. We note that when $\V$ and $\W$ are isomorphic irreducible real
representations and $\V$ is not absolutely irreducible, the space
$\operatorname{Hom}_{\G}(\V,\W)$ of equivariant linear mappings is
either $2$ or $4$ dimensional \citep{poonen2016real}. In this setting, using an automatic computational method to find all 2-4 equivariant layers may be beneficial \citep{finzi}.

\paragraph{Layer characterization using Schur's Lemma}
Equivariant layers from a representation $\V$ to itself can be characterized using Schur's lemma via two steps. The first step is decomposition, where we identify the decomposition of $\V$ into irreducible representations $\V=\oplus_{i=1}^s \V_i$. Any $x\in \V$ can be written uniquely as a sum $x=\sum_{i=1}^s x_i$ where each $x_i$ is in $\V_i$. The decomposition step also requires an algorithm to compute this decomposition. Next, we will need to identify which of the space pairs $\V_i$ and $\V_j$ are isomorphic, and if they are, specify an isomorphism $L_{ij} $. Finally, since we are working over the reals, we will need to verify that all irreducible representations are absolutely irreducible. 

Once the decomposition step is carried out, the second step uses Schur's lemma. Equivariant mappings $T:\V \to \V $   can be written as $T=\sum_{i,j=1}^s T_{ij} $, where $T_{ij}:\V_i \to \V_j$ are equivariant. By Schur's lemma if $\V_i$ and $\V_j$ are isomorphic then $T_{ij}=\lambda_{ij}L_{ij} $ for some $\lambda_{ij}\in \RR$, and otherwise $T_{ij}=0$. All in all, we obtain that $T$ is equivariant if and only if for some choice of the scalar $\lambda_{ij}$ we have
\begin{equation}\label{eq:Tgeneral}
T(x_1+\ldots+x_s)=\sum_{(i,j), \V_i \cong \V_j} \lambda_{ij}L_{ij}(x_i) .\end{equation}

The number of parameters in the expression above depends on the isomorphism relations between the irreducibles. We can assume without loss of generality that the first $1\leq t\leq s$ irreducible spaces are not isomorphic, and by identifying isomorphic irreducibles, we can rewrite the decomposition above as $\V=\oplus_{j=1}^t \V_j^{\oplus \alpha_j} $ (Where $\V_j^{\oplus \alpha_j}$ is defined to be the direct sum of $\alpha_j$ copies of $\V_{j}$)
where  $\alpha_1+\ldots+\alpha_t=s$.
The number of parameters $\lambda_{ij} $ in the decomposition is then $\sum_{i=1}^t \alpha_i^2 $. 
This process can easily be generalized to compute mappings between representations $\V$ and $\W$ by decomposing both spaces into irreducibles.

We note that the cornerstones of this methodology:  decomposition into irreducibles and Schur's Lemma, are applicable for all finite dimensional representations of finite groups (and also for compact infinite groups like $SO(d)$). The main challenge in this approach is characterizing and computing the decomposition into irreducibles. This needs to be done on a case to case basis. Much of the remainder of the paper will be devoted to computing these decompositions for important equivariant learning scenarios.

\section{Computing linear equivariant layers}
We now demonstrate how to derive the results in \cite{DeepSets,IGN, DWS} via decomposition into irreducibles.

\subsection{Deep Sets}
\label{sub:deepsets}
We begin with the simple case of the action of the permutation group $S_n$ on  $\V = \mathbb{R}^{n}$ by permutation of elements (we assume $n\geq 2$). Here, there are two non-zero invariant spaces: 
\begin{equation}\label{eq:deepsets}
    \Scalar = \{\alpha \cdot 1_n| \alpha \in \mathbb{R}\} , \quad \Vector(n) =\{x\in \mathbb{R}^{n}:\sum^{n}_{i=1}x_{i} = 0\}.
\end{equation}
    As the first space is one-dimensional, it's irreducible. By \cite{Hinton06}, the second one is also irreducible. These two spaces are not isomorphic since the action of $S_n$ on $\Scalar$ is trivial, but the action of $S_n$ on $\Vector(n)$ is not (also they do not have the same dimension when $n\geq 3$). 
    
The decomposition of $x\in \RR^n$ to a sum of elements from the irreducible spaces can be computed as 
$$x=\bar x 1_n+(x- \bar x 1_n) $$
where $\bar x$ is the average of $x$, and $1_n$ is the all one vector. By \cref{eq:Tgeneral},  Schur's lemma, and the fact that all irreducibles of the permutation group are absolutely irreducible (see Appendix \ref{app:absolutely}), the linear equivariant mappings $T:\V \to \V$ are characterized by two parameters $a,b\in \RR$, that is: 
\begin{align*}
Tx=a\bar x 1_n+b(x-\bar x 1_n).
\end{align*}
This is exactly the result described in DeepSets \citep{DeepSets}.

\subsection{Equivariant graph layers}\label{sub:graph}
We next consider the setting where $\V = \mathbb{R}^{n \times n},\G = S_{n}$ and the group action is defined by $(\tau \cdot X)_{ij} = X_{\tau^{-1}(i),\tau^{-1}(j)}$. This setting is natural for graph neural networks, as two adjacency matrices $A, B\in \RR^{n\times n}$ represent isomorphic graphs if and only if  $A=\tau \cdot B $ for some permutation $\tau$. This setting was considered in \cite{IGN} using a parameter sharing scheme, and we show how to obtain similar results using irreducibles (note that \cite{IGN} also discussed general mappings between $k$-order and $\ell$-order tensors. In contrast, we only discuss the $k=\ell=2$ case). 

We claim that, when $n\geq 4$, the space $\RR^{n\times n}$ can be written as a sum of seven irreducible permutation invariant sub-spaces. The behavior for $n<4$ is discussed in Appendix \ref{app:small}.

The first two spaces are one-dimensional representations on which the action of $S_n$ is trivial: diagonal matrices with identical diagonal entries and matrices with zero diagonal and identical off-diagonal entries:
\begin{equation} \label{eq:graphTrivial}
    \V_{0} = \{ a \cdot I_n|a\in \R \}, \quad     \V_{1} = \{a \cdot (1_{n\times n} -   I_n)|a\in \R \}.
\end{equation}

Next, we have three spaces of dimension $n-1$ which are isomorphic to $\Vector(n)$ from \eqref{eq:deepsets}. The first space is the space $\V_{2}$ of diagonal matrices whose diagonal sums to zero
and the next two $n-1$ dimensional spaces are the space of matrices whose rows (respectively columns) are constant, and columns (respectively rows) sum to zero:
$$\V_3=\{r1_n^T|  \sum_{i=1}^n r_i=0 \}, \quad \V_4=\{1_nc^T|  \sum_{i=1}^n c_i=0 \}  $$
Finally, there are two larger irreducible spaces:
\begin{equation*}
	\V_5=\{A| \quad A=-A^T, A1_n=0_n\}, \quad 
	\V_6=\{A| \quad A=A^T, A1_n=0_n,  A_{ii}=0, \forall i=1,\ldots,n\}
\end{equation*}

The dimension of $\V_{5}$ is $\binom{n-1}{2}=\frac{n^{2}-3n}{2}+1$, and the dimension of $\V_{6}$ is $\binom{n-1}{2}-1=\frac{n^{2}-3n}{2}$.
So, we have the following decomposition:
\begin{align*}
    \V = \V_0\oplus\V_{1} \oplus \V_{2} \oplus \V_{3} \oplus \V_{4} \oplus \V_{5} \oplus \V_{6}  \cong \V_0^{\oplus 2}\oplus \V_3^{\oplus 3} \oplus \V_{5} \oplus \V_{6}
\end{align*}
Accordingly, using Schur's lemma, each linear mapping can be characterized by $3^{2}+2^{2}+1+1 = 15$ parameters, the same result from \cite{IGN}, so we have the expected number of parameters. In appendix \ref{decomposition}, we give a formal proof of these arguments and present an algorithm to decompose an input matrix $\X$ into its seven irreducible spaces with linear complexity in the matrix size $n^2$. These results are summarized in the following theorem.
\begin{restatable}{theorem}{graphs}\label{thm:graphs}
For all $n\geq 4$, the space $\RR^{n\times n} $ can be written as a direct sum of the spaces $\V_0,\ldots,\V_6$. These spaces are invariant and irreducible, and the isomorphism relations between them are given by 
$\V_0 \cong \V_1, \V_2\cong \V_3\cong \V_4. $
\end{restatable}

\subsection{Deep weight spaces}
Recently, there has been growing interest in devising neural operators: neural networks that operate on an input, which is itself a neural network. This type of problem is of interest for various tasks involving post-processing and synthesis of multiple trained neural networks, as well as for processing Implicit Neural Representations (INRs), which are a popular alternative for representing certain standard data structures.  (see e.g. \cite{scaleEqui} for further discussion). 

A key concept in the design of 'neural operators' has been the requirement that they are equivariant to the input neural data \citep{scaleEqui,kofinas2024graph,zhou2024universalneuralfunctionals,zhou2023neural,lim2024graph}. 
In this section we consider the setting discussed in \cite{DWS,zhou2024permutation}, where  the neural data  is a collection of weights and biases $(W_m,b_m)_{m=1}^M$ representing an MLP of depth $M$, and a neural operator is constructed by composing standard activation with linear mappings  which are equivariant with respect to the multi-permutation action we will now describe:

The output of an MLP architecture is invariant to the permutation of its hidden neurons. For example, the MLP defined by $W_2\cdot ReLU \cdot (W_1x) $ will remain the same function if we replace weights $W_2,W_1$ with the weights $W_2P,P^TW_1 $. 
 
To define the symmetries of learning on MLPs in full generality, we will adapt the notation from \cite{DWS}. We consider the space of MLP parameters with a given fixed depth $M$ and layer dimensions $d_0,\ldots,d_{M}$ ($M$ and all $d_j$ are assumed to be larger than one). These are parameterized by the vector space
$\V=\bigoplus_{m=1}^M \left( \mathcal{W}_m\oplus \mathcal{B}_m \right)$
where $\mathcal{W}_m:=\mathbb{R}^{d_{m}\times d_{m-1}}$ and $\mathcal{B}_m := \mathbb{R}^{d_m}$ represent the weights and biases of the $m$-th layer. 
The symmetry group of the weight space is the direct product of symmetric groups 
$
    \G 
    = S_{d_1}\times \cdots\times S_{d_{M-1}}.
$
An element $g=(\tau_1,\ldots,\tau_{M-1})$ in the group acts on an element $v=[W_m,b_m]_{m\in[M]}$ from  $\mathcal{V}$ as follows: 
\begin{subequations}\label{eq:action}
\begin{equation}
    \rho(g)v 
    = [W'_m,b'_m]_{m\in[M]}, 
\end{equation}
\begin{equation}
            W_1'=  P_{\tau_1}^T W_1, 
            ~ b_1' = P_{\tau_1}^Tb_1, 
\end{equation}
\begin{equation}
            W_m' =  P_{\tau_m}^T W_m P_{\tau_{m-1}}, 
            ~ b_m' = P_{\tau_m}^T b_m, ~m\in [2,M-1]
\end{equation}
\begin{equation}
            W_{M}' =   W_M P_{\tau_{M-1}}, 
            ~ b_{M}' = b_M. 
\end{equation}
\end{subequations}

where $P_{\tau_m}\in \mathbb{R}^{d_m \times d_m}$ is the permutation matrix of $\tau_m\in S_{d_m}$.

Previous work \citep{DWS,zhou2024permutation} has already characterized all linear equivariant functions from $\V$ to itself. However, this characterization requires tedious bookkeeping and division into a large number of different cases. Here we show how to decompose $\V$ into irreducibles in a rather straightforward way, and as a result obtain an (arguably) simpler  characterization of all linear equivariant layers.

We claim that $\V$ is a direct sum of multiple copies of $2M-2$ irreducible representations of $\G$: 
\begin{enumerate}[wide, labelwidth=!, labelindent=5pt]
    \item The first irreducible representation is the trivial scalar representation $\Scalar=\RR$ with the trivial action $gx=x$. 
    \item The next $M-1$ representations are the vector representations $\Vector_m, m=1,\ldots,M-1$ of vectors in $\RR^{d_m}$ which sum to zero (the spaces $\Vector(d_m)$ from \eqref{eq:deepsets}), with the action of $g$ being permutation of the vector by the $m$-th permutation $\tau_m$.
    
    \item The last $M-2$ representations are representations $\Matrix_m, \quad m=2,\ldots,M-1 $ of $d_m\times d_{m-1} $ matrices whose rows and columns sum to zero, where the action of $g$ on a matrix $W$ in this space is given by $P_{\tau_m}^TWP_{\tau_{m-1}}$. 
\end{enumerate}

The linear equivariant layers from $\V$ to itself can be inferred from the following theorem:
\begin{theorem}\label{thm:dws}
The spaces
$\Scalar, \Vector_1,\ldots,\Vector_{M-1},
\Matrix_2,\ldots,\Matrix_{M-1}$
are absolutely irreducible and pairwise non-isomorphic.

If $M\geq 3$, then
\[
\V \simeq
\Scalar^{\oplus \alpha}
\oplus
\left(
\bigoplus_{m=1}^{M-1}
\Vector_m^{\oplus \beta_m}
\right)
\oplus
\left(
\bigoplus_{m=2}^{M-1}\Matrix_m
\right),
\]
where
\begin{equation}\label{eq:sizeDWS}
\alpha=d_0+2d_M+2M-3,
\qquad
\beta_1=d_0+2,
\qquad
\beta_{M-1}=d_M+2,
\qquad
\beta_m=3,\quad m=2,\ldots,M-2.
\end{equation}
Consequently, the space of equivariant linear mappings from $\V$ to
itself has dimension
\[
\alpha^2+\sum_{m=1}^{M-1}\beta_m^2+(M-2).
\]

If $M=2$, then
\[
\V \simeq
\Scalar^{\oplus \alpha}
\oplus
\Vector_1^{\oplus \beta_1},
\]
where
\[
\alpha=d_0+2d_2+1,
\qquad
\beta_1=d_0+d_2+1,
\]
and hence the space of equivariant linear mappings from $\V$ to itself
has dimension
\[
\alpha^2+\beta_1^2.
\]
\end{theorem}
\begin{proof}
The spaces $\Scalar, \Vector_1,\ldots,\Vector_{M-1},\Matrix_2,\ldots,\Matrix_{M-1}$ are not isomorphic (even when they have the same dimension). The action of $\G$ on $\Scalar$ is trivial while the action on the other spaces is not. The other spaces are pairwise non-isomorphic because different factors of
$\G$ act non-trivially on them: $\Vector_m$ is acted on non-trivially only
by $S_{d_m}$, whereas $\Matrix_m$ is acted on non-trivially by
$S_{d_{m-1}}$ and $S_{d_m}$. Hence, these representations have different
kernels and therefore cannot be isomorphic.
The spaces $\Scalar$ and $\Vector_m$ are absolutely irreducible, as we discussed in Subsection \ref{sub:deepsets}. We will explain why $\Matrix_m$ are absolutely irreducible in Appendix \ref{app:dws}.

Obtaining the decomposition of $\V$ is rather straightforward. First, following \cite{DWS}, we identify the weight spaces $\W_m $ and bias spaces $\B_m$ with subspaces $\hat{\W}_m $ and $\hat{\B}_m$ of $\V$ by zero padding, and note that each one of these subspaces is $\G$-invariant and that their direct sum gives us the full parameter space $\V$. We can then decompose each one of these spaces into irreducibles to obtain
\begin{subequations}\label{eq:dws}
\begin{align}
\hat{\B}_m &\cong \Scalar \oplus \Vector_m, \quad m=1,\ldots,M-1\\
\hat{\B}_M &\cong \Scalar^{\oplus d_M}\\
\hat{\W}_1 &\cong \Scalar^{\oplus d_0} \oplus \Vector_1^{\oplus d_0}\\
\hat{\W}_m &\cong \Scalar \oplus \Vector_{m-1}\oplus \Vector_{m} \oplus \Matrix_m, \quad m=2,\ldots,M-1 \label{most_challenging}\\
\hat{\W}_M & \cong \Scalar^{\oplus d_M} \oplus \Vector_{M-1}^{\oplus d_M}
\end{align}
\end{subequations}

The multiplicities of each irreducible in $\V$, specified in \eqref{eq:sizeDWS}, can now be found by simply counting how many times each irreducible appears in the decomposition above. 

The decomposition in \eqref{eq:dws} can actually be easily obtained from the 'deepsets' decomposition of the action of $S_n$ on $\RR^n$ described previously: the action of $\G$ on $\hat \B_m$ with $m<M$ is isomorphic to the action of $S_{d_m}$ on $\RR^{d_m}$ and hence we get the exact 'DeepSets' decomposition from  \cref{sub:deepsets}. The action of $\G$ on $\hat{\B}_M$ is trivial and hence can be written as a direct sum of $d_M$ trivial one dimensional $\Scalar$ spaces. The action on $\hat{\W}_1$ (and $\hat{\W}_M$) multiplies a matrix by a permutation from the left (right), and hence can be seen as a direct sum of ‘deep-sets’ actions  on the columns (rows) of the matrix. Finally, the action of $\G$ on $\hat{\W}_m$ for $m=2,\ldots,M-1$ is a tensor product of the natural action of $S_{d_{m-1}}$ on $\RR^{d_{m-1}}$ and the action of $S_{d_{m}}$ on $\RR^{d_{m}}$, and therefore its irreducible decomposition can be obtained by taking tensor products of the irreducible representations of $\RR^{d_{m}}$ and $\RR^{d_{m-1}}$, as explained in more detail in the full proof.
\end{proof}
We note that \eqref{eq:dws} includes almost all information required to compute linear equivariant maps from $\V$ to itself. All that is left is the decomposition algorithm for writing each
$(W_m,b_m)_{m\in [M]}$ as a direct sum of elements in the irreducible decomposition. This can be done independently for each subspace $\W_m$ and $\B_m$. The decomposition for  $\hat{\W}_m$ in \eqref{most_challenging} is not immediate and we will explain it in Appendix \ref{app:Wdecompose}.

The decomposition in \eqref{eq:dws} and \eqref{eq:sizeDWS} also provides substantial additional information. For example, we can immediately see that there are $\alpha$  invariant maps from $\V$ to $\RR=\Scalar$ , which correspond to the number of copies of the trivial representation $\Scalar$ in $\V$. Moreover, if we are interested in the equivariant maps from a bias space $\hat{\B}_i$ (or weight space $\hat{\W}_i$)  to another bias space $\hat{\B}_j$ (or weight space $\hat{\W}_j$), we can easily infer the equivariant mappings from the decomposition in \eqref{eq:dws}. For example, when $i=j=M$ there will be $d_M^2 $ mappings since $\hat{\B}_M$ consists of $d_M$ isomorphic representations,  and when $i<j=M$ there will be $d_M$ mappings since $\Scalar$ appears a single time in $\hat{\B}_i$ and $d_M$ times in $\hat{\B}_M$. Continuing in this way we can reconstruct all the different bias-to-bias, bias-to-weight, weight-to-bias, and weight-to-weight cases analyzed in Tables 5-8 of \cite{DWS}, and Tables 8-11 in \cite{zhou2024permutation}. More importantly, these tables which were necessary for implementing weight space layers in previous work, are not necessary when implementing these layers using Schur's lemma as we suggest.

\section{Sets of unaligned symmetric elements}
\label{sec:wreath}
Next, we consider the setting where our data is a $k$-tuple of 'unaligned objects' $(v_1,\ldots,v_k)$, each coming from a representation $\V$ of $\G$, and we want to learn functions which are equivariant with respect to the joint action of a $k$-tuple of group elements $(g_1,\ldots,g_k)$ on each coordinate independently, and to a permutation $\tau\in S_k$ of the $k$-tuple.

We define
\[
G \wr S_k := G^k \rtimes S_k,
\]
where $S_k$ acts on $G^k$ by permuting its coordinates. Writing an element
of $G \wr S_k$ as $((g_1,\ldots,g_k),\tau)$, its action on $V^k$ is given by
\begin{equation}\label{eq:wreath_action}
((g_1,\ldots,g_k),\tau)\cdot(v_1,\ldots,v_k)
=
\left(
g_1\cdot v_{\tau^{-1}(1)},
\ldots,
g_k\cdot v_{\tau^{-1}(k)}
\right).
\end{equation}
The group for which this action is defined is also called the restricted wreath product of $\G$ and $S_k$. For more details, see \cite{wreath}.

This type of 'wreath-equivariant-structure' arises in several settings. One is 'alignment problems', where our goal is, given a pair of elements $(v_1,v_2)\in \V^k, k=2$,  to find the group element $g^*=g^*(v_1,v_2)$ which makes $v_1$ 'as similar as possible' to $v_2$. This task is equivariant to application of $\G$ elements to each coordinate because if 
$g^*v_1\approx v_2$ then $g_2g^*g_1^{-1} (g_1v_1)\approx g_2v_2 $. Similarly, this task is equivariant to permuting $v_1$ and $v_2$, because if $g^*v_1 \approx v_2$ then $(g^*)^{-1}v_2\approx v_1 $. For a more detailed derivation, see \cite{Samanta,DWS-align}, which discussed these problems for sets and weight spaces, respectively. Additional examples of 'wreath-equivariant-problems' are the anomaly detection problem discussed in the experimental section, and problems with hierarchical structures as discussed in \cite{wreath}. 

\paragraph{Wreath-equivariant layers.} Our aim is to characterize all $\G \wr S_k$ equivariant mappings from $\V^k$ to itself. We note that any linear $\G$-equivariant mapping $\hat L:\V \to \V $ induces a 'Siamese'  $\G \wr S_k$ equivariant mapping defined by 
$$L(v_1,\ldots,v_k)=(\hat L(v_1),\ldots,\hat L(v_k)) .$$

The interesting question is how many additional mappings are present. 
This problem was previously studied in \cite{maron2020learning,wreath} when $\G$ is a finite group acting on $\RR^n$ \emph{transitively} by permutations (this means that the action of $\G$ on $[n]$ has a single orbit). In this setting, the equivariant mappings are composed of the Siamese mappings and a single additional non-Siamese mapping. However, the transitivity assumption does not hold in many examples of interest, such as the graph and weight space examples discussed in this paper. In our analysis, we will relax the transitivity assumption, and allow $\G$ to be a general finite group. In some cases, this will lead to a substantial number of non-Siamese networks. 

To characterize $\G \wr S_k $ equivariant functions, we first aim to characterize all invariant irreducible sub-spaces of $\V^{k}$, assuming we know all irreducible sub-spaces of $\V$. An important role will be played by \emph{trivial representations}: representations $\Scalar$ of $\G$ such that $gv=v$ for all $g\in \G$ and $v\in \Scalar$. 

\begin{theorem}\label{thm:wreath_irreps}
Let $k\geq 2$, and let $\V$ be a real representation of $\G$, with irreducible decomposition 
\begin{equation}\label{eq:trivialPlusNon}
\V=\left(\oplus_{i=1}^s \Scalar_i\right) \oplus \left( \oplus_{j=1}^t \V_j \right) \end{equation}
where $\Scalar_i$ are trivial representations and $\V_j$ are non-trivial irreducible representations. Then an irreducible decomposition for $\V^k$ with respect to the action of $\G \wr S_k$ is given by
\[
\V^k
=
\left(\oplus_{i=1}^s \Scalar_{i,0}^k \right)
\oplus
\left(\oplus_{i=1}^s \Scalar_{i,1}^k \right)
\oplus
\left(\oplus_{j=1}^t \V_j^k \right),
\]
\[
\text{where}\quad
\Scalar_{i,0}^k
=
\left\{
(s_1,\ldots,s_k)\in \Scalar_i^k :
\sum_{\ell=1}^k s_\ell=0
\right\},
\qquad
\Scalar_{i,1}^k
=
\left\{
(s,\ldots,s)\in \Scalar_i^k
\right\}.
\]
\end{theorem}
\begin{proof}
The fact that $\V_i $ is $\G$ invariant implies easily that  $\V_i^k$ is $\G \wr S_k$ also invariant. In the appendix \ref{power_irred}, we will show that if the action of $\G$ on $\V$ is not trivial then $\V^k $ is irreducible representation of $\G \wr S_k$. 
In contrast, for the spaces $\Scalar_i^k$ the action of $\G$ is trivial, so this representation can be identified with the standard representation $\RR^k$ of $S_k$. The decomposition to $\Scalar_{i,0}^k$ and $\Scalar_{i,1}^k$ then follows from the 'DeepSets decomposition' discussed in \cref{sub:deepsets}. \end{proof}
To count the number of linear equivariant mappings $L:\V^k \to \V^k $, we note that 
$$\Scalar_{i,0}^k \cong \Scalar_{j,0}^k, \quad \Scalar_{i,1}^k \cong \Scalar_{j,1}^k, \quad \Scalar_{i,0}^k \not \cong \Scalar_{i,1}^k, \quad \forall 1\leq i<j\leq s $$
Thus, the dimension of the space of linear equivariant mappings from the ``trivial part'' of $\V^k$, that is,
\[
\left(\oplus_{i=1}^s \Scalar_{i,0}^k \right)
\oplus
\left(\oplus_{i=1}^s \Scalar_{i,1}^k \right),
\]
to itself is $2s^2$, while the number of corresponding Siamese mappings,
induced by linear equivariant mappings from the ``trivial part'' of $\V$
to itself, is $s^2$. As a result, we obtain $s^2$ additional non-Siamese
$\G \wr S_k$-equivariant basis maps. 

\paragraph{Examples.}
We found that the number of non-Siamese layers is $s^2$, where $s$ is the number of trivial representations in $\V$. Let us consider the implications for the three examples we have discussed earlier:
\begin{enumerate}[wide, labelwidth=!, labelindent=1pt]
\item \textbf{Deep-sets.} In the DeepSets setting where $\V=\RR^n, \G=S_n$, there is a single trivial representation of constant vectors, and therefore in this setting there is a unique non-Siamese layer for $\V^k$. Indeed, this is the case for any group acting transitively by permutations on $\RR^n$, and thus we obtain the results for transitive actions from \cite{maron2020learning,wreath}. 
\item \textbf{Graphs.} In the graph setting where $\V=\RR^{n\times n}, \G=S_n$ there are $s=2$ trivial representations (see \eqref{eq:graphTrivial}), and therefore $s^2=4$ non-Siamese layers.
\item \textbf{Weight Spaces.} In the weight space examples the number $s$  of trivial representations is rather large,  $\alpha=d_0+2d_M+2M-3 $ from \eqref{eq:sizeDWS}, and there are $\alpha^2$ non-Siamese mappings. 
\end{enumerate}

In the next theorem we give an explicit characterization of the non-Siamese layers of $\V^k$. For this characterization, we note that the direct sum of all trivial representations of $\V$ (as in \eqref{eq:trivialPlusNon}) is the vector space 
$\V_{fixed}=\{v\in \V | gv=v, \forall g \in \G \}. $
Alternatively, every basis $e_1,\ldots,e_s$ of $\V_{fixed}$ defines a decomposition of $\V_{fixed}$ into one-dimensional irreducible invariant sub-spaces $\S_i=\{ce_i|  c \in \RR\} $. Accordingly, our characterization of non-Siamese layers is based on such bases:
\begin{restatable}{theorem}{wreath}
\label{characterization}
Let $\V$ be a real representation of a finite group $\G$. and
let $e_1,\ldots,e_s$ be a basis for the subspace 
$\V_{fixed}$. Let $\langle \cdot, \cdot \rangle$ be a $\G$ invariant inner product on $\V$.
Then every linear equivariant map $L:\V^{k} \to \V^{k} $ is of the form

\begin{equation}\label{eq:form}
L(v_1,\ldots,v_k)=\sum_{i,j=1}^s a_{ij}\left(\sum_{\ell=1}^k \langle v_\ell,e_i \rangle e_j, \ldots, \sum_{\ell=1}^k \langle v_\ell,e_i \rangle e_j\right) 
+ \left(\hat{L}(v_1),\ldots ,\hat{L}(v_k) \right)  \end{equation}
where $\hat L:\V \rightarrow \V$ is a linear equivariant map, and $a_{ij}$ are real numbers. Conversely, every linear mapping of the form defined in \eqref{eq:form} is equivariant. 
\end{restatable}
The theorem is proven in  Appendix \ref{app:wreath}. We note that an invariant inner product on $\V$ is an inner product satisfying $\langle gv,gu \rangle=\langle v,u \rangle$ for all $g\in \G$ and $u,v\in \V$. When $\G$ is finite, an invariant inner product always exists: it can be obtained by starting from an arbitrary inner product and then averaging over the group \citep{fulton2013representation}. 
In the examples we consider in this paper, the group $\G$ acts by permutations. In this case, the standard $\ell_2$ inner product is invariant. Implementing the non-Siamese layers defined in \eqref{characterization} only requires finding a basis for $\V_{fixed}$. In particular, if $\V$ is one of the spaces discussed previously, the basis $e_i$ is just a choice of an element from each of the trivial spaces $\Scalar_i$ in the decomposition of $\V$. 

\paragraph{Beyond $S_k$.} So far, we have considered the action of $\G \wr S_k$ where $\G$ is finite. We can also generalize these results to the setting $\G \wr H $, where $H$ is a subgroup of $S_k$ which acts transitively on $\{1,\ldots,k\} $. By doing this, we are generalizing the results in \cite{wreath}, which assumes that both $\G$ and $\H$ act transitively. 

The full generalization is described in Appendix \ref{app:beyond}. The general idea is this: Since $\G$ acts trivially on $\V_{fixed}$, the action of $\G \wr H $ on $\V_{fixed}^k$ can be identified with the action of $H$ on $\V_{fixed}^k$. In turn, $\V_{fixed}^k$ is a direct sum of $s$ different copies of $\RR^k$. Accordingly, the number of $H$ equivariant maps from $\V_{fixed}^k$ to itself is $s^2\cdot h$, where $h$ is the number of $H$ equivariant maps from $\RR^k$ to itself. One of these $h$ maps is the identity map, which is Siamese; therefore, the total number of non-Siamese equivariant maps is $(h-1)\cdot s^2 $. In the case where $H=S_k$, we have that $h=2$, and therefore, in our analysis above, we have found $s^2$ non-Siamese maps.
\section{Experiments}

In this section, we consider problems with a wreath-equivariant structure. We compare networks implementing the complete basis of equivariant layers that we have found, with networks that only use Siamese layers or combine a partial list of non-Siamese layers suggested in previous works. Implementation details of all experiments are described in \autoref{implementation_details}.

\paragraph{Graph Anomaly Detection.}
We begin with a synthetic wreath-equivariant task in which Siamese networks will fail by design: we consider a graph anomaly detection problem, where the input is $k$ graphs with $n$ nodes, $(\mathbb{G}_1,\ldots,\mathbb{G}_k) $, where most graphs are similar, and one is an 'anomaly.'  The output is a $k$ dimensional probability vector, where the $i$-th entry denotes the probability that $\mathbb{G}_i$ is an anomaly. This task is $\G \wr S_k$ equivariant (where $\G=S_n$), where the action on the input space is as in \eqref{eq:wreath_action}, and the action on the output space is just permuting the entries of the probability vector.
We generate data for this problem as follows: We randomly generate two graphs $\mathbb{G}$ and $\hat{\mathbb{G}}$ by sampling each entry of their adjacency matrices independently from a Bernoulli distribution. We take $k-1$ copies of the graphs $(\mathbb{G}_1,\ldots,\mathbb{G}_{k-1})$ to be permuted copies of $\mathbb{G}$, and one of them to be $\hat{\mathbb{G}}$ and insert it in a random location. We add Gaussian noise with standard deviation $\eta$ to the permuted copies of
$\mathbb{G}$ and to the anomalous graph $\hat{\mathbb{G}}$.
                              
We consider equivariant models for this task, composed of several linear wreath-equivariant layers and point-wise ReLU activations. The final layer is a point-wise summation of each of the $k$ graphs to obtain a final vector in $\RR^k$, followed by a softmax layer to obtain a probability vector. For the linear wreath-equivariant layers, we consider several alternatives:
Siamese layers only, adding the single additional non-Siamese layer suggested in \cite{wreath,maron2020learning} denoted by DSS, and the full model we suggest, which has four non-Siamese layers (we name our model SchurNet). The Siamese layers are implemented using the decomposition computed in  Subsection \ref{sub:graph}.


\begin{table}[ht]
    \centering
    \caption{Anomaly detection accuracy under different noise levels.}
    \label{tab:anomaly_acc}
    \begin{tabular}{lcc}
        \toprule
        \textbf{Model} & \(\boldsymbol{\eta = 0.0}\) & \(\boldsymbol{\eta = 0.1}\) \\
        \midrule
        Siamese            & 48\%           & 28\%          \\
        DSS                & 97.5\%           & 92.0\%          \\
        \textbf{SchurNet (Ours)}
                           & \textbf{100.0\%} & \textbf{97.0\%} \\
        \bottomrule
    \end{tabular}
\end{table}

As shown in ~\autoref{tab:anomaly_acc}, the Siamese model achieves only
$48\%$ accuracy without noise, and its performance further decreases to
$28\%$ when noise is added. In contrast, DSS achieves $97.5\%$ and $92.0\%$
accuracy, while SchurNet achieves perfect accuracy in the noiseless setting
and $97.0\%$ accuracy under noise. This is to be expected since Siamese features can be useful to differentiate between $\mathbb{G}$ and $\hat {\mathbb{G}}$, but not to determine how many times each one of them occurs in a $k$-tuple.
In contrast, networks with non-Siamese layers attain significantly better performance, whereas our method, which contains the maximal number of non-Siamese layers, attains the best performance. 

\begin{table}[ht]
    \centering
    \small
    \setlength{\tabcolsep}{6pt}
    \renewcommand{\arraystretch}{1.05}

    \caption{Comparison of SchurNet and NProductNet. 
    The best result in each setting is shown in bold, and the second-best is underlined.}
    \label{table:performance_comparison}

    \begin{tabular}{llcc}
        \toprule
        \textbf{Dataset}
        & \textbf{Input size}
        & \textbf{SchurNet}
        & \textbf{NProductNet} \\
        \midrule

        \multirow{2}{*}{noisy-sphere-3}
            & \([100, 300]\) & \textbf{0.0389} & \underline{0.0460} \\
            & \([300, 500]\) & \textbf{0.1026} & \underline{0.1580} \\

        \addlinespace[3pt]
        \multirow{2}{*}{noisy-sphere-6}
            & \([100, 300]\) & \underline{0.0217} & \textbf{0.0150} \\
            & \([300, 500]\) & \underline{0.0795} & \textbf{0.0490} \\

        \addlinespace[3pt]
        \multirow{2}{*}{uniform}
            & \(256\)        & \underline{0.0974} & \textbf{0.0970} \\
            & \([200, 300]\) & \textbf{0.1043} & \underline{0.1089} \\

        \addlinespace[3pt]
        \multirow{2}{*}{ModelNet-small}
            & \([20, 200]\)  & \textbf{0.0623} & \underline{0.0840} \\
            & \([300, 500]\) & \textbf{0.0738} & \underline{0.1110} \\

        \addlinespace[3pt]
        \multirow{2}{*}{ModelNet-large}
            & \(2048\)         & \textbf{0.0468} & \underline{0.1400} \\
            & \([1800, 2000]\) & \textbf{0.0551} & \underline{0.1620} \\

        \addlinespace[3pt]
        \multirow{2}{*}{RNA-seq}
            & \([20, 200]\)  & \underline{0.0123} & \textbf{0.0120} \\
            & \([300, 500]\) & \textbf{0.0334} & \underline{0.2920} \\

        \bottomrule
    \end{tabular}
\end{table}

\paragraph{Wasserstein Distance Computation}
We next consider the task of learning the Wasserstein distance, as discussed in \cite{amir2024injective,Samanta,haviv2024wasserstein}.  The Wasserstein distance between two unordered multisets of vectors in $\RR^d$, denoted by $\{x_1,\ldots,x_n \} $ and $\{y_1,\ldots,y_n \} $, is defined to be the minimal distance between the sets, under the optimal permutation giving the best alignment between them (as in the alignment problem discussed in the beginning of Section \ref{sec:wreath}). Computing Wasserstein distances can be computationally intensive, so this task aims to devise a neural network to learn the distance instead. 
Learning Wasserstein distances is a $S_n \wr S_2 $ invariant task. In \cite{Samanta}, a Siamese approach named NProductNet is suggested to address this problem.
We compare SchurNet to \cite{Samanta} to demonstrate that adding our non-siamese layers enhances the performance of standard Siamese layers. We note that \cite{amir2024injective} achieved state-of-the-art results for this task with a very different method, which does not follow the conventional paradigm of a stack of linear layers and non-linear activation functions. Therefore, we have no direct way of adding our non-siamese layers to their method for comparison.
In our experiments, we add the additional  $d^2$  non-Siamese layers to their implementation and compare performance on the datasets addressed in their paper. The results,  reported in Table \ref{table:performance_comparison}, show that adding non-Siamese layers improves performance in most cases. We note that each dataset has two settings: the top row corresponds to
test sets whose cardinalities are within the range seen during training,
while the bottom row evaluates generalization to unseen set sizes.
\paragraph{Weight Space Alignment.}
We consider the alignment task discussed above in the setting where $v_1,v_2$ are elements in weight spaces, and the task is to find the group element that optimally aligns $v_1$ and $v_2$.  One interesting application of this problem is model merging: in \cite{rebasin}, it was shown that linear interpolation of $v_1,v_2$ gives a new network whose performance is considerably worse, but interpolation after alignment leads to much better results. 

The weight space alignment problem was considered in \cite{rebasin,NEURIPS2020_aecad423,pena2023re}. Recently, \cite{DWS-align} outperformed these methods using a learning approach based on the DWS layers from \cite{DWS}, applied to $v_1,v_2$ in a Siamese fashion. This experiment aims to check whether adding non-Siamese layers improves their results. 
We take only part of the described layers in
\autoref{characterization} layers: the mappings from each weight/bias space to itself, excluding, for example, non-Siamese weight to other weight or weight to bias. This added fewer parameters and generalized better than taking all mappings.
\begin{table}[h!]
\centering
\begin{tabular}{lcccc}
\toprule
\textbf{Model} & \multicolumn{2}{c}{\textbf{MNIST}} & \multicolumn{2}{c}{\textbf{CIFAR10}} \\
\cmidrule(lr){2-3} \cmidrule(lr){4-5}
               & \textbf{Acc(↓)} & \textbf{Loss (↓)} & \textbf{Acc(↓)} & \textbf{Loss (↓)} \\
\midrule
\textbf{SchurNet (Ours)}    & \textbf{1.25e-5}   & \textbf{0.251346}  & \textbf{0.0}   & \textbf{1.7822}  \\
\textbf{Siamese}  & 1.5e-5    & 0.262913    & 1.0e-4        & 1.7876   \\
\bottomrule
\end{tabular}
\caption{Comparison of SchurNet and Siamese models on MNIST and CIFAR10.}
\end{table}
\vspace{-10pt}

We ran two sets of experiments from \cite{DWS-align}: one on MLPs trained on MNIST \citep{mnist} and one on MLPs trained on the CIFAR10 \citep{cifar10} dataset. We trained both models for 100 epochs and reported the test accuracy and the reconstruction loss. We ran several hyper-parameter configurations for both methods to ensure a fair comparison and reported the results with the best reconstruction loss.  As we can see, our model with non-Siamese layers outperforms the Siamese network from \cite{DWS-align} in all settings.

\paragraph{Conclusion.}
In this paper, we revisited the idea of using irreducible representations instead of parameter-sharing to characterize equivariant linear layers for representations of permutations and related groups. Using this approach, we obtained alternative derivations for the characterizations of equivariant layers from DeepSets and $2$-IGN, and a significantly simplified derivation of deep weight spaces equivariant layers. We have also obtained a previously unknown characterization for the linear equivariant layers of wreath products $\G \wr S_k$, and showed the benefits of using the full characterization for several wreath-equivariant tasks. 
In general, looking forward to other yet unknown applications, what could be the benefit of using the irreducible approach? One answer is reduced book-keeping, as exemplified in the deep weight space example. In general, the number of equivariant layers can be, in the worst case,  quadratic in the number of irreducibles. When this happens, the irreducible approach is expected to lead to simpler characterizations that are easier to implement. Additionally, networks like $k$-IGN \citep{IGN}, which are based on tensor representations, use intermediate features in $n^k$, and using irreducibles could lead to new equivariant models with intermediate irreducible features of lower dimensions. Finding the irreducible decompositions of $k$-IGN when $k>2$ and the potential applications of this decomposition is an interesting avenue for future work.


\section*{Appendix}
\phantomsection
\addcontentsline{toc}{section}{Appendix}
\addtocontents{toc}{\protect\setcounter{tocdepth}{0}}
\setcounter{section}{0}
\renewcommand{\thesection}{\Alph{section}}
\renewcommand{\theHsection}{appendix.\thechapter.\Alph{section}}

\section{Implementation details}
\label{implementation_details}
In this section, we detail the hyper-parameters used through training.
In all our models we used Adam or AdamW optimizers.
\paragraph{Graph Anomaly Detection}
In this set of experiments, we ran each model with 27 different hyper-parameter configurations and reported the best result for each model. We searched over different learning rates, weight decays, and learning-rate decay factors. In our model, we used four layers, each with a hidden dimension of $256$ and ReLU activation.
\paragraph{Deep weight space alignment}
In this section we took the model of \cite{DWS-align} and added our non-Siamese common layers. We searched for each model and dataset over four hyper-parameters and for each reported the best performance. We trained for $100$ epochs and reported the reconstruction loss.
\paragraph{Wasserstein Distance computation}
In this set we took the model of \cite{Samanta} and added the non-Siamese layers. We trained for $200$ epochs and reported the relative absolute mean error. In most of the experiments we used a learning rate of $1e^{-4}$ and weight decay of $1e^{-5}$.

\section{Absolutely Irreducible Real Representations}\label{app:absolutely}
Here we give some more details on the concept of absolutely irreducible real representations. 

As discussed in the main text, a real irreducible representation $\V$ is called \emph{absolutely irreducible} if its complexification is irreducible over $\CC$. The complexification is denoted by $c\V$. It can be defined by choosing a basis $v_1,\ldots,v_n$ of $\V$ and then defining 
$$c\V=\{\sum_{i=1}^n c_iv_i, \quad c_i \in \CC \} .$$

Let $T:\V \to \V$ be an equivariant non-zero mapping, and assume $\V$ is absolutely irreducible. Then $Tv=\lambda v$ for some real $\lambda$. This is because $T$ can be linearly extended to a non-zero linear mapping $T:c\V \to c\V$. Similarly, the group action of $\G$ which is linear can also be extended to $c\V$. The extension of $T$ to $c\V$ will be equivariant, and therefore, since $c\V$ is irreducible by assumption,  Schur's Lemma for complex representations implies that $T=\lambda I$ for $\lambda \in \CC$. Since the original map $T$ maps $\V$ into itself, for every non-zero
$v\in\V$ we have $Tv=\lambda v\in\V$, which implies that
$\lambda\in\RR$.

\paragraph{Real irreducible permutation representations are absolutely irreducible.}
Classical representation theory results (see e.g., 
\cite{fulton2013representation}) characterize all complex irreducible representations of the permutation group, up to isomorphism. In this process, they show that all these representations can be defined over the rational numbers. This means that, for any complex irreducible representation $\U$ of the permutation group $S_n$, there exists a basis in which for every $g\in \S_n$, the matrix representing the linear mapping defined by $g$ will be rational.

As explained, e.g., in   \cite{overflow}, this fact implies that every irreducible $s$ dimensional real representation $\V$ is absolutely irreducible. Indeed, if $c\V$ were not irreducible over $\CC$, then $c\V=\V_1\oplus \V_2$ where $\V_1$ is an invariant (complex) irreducible subspace, and the dimension $t$ of $\V_1$ satisfies $0<t<s$. In an appropriate basis $u_1,\ldots,u_t$ of $\V_1$ we will have that the matrix representing each $g\in S_n$ is rational. We can write each $u_j$ as
\[
u_j=\sum_{k=1}^s (a_{jk}+ib_{jk})v_k.
\]
Consider the real subspaces
\[
A=\operatorname{span}_{\mathbb R}
\left\{
\sum_{k=1}^s a_{jk}v_k : j=1,\ldots,t
\right\},
\qquad
B=\operatorname{span}_{\mathbb R}
\left\{
\sum_{k=1}^s b_{jk}v_k : j=1,\ldots,t
\right\}.
\]
Since the matrices representing the action of $G$ on $V_1$ in the
basis $u_1,\ldots,u_t$ have real entries, both $A$ and $B$ are
$G$-invariant real subspaces of $V$. Moreover, at least one of them is
non-zero, while both have dimension at most $t<s$. Thus $V$ contains a
non-zero proper invariant real subspace, contradicting its
irreducibility.

\section{graph networks irreducible decomposition}
\label{decomposition}
In the main text, we claim that, for $n\geq 4$, the representation $\RR^{n\times n}$ of the permutation group $S_n$ can be decomposed into a direct sum of seven irreducible representations:
\begin{align*}
    \V_{0} &= \{ a \cdot I_n|a\in \R \}\\
    \V_{1} &= \{a \cdot (1_{n\times n} -  I_n)|a\in \R \}\\
    \V_{2} &= \{ Diag(\alpha_{1},..,\alpha_{n})|  \sum^{n}_{i=1} \alpha_{i} = 0 \}\\
\V_3&=\{r1_n^T|  \sum_{i=1}^n r_i=0 \}\\
\V_4&=\{1_nc^T|  \sum_{i=1}^n c_i=0 \} \\
\V_5&=\{A|  A=-A^T, A1_n=0\}\\
\V_6&=\{A| A=A^T, A1_n=0,  A_{ii}=0, \forall i=1,\ldots,n\}.\\	
\end{align*}

We begin by giving an algorithm describing how every $n$ by $n$ matrix can be decomposed into a sum of matrices from these spaces, with a computational complexity of  $O(n^2)$. This is necessary for a 2-IGN implementation which is based on irreducibles, and is also a first step towards a full proof of \cref{thm:graphs}. This full proof will be presented after the algorithm.

\subsection{Decomposition algorithm}
\label{sub:decomposition}
We will give an algorithm to decompose a matrix $A\in \RR^{n\times n}$ into a sum of matrices from these seven sub-spaces. We assume $n \geq  4$ (actually, when $n=3$ the algorithm works as well, but in this case $\V_6$ is zero dimensional as discussed in the main text). 

The first step of the algorithm is to write 
$$A=B+\begin{pmatrix}
	0 & a & \ldots &a\\
	a & 0 & \ldots &a\\
	\vdots & & & \\
	a & a & \ldots &0\\
\end{pmatrix}
+\begin{pmatrix}
	b &  &  &\\
	 & b &  &\\
	\vdots & & & \\
	 &  & \ldots &b\\
\end{pmatrix}
 $$	
 where the two matrices from the right are in $\V_0$ and $\V_1$, and $a$ and $b$ are the average of off-diagonal and diagonal elements of $A$, respectively, so that the diagonal and off-diagonal elements of $B$ both sum to zero. It remains to find a decomposition for $B$. The crucial part of this is the following lemma
 \begin{lemma}
 Let $n>2$ be a natural number, and let $B\in \RR^{n\times n}$ be a matrix whose diagonal elements sum to zero, and off-diagonal elements sum to zero. Then there exists a matrix $C$ in 
  \begin{equation}\label{eq:hatV}
  \hat \V=\{C\in \RR^{n\times n}| \quad, C1_n=0, C^T1_n=0 \text{ and }  C_{ii}=0, \forall i=1,\ldots,n  \} \end{equation}
  and vectors $r,c,d\in \RR^n$ which all sum to zero, such that
 \begin{equation}\label{eq:mainDecomposition}
 C=B+r1_n^T+1_nc^T+diag(d). 
 \end{equation}
 Moreover, this decomposition of $B$ is unique.
 \end{lemma}

Stated differently, the matrix $B$ will be a linear combination of matrices
$r1_n^T$, $1_nc^T$, and $diag(d)$, which are in $\V_3$, $\V_4$, and $\V_2$,
respectively, and the matrix $C$, which is in $\hat \V$. Once we prove the lemma, we will conclude by showing that a matrix in $\hat \V$ can be written as a sum of matrices in $\V_5$ and $\V_6$, which will conclude the argument.
 \begin{proof}[Proof of the Lemma.]
Let us denote the sum of the $i$-th row and column of $B$ by $r_i^B $ and $c_i^B$ respectively. Denote $d_i^B=B_{ii}$. Note that the vectors $r^B,c^B$, and $d^B$ all sum to zero by our assumption on $B$. 

We need to show there exist vectors $r,d,c\in \RR^n$ satisfying \eqref{eq:mainDecomposition} for an appropriate $C \in \hat \V$. This will occur if and only if  the following equations are satisfied 
 
 \begin{align*}
 \sum_{i=1}^n r_i&=0\\
 \sum_{i=1}^n c_i&=0\\
 \sum_{i=1}^n d_i&=0\\
 r_i+c_i+d_i&=-d^B_i, \quad \forall i=1,\ldots,n\\
 nr_i+\sum_{j=1}^nc_j+d_i&=-r^B_i,  \quad \forall i=1,\ldots,n\\
  nc_i+\sum_{j=1}^nr_j+d_i&=-c^B_i,  \quad \forall i=1,\ldots,n.\\
 \end{align*}
 Here the first three constraints are the requirements that the vectors $r,c,d$ all sum to zero, and the last three constraints (each of which actually represents $n$ constraints) follow from our requirement that the diagonal elements of $C$ will be zero and that the rows and columns of $C$ sum to zero. 
  
Next, we note that since the vectors $r,c,d,r^B,c^B,d^B$ all sum to zero, if the last three equations above are satisfied for $i=1,\ldots,n-1$, they will automatically be satisfied for $i=n$. That is, these equations are equivalent to
 \begin{align*}
 	\sum_{i=1}^n r_i&=0\\
 	\sum_{i=1}^n c_i&=0\\
 	\sum_{i=1}^n d_i&=0\\
 r_i+c_i+d_i&=-d^B_i, \quad \forall i=1,\ldots,n-1\\
 nr_i+d_i&=-r^B_i,  \quad \forall i=1,\ldots,n-1\\
 nc_i+d_i&=-c^B_i,  \quad \forall i=1,\ldots,n-1\\
 \end{align*}
 This gives us a $3n$ linear equations system in $3n$ variables. We will show this equation has a unique solution and compute it explicitly. First, we eliminate the variables $r_n,c_n$ and $d_1,\ldots,d_n$ by setting
 \begin{equation}\label{eq:d}
 	d_i=-d_i^B-r_i-c_i, \quad i=1,\ldots,n-1 \end{equation}
 and 
 \begin{equation}\label{eq:rest}
 r_n=-\sum_{j=1}^{n-1}r_j, \quad c_n=-\sum_{j=1}^{n-1}c_j, \quad d_n=-\sum_{j=1}^{n-1}d_j
 \end{equation}
This ensures that the first four equations above are satisfied, and we are left with the task of choosing $r_1,\ldots,r_{n-1},c_1,\ldots,c_{n-1}$ satisfying the last two equations (which are actually $2n-2$ equations). By replacing the eliminated variables $d_i$ we obtain the equations
\begin{align*}
	 (n-1)r_i-c_i&=-r^B_i+d_i^B,  \quad \forall i=1,\ldots,n-1\\
	(n-1)c_i-r_i&=-c^B_i+d_i^B,  \quad \forall i=1,\ldots,n-1\\
	\end{align*}
	For every $i$, there are two equations in $r_i$ and $c_i$. Since we assume $n\geq 4$, this equation has a unique solution
	\[
    \begin{pmatrix}
    r_i\\
    c_i
    \end{pmatrix}
    =
    \frac{1}{n^2-2n}
    \begin{pmatrix}
    nd_i^B-(n-1)r_i^B-c_i^B\\
    nd_i^B-(n-1)c_i^B-r_i^B
    \end{pmatrix}.
    \]
 And we can derive
 \begin{align*}
	\begin{pmatrix}
	r_i\\
	c_i
	\end{pmatrix}=
    \frac{1}{n^2-2n}
	\begin{pmatrix}
	n-1 & 1\\
	1 & n-1\\	
		\end{pmatrix}
	\begin{pmatrix}
	-r^B_i+d_i^B,\\
	-c^B_i+d_i^B,
	\end{pmatrix}
 \end{align*}
 
 We can then reconstruct $d_1,\ldots,d_{n-1}$ from \eqref{eq:d} and $c_n,r_n,d_n$ from \eqref{eq:rest}. This concludes the proof.
 \end{proof}

Given the proof, it remains to decompose a matrix $C\in \hat{\V}$ into a sum of two matrices in $\V_5, \V_6$. This can be simply done by writing 
$$C=\frac{1}{2}(C+C^T)+\frac{1}{2}(C-C^T) $$
and noting that the matrices $\frac{1}{2}(C-C^T)$ and $\frac{1}{2}(C+C^T)$ are in $\V_5$ and $\V_6$. 

\subsection{isomorphism types}
We now restate and prove the theorem on 2-IGN stated in the main text
\graphs*
\begin{proof}
The decomposition given in Subsection \ref{sub:decomposition} is unique when $n\geq 4$, and thus we see that 
$$\RR^{n \times n}=\V_0\oplus \ldots \oplus \V_6.  $$
It is also clear that all the spaces $\V_i$ are invariant, that $\V_0,\V_1$ are isomorphic to the irreducible trivial space, and $\V_2,\V_3,\V_4$ are isomorphic to the irreducible space $\V(n)$. It remains to show: (a) That $\V_5,\V_6$ are not zero-dimensional, (b) that they are irreducible, and (c) that both $\V_5$ and $\V_6$ are not isomorphic to any of the other spaces.

We handle (a) by computing the dimension of $\V_5$ and $\V_6$. We note that a matrix $A \in \V_5$ can be parameterized by freely choosing all upper diagonal elements $A_{ij}$ with $j-i>1$. The elements $A_{ji}$ with $j-i>1$ are then determined  by the constraint $A_{ji}=-A_{ij}$, and then the coordinates $A_{12}=-A_{21},A_{23}=-A_{32},\ldots$ are determined recursively by the constraint that the first $n-1$ rows of $A$ will sum to zero. The obtained matrix $A$ satisfies $A=-A^T$ and its first $n-1$ rows sum to zero. It follows from $A=-A^T$ that the sum of all elements of $A$ is zero and hence the last row of $A$ must also sum to zero and $A$ is indeed in $\V_5$.

To summarize, since $\V_5$ is parameterized by  the upper diagonal elements $A_{ij}$ with $j-i>1$ we deduce that 
$$\dim(\V_5)=(n^2-n)/2-(n-1)=\frac{n^2-3n}{2}+1. $$
We compute the dimension of $\V_6$ by noting that $\hat \V$ from \eqref{eq:hatV} is a direct sum of $\V_5$ and $\V_6$. The space $\hat \V$ can be parameterized by choosing an $(n-1)\times (n-1)$ matrix $A$ with zero on the diagonal, and whose elements sum to zero. There is then a unique extension of this matrix to an $n$ by $n$ matrix $\hat A$ which is in $\hat \V$ and satisfies $\hat A_{ij}= A_{ij}$ for all $1\leq i,j \leq n-1$. Accordingly, the dimension of $\hat \V$ is $(n-1)^2-(n-1)-1=n^2-3n+1$ and 
$$\dim( \V_6)=\dim(\hat \V)-\dim(\V_5)=\frac{n^2-3n}{2} .$$
It follows that when $n\geq 4$, both $\V_5$ and $\V_6$ have positive dimension. Moreover, $\V_5$ and $\V_6$ are not isomorphic as they do not have the same dimension. The same argument shows that $\V_6$ is not isomorphic to any of the other spaces $\V_i$, for $i=0,\ldots,4$, since they do not have the same dimension for any $n$. In contrast $\V_5$ does have the same dimension as $\V_2$
when $n=4$, as we get
$\dim(\V_5)=3=\dim(\V_2)$. Nonetheless, the spaces are still not isomorphic since the action of $S_n$ on the spaces is different. For example, the subspace of fixed points of the permutation $\tau=(1,2)$ in the space $\V_2\cong \V(n)$ is all $n=4$ dimensional vectors $x$ which sum to zero and satisfy $x_1=x_2$. This space is two dimensional. In contrast, one can verify that the subspace of fixed points of $\tau$ in $\V_5$ is one dimensional. Hence they cannot be isomorphic.

It remains to show that $\V_5$ and $\V_6$ are indeed irreducible. Let $\tilde \V_5 \subseteq \V_5$ and $\tilde \V_6 \subseteq \V_6$ be non-zero invariant subspaces. We need to show that $\tilde \V_j=\V_j$ for $j=5,6$. To show this, we consider the space 
$$\U=\V_0 \oplus \V_1 \oplus \ldots \V_4 \oplus \tilde \V_5 \oplus \tilde \V_6. $$
It is sufficient to show that $\U=\RR^{n\times n}$. Since $\U$ contains all diagonal matrices which are given by $\V_0\oplus \V_2 $, and since it is an invariant subspace, it is sufficient to show that the  matrix $E^{1,2}$ which is zero in all coordinates except for a single coordinate $E_{1,2}^{1,2}=1 $, is in $\U$. 

To show that $E^{1,2}$ is in $\U$,  let us choose a non-zero matrix $A \in \tilde {\V}_5 $ and $\B \in \tilde {\V}_6$. Both matrices have a non-zero off diagonal matrix element, and by applying an appropriate permutation and rescaling we can assume that $A_{12}=1=B_{12} $. Next, let us average over the group $$stab(1,2)=\{\tau\in S_n, \tau(1)=1, \tau(2)=2 \} $$
to obtain new matrices
$$\hat A=\frac{1}{|stab(1,2)|}\sum_{\tau\in stab(1,2)} \tau \cdot A, \quad \hat B=\frac{1}{|stab(1,2)|}\sum_{\tau\in stab(1,2)} \tau \cdot B  $$
Note that $\hat A$ (respectively $\hat B$) is in $\tilde{\V}_5$ (respectively $\tilde{\V}_6$) and satisfy $\hat A_{12}=1=\hat B_{12}$ and
$$\hat A_{1j}= \hat A_{1k}, \hat B_{1j}= \hat B_{1k}  \forall j,k>2 $$
and 
$$\hat A_{jk}=\hat A_{st}, \hat B_{jk}=\hat B_{st} $$
for all $j,k,s,t$ which are all larger than $2$, and such that $j\neq k, s\neq t$. Defining for convenience
$$a=\frac{1}{n-2}, b=\frac{2}{(n-2)(n-3)} ,$$
it follows that 
\begin{equation*}
\hat A=
\begin{pmatrix}
0 & 1& -a& \ldots & -a\\
-1& 0& a& \ldots & a\\
a & -a & 0 & \ldots & 0\\
\vdots & & & &\\
a & -a & 0 & \ldots & 0
\end{pmatrix}
\quad
\hat B=
\begin{pmatrix}
0 & 1& -a& \ldots & -a\\
1& 0& -a& \ldots & -a\\
-a & -a & b & \ldots & b\\
\vdots & & & &\\
-a & -a & b & \ldots & b
\end{pmatrix}
-diag(0,0,b,\ldots,b)
\end{equation*}
It is not difficult to show that any matrix which is constant along the rows is in $\V_0\oplus \ldots \oplus \V_4 \subseteq \U $. The same is true for any matrix which is constant along the columns. Let $C(i)$ denote the matrix with $C_{ij}(i)=1$ for all $ j$ and $C_{kj}(i)=0$ for all $j$ and all $k\neq i$. Then 
\begin{align*}\tilde A:=&\hat A+aC(1)-aC(2)-aC^T(1)+aC^T(2)\\
&= 
\begin{pmatrix}
0 & 1+2a& 0& \ldots & 0\\
-1-2a& 0& 0& \ldots & 0\\
0 & 0 & 0 & \ldots & 0\\
\vdots & & & &\\
0 & 0 & 0 & \ldots & 0
\end{pmatrix}\\
\tilde B:=& \hat B+diag(0,0,b,\ldots,b)-b1_{n\times n}+(a+b)(C(1)+C(2)+C^T(1)+C^T(2))\\
&=
\begin{pmatrix}
2a+b & 1+2a+b& 0& \ldots & 0\\
1+2a+b& 2a+b& 0& \ldots & 0\\
0 & 0 & 0 & \ldots & 0\\
\vdots & & & &\\
0 & 0 & 0 & \ldots & 0
\end{pmatrix}
\end{align*}
where both $\tilde A$ and $\tilde B$ are in $\U$. Since $a,b$ are positive numbers, we can take the following linear combination which zeros out the $(2,1)$ entry and sets the $(1,2)$ entry to one:
$$\frac{1}{2}\left(\frac{1}{1+2a}\tilde A+ \frac{1}{1+2a+b}\tilde B \right)=\begin{pmatrix}
c & 1& 0& \ldots & 0\\
0& d& 0& \ldots & 0\\
0 & 0 & 0 & \ldots & 0\\
\vdots & & & &\\
0 & 0 & 0 & \ldots & 0
\end{pmatrix} $$
for appropriate $c,d$. Since diagonal matrices are in $\U$, we can remove the diagonal matrix $diag(c,d,0,\ldots,0)$ to finally obtain that the matrix $E^{1,2}$ is in $\U$. 
\end{proof}

\subsection{Small $n$}\label{app:small}
 As discussed in \cite{finzi,pearce2022connecting}, when $n<4$ the dimension of the space of equivariant mappings is smaller than 15: it is $14$ for $n=3$ and $8$ for $n=2$. We now explain this according to our derivation. 

For $n=3$, the space $\V_6$ is zero-dimensional. In this case, there are only $2^2+3^2+1=14$ equivariant linear mappings. When $n=2$, both $\V_5$ and $\V_6$ are zero dimensional. Additionally, we can directly verify that in this case, we can "drop" $\V_4$ as well because $\RR^{2\times 2}=\V_0\oplus \V_1 \oplus \V_2\oplus \V_3$. Since $\V_0 \cong \V_1$ and $\V_2 \cong \V_3$ we have  $2^2+2^2=8 $ equivariant linear mappings.

\newpage
\section{Proof for Deep Weight Space decomposition}\label{app:dws}
In this section we fill in some details on the tensor product of representations which were omitted from the proof of \cref{thm:dws} in the main text. 

At the base of this discussion is the following lemma \citep{serre1977representations}:
\begin{lemma}
Let $\G_1,\G_2$ be finite groups, and let $\V_1,\V_2$ be (real or complex) finite dimensional absolutely irreducible representations, with an action denoted by $\rho_1,\rho_2$. Then the action $\rho_1\otimes \rho_2 $ on $\V_1\otimes \V_2$ is an absolutely irreducible representation of $\G_1 \times \G_2 $.
\end{lemma}

Firstly, we explain the decomposition of $\hat \W_m$ into irreducibles. We note that the representation $\hat \W_m$ of $\G$ can be thought of as a representation of $S_{d_{m-1}}\times S_{d_m} $, which is a tensor product of the representation $\RR^{d_{m-1}}$ of $S_{d_{m-1}}$  and the representation $\RR^{d_m}$ of $S_{d_m}$. Therefore \citep{serre1977representations}, its irreducible decomposition is given by the tensor product of the irreducible decomposition of its factors, namely
 \begin{align*}
\hat \W_m &\cong \RR^{d_{m-1}}\otimes \RR^{d_{m}} \cong (\Scalar \oplus \Vector(d_{m-1})) \otimes (\Scalar \oplus \Vector(d_{m}))\\
& \cong (\Scalar \otimes \Scalar) \oplus (\Scalar \otimes \Vector(d_{m})) \oplus (\Vector(d_{m-1}) \otimes \Scalar) \oplus (\Vector(d_{m-1}) \otimes \Vector(d_{m}))\\
&\cong \Scalar \oplus  \Vector(d_{m}) \oplus \Vector(d_{m-1})  \oplus \Matrix_m.
 \end{align*}
All components in the decomposition are irreducible as a tensor product of irreducibles. 

We also need to show that all irreducibles in the decomposition are absolutely irreducible. For $\Scalar$ this is obvious. For all representations  $ \Vector(d_{m}) $ this follows from the fact that these representations can be identified with representations of the permutation group $S_{d_{m}}$, which are always absolutely irreducible. To see that $\Matrix_m $ is absolutely irreducible, we note that
$\Matrix_m$ is the set of $d_{m}\times d_{m-1} $ real matrices whose rows and columns all sum to zero, and its complexification $c\Matrix_m$ is the set of \emph{complex} matrices whose rows and columns all sum to zero. $c\Matrix_m$ is the tensor product of $ c\Vector(d_{m-1}) $ and $ c\Vector(d_{m}) $, the complexification of $\Vector(d_{m-1}) $ and $ \Vector(d_{m}) $, respectively. Since these spaces are irreducible, so is their tensor product.

\subsection{Decomposition of weight spaces}  \label{app:Wdecompose}
We explain here how each matrix $A \in \W_m, m=2,\ldots,M-1$ can be written as a sum of elements from its irreducible decomposition. For convenience denote $a=d_{m}, b=d_{m-1} $. We want to write $A$ as a sum of elements from the spaces
$$\Scalar \cong \{c1_{a\times b} \}, \quad \V_{m-1}\cong \{1_ac^T| c\in \RR^b, c^T1_b=0 \}, \quad \V_{m}\cong \{r1_b^T| r\in \RR^a, r^T1_a=0 \} $$
and 
$$\Matrix_m=\{C\in \RR^{a\times b}, C1_b=0_b, C^T1_a=0_a \} $$

As a first step, we take our given $A\in \W_m$ and write it as 
$$A=\bar{A}1_{a\times b}+(A-\bar{A}1_{a\times b}) $$
where $\bar{A}$ is the average of all elements of $A$, and $\bar{A}1_{a\times b}$ is in $\Scalar$. Next, we need to decompose $(A-\bar{A}1_{a\times b}) $ which we denote by $B$. Note that $\sum_{ij}B_{ij}=0$. 

Let $r$ and $c$ be the vectors describing the average of the rows and columns of $B$, respectively: 
$$r=\frac{1}{b}B1_b, \quad c=\frac{1}{a}B^T1_a $$
Note that $r^T1_a=\frac{1}{b}\sum_{ij}B_{ij}=0 $ and similarly $c^T1_b=0$ and thus $1_ac^T\in \V_{m-1} $ and $r1_b^T \in \V_m $. Now let us define
$$C=B-1_ac^T- r1_b^T$$
We can directly verify that $C1_b=0_b$ and $C^T1_a=0_a$. Therefore $C$ is in $\Matrix_m$ and we obtained the decomposition we wanted.
 \newpage
\section{Proofs for Wreath Products}\label{app:wreath}
To conclude the proof of \cref{thm:wreath_irreps}, we need to prove the following lemma
\begin{lemma}
\label{power_irred}
Let $\V$ be an irreducible representation of $\G$, and assume the action of $\G$ on $\V$ is not trivial. Then $\V^{ k}$ is an irreducible representation of $\G \wr S_k$. 
\end{lemma}
\begin{proof}

The space $\V^{ k}$ is invariant. Let $\W \subseteq \V^{ k}$ be a non-zero $G\wr S_k$ invariant subspace. We need to show that $\W=\V^{ k}$. 

We first claim that there exists an element in $\W$ of the form $(u,0,...,0)$ with $ u \neq 0$.
Let $(v_{1},v_{2},..,v_{k})$ be a non-zero element in $\W$. By applying a permutation if necessary, we can assume without loss of generality that  the first element is non-zero.  If $v_{i}=0, \forall 2\leq i \leq k$, we are done. Otherwise, there exists some $g\in G$ such that $gv_{1}\neq v_{1}$ and so 
\begin{align*}
    (g\cdot v_{1},v_{2},..,v_{k})- ( v_{1},v_{2},..,v_{k})= (gv_{1} - v_{1},0,..,0)\in \W
\end{align*}

The space $\{v\in \V|  (v,0,..,0) \in \W\}$ is invariant under the action of $G$,  and we just saw it is not zero. Since $\V$ is irreducible, this space equals $\V$, and thus $\W$ contains $\V\oplus0...\oplus 0$. By $S_{k}$ invariance it follows that $ \W $ contains all $k$-tuples with a single non-zero entry, and since such tuples span all of $\V^k$ we deduce that $\W=\V^k$. 
\end{proof}

\wreath*
\begin{proof}
It is straightforward to check that every mapping of the form \eqref{eq:form} is equivariant. We would like to prove the converse statement.

It is also straightforward to verify that the space spanned by the non-Siamese  mappings 
$$(v_1,\ldots,v_k)\mapsto(\sum_{\ell=1}^k \langle v_\ell,e_i \rangle e_j,\ldots,\sum_{\ell=1}^k \langle v_\ell,e_i \rangle e_j $$
does not depend on the choice of the basis $e_1,\ldots,e_s$ for $\V_{fixed}$. In particular, we can assume without loss of generality that $e_1,\ldots,e_s$ are an \emph{orthonormal} basis for $\V_{fixed}$.

Note that $\V$ can be written as a direct sum of two $\G$ invariant spaces $\V=\V_{fixed}\oplus \V_{\perp} $, where $\V_{\perp}$ is the space of vectors in $\V$ which are orthogonal to $\V_{fixed}$ with respect to the given $\G$ invariant inner product. It follows that $\V^k=\V_{fixed}^k\oplus \V_{\perp}^k $ is a decomposition of $\V^k$ into two $\G^k$ invariant spaces.

In this situation, we know that any linear equivariant $L:\V^k \to \V^k$ can be written as $L=L_{fixed}+L_{\perp} $, where $L_{fixed}:\V_{fixed}^k \to \V_{fixed}^k  $ and 
$L_{\perp}:\V_{\perp}^k \to \V_{\perp}^k  $ are equivariant. We know that there are no non-zero linear equivariant maps from
$\V_{fixed}^k$ to $\V_{\perp}^k$ or vice versa, since
$\V_{fixed}^k$ is a direct sum of trivial $\G^k$-representations,
whereas $\V_{\perp}^k$ contains no trivial $\G^k$-subrepresentation.

Let $L:\V_{\perp}^K \to \V_{\perp}^k  $ be a linear equivariant map. We want to show that it is a 'Siamese mapping'. Denote 
$$L(v_1,\ldots,v_k)=(L_1(v_1,\ldots,v_k), \ldots,L_k(v_1,\ldots,v_k))  ,$$
We define $\hat L:\V_{\perp} \to \V_{\perp}$ by 
$$\hat L(v)=L_1(v,0,\ldots,0). $$
Note that $\hat L$ is $\G$ equivariant because 
$$\hat L(gv)=L_1(gv,0,\ldots,0)=g L_1(
v,0,\ldots,0)=g\hat L(v), \quad \forall g\in \G .$$
Next, note that for every $v_2,\ldots,v_k \in \V_{\perp}$, we have that 
\[
L_1(0,v_2,\ldots,v_k)
=
L_1(g\cdot 0,v_2,\ldots,v_k)
=
g\cdot L_1(0,v_2,\ldots,v_k).
\]
Thus,
\[
L_1(0,v_2,\ldots,v_k)\in \V_{fixed}.
\]
On the other hand, since $L:\V_{\perp}^k\to\V_{\perp}^k$, we also have
\[
L_1(0,v_2,\ldots,v_k)\in \V_{\perp}.
\]
Therefore,
\[
L_1(0,v_2,\ldots,v_k)
\in \V_{fixed}\cap\V_{\perp}
=
\{0\}.
\]
We deduce that 
\[
L_1(v_1,\ldots,v_k)
=
L_1(v_1,0,\ldots,0)
=
\hat L(v_1).
\]
By $S_k$ equivariance we can deduce that 
\[
L_j(v_1,\ldots,v_k)
=
\hat L(v_j),
\]
and we can also show, as we did for $L_1$, that 
\[
L_j(v_1,\ldots,v_k)
=
L_j(0,\ldots,0,v_j,0,\ldots,0).
\]
Thus $L$ is a Siamese mapping induced by $\hat L$, that is,
\[
L(v_1,\ldots,v_k)
=
\left(\hat L(v_1),\ldots,\hat L(v_k)\right).
\]

Now, let us consider linear equivariant mappings from $\V_{fixed}^k$ to itself. Denote
\[
\Scalar_i=\{ce_i : c\in\RR\},
\]
and note that an irreducible decomposition of $\V_{fixed}$ is given by  
\[
\V_{fixed}
=
\Scalar_1\oplus\cdots\oplus\Scalar_s.
\]
Accordingly, $\V_{fixed}^k$ can be written as a direct sum of the invariant subspaces $\Scalar_i^k$ for $i=1,\ldots,s$. Since $\G^k$ acts trivially on $\V_{fixed}^k$, the action of $\G\wr S_k$ on each $\Scalar_i^k$ reduces to the action of $S_k$, which is isomorphic to the standard permutation action of $S_k$ on $\RR^k$. The isomorphism
\[
\psi_i:\Scalar_i^k\to\RR^k
\]
is given by
\[
\psi_i(v_1,\ldots,v_k)
=
\left(
\langle v_1,e_i\rangle,\ldots,
\langle v_k,e_i\rangle
\right).
\]
Note that, since we assumed without loss of generality that $e_1,\ldots,e_s$ is an orthonormal basis, $\psi_i$ is zero on all other components of $\V_{fixed}^k$. Similarly, we define an isomorphism
\[
\tilde\psi_i:\RR^k\to\Scalar_i^k
\]
by 
\[
\tilde\psi_i(x_1,\ldots,x_k)
=
(x_1e_i,\ldots,x_ke_i).
\]

As discussed in \cref{sub:deepsets}, an equivariant mapping from $\RR^k$ to itself can be written as a linear combination of the identity mapping $x\mapsto x$ (which is a Siamese mapping), and the mapping 
\[
Tx
=
\left(
\sum_{\ell=1}^k x_\ell,\ldots,
\sum_{\ell=1}^k x_\ell
\right).
\]
Accordingly, an equivariant mapping from $\Scalar_i^k$ to $\Scalar_j^k$ is a linear combination of a Siamese map and the mapping
$\tilde\psi_j\circ T\circ\psi_i$, which is given by 
\[
\tilde\psi_j\circ T\circ\psi_i(v_1,\ldots,v_k)
=
\left(
\sum_{\ell=1}^k\langle v_\ell,e_i\rangle e_j,\ldots,
\sum_{\ell=1}^k\langle v_\ell,e_i\rangle e_j
\right).
\]
This concludes the proof.
\end{proof}

\subsection{Beyond $S_k$}\label{app:beyond}
We now generalize the results from Theorem \ref{characterization} to the case where $S_k$ is replaced with any subgroup $H\leq S_k$ which acts on $\{1,\dots,k\}$ transitively:
\begin{theorem}[Generalization of Theorem \ref{characterization}]
Let $\V$ be a real representation of a finite group $\G$. and
let $e_1,\ldots,e_s$ be a basis of the subspace 
$\V_{fixed}$. Let $\langle \cdot, \cdot \rangle$ be a $\G$ invariant inner product on $\V$.

Let $H$ be a subgroup of $ S_k$ which acts on $\{1,\dots,k\}$ transitively. Let 
$T_1, T_2,\ldots,T_h $ be a basis for the $H$-equivariant mappings from $\RR^k$ to itself, where $T_h $ is the identity mapping.

Then every $\G \wr H $ linear equivariant map $L:\V^{k} \to \V^{k} $ is of the form
\begin{align*}
L(v_1,\ldots,v_k)&=\sum_{\ell=1}^{h-1}\sum_{i,j=1}^s a_{ij\ell} \cdot (\tilde \psi_j \circ T_\ell \circ \psi_i)(v_1,\ldots,v_k)\\
&+ \left(\hat{L}(v_1),\ldots ,\hat{L}(v_k) \right)\end{align*}
where $\hat L:\V \rightarrow \V$ is a linear equivariant map, and $a_{ij\ell}$ are real numbers. Conversely, every linear mapping of the form defined in is equivariant. 
\end{theorem}
The proof is identical to the proof of \cref{characterization}.

\addtocontents{toc}{\protect\setcounter{tocdepth}{1}}
\printbibliography[
    heading=subbibintoc,
    title={References}
]

\endgroup

\end{refsection}
\chapter{Monotone and Separable Set Functions: Characterizations and Neural Models}
\label{chap:paper3}

\noindent
\textbf{Soutrik Sarangi, Yonatan Sverdlov, Nadav Dym, and Abir De.}\\
Advances in Neural Information Processing Systems (NeurIPS), 2025.\\
Soutrik Sarangi and Yonatan Sverdlov contributed equally.

\bigskip


\begin{refsection}[papers/paper3/refs.bib]

\begingroup

\graphicspath{{papers/paper3/}}


\def\R{\mathbb{R}}
\def\RR{\mathbb{R}}
\def\NN{\mathbb{N}}
\def\B{\mathcal{B}}
\def\Bcal{\mathcal{B}}
\def\W{\mathbb{W}}
\def\Wcal{\mathcal{W}}
\def\S{\mathcal{S}}
\def\St{\mathcal{S}}
\def\Gcal{\mathcal{G}}
\def\Mcal{\mathcal{M}}

\def\Ps{\mathcal{P}_{<\infty}}
\def\Pvinf{\mathcal{P}_{<\infty}(V)}
\def\Pvk{\mathcal{P}_{\leq k}(V)}

\def\Exp{\mathbb{E}}
\def\Prob{\mathbb{P}}
\def\Unif{\mathrm{Unif}}
\def\Normal{\mathcal{N}}

\def\mas{MAS\xspace}
\def\inexact{inexact\xspace}
\def\our{\textsc{MASNet}\xspace}

\def\mm{m^{\star}}
\def\model{F}

\def\mtwo{\mathrm{M}_2}
\def\mone{\mathrm{M}_1}
\def\mtwoneural{M_{\theta_2}}
\def\moneneural{M_{\theta_1}}

\def\hhat{\text{Hat}\xspace}
\def\hatmas{\our-\hhat}
\def\integralmas{\our-INT\xspace}
\def\relumas{\our-{\sf ReLU}\xspace}

\def\tri{\mathrm{TRI}}
\def\trimas{\our-$\tri$}

\def\relu{\mathrm{ReLU}}
\def\emd{\mathrm{EMD}}
\def\act{\sigma}
\def\das{\Delta}

\def\ba{\bm{a}}
\def\bx{\mathbf{x}}
\def\bv{\mathbf{v}}
\def\bq{\mathbf{q}}
\def\bk{\mathbf{k}}
\def\boldA{\bm{A}}
\def\boldb{\bm{b}}

\def\tran{^\top}
\def\dash{^\prime}
\def\dif{\,\mathrm{d}}

\def\leftend{\gamma_1}
\def\rightend{\gamma_2}

\def\sizes{M}
\def\sizet{N}
\def\twok{2^{k}}
\def\labelst{y(S,T)}

\def\set#1{\left\{#1\right\}}
\def\curly#1{\left\{#1\right\}}
\def\round#1{\left(#1\right)}
\def\sbrac#1{\left[#1\right]}
\def\abs#1{\left|#1\right|}
\def\tuple#1{(#1)}

\def\indic#1{\mathbf{1}_{\{#1\}}}
\def\supp{\text{supp}}

\def\lms{\{\!\!\{}
\def\rms{\}\!\!\}}

\def\Id{\mathbb{I}_d}
\def\eg{{\em e.g.}}
\def\ie{{\em i.e.}}
\def\mhs{MHS\xspace}

\DeclarePairedDelimiter{\norm}{\lVert}{\rVert}

\let\citet\textcite
\let\citep\parencite


\begin{abstract}

Motivated by applications for set containment problems, we consider the following fundamental problem: can we design set-to-vector functions so that the natural partial order on sets is preserved, namely $S\subseteq T \text{ if and only if } F(S)\leq F(T) $. We call functions satisfying this property \emph{Monotone and Separating (MAS)} set functions.  
We establish lower and upper bounds for the vector dimension necessary to obtain MAS functions, as a function of the cardinality of the multisets and the underlying ground set. In the important case of an infinite ground set, we show that MAS functions do not exist, but provide a model called \our\ which provably enjoys a relaxed MAS property we name ``weakly MAS'' and is stable in the sense of Holder continuity. We also show that MAS functions can be used to construct universal models that are monotone by construction and can approximate all monotone set functions. 
Experimentally, we consider a variety of set containment tasks. The experiments show the benefit of using our \our\ model, in comparison with standard set models which do not incorporate set containment as an inductive bias. Our implementation is available in \url{https://github.com/structlearning/MASNET}.
\end{abstract}

\section{Introduction}
\label{sec:Intro}
A multiset $\lms x_1,\ldots,x_n \rms$ is an unordered collection of vectors, where order does not matter (like sets) and repetitions are allowed (unlike sets). In recent years, there has been increased interest in neural networks that can map multisets to vectors, with applications for physical simulations \cite{huang2021geometrybackflowtransformationansatz}, processing point clouds \cite{qi2017pointnet}, and graph neural networks \cite{xu2018how}. 
Another important application of multiset-to-vector maps is set-containment search~\cite{roy2023locality,singh2020explaining,engels2023dessert}. Here the goal is to check whether a given multiset $S$ is (approximately)  a subset of $T$, and this is often carried out by learning a multiset-to-vector mapping $F$, and then checking whether  $F(S)\leq F(T)$ (element-wise), in which case, one deduces that $S$ is a subset of $T$. 
In this paper, we look into this problem from a theoretical perspective.

\paragraph{Monotone and Separable functions}We begin our analysis with some definitions: we say that a function $F$ mapping sets to vectors is \emph{monotone} if  $S \subseteq  T \subseteq V$ implies that $F(S)\leq F(T)$, and we will say that $F$ is \emph{separable} if $F(S)\leq F(T)$ implies that $S \subseteq T$. When  $F$ is simultaneously \emph{monotone and separable}, we call it a MAS function. When $F$ is a MAS function, we can safely test whether $F(S)\leq F(T) $, and this will be fully equivalent to checking whether $S\subseteq T$. 

\paragraph{Motivation for MAS functions}It is natural to ask why we're interested in functions that are MAS instead of just being monotone or just separable. For this, we begin with a couple of real-life applications based on the set-containment task: (I) In recommendation system design, one problem is to recommend item having a particular set of features, $S$. Here, we represent each item as a set of corresponding features $T$, and the problem is to find all sets $T$ such that approximately, $S \subseteq T$. (II) Text entailment where we are given a small query sentence $q$ and the goal is to find the set of corpus items $c$ from a large corpus $C$, where $q \implies c$ or $c$ entails $q$. Here, $q$ and $c$ are typically represented as sets of contextual embeddings ($S$ for $q$ and $T$ for $c$) and the entailment problem can be cast as the problem of checking if $S \subset T$.

In a neural network setting, such problems require us to design a function $F$ such that the "order" between $F(S)$ and $F(T)$ can serve as a boolean test for whether $S$ is a subset of $T$. For this, $F$ needs to satisfy two conditions: (A) $F(S) \leq F(T)$ implies $S \subseteq T$ (B) $F(S) \not \leq F(T)$ implies $S \not \subseteq T$. Here, we like to emphasize that, for high accuracy, $F$ needs to satisfy both conditions A and B as above. Monotone functions satisfy condition (B) without necessarily satisfying condition (A). As a result, if we predict $S \subseteq T$ based on $F(S) \leq F(T)$ when $F$ is monotone but not separable, it will give large number of false positives. Similarly, if $F$ is separable but not monotone, then $F$ satisfies condition (A) without satisfying condition (B). Thus, using such for checking $S \subseteq T$ results in a large number of false negatives. Hence, both monotonicity and separability are necessary to develop an accurate test for set-containment

\paragraph{Related works}
Monotone set functions have been widely studied in the theory of capacities~\cite{choquet1953capacities}, fuzzy measures~\cite{agahifuzzy,monotonefuzzy}, game theory and economics~\cite{grabisch_set_func_game}, combinatorial auctions~\cite{lehmann2001combinatorial,dobzinski2005approximation,feige2009maximizing}, and learning theory~\cite{balcan2012learning,you2017deep,blum1988training,odonnell2003learning,bshouty1993exact,li2025monotonic}. These works primarily focus on scalar-valued set functions or vector-valued models without structural order constraints.

To the best of our knowledge, jointly monotone and separable multiset (MAS) functions, i.e., multiset-to-vector maps that preserve the partial order induced by multiset containment—have not been previously formalized. Classical results on partially ordered sets, such as order dimension~\cite{dushnik1941partially}, provide relevant mathematical background but do not address learnable or differentiable constructions. Our goal is to characterize the existence of MAS functions and develop models that satisfy MAS constraints by construction for multiset containment tasks.

This work is also related to injective multiset representations. Foundational results on permutation-invariant functions include~\cite{zaheer2018deepsets,xu2018how}, with subsequent work analyzing the latent dimension required for injectivity~\cite{amir_finite,wagstaff2019limitations,wang2023polynomial} and proposing differentiable injective or bi-Lipschitz embeddings~\cite{amir2024fourierslicedwassersteinembeddingmultisets,balan2022permutation,sverdlov}. In this work, we study MAS functions, which impose strictly stronger constraints than injectivity (see Subsection~\ref{subsec:inj}).

 An additional goal of this work is stability: in most learning scenarios, we are looking for $S$, which is only approximately a subset of $T$. Accordingly, we would like to design functions $F$ that are not only MAS but also stable, in the sense that when $S$ is approximately a subset of $T$, then  $F(S)$ is approximately dominated by $F(T)$. 

\paragraph{Summary of our goal}
In summary, in this work, we aim to characterize (\mas) set functions that output finite-dimensional set representations and subsequently design neural networks for such functions.  At a high level, we seek to address the following questions:
\textbf{(1)} What are the conditions for existence of a finite dimensional \mas functions?
\textbf{(2)} What are the permissible relaxations to monotonicity and separability if the existential conditions identified in (1) are not satisfied?
\textbf{(3)} What are the possible neural architectures for these functions? \textbf{(4)} Are these functions stable?
We expect that the design of trainable models with built-in guarantees of monotonicity, separability, and stability will significantly enhance the inductive bias for set containment applications. 

\vspace{-4pt}
\subsection{Main Results}
\vspace{-1mm}
 
We address the goal specified in the previous subsection by providing a detailed characterization and neural architecture for monotone and separable set functions. Our main results are as follows: 

\paragraph{Existential characterization of \mas functions}
Our first objective is to determine whether it is possible to obtain a \mas\ set function with finite output dimension $m$. We begin our analysis in the case where set elements are assumed to be taken from a finite ground set $V$, and show that in this case a \mas function exists if, and only if,  $m \geq |V|$. Next, we add the assumption that the cardinality of the input multisets is bounded by $k$, and provide lower and upper bounds for $m$ in this case. When the ground set is infinite,  we show that  MAS set functions do not exist.

\paragraph{Weakly MAS functions} 
In most applications, the ground set is $V = \mathbb{R}^D$, and our analysis shows that in this case, MAS  functions do not exist in general. To address this, we introduce the notion of weakly MAS functions by extending the set functions to parametric set functions $F(S, w)$, where $w$  is a parameter. Weakly  MAS functions require that (i) $F(:,w)$ is monotone, for all $w$, and (ii) for any $S \not\subset T$, there exists at least one $w \in \mathcal{W}$ such that $F(:, w)$ separates $S$ and $T$. We explain how, by choosing suitable activations, it is possible to construct simple deep sets \cite{zaheer2018deepsets} models which are weakly MAS, and name the resulting model \our.

\paragraph{Stability}
We define the notion of stability as discussed above, separability notions are Boolean and fail to capture graded distance. We would like to guarantee that $F(T)$ is 'almost' larger than $F(S)$ when $S$ is 'almost' a subset of $T$. To address this, we propose a novel asymmetric set distance and present a Holder separability condition~\cite{davidson2025holderstabilitymultisetgraph} which ensures stability in terms of this asymmetric distance.

\paragraph{Monotone Functions} Finally, we show that  MAS functions can be used to provide universal models for computing set-to-vector functions. Thus, MAS functions may be a useful concept also outside of the set containment setting. 

\paragraph{Experiments}  We provide a number of experiments showing that for set containment problems, using our weakly MAS model leads to stronger results in comparison with standard multiset models like DeepSets and SetTransformer, which do not incorporate MAS considerations as an inductive bias.   

\paragraph{Summary of contributions} Our {main contributions} in this paper are:
\begin{enumerate*}[label=\textbf{(\arabic*)}]
   \item We introduce the novel notion of \emph{monotone and separating} (MAS) multiset  functions.
   \item We discuss lower and upper bounds for the embedding dimension for MAS functions, and suggest a model \our, which is weakly MAS.
   \item We show the stability of \our.
   \item Experimentally, we prove the effectiveness of \our for set containment tasks.
\end{enumerate*}
\vspace{-5pt}
\section{Existential characterization of MAS functions}\label{sec:exist}
\vspace{-4pt}
We begin by stating the notation we will use for our paper.

\paragraph{Notation} We denote by $V$ the ground set and by $\Ps(V)$ the collection of all multisets from $V$ with finite cardinality $S=\lms x_1, \cdots, x_s\rms$  where $x_i\in V$.  We denote the space of all multisets with at most $k$ elements by $\Pvk$. 
We use $[d]$ to denote the set $\set{1, \cdots, d}$. Given two vectors $v, u \in \R^d$, we say $v\le u$  if $v[i] \le u[i]$ for all $i\in [d]$. Also, for an element $v \in V$ and a multiset $S \in \Ps(V)$, we use $c_S(v)$ to denote the number of times $v$ occurs in $S$. We say $S\subseteq T$ if $c_S(v) \le c_T(v)$ for all $v \in V$. 

\paragraph{Proofs}  All proofs for the results stated in the paper are provided in Appendix~\ref{app:proofs}.

In this section, our goal is to understand when an MAS function $F:\Pvk \to \RR^m$ or $F:\Ps(V)\to \RR^m$ exists, and if it does, what is the smallest $m$ for which such a mapping exists.  
We denote this minimal dimension as $\mm(V, k)$, and $\mm(V,\infty) $, respectively. 
\vspace{-6pt}
\subsection{Prologue: Relation to injectivity}
\vspace{-6pt}
\label{subsec:inj}
Our inquiry is related to the notion of injective multiset functions, as every MAS function $F$ is, in particular, injective. Indeed,  suppose for two sets $S, T$ we have $F(S) = F(T)$. Then, we have both $F(S) \leq F(T)$ and $F(T) \leq F(S)$. Hence, we have both $S \subseteq T$ and $T \subseteq S$ by separability, which implies $S = T$. Thus, injectivity follows from separability. 

Injective multiset functions $f:\Pvk \to \RR^m $ are attainable even for $m=1$, provided that $V$ is finite or even infinite \cite{zaheer2018deepsets,xu2018how,amir_finite,amir2024fourierslicedwassersteinembeddingmultisets}. As we will see, attaining the strong condition of MAS multiset functions requires high dimension, and in some cases (infinite $V$), it does not even exist. This can be seen even in the following very simple example:


\begin{example}\label{ex:simple} Assume the ground set consists of only two elements $V=\{0,1\}$, and we only look for multisets of cardinality $\leq k=1$. In this case, there are only three multisets in the space $\mathcal{P}_{\leq 1}(V)$: the empty set, $S=\{0\}$, and $T=\{1\}$. As discussed previously, there exists a multiset function $F:\mathcal{P}_{\leq 1}(V)\to \RR $ which is injective. However, no such function can be separated. This is because neither $S$ nor $T$ is a subset of the other. However, because real numbers are totally ordered, we have either $F(S) \le F(T)$ or $F(T)\le F(S)$, violating the separability condition.

We note that the monotonicity can be satisfied by using functions of the form $F(S)=\sum_{x\in S} f(x)$ for non-negative $f$. Thus, achieving separation is harder than achieving monotonicity. 
\end{example}

The key reason for the non-existence of a scalar MAS set function is that multisets (or sets) are not totally ordered, but scalars are.
However, this issue does not arise for multidimensional set functions, where vectors, similar to sets, are partially ordered. So, when mapping to vectors, when and how can MAS functions be constructed? We will now discuss this.

\subsection{Existence of \mas\ functions:  finite ground set, unbounded cardinality}
\label{subsec:mas-finite}


We study \mas functions $F: \Pvinf \to \RR^m$, where input multisets may have an arbitrarily large finite cardinality, and the ground set is finite $|V|=n$. In this case, we can show that the  smallest possible dimension $\mm(V,k)$ of a \mas function is exactly $n$:
\begin{restatable}[Dimension of MAS Function]{theorem}{finiteground}
\label{thm:dim_geq_n}
For a finite ground set $V$ of size $n$, there exists a \mas function $F: \Pvinf \to \R^n$. In addition, any \mas function must have a dimension of at least $n$. In other words, $\mm(V, \infty) = n$. 
\end{restatable}
\paragraph{Proof Sketch} 
The construction of a monotone embedding with $n=|V|$ simply uses one-hot encoding. Namely, we identify $V$ with $[n]$.  For every $S\in \Pvinf$, we define
\begin{align}
    \label{eq:onehot}
     F(S) = \sum_{s\in S}e_s\in \mathbb{R}^{n}\textstyle 
\end{align}
where $e_s\in \RR^n$ is the vector with $e_s[s]=1$ and $e_s[j]=0$ for all $j\neq s$. It's clear $F$ satisfies all conditions.
For the second part, assume we have an embedding of dimension $m$, for every output dimension $i \in [m]$, 
there is a "maximal singleton element" $v^*_i \in V$  such that $F(\set{v^*_i})[i] \geq F(\set{v})[i], \forall v \in V$.  The value of the set function $F$ applied to $T=\set{v^*_1,\ldots,v_m^*}$  will dominate any singleton, due to the monotonicity of $F$. However, if $m < |V|$, then we can select a $u\in V \setminus T$ and then $F(\set{u}) \leq F(T)$, which contradicts separability. 



\subsection{Existence of MAS functions: finite ground set, finite cardinality} 
We now consider the case where the ground set $V$ is finite, as before, but now the multisets have bounded cardinality  $k < |V|$. Indeed, in many practical applications, input multiset cardinality is much smaller than $|V|$, even when $|V|$ is large. In this setting, we show that we can get a lower embedding dimension $m$ than in Theorem \ref{thm:dim_geq_n}.  In fact, we show the minimal output dimension $m$ of a \mas function $F: \Pvk \to \mathbb{R}^m$ can scale logarithmically with the size of the ground set $|V|$. However, this comes at the price of an exponential dependence on $k$: 
\begin{restatable}[Upper Bound on $\mm(V,k)$]{theorem}{finitegroundsetrestrictedcardinality}
\label{thm:upper_bound_V}
Let $k<n$ be natural numbers, and let $V$ be a ground set with $|V|=n$.   Then there exists a MAS function $F:\Pvk\to \R^m$ with embedding dimension $m= (k+2)^{k+2} \log(n)$. 
\end{restatable}
\paragraph{Proof sketch}We construct the MAS function by taking the one-hot embedding $F:\Pvk \to \RR^n$ from \eqref{eq:onehot}, and then applying $m$ random projections defined by vectors in $\RR^n$ with non-negative entries. The non-negativity ensures monotonicity, and we prove that for the stated value of $m$, the probability of achieving a MAS function using this procedure is strictly positive. 

The theorem shows  that $\mm(V,k)\leq (k+2)^{k+2}\log(n)$.
We now give two lower bounds on the embedding dimension: we show that $m$ must depend at least linearly on $k$, and at least double logarithmically on $n$:

\begin{restatable}[Lower bounds on $\mm(V,k)$]{theorem}{lowerBoundK}
\label{thm:lower_bound_k}
Let $k\geq 2$ and $n$ be natural numbers, and let $V$ be a ground set with $|V|=n$.  Then the smallest possible dimension $\mm(V,k)$ of a MAS function satisfies $\mm(V,k) \geq \log_2(\log_3 n)$. Moreover, if $k\leq \frac{n-1}{2} $ then $\mm(V,k)\geq 2k $
\end{restatable}
\paragraph{Proof Sketch}To obtain the first lower bound, we consider the sequence of singleton-set embeddings $\Mcal:= (F(\set{1}),\ldots,F(\set{n})$. By the Erd\"os-Szekers theorem~\cite{erdos}, every scalar sequence with $n$ elements has a monotone subsequence of length $\sim n^{\frac{1}{2}}$. Applying this recursively to vectors of length $m$ gets us a monotone subsequence in all $m$ coordinates of the vector, of length $\sim n^{\frac{1}{2^m}} $. If this subsequence is of length $\geq 3 $ then we will have three distinct elements $v_1,v_2,v_3\in V$ such that
$F(\{v_1\})[i]\leq F(\{v_2\})[i] \leq F(\{v_3\})[i] \text{ or } F(\{v_3\})[i]\leq F(\{v_2\})[i] \leq F(\{v_1\})[i], \forall i\in [m]$
Monotonicity implies that $F(\{v_2\})[i]$ is dominated by $F(\{v_1,v_3\})[i]$, for all $i\in [m]$, thus contradicting separability. It follows that MAS existence can only happen when $n^{\frac{1}{2^m}}\leq 2$, which leads to the double logarithmic lower bound. 

For the lower bound in terms of $k$, the argument is similar to the proof of ~\cref{thm:dim_geq_n}, where we select a  "maximal singleton element" $v^*_i$ for every dimension $i \in [m]$. Since $m < |V|-1$,  we can find elements $\set{u_1, u_2}$ disjoint from the collection $\bigcup_{i=1}^m\set{v^*_i}$, and WLOG $F(\set{u_1})[i] \geq F(\set{u_2})[i]$ for at least half of the indices in $[m]$. We then construct $S = \set{u_2}$ and $T$ to be union of $u_1$ and all the $v^*_i$ for all dimensions $i$, where $F(\set{u_1})[i] \leq F(\set{u_2})[i]$. Then monotonicity implies $F(S) \leq F(T)$, but $S \not \subseteq T$. This implies $k \leq |T|-1 \leq m/2$, as otherwise separability of $F$ in $\Pvk$ is violated. 

 \vspace{-5pt}
\subsection{Non-existence of MAS functions: for infinite ground set} 
For most practical applications, we deal with infinite (often uncountable) ground sets, for example $V=\R^d$. Thus, we would like to analyze whether MAS functions exist when $|V| = \infty$. The answer to this question is negative:
\begin{corollary}
    \label{impossibility}
    Given a ground set $V$ with $|V| = \infty$, and $k\geq 2$, there does not exist a MAS function  $F: \Pvk \to \R^m$ for any finite $m\in \NN$. 
\end{corollary}
This corollary is a simple consequence of the result in   Theorem~\ref{thm:lower_bound_k} that the embedding dimension cannot be larger than $\log \log |V|$ when $V$ is finite. We note that this result makes the minimal assumption $k\geq 2$. In Appendix \ref{app:proofs} we show this assumption is necessary, and that in the degenerate case $k=1$ there is a MAS function 
$F:\mathcal{P}_{\leq 1}(V) \rightarrow \mathbb{R}^{2}$ for $V=[-1,1] $.

\paragraph{Summary} In this section, we provided lower and upper bounds for the smallest value $\mm(V,k)$ for which an MAS function $F:\Pvk \to \RR^m $ exists. Our results are summarized in Table \ref{tab:m}. We note that some of these results require weak assumptions, which are stated in the theorems but not in the table.
\begin{table}[!!h]
    \centering
\resizebox{0.98\textwidth}{!}{
    \begin{tabular}{>{\centering\arraybackslash}m{6.2cm}|c||c|c} \hline  
        \multicolumn{2}{c||}{Ground set size $: \qquad \qquad$ $|V|=n < \infty$} & \multicolumn{2}{c}{$|V|=n = \infty$} \\ \hline
         Input multiset size $: \quad $  $k < \infty $ & $k=\infty$ & $k=1$ & $k\geq 2$ \\ \hline  \hline 
        $\mm(V,k)\leq \min\{n,(k+2)^{k+2}\log(n) \}$,
        $\mm(V,k)\ge \max\{2k,\log_2 \log_3 n\}$
       & $\mm(V,k)=n$ & $\mm(V,k)=2$ & Not possible \\  \hline
    \end{tabular}}
    \vspace{2mm}
    \caption{\it This table summarizes the lower and upper bounds for the smallest value $\mm(V,k)$ for which an MAS function $F:\Pvk \to \RR^m $ exists, where $k$ is maximal set cardinality and $n$ denotes the cardinality of the ground set $V$.  }
    \vspace{-6mm}
    \label{tab:m}
\end{table}
\vspace{-4pt}
\section{Relaxations of MAS functions}
\vspace{-4pt}
\label{sec:weak-mas}
Our definitions so far did not address differential parameterized set functions $F(S,w)$, which are the natural object of interest for set learning applications. One natural way to address this could be to require that there exists $w$ such that $F(\bullet,w)$ is a MAS function; however, since our theoretical results show that MAS functions do not exist when $|V| = \infty$, this requirement is too restrictive. Instead, we retain monotonicity as a hard constraint for every parameter $w \in \Wcal$, while treating separability as an existential condition over the parameter space. 
\vspace{-4pt}
\subsection{The notion of weakly \mas functions}
\vspace{-4pt}
Formally, given a ground set $V$, and a probability space $(\Wcal, \Bcal, \mu)$, we consider parametric set functions $F:\Pvinf \times \Wcal \to \mathbb{R}^m$, where $F(S,\bullet)$ is measurable for all $S$. We then define weakly \mas functions as: 
\begin{definition}[weakly  \mas function]
\label{def:weak-mas}
 The set function $F:\Pvinf\times \Wcal\to \mathbb{R} $ is a weakly  \mas function if the following two conditions are satisfied:\\
\textbf{(1) Pointwise monotonicity}  For any $w\in W$ and $S \subseteq T$, we have that  $F(S,w) \leq F(T,w)$.\\
\textbf{(2)  Weak separability}
If $S \not \subseteq T$, then there exists $w\in \Wcal$ such that  $F(S;w)>F(T;w)$. 
\end{definition}

\paragraph{Which condition, monotonicity or separability, should be relaxed?} In our definition, we relax the separability requirement so that different multiset pairs can be separated by different parameters, while we maintain strict requirements of pointwise monotonicity. This is because, as discussed in Example \ref{ex:simple}, constructing monotone set functions is straightforward, even in one dimension. In contrast, separability is a much stronger and more difficult condition to satisfy. This is a key reason why \mas functions do not exist when $|V| = \infty$. Indeed, as shown in Appendix~\ref{app:proofs}, there can be no continuous separable set function even without the monotonicity assumption when the ground set is $V = \RR^D$.
Therefore, separability is the natural condition for relaxation. Moreover, monotonicity remains a useful and often necessary constraint in many applications.




\paragraph{Are scalar deep sets weakly  MAS?}
Equipped with the new notion of weakly \mas functions, we ask the question of how to construct such a class of functions. As a first step to achieve this goal, we look into a simple instance of DeepSets \cite{zaheer2018deepsets}. DeepSets defines a set function by applying an 'inner' MLP $\mone$ to each set element, summing over the result, and then applying an 'outer' MLP $\mtwo$. We consider the case where $\mone$ is a shallow MLP $\mone(x)=\sigma(Ax+b)$ which gives us models of the form 
\begin{align}
F(S; (A,b)) =\textstyle \mtwo\big(\sum_{x\in S} \sigma(A x +b)\big)    \label{eq:parametric_setfn}
\end{align}
To ensure monotonicity for every parameter choice $A,b$, we will require that $\mtwo$ is a monotone vector-to-vector function, and the activation $\sigma$ is non-negative. While these choices automatically ensure monotonicity, they may not lead to weak MAS functions as for many activations there will be no separation:
\begin{proposition}
\label{prop:shallow_relu_not_sep}
   If the deep set model in \eqref{eq:parametric_setfn} is implemented with a vector-to-vector monotonously increasing function $\mtwo$ and a non-negative activation $\sigma$, then $F(\bullet; (A,b))$ is monotone for every $(A,b)$. If in addition $\sigma$ is monotone (increasing or decreasing), then there exists $S \not \subseteq T$ with $F(S; (A,b))\leq F(T; (A,b)) $ for all $A,b$.   
\end{proposition}
\textbf{Proof idea}
The monotonicity is rather straightforward. To prove lack of separation when $\sigma$ is monotone, choose 
$x\neq y, \ z=\frac{1}{2}(x+y), \ S=\{z\}, \ T=\{x,y\}, $
and then $Az+b$ is the average of $Ax+b$ and $Ay+b$, which can be used to show that $F\left(T;(A,b)\right)\geq F\left(S;(A,b)\right)  $ for all $A,b$.

\paragraph{Are Set Transformers weakly MAS?}Now we ask the question if Set Transformers~\cite{lee2019settransformerframeworkattentionbased} are weakly MAS. We notice in the following that Set Transformers are not even monotone in the following result. For this, we consider Set Transformer with sum-pooling, namely $F(S; W_Q, W_K, W_V) = \text{SumPool}(\text{Attn}(S))$. Explicitly, this is defined for a given multiset $X = \set{\bx_1, \cdots, \bx_n}$ by first defining query, key, values per-point: $\bq_i = W_Q\bx_i, \bk_i = W_K\bx_i, \bv_i = W_V\bx_i$, and then taking a weighted average, defined by weights: $\alpha_{i, j} = \frac{e^{\bq_i\cdot \bk_j}}{\sum_s e^{\bq_i \cdot \bk_s}}$ to obtain a permutation-invariant function $F(S; W_Q, W_K, W_V):= \sum_{i,j}\alpha_{i,j}\bv_i$

\begin{restatable}[]{proposition}{STnonmon}
    Let $V = \R^d$ and let $F: \Pvinf \to \R^D$ be a Set Transformer defined by $F(S; W_Q, W_K, W_V) := \text{SumPool}(\text{Attn}(S))$. Then $F$ is not a point-wise monotone function.
\end{restatable}
\vspace{-5pt}
\paragraph{Proof Idea} We consider full rank matrices $W_Q = W_K$ and any non-zero $W_V$. We choose $\bx_1 \in \R^d$ such that $\bv_1 := W_V \bx_1 \neq 0$, and consider the sets $S = \set{\bx_1}, T = \set{\bx_1, \mathbf{0}_d}$. Then, $F(S; W_Q, W_K, W_V) = \bv_1$ and $F(T; W_Q, W_K, W_V) = (\alpha_{1,1} + \alpha_{1,2})\bv_1$. Note that, $\alpha_{1,1} + \alpha_{1,2} > 1$ and thus, for any negative $\bv_1[j]$, we have $F(T; W_Q, W_K, W_V)[j] < F(S; W_Q, W_K, W_V)[j]$ and thus, monotonicity is violated.

Thus, unlike Deep Sets, it's not obvious how to make Set Transformers monotone, let alone weakly separable. Thus, in the following sections, we shall focus on obtaining weakly-MAS functions from DeepSet-like models only. 
\vspace{-4pt}
\subsection{The \hhat activation class}
\vspace{-4pt}
Note that most commonly used non-negative activation functions, such as $\relu $ or sigmoid, are monotonically non-decreasing and therefore, from Proposition~\ref{prop:shallow_relu_not_sep}, using them as $\sigma$ above will fail weak separability. 
Here, we consider a novel class of activation functions we call  'Hat Activations', which will all make $F$  in Eq.~\eqref{eq:parametric_setfn} weakly MAS. 

\begin{definition}[The \hhat\  activation function]
\label{def:hhat}
We call a function $\act: \R \to \R$  a hat activation if it is (a) non-negative  (b) compactly supported (c) not identically zero and (d) continuous. 
\end{definition}\vspace{-6pt}
Examples of hat functions are the third and fourth functions in Figure \ref{fig:activations}. 
\begin{figure}
    \centering
   \includegraphics[width=\linewidth]{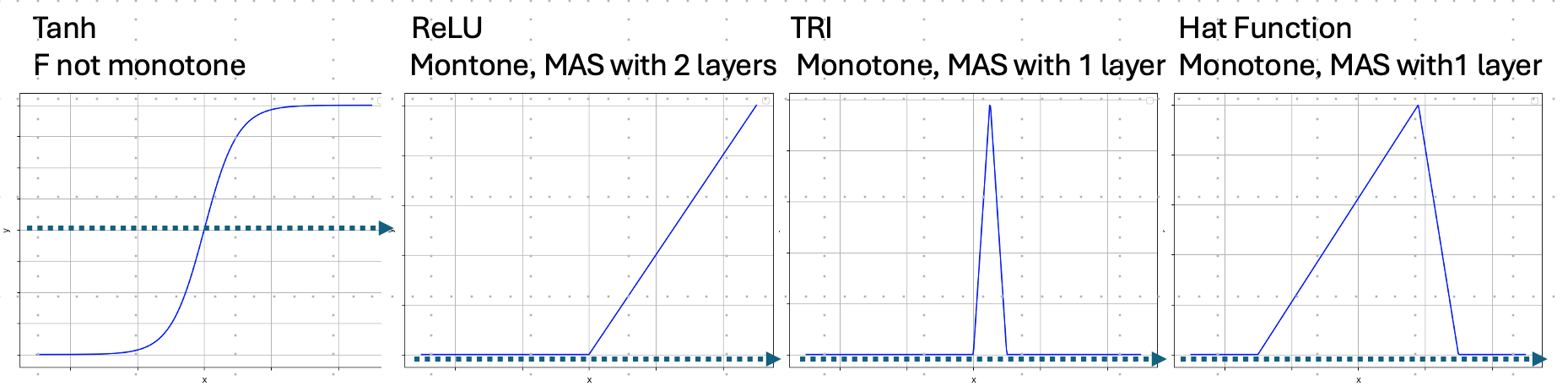}
    \caption{\it Using  a multiset model as in \eqref{eq:parametric_setfn} with activations which are not always non-negative, like $\sigma=\mathrm{Tanh}$ will not be monotone. ReLU will be monotone, but to be weakly MAS, two layers are required. TRI and more general hat functions are weakly MAS even with a single layer.}
    \label{fig:activations}
\end{figure}
When $\sigma$ is a hat function, \cref{eq:parametric_setfn} is weakly MAS, even when the output of $F$ is taken to be scalar. For this result, we will need to require that $\mtwo$ is strictly monotone. This can be handled by simply setting $\mtwo$ to be the identity. 
\begin{proposition}
    \label{prop:weakmas_hat}
    If $\mtwo$ is strictly monotone increasing, and $\sigma$ is a hat activation function, then $\mtwo(\sum_{x\in S}\sigma(a\tran x+b)) $ is a weakly \mas function. 
\end{proposition}
\paragraph{Proof idea}
If $S$ is not a subset of $T$, there is an element $s$ whose multiplicity in $S$ is larger than its multiplicity in $T$. We can then choose $a,b$ so that $s$ is in the support of  $\sigma(ax+b)$ but the elements of $T$ are not.  
\vspace{-4pt}
\subsection{Slightly larger ReLU networks}
So far, we have considered simple models as in \eqref{eq:parametric_setfn} which apply a single linear layer before applying activation and summation. In this case, we saw that $\relu$ and other activations do not create weakly \mas functions, and defined the notion of hat activations, which are weakly \mas functions. But what happens if we use $\relu$ activations but allow deeper networks? 

We can use our previous analysis to show that very small $\relu$ networks with two layers can already be \mas functions. To see this, we consider a candidate from the hat function class, namely the $\tri$ function from Figure \ref{fig:activations},  and observe that: $\tri(x)=  \relu\left(\relu(2x)+\relu(2x-2)-\relu(4x-2) \right)$
We use this to deduce the following result:

\begin{restatable}[]{proposition}{proprelumas}\label{prop:relu_mas}
Given $V \subseteq \R^d$, we consider affine transformations $\mathcal{A}_2:\RR^d \to \RR^3, \mathcal{A}_1:\RR^3 \to \RR$. Thus, here $\mathcal{A}_2(t) := A_2 t + b_2$ and $\mathcal{A}_1(z) = a_1\tran z + b_1$, where $A_2 \in \R^{3 \times d}, b_2 \in \R^3,a_1 \in \R^3, b_1 \in \R$ are the respective parameters. Then, the set functions of the form 
\begin{equation}
    \label{eq:relu_masnet}
    F: \Pvinf \to \R, \textstyle F(S;\mathcal{A}_1,\mathcal{A}_2)=\mtwo\left(\sum_{x\in S} \relu \circ \mathcal{A}_1 \circ \relu \circ \mathcal{A}_2(x) \right)
\end{equation}
 are weakly \mas functions.
\end{restatable}

Our results so far are summarized in Figure \ref{fig:activations}. If $\sigma$ is not a non-negative function, then $F$ will not be monotone. We can attain weakly MAS functions with a two-layer ReLU network, or with a one-layer network with hat activations. 
\subsection{Asymmetric distance induced Holder separability}
\paragraph{Why weak separability is not sufficient} 
While weak separability is a key relaxation for weakly \mas functions~(Definition~\ref{def:weak-mas}), it has two key limitations:
\begin{enumerate*}[label=\textbf{(\arabic*)}]
\item Separation should ideally hold over a non-negligible subset of $\Wcal$, not just a single point $w$;  \item it treats separability as a boolean condition, ignoring approximate containment. We would like to guarantee that if $S$ is almost a subset of $T$, then $F(S,w)$ will be close to dominated by $F(T,w)$.
This motivates us to create an \emph{asymmetric (pseudo) metric} to quantify the size of set difference and create stronger notions of separability based on that.

\end{enumerate*}

\paragraph{Asymmetric set distance}
For a ground set $V \subseteq \R^d$ and two sets $S, T \in \Pvinf$ such that $|S| \leq |T|$.
We define the asymmetric distance from $S$ to $T$ as the Earth Mover Distance (EMD) between $S$ and the subset of $T$ that is \emph{closest} to $S$ with the same cardinality as $S$.
\begin{equation}
    \label{eq:asymmetric_dist}
    \das(S, T) =  \min_{T'\subseteq T: |S|=|T'|}\emd(S,T')
\end{equation}
$\das(\bullet,\bullet)$  captures graded set containment:  if $S \not \subseteq T$, but is "very close" to a subset of $T$ in EMD metric, then $\das(S,T)$ is small. Moreover, $\das(S, T) = 0 \iff S \subseteq T$. Based on $\das$, we propose a newer separability notion for parametric functions.

\paragraph{Lower H\"older separability of set functions}
\citet{davidson2025holderstabilitymultisetgraph} introduced H\"older continuity through expectation over parameters. Unlike their symmetric EMD, we define H\"older separability using the asymmetric distance $\das$\eqref{eq:asymmetric_dist}.
\begin{definition}[Lower H\"older separability]
    \label{def:lower-holder}
    Let $V \subseteq \R^d$ be the ground set and $(\Wcal,\Bcal,\mu)$ be a probability space, and a constant $\lambda>0$.  A parametric set function $F(\bullet,w) : \Pvk \to \R^m$ with $w\sim \mu(\cdot)$ is $\lambda$ lower H\"older separable if there exists $c > 0$ such that:
    \begin{equation}
        \label{eq:lower_holder}
         \Exp_{w \sim \mu(\cdot)}\norm{\sbrac{F(S; w) - F(T; w)}_+}_1 \geq c \cdot \das(S, T)^\lambda, \\ \ \text{for all } S, T \in \Pvk
    \end{equation}
\end{definition}

\subsection{H\"older separable Set functions}
\label{sec:holder_separable_discussion}
\paragraph{H\"older separability using \hhat activation}Proposition \ref{prop:weakmas_hat} shows how to construct  weakly \mas\ functions using \hhat activations.
We now show that these constructions are also lower H\"older, under additional weak assumptions on the hat functions and $M_2$: 

 \begin{restatable}[$F$ is lower H\"older]{theorem}{flowerholder}
   \label{thm:F_lower_holder}
     Let $V \subset \R^d$ be a compact set, and $\sigma$ a \hhat\ activation function which is piecewise continuously differentiable supported in some interval $[\leftend, \rightend]$, and satisfying the condition: $\lim_{t \to \leftend^+}\diff{\sigma}{t} > 0 $. Let  $\mtwo: \R \to \R$ be a lower Lipschitz function. Consider the function
     $F(S; \tuple{a, b, c})= \mtwo\big(\sum_{x \in S}\sigma\big(\frac{a\tran x + b}c\big)\big),$ 
     where the multisets $S$ come from $\Pvk$, and $a \sim \Unif(\S^{d-1}), b \sim \Unif([-1,1]), c \sim \Unif((0,2])$. Then, $F(\bullet,\tuple{a, b, c})$ is Monotone H\"older separable with exponent $\lambda=2$.
\end{restatable}
\vspace{-2pt}

\paragraph{Probability of successful separation} Given $S, T \subseteq V$, we can consider $[F(S; \tuple{a, b, c}) - F(T; \tuple{a, b, c})]_+$ to be a real-valued non-negative random variable, where the randomness is over the parameter space $\tuple{a, b, c}\in \Wcal =\RR^d \times \RR\times \RR$ equipped with a probability measure. The above result on lower H\"older stability 
gives us a lower bound on the probability of a parameter tuple $\tuple{a, b, c}$ separating two non-subsets $S, T$ proportional to the set distance $\das(S, T)$. Moreover, the functions we have considered so far are scalar-valued set functions. By considering independent copies of the parameters across multiple dimensions, we can increase the probability of separation, since $F(S), F(T)$ are not separated iff $F(S)[i] \leq F(T)[i]$ across each embedding dimension $i$.  We formalize the above observations as follows:

\begin{restatable}[Probability bounds on separation]{theorem}{probabilityseparation}
\label{thm:probabilistic_implications_of_lower_holder}
 Let $V \subseteq \R^d$ and $\sigma$ be as in Theorem \ref{thm:F_lower_holder}, and let $A\in \RR^{m\times d},b\in \RR^m, c\in \RR^m$ whose $m$ columns (respectively entries) are drawn independently from the distribution on $a_j,b_j,c_j$ described in Theorem \ref{thm:F_lower_holder}, and consider the function 
 \begin{equation} \label{eq:hadamard}
 \textstyle F(S;A,b,c)=\sum_{x\in S} \sigma\left(  c^{-1} \odot (Ax+b)\right) .\end{equation}
Then there exists $C>0$, so that for all $S \not \subseteq T$, 
$\Prob\left( F(S)\leq F(T) \right) \leq \round{1 - C\das^2(S,T)}^m.$
 \end{restatable}
In Equation \eqref{eq:hadamard} $c^{-1}$ stands for the elementwise inverse and $\odot$ stands for elementwise multiplication. 

Note that the probability of failed separation goes to zero exponentially as $m$ increases, and also becomes smaller as $\das(S,T)$ increases. This supports the idea that a larger measure of parameters separates sets with larger asymmetric distance and that separability becomes easier with larger embedding dimensions (which we saw previously in the existential results).

\paragraph{H\"older separable set functions using ReLU} As seen in Proposition~\ref{prop:relu_mas}, one can construct weakly \mas functions with a two layer neural network with $\relu$ activation. Our proof used that two layer $\relu$ networks can basically "simulate" a \hhat function. Under the light of the above results on H\"older separability, one can argue that two-layer $\relu$ networks would also have such guarantees. Also, we show in Appendix~\ref{app:proofs} that under certain structure on the ground set $V$ (for example, $V$ being a hypersphere), even with \emph{one layer} $\relu$ networks, functions of the form in ~\cref{eq:parametric_setfn}  will also have Lower H\"older separability guarantees. 

\paragraph{Upper Lipschitz bounds} In Appendix.~\ref{app:proofs} we complement our results on \emph{lower} Holder stability by showing that the construction in Theorem \ref{thm:probabilistic_implications_of_lower_holder} is also \emph{upper} Lipschitz in an appropriate sense. 
\vspace{-4pt}
\subsection{Monotone Universality and the role of $M_2$}
\vspace{-4pt}
Besides set containment-based applications, there are other interesting scenarios where one would like to construct monotone multiset-to-vector functions. The following theorem shows that on finite ground sets,  the combination of a \mas function and a universal monotone vector-to-vector (such as \cite{sill1997monotonic})  can approximate \emph{all} monotone multiset-to-vector functions. 

\begin{restatable}[Universality]{theorem}{universal}\label{thm:universal_approx}
Let $V$ be a finite ground set, and let $F:\Pvk \to \RR^m$ be a \mas\ function. Then for every 
multiset-to-vector monotone function $f:\Pvk \to \RR^s $, there exists a vector-to-vector monotone function $M:\RR^m \to \RR^s$ such that $F(S)=M\circ f(S) $.
\end{restatable}
\vspace{-10pt}
\section{\our: Neural Modeling of \mas functions}
\vspace{-5pt}
\label{sec:model}
Now, our goal is to leverage our theoretical analysis to design neural network-based multiset-to-vector models that preserve monotonicity and weak separability. Motivated by the formulation in DeepSets and our analysis so far, we wish to design neural set functions of the form:  
\begin{equation}
    \label{eq:masnet}
    \our(S) = \textstyle\mtwoneural\round{\sum_{x \in S}\sigma\round{\moneneural(x)}}
\end{equation}
In our analysis, we required $\mtwoneural$ to be a monotone vector-to-vector function. In most of our experiments, we enforce this simply by choosing $M_2(x)=x$, but this can also be enforced by using monotone activations and non-negative parameters. 
\vspace{-0.5pt} 

\paragraph{\hatmas} To enforce weakly \mas in~\cref{eq:masnet}, our results from Prop.~\ref{prop:weakmas_hat} and Theorem~\ref{thm:F_lower_holder} suggest using a \hhat activation as $\sigma$. In our experiment, we do it by not choosing $\sigma$ to be a specific hat activation, but rather the parametric form:    
\begin{equation}\label{eq:simpleHat}
\textstyle \sigma_{\alpha,\beta,\gamma}(x)=\relu \round{ \frac{x-\alpha}{\gamma \cdot \beta}}
+\relu \round{ \frac{x-(\alpha+\beta)}{(1-\gamma) \cdot \beta}}
-\relu \round{ \frac{x-(\alpha+\gamma \cdot \beta)}{\gamma \cdot (1-\gamma) \cdot \beta}}
\end{equation}
For all $\alpha$, $\beta>0$, and $\gamma\in (0,1)$, $\sigma_{\alpha,\beta,\gamma}$ is a hat function (Definition \ref{def:hhat}) with support $[\alpha,\alpha+\beta]$ and peak at $\alpha+m\beta$. Examples include the third and fourth functions in Figure \ref{fig:activations}. Applying this to an $m$-dimensional output with independent $\alpha, \beta, \gamma$ per dimension yields $3m$ parameters total.
\vspace{-0.5pt}

\paragraph{\our with $\relu$}
Motivated by our theoretical results from ~\cref{prop:relu_mas} of 2-Layer $\relu$ networks being weakly \mas and from our discussion of $\relu$ networks being H\"older separable (under appropriate assumptions) in Section~\ref{sec:holder_separable_discussion}, we propose using $\relu$ in \our, which case $M_{\theta_1}$ will have to be a network with at least 2 layers to ensure weakly \mas as in \cref{prop:relu_mas}. This gives one more way to design \our and we refer to this option as \relumas.
\vspace{-0.5pt}

\paragraph{Other variants of\ \our} Definition \ref{def:hhat} establishes a general hat function class extending beyond the piecewise linear functions as in \eqref{eq:simpleHat}. We present \integralmas in Appendix \ref{app:model}, which achieves universal approximation of the \hhat class by modeling its derivative via neural networks and approximating the integral. Additionally, using the specific hat activation $\tri$ (Figure \ref{fig:activations}) as $\sigma$ yields an alternative formulation called \trimas.
\vspace{-0.5pt}

\paragraph{Which \our to choose and when?} We have given multiple recipes of \our in the above discussion, and the question arises: which one to choose? We show via our experiments: For parameter-constrained scenarios requiring single-layer $M_{\theta_1}$ networks, \hatmas provides optimal separability (both theoretically and empirically). For deeper $M_{\theta_1}$ ($\geq 2$ layers), both \relumas and \hhat-based variants perform similarly, with \relumas often providing more stable training and better performance overall.

\vspace{-5.8pt}
\section{Experiments}
\vspace{-3.8pt}
 \label{sec:expts}
We evaluate the \our\ variants (Section~\ref{sec:model}) on synthetic, text and point-cloud datasets to characterize monotonicity and separability. Specifically, we focus on (exact and approximate) set containment. Given sets $S$ and $T$, with binary label $y(S, T) \in \{0, 1\}$ indicating (exact or approximate) set containment, we evaluate how accurately \our\ predicts $y(S, T)$.

\vspace{-4pt}
\subsection{Set Containment}
\vspace{-3pt}
We perform experiments on a synthetically generated dataset, four text datasets, and one image dataset. In each case, we split the dataset into 5:2:2 train, test, and dev folds. We minimize the following fixed-margin hinge loss that enforces vector dominance, to train the parameters of  \our.
\begin{align}
     \sum_{S,T} (1-y(S,T)) \min_{i\in[m]}\sbrac{\model(S)[i] - \model(T)[i] +\delta}_+ 
    + y(S,T)\max_{i\in[m]}\sbrac{\model(S)[i] - \model(T)[i] +\delta}_+ 
\vspace{-2mm}
\end{align}
 %

\begin{wraptable}[9]{r}[-10pt]{0.42\linewidth}
    \vspace{-3mm}
    \centering
    \captionsetup{font=small, skip=3pt}

    \begin{adjustbox}{max width=\linewidth}
        \begin{tabular}{@{}cc@{\hspace{9pt}}cccc@{}}
            \toprule
            \multicolumn{2}{c}{\textbf{Set sizes}}
            & \multicolumn{4}{c}{\textbf{Model}} \\
            \cmidrule(lr){1-2}
            \cmidrule(l){3-6}
            \(\lvert S\rvert\)
            & \(\lvert T\rvert\)
            & DS
            & ST
            & M-ReLU
            & M-\(\hhat\) \\
            \midrule
             1 &   2
               & \underline{0.98}
               & \underline{0.98}
               & \textbf{0.99}
               & \textbf{0.99} \\

             1 &  10
               & 0.59
               & 0.59
               & \textbf{0.99}
               & \textbf{0.99} \\

            10 &  30
               & \underline{0.89}
               & 0.80
               & \textbf{1.00}
               & \textbf{1.00} \\

            10 & 100
               & 0.54
               & 0.57
               & \underline{0.98}
               & \textbf{0.99} \\
            \bottomrule
        \end{tabular}
    \end{adjustbox}

    \caption{Accuracy on the synthetic dataset.}
    \label{tab:synth}
    \vspace{-2mm}
\end{wraptable}

\paragraph{Analysis on synthetic datasets} To generate our synthetic dataset, we first sample a target set $T \subset \R^d$, where each element  $x\in T$ is drawn from $\Normal(0, \bm{I})$, i.i.d. Given $T$, we compute $S$ with a fixed size $|S|=s$ as follows. We first obtain $S$ with positive labels $y(S,T)=1$ by drawing a subset from $T$, uniformly at random, without replacement. To generate $S$ with $y(S,T)=0$, we sample $s$ from $\Normal(0,\bm{I})$ independently. We chose $d=4$ and the set embeddings have $m=256$ dimensions. The class-ratio of subsets to non-subsets was taken to be 1:1. 

We compare two variants of \our--- \our-ReLU (M-ReLU) and \our-Hat (M-Hat)--- against Deep Sets (DS) \cite{zaheer2018deepsets} and Set Transformer (ST) \cite{lee2019settransformerframeworkattentionbased}. Table~\ref{tab:synth} summarizes the results
\begin{wrapfigure}[11]{r}{0.37\textwidth} 
\vspace{-1mm}
    \centering
     \includegraphics[width=0.35\textwidth]{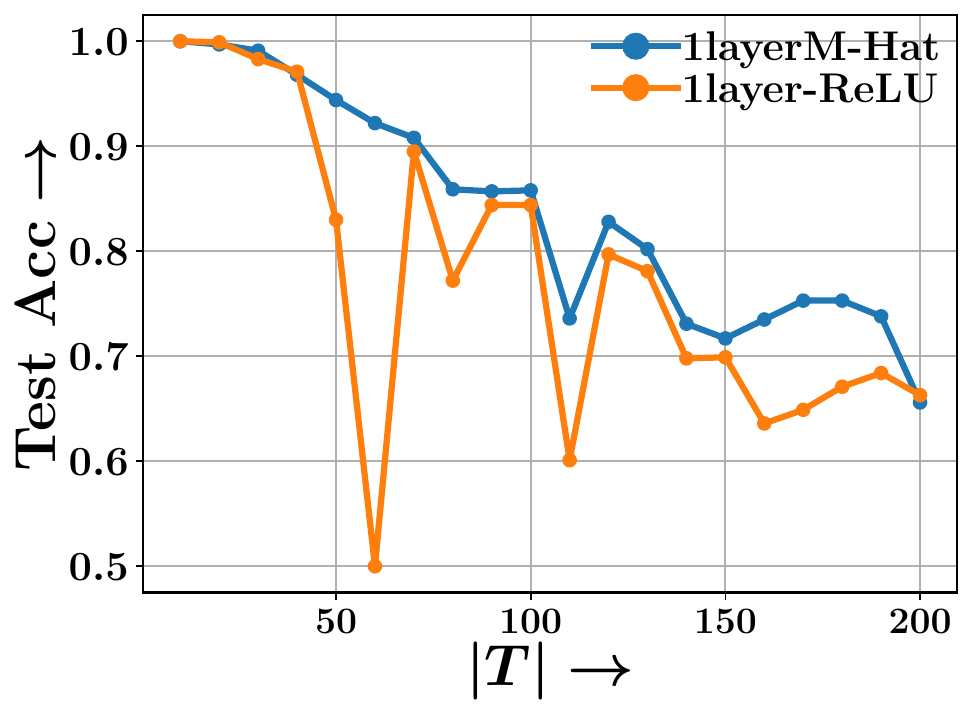}
    \caption{Acc vs $|T|$ for 1-layer MLP}
    \label{fig:shallow}
    \vspace{-15pt}  
\end{wrapfigure}
for varying values of $|S|$ and $|T|$. We observe that:
(1) Both variants of our method outperform the baselines;
(2) the baselines degrade significantly as $|T|$ increases (with fixed $|S|$), as predicting separability becomes harder as the gap between $|T|$ and $|S|$ increases;
(3) The performance of M-ReLU and M-Hat is comparable. We now compare 1-layer pointwise models ending 
 in $\relu$ and \hhat activation respectively, followed by aggregation: we call them 1layer-$\relu$ and 1layerM-\hhat. We see in Fig.~\ref{fig:shallow} that: (1) 1layerM-\hhat performs significantly better than 1layer-$\relu$, validating the results from Propn.~\ref{prop:shallow_relu_not_sep}. (2) acc. declines for both as $|T|$ increases, consistent with the intuition that larger $|T\setminus S|$ gaps make the task harder.

\paragraph{Set containment on Text datasets} We evaluate on MSWEB, MSNBC, and Amazon Registry datasets, which exhibit natural set containment from user behavior. Each datapoint is a wordbag; we compute BERT embeddings for all unique words to form the ground set. $(S, T)$ pairs are sampled from these  (Appendix~\ref{app:addl-details}), and labeled via set containment. For \inexact containment,  small 
\begin{table}[t]
    \centering
    \small
    \captionsetup{font=small, skip=3pt}
    \setlength{\tabcolsep}{4pt}
    \renewcommand{\arraystretch}{1.05}

    \begin{tabular}{@{}lcccc@{}}
        \toprule
        \textbf{Model}
        & \textbf{Bedding}
        & \textbf{Feeding}
        & \textbf{MSWEB}
        & \textbf{MSNBC} \\
        \midrule
        DeepSets
        & 0.51 & 0.50 & 0.88 & 0.67 \\

        Set Transformer
        & 0.77 & 0.79 & 0.90 & 0.94 \\

        FlexSubNet
        & 0.88 & 0.85 & 0.92 & 0.91 \\

        Neural SFE
        & 0.52 & 0.53 & 0.88 & 0.66 \\

        \midrule
        \relumas
        & \textbf{0.98}
        & \textbf{0.98}
        & \textbf{0.99}
        & \textbf{0.97} \\

        \hatmas
        & \underline{0.94}
        & \underline{0.93}
        & \underline{0.97}
        & \underline{0.95} \\
        \bottomrule
    \end{tabular}

    \caption{Accuracy on text datasets with a 1:1 class ratio.}
    \label{tab:noisy_accuracy}
\end{table}

Gaussian noise is added to $S$ without changing $\labelst$.
We now compare the variants of \our: \relumas and \hatmas against several baselines, DeepSets~\cite{zaheer2018deepsets}, Set Transformers~\cite{lee2019settransformerframeworkattentionbased}, FlexSubNet~\cite{de2022neuralestimationsubmodularfunctions}, Neural SFE~\cite{sfekaralias2022neural}, for \inexact set containment on the datasets: MSWEB, MSNBC, Amazon-Feeding, Amazon-Bedding. Gaussian noise with $0.01$ std was used to generate noisy pairs. The class ratio of subsets to non-subsets was taken to be 1:1.  Here, set embeddings have dim $m=50$, and set elements have dim $d=768$. The results in table~\ref{tab:noisy_accuracy} show: (1) \relumas and \hatmas have similar acc., with \relumas being marginally better. (2) Both are generally better than baselines.

\paragraph{Set containment on point clouds} We use ModelNet40~\cite{modelnetwu20153dshapenetsdeeprepresentation}, a dataset of 3D CAD models across 

\begin{wraptable}{r}{0.40\textwidth}
    \vspace{-4pt}
    \centering
    \scriptsize
    \captionsetup{font=small, skip=2pt}

    \begin{adjustbox}{max width=\linewidth}
        \begin{tabular}{@{}lccc@{}}
            \toprule
            \textbf{Model}
            & \multicolumn{3}{c}{\(\lvert S\rvert\)} \\
            \cmidrule(l){2-4}
            & \(128\) & \(256\) & \(512\) \\
            \midrule

            DeepSets
            & 0.52 & 0.51 & 0.50 \\

            Set Transformer
            & 0.62 & 0.51 & 0.50 \\

            FlexSubNet
            & 0.84 & 0.75 & 0.60 \\

            Neural SFE
            & 0.52 & 0.51 & 0.50 \\

            \midrule
            \relumas
            & \textbf{0.98}
            & \textbf{0.94}
            & \textbf{0.72} \\

            \hatmas
            & \underline{0.87}
            & \underline{0.81}
            & \underline{0.65} \\
            \bottomrule
        \end{tabular}
    \end{adjustbox}

    \caption{Point-cloud accuracy for different set sizes \(\lvert S\rvert\), using a 1:1 class ratio.}
    \label{tab:pointcloud}
    \vspace{-4pt}
\end{wraptable}

40 categories. For each object in category $C_1$, we sample 1024 points to form $T$. For true subsets, we sub-sample $S \subseteq T$ and for non-subsets, $S$ is sampled from a different category $C_2$.  Small white noise(std=0.005) is added to $S$ for the task of \inexact containment. We use a pointcloud encoder(such as pointnet) followed by a set-to-vector embedding model to produce $m=50$ dimensional embeddings, and each point in the dataset is a 3D coordinate with $d=3$ dimensions. The class ratio of subsets to non-subsets was taken to be 1:1.  Results are in table~\ref{tab:pointcloud}. We observe that: (1) \relumas and \hatmas are generally better than baselines, with \relumas being slightly better. (2) The accuracy of \our decreases as $|S|$ decreases, with the separability becoming hard for monotone models.
 %

\vspace{-3pt}
\subsection{Monotone Set function approximation}
\vspace{-5pt}
\begin{wraptable}[5]{r}{0.40\textwidth}
    \vspace{-5mm}
    \centering
    \scriptsize
    \captionsetup{font=small, skip=2pt}
    \setlength{\tabcolsep}{6pt}
    \renewcommand{\arraystretch}{1.05}

    \begin{tabular}{@{}lc@{}}
        \toprule
        \textbf{Model} & \textbf{MAE} \\
        \midrule
        DeepSets
        & \(0.01024 \pm 0.00030\) \\

        Set Transformer
        & \(0.01026 \pm 0.00033\) \\

        \relumas
        & \(\underline{0.01023 \pm 0.00107}\) \\

        \hatmas
        & \(\mathbf{0.00959 \pm 0.00016}\) \\
        \bottomrule
    \end{tabular}

    \caption{Monotone function approximation.}
    \label{tab:mae_results}
    \vspace{-5pt}
\end{wraptable}

In this experiment, we examine if a scalar variant of \our\ can approximate a monotone function $F^*$, by getting trained from set, value pairs in the form of $\set{(S,F^*(S))}$. Once trained, we measure $|\mas(S)-F^*(S)|$ for each $S$ and average over them to compute MAE. 
We compare \our against baselines: DeepSets, SetTransformer. We trained using MSE loss, and report test MAE results in Table \ref{tab:mae_results} where we see: (1) \hatmas is best performing, followed by \relumas. (2) Both are better than baselines.
\vspace{-5pt}
\section{Conclusion and Future Work}
\vspace{-5pt}
In this work, we study the design of set functions that are useful in set containment through Monotone and Separating (MAS) properties. We derive bounds on the embedding dimensions required for MAS functions and show their nonexistence in infinite domains. To address this, we introduce a relaxed model, \our, which satisfies a weak MAS property and is provably stable. Experiments demonstrate that \our\ outperforms standard models on set containment tasks by leveraging monotonicity as an inductive bias. It would be interesting to check if our method can be used for applications in graphs, \eg, subgraph matching. Another potential direction is to consider other possible relaxations of separability.
\vspace{-5pt}
\section{Limitations}
\vspace{-5pt}
\label{app:impact}
Our work provides a theoretical analysis of monotonicity and separability for set functions in the context of set containment, and introduces neural models that outperform standard set-based architectures such as DeepSets and Set Transformers on set containment tasks. Our work has the following limitations, addressing which are interesting directions for future work:

\begin{itemize}
    \item After proving the impossibility of constructing \mas functions over infinite ground sets, we proposed parametric set functions defined with respect to a probability measure over the parameter space. We provided probabilistic guarantees: for non-subsets $S, T$, if their asymmetric distance $\das(S, T)$ is large, then a randomly sampled function from the parameter space will separate them with high probability. However, these guarantees hold only in the randomized setting---analogous to guarantees for neural networks at random initialization---and do not extend to models after training via gradient-based optimization.

    \item Our universality result, which shows that \mas functions composed with monotone vector-to-vector mappings can approximate all monotone set-to-vector functions, was only established for finite ground sets. Extending this result to infinite ground sets remains an open problem.

    \item We currently do not generalize our analysis to the subgraph isomorphism problem, which is a natural and strictly harder extension of the set-containment task.
\end{itemize}
\vspace{-5pt};
\newpage

\newpage
\section*{Appendix}
\phantomsection
\addcontentsline{toc}{section}{Appendix}
\addtocontents{toc}{\protect\setcounter{tocdepth}{0}}

\setcounter{section}{0}
\renewcommand{\thesection}{\Alph{section}}
\renewcommand{\theHsection}{appendix.\thechapter.\Alph{section}}

\begin{center}
\textbf{\large Appendix}
\end{center}
\section{More discussions on related work}
\label{app:related_works}
This work lies at the intersection of representation learning for sets and multisets, monotone and submodular set functions, and theoretical limits of learning under constraints on some partial order. Below we review related literature across several communities, such as machine learning, optimization, theoretical computer science, fuzzy measures, and physics. We highlight how these threads connect to our problem setting.

Multiset to vector functions, and permutation invariant functions (including symmetric and anti-symmetric polynomials) have been used in a bunch of real life modelling tasks. Anti symmetric polynomials in the context of quantum many-body physics were used in~\cite{huang2021geometrybackflowtransformationansatz}. For point cloud datasets, such set-to-vector embeddings are widely used such as in~\cite{qi2017pointnet}. In the context of search and retrieval, set-containment was directly used in~\cite{roy2023locality}. On a somewhat related aspect, \cite{singh2020explaining} deals with monotone ranking functions on a vector. This is specifically a monotone function from multisets to reals if the vector P denotes the weight/frequency of each object in the set. Moreover, since this is a vector-to-real monotone function, composing a set-to-vector monotone function with a monotonic ranking function preserves monotonicity. These are relevant to the recent usage of set containment in the context of information retrieval. Another related usage is in vector search~\cite{engels2023dessert}, where relevance based retrieval is done after converting objects into high-dimensional embeddings. 

More specifically monotone (multi)set to vector functions have been of interest in the form of theory of capacities~\cite{choquet1953capacities} which is a probability measure based on monotonicity. In fuzzy set theory, where partial membership is allowed, any set-to-vector monotone function is equivalent to a vector-to-vector monotone function. Measures and integrals on such structures have been explored in the form of Choquet integrals~\cite{agahifuzzy, monotonefuzzy} etc. In the context of game theory and economics, monotone and submodular functions are used to model diminishing marginal utilities and in combinatorial auctions, such as in~\cite{lehmann2001combinatorial,dobzinski2005approximation,feige2009maximizing}. In learning theory, understanding PAC learning and hardness of vector-to-vector monotone functions have always received a lot of interest. And using the membership vector of multisets, this is basically equivalent to learning multiset-to-vector monotone functions. Some works such as~\cite{blum1988training} uses set containment and monotone set functions as a tool in the proof. Specifically, they perform a reduction of neural network training to the set splitting problem to prove the hardness result. This is the task of partitioning a set $S = S_1 \cup S_2$ such that for a collection of subsets $\mathcal{C} = \{C_i: C_i \subseteq S, \forall i\}$ we must have $C_i \not\subseteq S_1$ and $C_i \not\subseteq S_2$, $\forall i$. This is basically preventing set containment for a subset family. More direct applications include~\cite{you2017deep} which discusses the challenges of learning vector to vector monotonic functions or~\cite{balcan2012learning} that discusses learning monotone valuation functions. Works like~\cite{bshouty1993exact, odonnell2003learning} deal with learning monotone boolean functions to any constant accuracy, under the uniform distribution in polynomial time. Since any boolean vector is basically one-hot encoding of a set on a fixed size vocabulary, this paper essentially describes learning monotone set functions. Works on monotone learning such as~\cite{li2025monotonic} describe learning processes in which expected performance consistently improves
as the amount of training data increases. Thus, a trained model is a monotone set function from the training dataset to reals, where the output is a valuation metric.

So far, we have discussed works that have used multiset to vector embeddings and analyzed them. We have taken special care to discuss works that have involved the structure of monotonicity and applied it to various domains. However, to the best of our knowledge, there has been no previous work that specifically deals with partial order preserving set-to-vector maps. One related work is~\cite{dushnik1941partially}, which has studied the order dimension of a partially ordered structure. Now, note that partial order preserving sets are injective as discussed before, and injectivity is in fact a weaker condition that the MAS property. Here, we discuss a number of works on injective functions on multisets.  The paper~\cite{amir_finite} shows that moments of neural networks
do define injective multiset functions, provided that an analytic non-polynomial
activation is used. They also state and
prove a finite witness theorem, which is of independent interest. There are a series of works such as~\cite{amir2024fourierslicedwassersteinembeddingmultisets, balan2022permutation, sverdlov} that design bi-Lipschitz(hence, injective) embeddings on sets and graphs. On the other hand, works like~\cite{wagstaff2019limitations, wang2023polynomial} investigate the impact of latent dimension on the expressive power of set functions such as DeepSets.

\newpage
\section{Proofs of the technical results}
\label{app:proofs}
\subsection{Proofs of existential results on \mas functions}
\finiteground*
\begin{proof}
\label{proof-of-1}
The construction of a monotone embedding with $n=|V|$ simply uses one-hot encoding. Namely, we identify $V$ with $[n]$.  For every $S\in \Pvinf$, we define
\begin{align}
    F(S) = \sum_{s\in S}e_s\in \mathbb{R}^{n}
\end{align}
where $e_s\in \RR^n$ is the vector with $e_s[s]=1$ and $e_s[j]=0$ for all $j\neq s$. We now show it satisfies all conditions:
Let $S\subseteq T$. Then, by definition, $\forall v\in V, c_S(v)\leq c_T(v)$, then $F(S) =\sum_{s\in S}e_s\leq \sum_{t \in T} e_t=F(T)$ thus $F(S)\leq F(T)$. On the other hand, if $F(S)\leq F(T)$ then $\forall v\in V, c_S(v)\leq c_T(v)$, and thus $S\subseteq T$.

For the second direction, assume there exists \mas $F:V \to \R^m$ for $m\leq n -1$.
For every output dimension $i \in [m]$, 
there is a "maximal singleton element" $v^*_i \in V$  such that $F(\set{v^*_i})[i] \geq F(\set{v})[i], \forall v \in V$.  The value of the set function over the union of such $v^*_i$-s across $i\in [m]$ will dominate any singleton that is disjoint from this collection, thanks to the monotonicity of $F$. As $m < |V|$, then $T:=\bigcup_{i=1}^m \set{v^*_i}$ does not cover $V$. Thus, we can select an element disjoint from this collection named $s\in V$, which gives disjoint sets $S=\{s\}, T$ with $F(S) \leq F(T)$, which contradicts separability. 
\end{proof}
\finitegroundsetrestrictedcardinality*
\begin{proof}
\label{proof_of_probablistic_construction}
For convenience of notation, in this proof we will assume without loss of generality that $V=[n]$. 
As discussed in the main text, the function $F_{a_1,\ldots,a_m}$ will be weakly monotone for any choice of non-negative vectors. Our goal is to show that there exists a set of parameters $a_1,\ldots,a_m\in [0,1]^n$, such that the obtained $F_{a_1,\ldots,a_m}$ is a monotone embedding, which means that, if $S \not\subseteq T$, then there exists some $j$ such that $h_j(S)>h_j(T)$, or in other words $\langle f(S),a_j \rangle>\langle f(T),a_j \rangle$. We note that to prove this, it is sufficient to consider pairs $(S, T) $ which we will call \emph{extreme pairs}. This means that $S$ consists of a single element  $x\in V$, and $T$ consists of $k$ elements in $V$ but does not contain $x$ (but repetitions are allowed). Indeed, if the claim is true for such pairs, and $S,T\in \Pvk $ such that $S \not\subset T$, then there exists an element $x$ which is in $S$ more times than it's in $T$. First, remove all copies of $x$ from $T$, and we know that at least one copy of $x$ will be in $S$, note that the difference between the function values remains the same. Then, we will 'replace' $S$ with a smaller set $\tilde S=\{x\}$, and $\tilde T$ from adding more elements to $T$ from $V \setminus \{x\}$ to obtain a new set $\tilde{T}$ of maximal cardinality $k$. We then will have that $(\tilde S,\tilde T)$ is an extreme pair, and so for some $j$ we will have  that $h_j(\tilde{S})> h_j(\tilde{T})$. Therefore, using the weak monotonicity of the function $h_j$, we will have 
$$h_j(S)\geq h_j(\tilde{S})> h_j(\tilde{T})\geq h_j(T).$$

Given an extreme pair $(S,T)$ we consider the set of all 'bad' $a$, namely
\begin{align*}
    B_{S,T}:=\{a\in [0,1]^n, \langle f(S),a \rangle > \langle f(T),a \rangle \}
\end{align*}

Our first goal will be to bound the Lebesgue measure of this set. Recall that $S$ is a singleton $S=\{s\}$, and $T$ does not contain this singleton. Accordingly,
\begin{align*}
B_{S,T}=\{a\in [0,1]^n| \quad a_{s}\leq \sum_{t \in T} a_t\}
\end{align*}
We will compute the probability of the complement of this set, namely vectors $a$ such that $a_s > \sum_{t \in T} a_t$. 
First denote by
\begin{align*}
    E_{S,T}=\{a\in [0,1]^n|\forall t\in T, a_t < \frac{a_s}{k}\}
\end{align*}
Note, that 
\begin{align*}
  E_{S,T} \subseteq \{a\in [0,1]^n| \quad a_{s} > \sum_{t \in T} a_t\} 
\end{align*}
So let's compute the measure of $E_{S,T}$:

\begin{align*}
     \mathrm{Leb}(E_{S,T}) = \mathrm{Leb}(\{a\in [0,1]^n|\forall t\in T, a_t\leq \frac{a_s}{k}\}) = \int_0^1 (\frac{a_s}{k})^k da_s = \frac{1}{k^k\cdot (k+1)} \geq \frac{1}{(k+1)^{k+1}} 
\end{align*}
So, 
\begin{align*}
    \mathrm{Leb}(B_{S,T})\leq 1-\mathrm{Leb}(E_{S,T})  \leq 1 - \frac{1}{(k+1)^{k+1}}
\end{align*}

Now, consider the set 
$$B^m_{S,T}=\{a_1,\ldots,a_m \in B_{S,T} \},$$
Then we have 
$$ \mathrm{Leb}(B^m_{S,T})\leq\left(1 - \frac{1}{(k+1)^{k+1}}\right)^m$$
The vectors $a_1,\ldots,a_m $ will not define a monotone embedding, if there exists an extreme pair $S,T$  which is in $B^m_{S,T} $. The probability of this happening can be bounded by a union bound. There are $n \cdot {n-1 \choose k} $ extreme pairs, and for simplicity we will replace this number with a larger but simpler expression $n^{k+2}>n \cdot {n-1 \choose k} $. The union bound will then give us 
$$\mathrm{Leb}\{a_1,\ldots,a_m \text{ do not define a monotone embedding } \}\leq n^{k+2} \left(1-\frac{1}{(k+1)^{k+1}}\right)^m  $$
It is sufficient to show that this expression is smaller than $1$, for our $m$, which, by taking a logarithm, is equivalent to requiring that 
$$(k+2)\ln \left( n \right)+ m\ln \left( 1-\frac{1}{(k+1)^{k+1}} \right)<0 $$
Using Taylor's expansion it can be shown that $\ln(1-x)<-x$ for $x\in (0,1)$, and as a result it is sufficient to choose $m$ so that
$$(k+2)\ln \left( n \right)-  \frac{m}{(k+1)^{k+1}} <0 $$
or equivalently 
$$m> (k+1)^{k+1}\cdot (k+2) \cdot  \ln(n). $$
For convenience we slightly enlarge this lower bound to obtain $m\geq (k+2)^{k+2} \ln(n) $
\end{proof}
\lowerBoundK*
\label{lower_bound1}
\begin{proof}
The proof is partially inspired by the proof in \cite{dushnik1941partially}.

\paragraph{Lower bound on $k$:}
We want to show the lower bound of $\mm(V,k) \geq \min\round{2k, n-2}$. As in the proof of \cref{thm:dim_geq_n}, we select a  "maximal singleton element" $v^*_i \in V$ for every dimension $i \in \set{1, 2, \cdots, \mm(V, k)}$. Thus we must have: \begin{align*}
    \text{for each }i \in [\mm(V, k)]: \ F(\set{v^*_i})[i] \geq F(\set{v})[i], \forall v \in V
\end{align*} 
Now, if $\mm(V, k) \leq |V|-2$,  we can find two elements $\set{u_1, u_2}$ disjoint from the collection $\mathcal{V} := \bigcup_{i=1}^{\mm(V,k)}\set{v^*_i}$.  Without loss of generality, we can assume that $F(\set{u_1})[i] \geq F(\set{u_2})[i]$ for most indices in $[\mm(V,k)]$. We can then construct a set $T$ containing $u_1$ and all the $v^*_i$ for all indices  $i$ for which $F(\set{u_1})[i] < F(\set{u_2})[i]$, thus we define:
\begin{align*}
    T := \set{u_1}\cup\set{v_j^*: F(\set{u_1})[j] < F(\set{u_2})[j]}
\end{align*}
We then have that:
\begin{align*}
    &F(\set{u_2})[i]\leq \max\{  F(\set{u_1})[i], F(\set{v^*_i})[i]  \} \leq F(T)[i], \forall i \in [\mm(V,k)]
\end{align*}
where the last inequality uses the monotonicity of $F$. This implies that $F(S) \leq F(T)$ However,  $S $ is not a subset of $T$. It follows that  $k \leq |T|-1 \leq \mm(V,k)/2$, as otherwise separability of $F$ in $\Pvk$ is violated. Hence, we get that: $\mm(V,k) \geq 2k$

\paragraph{Lower bound on $n$: }
For this proof, we will use the Erd\"os-Szekeres theorem, which in particular states, for a natural $N\geq 2$, that a sequence of real numbers with $\geq N^2$ elements has a subsequence of cardinality $N$ which is monotone (either monotonously increasing or monotonously decreasing). 
Assume we have $F:\Pvk\rightarrow \mathbb{R}^{m}$ a monotone embedding, where $V=[n]$. Assume by way of contradiction that: 
\begin{align*}
    \log_2(\log_3 n)>m  \text{ or equivalently }  n > 3^{2^m} 
\end{align*}

Consider the first coordinate and look at the sequence
\begin{align*}
     F(\{1\})[1],\ldots,F(\{n\})[1]
\end{align*}
This is a real-valued sequence, and so there is  a monotone subsequence $a_1<a_2<\ldots,<a_\ell $ of the original sequence $1,\ldots,n$, with  $\ell=\sqrt{3^{2^m}}=3^{2^{m-1}}$, such that:
\begin{align*}
    F(a_1)[1]\leq F(a_2)[1]\leq \ldots \leq F(a_\ell)[1] \text{ or } F(a_1)[1]\geq F(a_2)[1]\geq \ldots \geq F(a_\ell)[1]
\end{align*}
 Next, we consider the second coordinate of this subsequence, namely $F(a_1)[2],\ldots,F(a_\ell)[2]$ and obtain a new subsequence $\{a_j\}$ such that both the sequences: $\{F(a_j)[1]\}$ and  $\{F(a_j)[2]\}$ are ordered monotonically, and the size of the subsequence is the square root of the previous one, namely $3^{2^{m-2}}$.  After doing this  $m$ times, we have a subsequence of size $3$. Namely, we have three distinct elements $u,v,w \in V$ such that for all $i=1,\ldots,m$,
 \begin{align*}
 \text{either } F(\{u\})[i]\leq F(\{v\})[i] \leq F(\{w\})[i] \text{ or }  F(\{u\})[i]\geq F(\{v\})[i] \geq F(\{w\})[i]
 \end{align*}

It follows that for all $i$, we have:
\begin{align*}
    F(\{v\})[i]\leq \max\left(F(\{u\})[i],F(\{w\})[i] \right) \leq F(\{u,w\})[i]
\end{align*}
Thus, $F(\{v\}) \leq F(\{u,w\})$, which is a contradiction since $\{v\}$ is not contained in $\{u,w\}$. 
\end{proof}

\paragraph{A refined lower bound:}Now, we may extend the above proof idea to generalize the lower bound on $\mm(V,k)$, which is useful especially when $|V| >> k$. In such cases, the following gives a tighter lower bound on $\mm(V, k)$ which is quadratic in $k$.
\begin{lemma}
    \label{lemma:refined_bound_k}
    For all $\ell \in [k]$, we have $\mm(V, k) \geq \min\round{|V| - \ell, \ell k +2\ell - \ell^2}$. More specifically, if $\ell = \frac{k+2}2$, then $\mm(V, k) \geq \min\curly{|V| - \frac{k+2}2, \round{\frac{k+2}2}^2}$
\end{lemma}
\begin{proof}
    Consider any $\ell \in [k]$. Like in the proof just before, we consider a "maximal singleton element" $v^*_i \in V$ for every dimension $i \in \set{1, 2, \cdots, \mm(V, k)}$ and obtain the collection $\mathcal{V} := \bigcup_{i=1}^{\mm(V,k)}\set{v^*_i}$. Let us assume $\mm(V, k) \leq |V| - \ell$. Thus, we can produce the collection $\mathcal{U} := \set{u_1, u_2, \cdots, u_\ell}$ disjoint from $\mathcal{V}$, i.e $\mathcal{V} \cap \mathcal{U} = \emptyset$. For each $u_q \in \mathcal{U}$, let $m_q$ be the number of co-ordinates $j \in [\mm(V,k)]$ such that $F(\set{u_q})[j] > F(\set{u_r})[j], \forall r \in [\ell]\setminus\set{q}$. Clearly, $\sum_{q \in [\ell]}m_q \leq \mm(V, k)$, thus, $\exists \ell_0 \in [\ell]\setminus$ such that $m_{\ell_0} \leq \frac{\mm(V,k)}\ell$. Like before, we consider the sets $S = \set{u_{\ell_0}}$. We construct $T$ as follows: we select $\mathcal{U}\setminus\set{u_{\ell_0}}$ and take union with all $\set{v^*_j}$ for each dimension $j$ in which $F(\set{u_{\ell_0}})[j] > F(\set{u_r}), \forall r \in [\ell]\setminus\set{\ell_0}$. Since $\set{v^*_j}$ is the maximal singleton for the dimension $j$, it follows that $F(\set{u_{\ell_0}})[j] \leq F(\set{v^*_j})[j]$. Hence, if we define:
    \begin{align*}
        T = \round{\mathcal{U}\setminus\set{u_{\ell_0}}} \bigcup \set{v^*_j: F(\set{u_{\ell_0}})[j] > F(\set{u_r}), \forall r \in [\ell]\setminus\set{\ell_0}}
    \end{align*}
    By the choice of $\ell_0$, we have that $\abs{\set{v^*_j: F(\set{u_{\ell_0}})[j] > F(\set{u_r}), \forall r \in [\ell]\setminus\set{\ell_0}}} \leq \frac{\mm(V, k)}\ell$ Now, we must have that, $\forall j \in [\mm(V, k)], \exists t \in T\text{ such that }F(\set{t})[j] \geq F(\set{u_{\ell_0}})[j]$. By monotonicity of $F$, we thus have that $F(\set{u_{\ell_0}})[j] \leq F(T)[j], \forall j \in [\mm(V,k)]$ which implies $F(S) \leq F(T)$. But by design, $S \cap T = \emptyset$, which contradicts separability of $F$. Thus, we must have that, $|T| > k$, i.e $|T| - 1 \geq k$. But as stated before, $|T| \leq \frac{\mm(V,k)}\ell + \ell - 1$. Combining these two, we get that: $\frac{\mm(V,k)}\ell \geq k + 2 - \ell$, which implies $\mm(V, k) \geq k\ell + 2\ell - \ell^2$, thus proving our result.

    Now, we already have a sufficient upper bound on $\mm(V,k)$ from~\cref{thm:upper_bound_V} that's $O\round{k^{k+2}\log(n)}$. Now, when $n >> k$ we have that $O\round{k^{k+2}\log(n)} < n - O(k)$, in which case we may apply $\ell = \frac{k+2}2$ in the above lemma. And since we already know that $O\round{k^{k+2}\log(n)} < n - O(k)$, we thus get that $\mm(V,k) < n - O(k)$. Thus, in this case the above lemma gives us $\mm(V,k) \geq \round{\frac{k+2}2}^2$, which is a tighter quadratic lower bound in $k$.
\end{proof}
\paragraph{When $k=1$}\label{app:k=1}
In  the degenerate case where only a single set element is allowed, $k=1$, it is possible to construct a MAS function even for the uncountable ground set $V=[-1,1]$ by defining
$
F:\mathcal{P}_{\leq 1}([-1,1]) \rightarrow \mathbb{R}^{2}
$
via       
\begin{align}
	\label{degenerate_case}
	F(S) =
	(-1, -1) \ \text{if } S = \emptyset\quad  \text{ and }\quad  F(S) =   
	(-x, x)  \  \text{if } S = \{x\}, 
\end{align}
\paragraph{Separable embeddings (even non monotone) don't exist:}Now, as discussed in the main text, we give a justification here on why we choose to relax separability and not monotonicity in the definition of weakly \mas embeddings. Here, we show that, with the added assumptions of injectivity and continuity, even non-monotone set-to-vector embeddings do not exist in any dimension for uncountable ground sets.
\begin{theorem}
\label{thm:separable_general}
    There does not exist a continuous injective separable set function taking values in $\R^m$ for any $m$ when the ground set $V$ is an open subset of $\R^d$
\end{theorem}
\begin{proof}
    \label{Impossibility_separating}
    We restrict ourselves to $\mathcal{P}_{=n}(V)$, i.e, only subsets of size $n$. Since $F$ is continuous and permutation-invariant, according to Theorem-7 of \cite{zaheer2018deepsets}, $f(\{x_1, \cdots, x_n\}) = \rho(\sum_{i=1}^n \phi(x_i))$, where $\rho, \phi$ are continuous and bijective(given in condition). Now, consider the space $\St \subset [0,1]^n, \St = \{(x_1, \cdots, x_n): x_1 < x_2 < x_3 < \cdots < x_n;  0 < x_i < 1, \forall i \in [n]\}$. Note that $\St$ is open in $[0,1]^n$ (to see this, let $\epsilon = \min_{i \in [n-1]} (x_{i+1} - x_i)$. The ball of radius $\frac{\epsilon}{2}$ is contained in the domain). Since $ F: \St \to \R^n$ is an injective, continuous map. Thus, by Invariance of Domain Theorem(\cite{munkres198479}), $F(\St)$ is open in $\R^n$ as well. Note that each element of $S$ uniquely corresponds to an element of $\mathcal{P}_{=n}(V)$. Thus, $F(\mathcal{P}_{=n}(V))$ is open in $\R^n$. Now, consider any set $A = \{a_1, \cdots, a_n\}$. Since $F$ is an open map, we have that $\exists \delta > 0$ such that $B(F(A), \delta) \subseteq F(\mathcal{P}_{=n}(V))$. Thus, $\exists B \in \mathcal{P}_{=n}(V)$ such that $F(B)_i = F(A)_i + \frac{\delta}{2n}, \forall i \in [n]$. But then, $F(B) > F(A)$ implies $A \subset B$ but $|A| = |B| = n$ by construction. This gives a contradiction.
\end{proof}

\subsection{Proofs of results on weakly \mas functions}

\paragraph {Proposition 6.}
   If the deep set model in \eqref{eq:parametric_setfn} is implemented with a vector-to-vector monotonously increasing function $\mtwo$ and a non-negative activation $\sigma$, then $F(\bullet; (A,b))$ is monotone for every $(A,b)$. If in addition $\sigma$ is monotone (increasing or decreasing), then there exists $S \not \subseteq T$ with $F(S; (A,b))\leq F(T; (A,b)) $ for all $A,b$.   
\begin{proof}
 Let $x<y$ and consider the two sets: 
 \begin{align*}
     S = \{\frac{x+y}{2}\},T = \{x,y\}
 \end{align*}
 And we claim that for all $A,b$, although $S\not\subseteq T$, $F(S,(A,b))\leq F(T,(A,b))$. Note that as $x<\frac{x+y}{2}<y$, then it must have been that 
 \begin{align*}
     Ax+b\leq A\cdot\frac{x+y}{2}+b\leq Ay+b 
 \end{align*}
 Or
  \begin{align*}
     Ay+b\leq A\cdot\frac{x+y}{2}+b\leq Ax+b 
 \end{align*}
 Then, by (weakly) monotonicity of the activation $\sigma$ and of $M_2$, it must be that 
 \begin{align*}
     \sigma(A\cdot\frac{x+y}{2}+b)\leq \sigma(A\cdot x+b)
 \end{align*}
 Or that 
  \begin{align*}
     \sigma(A\cdot\frac{x+y}{2}+b)\leq \sigma(A\cdot y+b)
 \end{align*}
 In any case, it's true that 
 \begin{align*}
     \sigma(A\cdot\frac{x+y}{2})\leq \sigma(A\cdot x + b)+\sigma(A\cdot y+b)
 \end{align*}
 Then, by monotonicity of $\mtwo$, 
 \begin{align*}
          F(S,(A,b))=\mtwo\cdot(\sigma(A\cdot\frac{x+y}{2}+b))\leq \mtwo\cdot(\sigma(A\cdot x+b)+ \sigma(A\cdot y+b))=F(T,(A,b))
 \end{align*}
 Thus, 
 \begin{align*}
     F(S,(A,b))\leq F(T,(A,b))
 \end{align*}
\end{proof}

\STnonmon*
\begin{proof}
We need to come up with a class of tuples of parameters $(W_Q, W_K, W_V)$ such that the function $S \mapsto F(S; W_Q, W_K, W_V)$ is not monotone. Choose $W_Q = W_K = W$ such that $W$ is full-rank. Choose any non-zero $W_V$. Now, choose some vector $\bx_1$ such that $W_V\bx_1$ is non-zero. By swapping $\bx_1$ with $-\bx_1$ if necessary, we can choose some index $j$ such that for $\bv_1 = W_V\bx_1$, the $j$-th index is negative, \ie $\bv_1[j] < 0$. By assumption we also have that $\bq_1 = \bk_1$ is a non-zero vector, and thus has a positive norm. Now let $S$ be the singleton set $S = \set{\bx_1}$, which is a subset of $T = \set{\bx_1, \mathbf{0}_d}$ where $\mathbf{0}_d$ is the $d-$dimensional all-zero vector. Note that, $F(S; W_Q, W_K, W_V) = \bv_1$. And since $\bq_2, \bk_2, \bv_2$ are all zero, we have: $F(T; W_Q, W_K, W_V) = (\alpha_{1,1} + \alpha_{1,2})\bv_1$. Note that, $\frac{\alpha_{1,1}}{\alpha_{1,1}+\alpha_{1,2}} = \frac{e^{\norm{\bq_1}^2}}{e^{\norm{\bq_1}^2} + 1} + \frac12 > 1$. Hence, it follows that $F(T; W_Q, W_K, W_V)[j] < F(S; W_Q, W_K, W_V)[j]$ and thus, the non-monotonicity follows
\end{proof}

\paragraph{Proposition 8.}
    If $\mtwo$ is strictly monotone increasing, and $\sigma$ is a hat activation function, then $F(S,(a,b))=\mtwo(\sum_{x\in S}\sigma(a\tran x+b)) $ is a weakly \mas function. 
\begin{proof}
    It's clear the function is a set function.
    Let $S\subseteq T$. Since $\sigma$ is positive, then for any $a,b$, 
    \begin{align*}
        \sum_{x\in S}\sigma(a^Tx+b)\leq \sum_{x\in S}\sigma(a^Tx+b) + \sum_{x\in T\setminus S}\sigma(a^Tx+b)=\sum_{x\in T}\sigma(a^Tx+b)
    \end{align*}
    So 
    \begin{align*}
        F(S,(a,b))\leq F(T,(a,b))
    \end{align*}
    Denote by $[s_1,s_2]$ the support of $\sigma$, and choose some point in it such that $s\in [s_1,s_2]:\sigma(s)>0$.
    Now, let $S\not\subseteq T$. Then there exists $z\in V:C_S(z)>C_T(z)$.
    Now, remove $z$ from both sets $C_T(z)$ times. This subtracts from both function values the same amount.
    Now, we have that $z\in S,z\not\in T$. Denote by $\epsilon =\frac{min_{t \in T}|z-t|>0}{2}$. Choose
    \begin{align*}
        a = \frac{s_2-s_1}{\epsilon}, b =-(a\cdot z - s)
    \end{align*}
    We claim that $\forall t\in T, \sigma(a\cdot t+b)=0$. Assume it's not zero, thus $s_1 < a\cdot t + b< s_2$. Thus
    \begin{align*}
         s_1 < \frac{s_2-s_1}{\epsilon}\cdot t - a\cdot z + s  < s_2
    \end{align*}
    Thus,
    \begin{align*}
        s_1-s < \frac{s_2-s_1}{\epsilon}\cdot t -\frac{s_2-s_1}{\epsilon}\cdot z  < s_2-s
    \end{align*}
    Thus, 
    \begin{align*}
        \frac{s_1-s}{s_2-s_1}\cdot \epsilon <  t - z  < \epsilon \cdot \frac{s_2-s}{s_2-s_1}
    \end{align*}
    But, 
    \begin{align*}
        |\frac{s_1-s}{s_2-s_1}|\leq 1 \\
        |\frac{s_2-s}{s_2-s_1}|\leq 1
    \end{align*}
    Thus, 
    \begin{align*}
        -\epsilon\leq t - z \leq \epsilon
    \end{align*}
    A contradiction to the definition of $\epsilon$. Thus, $F(T,(a,b))=\mtwo(0)$.
    But 
    \begin{align*}
         \sigma(a\cdot z + b) = \sigma(a\cdot z-a\cdot z +s_1) = \sigma(s_1)>0
    \end{align*}
    Thus 
    \begin{align*}
        F(S,(a,b)) = \mtwo(\sum_{x\in S}\sigma(a\tran x+b))\geq \mtwo(\sigma(s_1))>\mtwo(0)=F(T,(a,b))
    \end{align*}
    And we are done.
\end{proof}

\proprelumas*
\begin{proof}
    Consider the function $\tri$ as in the third plot of Fig.~\ref{fig:activations}. We know that $\tri$ belongs to the class of \hhat functions. From \ref{prop:weakmas_hat} we know that, for any $S \not \subseteq T \in \Pvinf$, $\exists \ba \in \R^d$ and $b \in \R$ such that $\sum_{x \in S} \tri(\ba\tran x + b) > \sum_{y \in T} \tri(\ba\tran y + b)$. Now, consider the following parameters:
    \begin{equation}
        \label{eq:relu_mas_params}
        \begin{split}
            & \boldA = \sbrac{\ba, \ba, \ba}\tran \in \R^{3 \times d}, \boldb = [b+1, b-1, b]\tran \in \R^3\\
            & \ba_1 = [1,1,-2]\tran \in \R^3, \text{ and }b_1 = 0
        \end{split}
    \end{equation}
   Then, for the above values of parameters, we have: 
   \begin{align*}
       \relu \circ \mathcal{A}_2(x) = \sbrac{\relu\round{(\ba\tran x + b) + 1}, \relu\round{(\ba\tran x + b) - 1}, \relu\round{\ba\tran x + b}}
   \end{align*}
   Thus, we have: 
   \begin{align*}
       &\ba_1\tran\round{\relu \circ \mathcal{A}_2(x)} + b_1 \\
       & =  \relu\round{(\ba\tran x + b) + 1} +  \relu\round{(\ba\tran x + b) - 1}- 2 \relu\round{\ba\tran x + b} \\
       &= \tri\round{\ba\tran x + b}
   \end{align*}
   Thus we have, $\relu \circ \mathcal{A}_1 \circ \relu \circ \mathcal{A}_2(x) = \relu \circ \tri\round{\ba\tran x + b} = \tri\round{\ba\tran x + b}$. Hence, $F(S) = \sum_{x \in S} \tri\round{\ba\tran x + b} > \sum_{y \in S} \tri\round{\ba\tran y + b} = F(T)$ and weak separability is proved. Since $\mtwo$ is monotone increasing and $\relu$ is non-negative, it follows that $F$ is monotone as well, hence weakly \mas.
\end{proof}
\subsection{H\"older separability of \mas functions}
\flowerholder*

\begin{proof}
    We consider $V \subset \R^d$ to be a compact set with maximum norm of $1$. Also, we note that, if $\sigma(x)$ is a \hhat function with support in $[\gamma_1, \gamma_2]$, then $\sigma(\frac{x - \gamma_1}{\gamma_2 - \gamma_1})$ is a \hhat function with support in $(0,1)$. For this proof, we thus consider $\sigma$ to have support in $[0,1]$. For support in general $[\gamma_1, \gamma_2]$ and a general norm bound of $V$, the distributions on $b, c$ need to be scaled and shifted to get same Lower H\"older results with the H\"older constants scaled appropriately. \footnote{If $\sup_{x \in V}\norm{x} \leq B$, and $\supp(\sigma)\subseteq[\gamma_1, \gamma_2]$ then distributions of $b, c$ should be shifted by $\gamma_1$ and linearly scaled by $\delta: = \frac{B}{\gamma_2 - \gamma_1}$ and to be: $b \sim \Unif(-\gamma_1-\delta, -\gamma_1+\delta)$ and $c \sim \Unif(0, 2\delta)$ respectively}. We recall the definition of $\das(S, T)$ from equation~\ref{eq:asymmetric_dist}. Let $S = \set{x_1, \cdots, x_\sizes}$ and let $T = \set{y_1, \cdots, y_\sizet}$, where $S, T \subset V$. Clearly by the definition of $\das(S, T)$, it is sufficient to consider $|S| \leq |T|$, i.e $\sizes \leq \sizet \leq k$.
    
    Using the definition of EMD gives us the following equivalent definition of $\das(S, T)$: If $\Omega_{\sizes, \sizet} := \set{ \tau : [\sizes] \to [\sizet], \tau \text{ is injective}}$ , then $\das(S, T) = \min_{\tau \in \Omega_{\sizes, \sizet}}\sum_{i=1}^\sizes\norm{x_i - y_{\tau(i)}}_2$. Let $\tau^*$ be the optimal coupling between $[\sizes] \text{ and } [\sizet]$, i.e $\tau^* = \arg\min_{\tau \in \Omega_{\sizes, \sizet}}\sum_{i=1}^\sizes\norm{x_i - y_{\tau(i)}}_2$, which means $\das(S, T):=\sum_{i=1}^\sizes\norm{x_i - y_{\tau^*(i)}}_2$. Now, $\forall i \in [\sizes]$, we define $\Delta_i := \norm{x_i - y_{\tau^*(i)}}_2$. For any $x \in \R^d, r > 0$, let $\B(x, r)$ denote the Euclidean ball centered at $x$ with radius $r$. We define $\Omega_i := \set{\ell \in [\sizes]: y_{\tau^*(\ell)} \in \B(x_i, \Delta_i)}, \forall i \in [M]$. With these notations, we make the following observation:

    \paragraph{Claim-1:}
        Consider any $ r \in (0, k)$. Then, $\exists i_0 \in [\sizes]$ such that $\sum_{\ell \in \Omega_{i_0}}\Delta_\ell \leq \Delta_{i_0}r$ and $\Delta_{i_0} \geq \frac{r^k}{k^{k+1}}\das(S, T)$ 
    
    \begin{proof}
        For the sake of contradiction, we assume otherwise, i.e $\forall i \in [\sizes]$, we have: either $\Delta_i \leq \frac{r^k}{k^{k+1}}\das(S, T)$ or $\sum_{\ell \in \Omega_i}\Delta_\ell \geq \Delta_i r$. WLOG, assume the following order on $\Delta_i$'s: $\Delta_1 \geq \Delta_2 \geq \cdots \geq \Delta_M$. We build the following directed graph $\Gcal$ on $[\sizes]$. Start with the node $v_0 = 1$ and add one node to $\Gcal$ at a time. At step $t$, we select $v_t := \arg\max_{u \in \Omega_{v_{t-1}}} \Delta_u$ and add the edge $(v_{t-1}, v_t)$ to $\Gcal$ along with the new node $v_t$. 
        
        Note that, by definition of $\das(S, T)$ we have that: $\das(S, T) = \sum_{i=1}^{\sizes}\Delta_i$. This, implies that: $\Delta_{v_0} \geq \frac{\das(S, T)}M$, as we take argmax over entire $[\sizes]$. Thus, $\Delta_1 \geq \das(S,T)/M \geq \das(S,T)/k$. Thus, by the contradictory assumption, we must have: $\sum_{u \in \Omega_{v_0}}\Delta_u \geq \Delta_1 r$. Since $v_1 = \arg\max_{u \in \Omega_{v_0}}\Delta_u$, it thus follows that: $\Delta_{v_1}\geq \frac rk\Delta_1 \geq \frac {r}{k^2}\das(S,T)$. Following the similar argument, we thus have, $\Delta_{v_2} \geq \frac rk\Delta_{v_1}$ and so on. Thus, $\forall t \in \NN$ we must have: $\sum_{u \in \Omega_{v_t}}\Delta_u \geq \Delta_{v_t}r$. Thus, for $v_{t+1}$, defined by: $\arg\max_{u \in \Omega_{v_t}}\Delta_u$, we must have: $\Delta_{v_{t+1}} \geq \frac rk \Delta_{v_t}$. Since this holds for every $t \in \NN$ and since $r > 0$ by hypothesis, we must have: $\Delta_{v_t} \geq \frac rk \Delta_{v_{t-1}} \geq \round{\frac rk}^2 \Delta_{v_{t-1}} \geq \cdots \geq \round{\frac rk}^t\Delta_{v_0}$. 
        
        As we work with finite $\sizes, \sizet$ either $\Gcal$ is a DAG and this process terminates; or $\Gcal$ has a cycle. In former case, at the termination step $\bm{t_t}$ we cannot add any more nodes, thus $\Omega_{\bm{t_t}} = \emptyset$. Since all nodes are unique in this case, we must have $\bm{t_t} \leq k$. So, $\Delta_{\bm{t_t}} \geq \round{\frac rk}^{\bm{t_t}}\Delta_{v_0} \geq \frac1k \round{\frac rk}^{\bm{t_t}}\das(S, T)$. Since $\frac rk \in (0,1)$ by hypothesis and since $\bm{t_t} \leq k$, we thus have: $\Delta_{\bm{t_t}} \geq \frac 1k \round{\frac rk}^k\das(S, T)$, and hence done with this case.
        
        For the second case, consider the smallest cycle in $\Gcal$. Let this cycle of length $p$ be $\mathcal{C} = \set{v_\beta \to v_{\beta+1} \to \cdots \to v_{\beta + {p-1}} \to v_\beta}$ for some $\beta \geq 0$. Then, we have: $\beta + i + 1 \in \Omega_{\beta+i}, \forall i \in \set{0, \cdots p-1}$, where addition of indices is modulo $p$. Thus we have: $\norm{x_{v_\beta + i} - y_{\tau^*(v_\beta + i +1)}}_2 < \norm{x_{v_\beta + i} - y_{\tau^*(v_\beta + i)}}_2, \forall i \in \set{0, \cdots, p-1}$. Taking sum, we have the following: $\norm{x_{v_\beta} - y_{\tau^*(v_\beta+1)}}_2 + \norm{x_{v_\beta+1} - y_{\tau^*(v_\beta+2)}}_2 + \cdots + \norm{x_{v_\beta + p-1} - y_{\tau^*(v_\beta)}}_2 < \sum_{j=0}^{p-1}\norm{x_{v_\beta+j} - y_{\tau^*(v_\beta+j)}}_2$. Hence, switching from the coupling $i \mapsto \tau^*(i)$ to $i \mapsto \tau^*(i+1)$ for $i \in \set{v_\beta, \cdots, v_{\beta + p-2}}$ and from $v_\beta + p - 1 \mapsto \tau^*(v_\beta+ p - 1)$ to $v_\beta + p - 1 \mapsto \tau^*(v_\beta)$ strictly reduces the value of the sum $\sum_{i=1}^\sizes\norm{x_i - y_{\tau^*(i)}}_2$, but we started with the optimal coupling $\tau^*$. This gives the reqd. contradiction. 
    \end{proof}
    We also make the following observation regarding this "optimal coupling" $\tau^*$:
    
    \paragraph{Observation-2:} For any given $i \in [\sizes]$, if $y_j \in T \cap \Omega_i$ for some $j \in [\sizet]$, then $\exists \ell \in [\sizes]$ such that $j = \tau^*(\ell)$
    \begin{proof}
        The proof follows from the following observation: if $y_j \in \Omega_i$ for any $i \in [\sizes]$, then we must have: $\norm{x_i - y_j}_2 \leq \Delta_i = \norm{x_i - y_{\tau(i)}}_2$. Thus, if $y_j$ was "free", i.e if $\not \exists \ell \in [\sizes]: j = \tau^*(\ell)$, then we can switch from $i \mapsto \tau^*(i)$ to $i \mapsto j$, and this would reduce the sum $\sum_{i=1}^\sizes\norm{x_i - y_{\tau^*(i)}}_2$, but $\tau^*$ was the optimal coupling. This gives a contradiction
    \end{proof}

    Now, as per the starting discussion, we consider $\supp(\sigma) = [0,1]$, and the proof for general support of $[\gamma_1, \gamma_2]$ requires the scaling and shifting as mentioned before.
    
    By assumption we have, $\lim_{ t\to 0^+}\sigma\dash(t) > 0$. We define $\kappa_1 := \frac{\lim_{ t\to 0^+}\sigma\dash(t)}2$. Since $\sigma$ is piecewise continuously differentiable as per assumption, we must have $\sigma\dash(t) \geq \kappa_1, \forall t \in (0, \omega)$ for some $\omega \in (0,1)$ by the continuity of $\sigma\dash$. Note that, by Lagrange's Mean Value Theorem, we have that:
    \begingroup
    \allowdisplaybreaks
    \begin{align*}
        &\forall x, y \in (0, \omega), \frac{\abs{\sigma(x) - \sigma(y)}}{\abs{x-y}} = \sigma\dash(z) , \text{ where } z \in (x,y)\\
        &\implies \forall x, y \in (0, \omega), \abs{\sigma(x) - \sigma(y)} \geq \kappa_1\abs{x-y}, \text{ since } z \in (0, \omega) \\
        &\implies \lim_{y \to 0^+} \abs{\sigma(x) - \sigma(y)} \geq \kappa_1\abs{x}, \text{ as }\sigma\text{ is cts.}\\
        &\implies \sigma(x) \geq \kappa_1x, \forall x \in (0, \omega), \text{ as }\sigma \geq 0
    \end{align*}
    \endgroup
    
     Also, $\sigma$ is upper Lipschitz with constant $\kappa_2 > 0$ per definition of Hat fn.  Now, consider the $i_0$ coming from the Claim as proved above. We define $\omega':= \min(\frac{\omega}{1-\omega}, 1)$. Let $P \in \NN$ be such that $\frac {\kappa_1}{P} < \frac 12$ and $\frac{\kappa_1}{\kappa_2 P} < k$ and $\omega' > \frac 2P$. Note that, such a $P \in \NN$ exists, since taking $P \to \infty$ satisfies all the 3 conditions. 
     
     Now, for a given $a \in \S^{d-1}$, consider the sets $S_a = \set{a\tran x: x \in S}, T_a = \set{a\tran y: y \in T}$.  Note that, since we have $\norm{x_i}, \norm{y_j} \leq 1, \forall i \in [M], j \in [N]$ as per earlier assumption, we get $S_a, T_a \subseteq (-1,1)$. For given $a \in \S^{d-1}$, we define the "optimal coupling" between the real-valued sets $S_a, T_a$ as: $\tau^*_a:= \arg\min_{\tau \in \Omega_{M, N}} \sum_{i=1}^M\abs{a\tran x_i - a\tran y_{\tau(i)}}$. Based on this, we define $\Delta^a_i := \abs{a\tran x_i - a\tran y_{\tau^*_a(i)}}$ and $\Omega^a_i = \set{\ell \in [M]: y_{\tau^*_a(\ell)} \in \B(a\tran x_i, \Delta^a_i)}$. We have: $\das(S_a, T_a) := \sum_{i=1}^M \Delta^a_i$. Let $i_0 \in [M]$ be the index that comes from applying the claim to the sets $S_a, T_a$. Now, we define the regions $\bm{B}_a := (a\tran x_{i_0} - \omega'\Delta^a_{i_0}, a\tran x_{i_0} - \frac{\Delta^a_{i_0}}P)$ and for a given pair $(a \in \S^{d-1}, b \in \R)$, we consider $\bm{C}_{a,b} := (\frac{a\tran x_{i_0}-b}\omega, a\tran x_{i_0}+\Delta^a_{i_0}-b)$. Thus, we have:
     \begin{align*}
         &\round{a\tran x_{i_0}+\Delta^a_{i_0}-b} - \round{\frac{a\tran x_{i_0}-b}\omega} \\
         = & \frac{(1-\omega)(b - a\tran x_{i_0}) + \omega\Delta^a_{i_0}}{\omega}\\
         = & \frac{1-\omega}\omega\round{(b- a\tran x_{i_0}) + \frac{\omega}{1-\omega}\Delta^a_{i_0}} \\
         \geq & \frac{1-\omega}\omega\round{b- (a\tran x_{i_0} - \omega'\Delta^a_{i_0})} > 0, \forall b > a\tran x_{i_0} - \omega'\Delta^a_{i_0}
     \end{align*}
    The above analysis shows that $\bm{C}_{a, b} \neq \emptyset, \forall b \in \bm{B}_a$, so the intervals are well defined. Now, if $b \in \bm{B}_a$ and $y < a\tran x_{i_0} - \Delta^a_{i_0}$ then we have: $y - b < a\tran x_{i_0} - \Delta^a_{i_0} - a\tran x_{i_0} + \omega'\Delta^a_{i_0} = (\omega' - 1)\Delta^a_{i_0} \leq 0$ by definition of $\omega'$. On the other hand, if $b \in \bm{B}_a$ and $ c \in \bm{C}_{a,b}$, for given any $ y > a\tran x_{i_0} + \Delta^a_{i_0}$ we have: $\frac{y - b}c \geq \frac{a\tran x_{i_0} - b + \Delta^a_{i_0}}{a\tran x_{i_0} - b + \Delta^a_{i_0}} =1$. Combining these, we get that if $\abs{y - a\tran x_{i_0}} \geq \Delta^a_{i_0}$, then $\frac{y - b}c \not \in (0,1), \forall b \in \bm{B}_a, c \in \bm{C}_{a,b}$. Thus, conditioning on a given $a \in \S^{d-1}$, we can write the conditional expectation taken over $b, c$ as follows:
    \begingroup 
    \allowdisplaybreaks
    \begin{align*}
        & \Exp_{b, c}\sbrac{\sum_{x \in S}\sigma\round{\frac{a\tran x-b}c}- \sum_{y \in T}\sigma\round{\frac{a\tran y-b}c}}_+\\
        \geq & \Exp_{b, c}\sbrac{\round{\sum_{x \in S}\sigma\round{\frac{a\tran x-b}c}- \sum_{y \in T}\sigma\round{\frac{a\tran y-b}c}}\indic{b \in \bm{B}_a, c \in \bm{C}_{a,b}}}_+, \text{using non-negative RV}\\
        \geq & \frac 14 \int_{b=a\tran x_{i_0} - \omega'\Delta^a_{i_0}}^{a\tran x_{i_0}}\int_{c=\frac{a\tran x_{i_0} - b}\omega}^{a\tran x_{i_0} + \Delta^a_{i_0} - b} \bigg[\sigma\big(\frac{a\tran x_{i_0}-b}c\big) - \\
        &\sum_{\ell \in \Omega^a_{i_0}} \bigg|\sigma\big(\frac{a\tran y_{\tau^*_a(\ell)}-b}c\big) \ - \sigma\big(\frac{a\tran x_\ell-b}c\big)\bigg|\bigg]_+\dif c\dif b\\
        \geq &\frac 14 \int_{b=a\tran x_{i_0} - \omega'\Delta^a_{i_0}}^{a\tran x_{i_0}}\int_{c=\frac{a\tran x_{i_0} - b}\omega}^{a\tran x_{i_0} + \Delta^a_{i_0} - b} \sbrac{\kappa_1 \frac{a\tran x_{i_0} - b}{c} - \kappa_2 \frac{\kappa_1}{P\kappa_2}\frac{\Delta^a_{i_0}}c }_+\dif c\dif b, \text{from Lipschitz cts. }\sigma\\
        =&\frac{\kappa_1}4\int_{b=a\tran x_{i_0} - \omega'\Delta^a_{i_0}}^{a\tran x_{i_0}}\sbrac{a\tran x_{i_0} - b - \frac{\Delta^a_{i_0}}P}_+\int_{c=\frac{a\tran x_{i_0} - b}\omega}^{a\tran x_{i_0} + \Delta^a_{i_0} - b}\frac{\dif c}{c} \dif b, \text{as }\relu(rx)=r\relu(x)\\
        =&\frac{\kappa_1}4\int_{b=a\tran x_{i_0} - \omega'\Delta^a_{i_0}}^{a\tran x_{i_0}-\frac{\Delta^a_{i_0}}P}\round{a\tran x_{i_0} - b - \frac{\Delta^a_{i_0}}P}\ln\round{\frac{\omega(a\tran x_{i_0}+\Delta^a_{i_0}-b)}{a\tran x_{i_0}-b}} \dif b\\
        =&\frac{\kappa_1}4\int_{z=\frac{\Delta^a_{i_0}}P}^{\omega'\Delta^a_{i_0}}\round{z - \frac{\Delta^a_{i_0}}P}\ln\round{\frac{\omega(z+\Delta^a_{i_0})}z}\dif z, \text{putting }z:=a\tran x_{i_0}-b\\
        \geq & \frac{\kappa_1}4\int_{z=\frac{\Delta^a_{i_0}}P}^{\frac{\omega'\Delta^a_{i_0}}2}\round{z - \frac{\Delta^a_{i_0}}P}\ln\round{\omega\round{1+\frac{\Delta^a_{i_0}}z}} \dif z. \text{using non-negativity of integrand}\\
        \geq & \frac{\kappa_1}4\ln\round{\omega\round{1 + \frac{2}{\omega'}}}\int_{z=\frac{\Delta^a_{i_0}}P}^{\frac{\omega'\Delta^a_{i_0}}2}\round{z - \frac{\Delta^a_{i_0}}P}\dif z\\
        \geq & \frac{\kappa_1}8\ln\round{2-\omega}\round{\frac{\omega'}2-\frac 1P}^2(\Delta^a_{i_0})^2, \text{ using } \omega' \leq \frac{\omega}{1-\omega}\\
        \geq& C \cdot \frac 1k \round{\frac{\kappa_1}{\kappa_2 P}}^k (\das^a(S, T))^2, \text{ where }C> 0\text{ as }\omega \in (0,1) \text{ and } P \geq \frac{2}{\omega'}\text{ by earlier choice}
    \end{align*}
    \endgroup
   Thus, we have shown that: $\Exp_{b, c}\bigg[\big[F(S)- F(T)\big]_+\big| a \bigg] \geq C'\cdot(\das^a(S,T))^2$. By law of total expectation, we have:
   \begingroup
   \allowdisplaybreaks
   \begin{align*}
       &\Exp_{a, b, c}\bigg[F(S) - F(T)\bigg]_+ = \Exp_a \Exp_{b, c}\bigg[\big[F(S)- F(T)\big]_+\big| a \bigg]\\
       \geq & C'\Exp_a \big(\das^a(S, T)\big)^2 \geq C'\round{\Exp_a \das^a(S,T)}^2, \text{by Jensen's ineq.}
   \end{align*}
   \endgroup
   Thus, to show Lower H\"older separability with exponent $\lambda =2$, it remains to show that, $\Exp_{a \sim \Unif(\S^{d-1})} \das^a(S,T) \geq c_1 \cdot \das(S, T)$, where $c_1 > 0$ is a constant. Now, we make the following observation:

   \paragraph{Observation-3:}Suppose we're given $\ell \in \NN$ and $\ell$ many non-zero vectors $u_1, u_2, \cdots, u_\ell \in \R^d$. If $ a \sim \S^{d-1}$, then $\exists \delta(\ell) > 0$ such that: $\Prob\curly{\abs{a\tran u_i} \geq \delta(\ell) \norm{u_i}, \forall i \in [\ell]} \geq \frac 12$
   \begin{proof}
       Firstly, we note that, for a fixed $i \in [\ell]$, $\abs{a\tran u_i}$ is a continuous non-negative real-valued random variable. Thus, $\exists t_i \in \R_+$ such that $\Prob\curly{\abs{a\tran u_i} \geq t_i} \geq \frac 12$. Now, Choosing $\delta_i := \frac{t_i}{\norm{u_i}} > 0$ gives us: $\Prob\curly{\abs{a\tran u_i} \geq \delta_i\norm{u_i}} \geq \frac 12, \forall i \in [\ell]$. Since $\delta_i > 0, \forall i \in [\ell]$ and $\ell$ is finite, we thus have: $\delta:= \min_{i \in \ell}\delta_i > 0$. For this particular choice of $\delta$, we have that $\Prob\curly{\abs{a\tran u_i} \geq \delta\norm{u_i}} \geq \frac 12, \forall i \in [\ell]$, and the observation is proved.
   \end{proof}.
   Now, consider all $\bm{\ell_0} := \binom{N}{M}M!$ vectors $(x_i - y_{\tau(i)}) \in \R^d$, where $\tau$ runs over all injective functions from $[M] \to [N]$. For this particular choice of $\bm{\ell_0}$, we get from the last observation that, $\exists \delta(N, M) > 0$ such that $\Prob\curly{\abs{a\tran\round{x_i - y_{\tau(i)}}} \geq \delta(N, M)\norm{x_i - y_{\tau(i)}}} \geq \frac 12$, for all possible injective $\tau: [M]  \to [N]$, we call this set of $a \in \S^{d-1}$ to be $A \subseteq \S^{d-1}$. Now, given a fixed $a \in \S^{d-1}$, we have a particular $\tau_a: [M] \to [N]$ that gives $\das^a(S, T) = \sum_{i=2}^M\abs{a\tran\round{x_i - y_{\tau_a(i)}}}$. Also, let $\tau^* : [M] \to [N]$ be the "optimal coupling" such that $\das(S, T) = \sum_{i=1}^M\norm{x_i - y_{\tau^*(i)}}$. Now, we can write:
   \begin{align*}
       & \Exp_{a \sim \Unif(\S^{d-1})} \das^a(S,T)\\
       = & \Exp_{a \sim \Unif(\S^{d-1})}\sbrac{\sum_{i=1}^M\abs{a\tran(x_i - y_{\tau_a(i)})}}\\
       \geq & \Exp_{a \sim \Unif(\S^{d-1})}\sbrac{\sum_{i=1}^M\abs{a\tran(x_i - y_{\tau_a(i)})}\bigg|a \in A}\Prob_a\curly{a \in A} \\
       \geq & \frac{\delta(N, M)}2 \Exp_a\sbrac{\sum_{i=1}^M \norm{x_i - y_{\tau_a(i)}}_2} \\
       \geq & \frac{\delta(N, M)}2 \Exp_a\sbrac{\sum_{i=1}^M \norm{x_i - y_{\tau^*(i)}}_2}, \text{as }\tau^*\text{ is optimal coupling}\\
       =& \frac{\delta(N, M)}2\round{\sum_{i=1}^M \norm{x_i - y_{\tau^*(i)}}_2} = \frac{\delta(N, M)}2\das(S, T)
   \end{align*}
   Hence, $F$ is H\"older separable with constant $\frac{C'\delta(N,M)}2$ and exponent $\lambda = 2$.
\end{proof}
\probabilityseparation*
\begin{proof}
    Denote by $f$ the scalar-valued set function for a co-ordinate of $F$; we proved it's a lower Holder function.
    Let $S\not\subseteq T$, then by lower continuity we know that,
    \begin{equation}
        \Exp_{w \sim \mu(\cdot)}\norm{\sbrac{f(S; w) - f(T; w)}_+}_1 \geq C \cdot \das(S, T)^2, \\ \ \text{for all } S, T \in \Pvk
    \end{equation}
    We know that 
    \begin{equation*}
    \begin{split}
        &c \cdot \das(S, T)^2\leq \Exp_{w \sim \mu(\cdot)}\norm{\sbrac{f(S; w) - f(T; w)}_+}_1 \\
        \leq&\Prob(f(S;w)\geq f(T;w))\cdot \sup_{w\in \W} |f(S;w)\geq f(T;w)|
    \end{split}
    \end{equation*}
    Now, note that as $f$ is a bounded function, denote by $M$ the supremum of $f$ over all sets $S$ and weights $w\in \W$. 
    Thus by the triangle equality, 
    \begin{equation*}
            c \cdot \das(S, T)^2 \leq 2M\cdot \Prob(f(S;w)\geq f(T;w))
    \end{equation*}
    Thus, we obtain that
    \begin{align*}
        \frac{c \cdot \das(S, T)^2}{2\cdot M} \leq \Prob(f(S;w)\geq f(T;w))
    \end{align*}
    Thus, the complement satisfies:
    \begin{align*}
       \Prob(f(S;w) < f(T;w))\leq 1 - \frac{c \cdot \das(S, T)^2}{2\cdot M}
    \end{align*}
    Now, taking $m$ independent copies of $f$, namely $F$, we have that 
    \begin{align*}
          \Prob(\forall i\in [m],F(S;w)_i < F(T;w)_i)\leq (1 - \frac{c \cdot \das(S, T)^2}{2\cdot M})^{m}
    \end{align*}
    Denoting by $C:=\frac{c}{2M}$ yield the desired:
    \begin{align*}
          \Prob(F(S;w) < F(T;w))\leq (1 - C \cdot \das(S, T)^2)^{m}
    \end{align*}
\end{proof}

\paragraph{Lower H\"older separability of $\relu$ networks} So far, we have only given results about Lower H\"older separability of \mas functions with Hat activations. However, as seen in Proposition \ref{prop:relu_mas}, we have seen that a two layer linear network with $\relu$ activation is weakly \mas, and as observed from the proof technique, such a 2 layer $\relu$ net can "simulate" a hat activation. In our arguments, we have shown Lower H\"older separability of Hat Activation based parametric set functions by finding a separating parameter point $w \in \Wcal$ for a given pair of sets $(S, T)$, and constructing an open set of non-zero measure, $\Wcal' \subseteq \Wcal$ around $w$. We lower bounded the expectation of the non-negative Random variable in the definition of Lower H\"older separability(in ~\cref{eq:lower_holder}) by restricting the expectation only to $\Wcal'$ and showing that it is proportional to $\das(S, T)^\lambda$. One can extend this exact same idea to construct the open set of parameters for the two later $\relu$ networks from the separable parameter point that we get from proof of Proposition \ref{prop:relu_mas}. 

Also, as obtained from Proposition \ref{prop:shallow_relu_not_sep}, we see that one layer $\relu$ networks are not even weakly \mas. But a key step  in that proof uses that one can find 3 collinear elements in the ground set $V$ and construct sets with them to obtain the necessary counter-example. Thus, a natural question is to ask: whether shallow 1-layer networks with $\relu$ activation are also weakly \mas with some additional assumptions on the domain. We show in the following result that, with the additional assumption of $V$ being a hypersphere, we may guarantee Lower H\"older separability for even shallow $\relu$ networks.

\begin{theorem}
    \label{thm:shallow_relu_lower_holder}
    Given a ground set $V \subseteq \S^{d-1}$. We define $F: \Pvk \to \R$ as $F(S) = \sum_{x \in S}\relu\round{a\tran x + b}$, where $a \sim \Unif(\S^{d-1})$ and $b \sim \Unif(-1,1)$. Then, for any $S, T \in \Pvinf$, we have: $\Exp_{a, b}\sbrac{F(S) - F(T)}_+ \geq C(k) \das(S, T)^{(d+3)2^k}$ for some constant $C(k) > 0$ depending only on $k$.
\end{theorem}
\begin{proof}
    Let $S = \set{x_1, \cdots, x_M}$ and $T = \set{y_1, \cdots, y_N}$, where $M \leq N \leq k$. And according to the definition of $\das(S, T)$ from~\cref{eq:asymmetric_dist} we can equivalently write $\das(S, T) = \min_{\pi \in \S_N}\sum_{i=1}^M\norm{x_i - y_{\pi(i)}}_2$. Let $\pi^*$ be the optimal solution of the above problem, then we can define the map $\tau: [M] \to [N]$ as follows: $\tau(i) = \pi^*(i), \forall i \in [M]$. Essentially, $x_i$ and $y_{\tau(i)}$ are coupled in the optimal alignment. Now, for any given $i \in [M]$, let $\Delta_i := \norm{x_i - y_{\tau(i)}}_2$. Then, consider the open Euclidean ball $\B(x_i, \Delta_i)$. Then if $j \in [N]$ such that $y_j \in \B(x_i, \Delta_i)$ then, $j =\tau(\ell)$ for some $\ell \in [M]$. Otherwise switching from $i \mapsto \tau(i)$ to $i \mapsto j$ reduces the cost, which violates optimal alignment. Let $\Omega_i := \set{\ell \in [M]: y_{\tau(\ell)} \in \B(x_i, \Delta_i)}$. Here, we make the following claim:
    \begin{claim}
    \label{claim:stupid_claim}
    $\exists \alpha \in [M]$ such that $\sum_{j \in \Omega_i}\norm{y_j - x_{\tau^{-1}(j)}}_2 \leq \frac{\Delta_\alpha^2}{16}$ and $\Delta_\alpha \geq \frac{16\das^{\twok}(S, T)}{(16k)^{\twok}}$
    \end{claim}
    \begin{proof}
        Suppose not. WLOG, assume the following order $\Delta_1 \geq \Delta_2 \geq \cdots \geq \Delta_M$. We build the following directed graph $G$ on $[M]$. Start with the node $v_0 = 1$ and add one node at a time. At step $t$, we select $v_t := \arg\max_{v \in \Omega_{v_{t-1}}} \Delta_v$ and add the edge $(v_{t-1}, v_t)$. By the assumption, $\frac{\Delta_{v_t}}{16k} \geq (\frac{\Delta_{v_{t-1}}}{16k})^2 \geq \cdots \geq (\frac{\Delta_{v_0}}{(16k)})^{\twok}$. Since we work with finite $M$, either $G$ is a DAG and this process terminates or $G$ has a cycle. In former case at termination step $T$ we cannot add any more nodes, thus $\Omega_T = \emptyset$, and $T \leq k$. So, $\Delta_T \geq \frac{16k}{k}\frac{\das^{2^T}(S, T)}{(16k)^{2^T}} \geq  \frac{16\das^{\twok}(S, T)}{(16k)^{\twok}}$ and we're done. Otherwise, consider the smallest cycle in $G$. Let this cycle of length $p$ be $\mathcal{C} = \set{v_\beta \to v_{\beta+1} \to \cdots \to v_{\beta + {p-1}} \to v_\beta}$. Then, we have $\norm{x_{v_\beta + i} - y_{\tau(v_\beta + i +1)}}_2 < \norm{x_{v_\beta + i} - y_{\tau(v_\beta + i)}}_2, \forall i \in \set{0, \cdots, p-1}$ and addition of sub-indices is modulo $p$. Thus, we have the following: $\norm{x_{v_\beta} - y_{\tau(v_\beta+1)}}_2 + \norm{x_{v_\beta+1} - y_{\tau(v_\beta+2)}}_2 + \cdots + \norm{x_{v_\beta + p-1} - y_{\tau(v_\beta)}}_2 < \sum_{j=0}^{p-1}\norm{x_{v_\beta+i} - y_{\tau(v_\beta+j)}}_2$. Hence, switching from the coupling $i \mapsto \tau(i)$ to $i \mapsto \tau(i+1)$ for $i \in \set{v_\beta, \cdots, v_{\beta + p-2}}$ and to $v_\beta + p - 1 \mapsto v_\beta$ strictly reduces the cost, but we started with the optimal coupling. This gives a contradiction. 
    \end{proof}
    Consider $\alpha$ as in the above claim. Then we define the region: $\mathcal{A} := \set{a \in \S^{d-1} : \norm{a - x_\alpha} < \frac{\Delta_\alpha}4}$. Then, $\forall a \in \mathcal{A}, \norm{a - x_\alpha}_2 < \norm{a - y_{\tau(\alpha)}}_2$. Thus, $a \tran x_\alpha > a \tran y_{\tau(\alpha)}$. Also, consider the region $\mathcal{B}_a := \set{b \in \R : -1 + \frac{\Delta_\alpha^2}{16} + \frac{\Delta_\alpha^2}{64} + \frac{\norm{a-x}^2}2 \leq b \leq -1 + \frac{\Delta_\alpha^2}8}$. For any given $a \in \mathcal{A}$, we have: $\frac{\norm{a-x}^2}2 \leq \frac{\Delta_\alpha^2}{32}$, thus $-1 + \frac{\Delta_\alpha^2}{16}+ \frac{\Delta_\alpha^2}{64} + \frac{\norm{a-x}^2}2 \leq -1 + \frac{7}{64}\Delta_\alpha^2 \leq -1 + \frac{\Delta_\alpha^2}8$, so $\mathcal{B}_a \neq \emptyset$. Also note that, $\forall a \in \mathcal{A}, b \in \mathcal{B}_a$, we have:
    \begingroup
    \allowdisplaybreaks
    \begin{align*}
        &\sbrac{F(S) - F(T)}_+ \geq \sbrac{\relu\round{a\tran x_\alpha + b} - \round{\sum_{\beta \in \Omega_\alpha}\relu\round{a\tran y_{\tau(\beta)}} - \relu\round{a\tran x_{\beta}}}}_+ \\
        & \geq  \sbrac{\relu\round{a\tran x_\alpha + b} - \sum_{\beta \in \Omega_\alpha}\abs{\relu\round{a\tran y_{\tau(\beta)}} - \relu\round{a\tran x_{\beta}}}}_+\\
        & \geq \sbrac{\relu\round{a\tran x_\alpha + b} - \sum_{\beta \in \Omega_\alpha}\abs{a\tran(x-y)}}_+ \\
        & \geq \sbrac{\relu\round{a\tran x_\alpha + b} - \sum_{\beta \in \Omega_\alpha}\norm{x_\beta - y_{\tau(\beta)}}}_+ \\
        & \geq \sbrac{a\tran x_\alpha + b - \frac{\Delta_\alpha^2}{16}}_+ = \sbrac{1 + b - \frac{\norm{a-x}^2}2 - \frac{\Delta_\alpha^2}{16}}_+ \geq \frac{\Delta_\alpha^2}{64}
    \end{align*}
    \endgroup
    Thus, we have: 
    \begingroup
    \allowdisplaybreaks
    \begin{align*}
        &\Exp_{a, b}\sbrac{F(S) - F(T)}_+ \geq \Exp_{a,b}\sbrac{(F(S) - F(T))\indic{a \in \mathcal{A}, b \in \mathcal{B}_a}}_+\\
        &\geq \frac{\Delta_\alpha^2}{64} \Prob\curly{a \in \mathcal{A}, b \in \mathcal{B}_a} \\
        &= \frac{\Delta_\alpha^2}{64S_{d-1}} \int_{a \in \S^{d-1} : a_1 \geq 1 - \frac{\Delta_\alpha^2}{32}} \frac{1}{2}\round{\frac{3}{64}\Delta_\alpha^2 - (a_1 - 1)} da \\
        & \geq \frac{\Delta_\alpha^4}{2^{13}S_{d-1}} \int_{a \in \S^{d-1} : a_1 \geq 1 - \frac{\Delta_\alpha^2}{32}} da \\
        & = \frac{\Delta_\alpha^4}{2^{13}}\int_{1 - \frac{\Delta_\alpha^2}{32}}^1 (1-x^2)^{\frac{d-3}2} dx \geq \frac{\Delta_\alpha^6}{2^{18}}\round{\frac{\Delta_\alpha^2}{32}}^{\frac{d-3}2}\round{2 - \frac{\Delta_\alpha^2}{32}}^{\frac{d-3}2}\\
        &\geq \frac{\Delta_\alpha^{d+3}}{2^{23}}
    \end{align*}
    \endgroup
    Now, using the bounds from~\ref{claim:stupid_claim}, we get that, $\Delta_\alpha \geq C(k) \das^{\twok}(S,T)$. Thus we have that, $\Exp\sbrac{F(S) - F(T)}_+ \geq C_1(k)\das(S,T)^{(d+3)\twok}$, and done
\end{proof}

\paragraph{Upper Lipschitz property of weakly \mas functions} We now show that, the \mas functions obtained by applying one layer neural network with \hhat activation for each element of the set, followed by sum aggregation are also Upper Lipschitz continuous in expectation with respect to the augmented Wasserstein distance, following the framework of~\citet{davidson2025holderstabilitymultisetgraph}. This shows stability of such \mas embeddings. 
We say that a parametric set function $F: \Pvk \times \Wcal \to \R^m$ is Upper Lipschitz in expectation if $\exists C > 0$ such that:
\begin{equation}
    \Exp_{w\in \Wcal}  \norm{F(S;w)-F(T;w)}_1 \leq C \cdot \W^{(k)}(S, T), \ \forall S, T \in \Pvk
\end{equation}Where $\W^{(k)}(., *)$ is the augmented-Wasserstein metric for sets in $\Pvk$, where a padding $z$ is added for $k - |S|$ times to any multiset $S$ of size less than $k$. We work with a compact ground set $V$, and we pick a padding $z$ that has a positive distance from $V$. For the following proofs, we work with $V \subseteq \R^d$ which is norm bounded by $1$, and we choose the padding element $z \in \R^d$ such that $\norm{z} \geq 3$. For a general $V$ whose norm is bounded by $B$, we can scale the padding $z$ by $B$. With this, we show that, real valued functions $F: \Pvk \to \R$ given by $F(S) = \sum_{x \in S}\sigma\round{\frac{a\tran x - b}c}$ are Upper Lipschitz w.r.t $\W_k$. Now, if we take independent parametric copies across $m$ output dimensions and if $\mtwo$ is a vector-to-vector Lipschitz function, then functions of the form $\mtwo \circ F$, where $F$ has the form in~\cref{eq:hadamard}, are Upper Lipschitz as well.
 This we state in the following theorem:
 \begin{restatable}[$F$ is upper Lipschitz]{theorem}{gupperlipshiz}
     Let $\sigma$ belongs to the class of \hhat activations, and $V \subset \R^d$ be a compact ground set such that $\sup_{v \in V}\norm{v} \leq 1$. We consider $F: \Pvk \to \R, F(S) = \sum_{x \in S}\sigma\round{\frac{a\tran x - b}c}$, with the distributions $a \sim \Unif(\S^{d-1}), b \sim \Unif(-1,1), c \sim \Unif(0, 2)$. Then $\exists \text{ a constant }C > 0$, such that for any $S, T \in \Pvk$:
    \begin{align*}
        \Exp_{a, b, c} \abs{F(S) - F(T)} \leq C  \cdot \W^{(k)}(S,T)
    \end{align*}
\end{restatable}
\begin{proof}
Consider $S = \{x_1, \cdots, x_M\}, T = \{y_1, \cdots, y_N\}$. Since $\abs{F(S) - F(T)}$ is symmetric in $S, T$, we consider WLOG that $M \leq N$. As in previous sections, we use $\tau$ to denote an injective function from $[M] \to [N]$, and we use $\Omega_{N, M}$ to denote the set of all injective functions from $[M] \to [N]$. Thus, we can write the augmented Wasserstein distance on $\Pvk$ with padding $z$ as:
\begin{align*}
    \W^{(k)}(S, T) &= \arg\min_{\tau \in \Omega_{N, M}}\curly{\sum_{i=1}^M \norm{x_i - y_{\tau(i)}} + \sum_{j \neq \tau(i)} \norm{y_j - z}}
\end{align*} 
Let $\tau^*$ be the argmin of the above expression that gives the $\W_k(S, T)$ Now, we can write:
\begin{align*}
    & \Exp_{a, b,c}\abs{F(S) - F(T)} = \Exp_{a,b,c} \bigg|\sum_{i=1}^M \sigma\big(\frac{a\tran x_i - b}{c}\big) - \sigma\big(\frac{a\tran y_{\tau^*(i)} - b}{c}\big)  - \sum_{j\neq \tau^*(i)} \sigma\big(\frac{a\tran y_j - b}{c}\big)\bigg| \\
    \leq & \sum_{i=1}^M\Exp_{a, b,c}\abs{\sigma\round{\frac{a\tran x_i - b}{c}} - \sigma\round{\frac{a\tran y_{\tau^*(i)} - b}{c}}} + \sum_{j\neq \tau^*(i)}\Exp_{a, b,c} \abs{\sigma\round{\frac{a\tran y_j - b}{c}}}
\end{align*}

We now separately analyze coupled and un-coupled terms: i.e terms of the form $\Exp \bigg|\sigma\bigg(\frac{a\tran y_j - b}{c}\bigg)\bigg|$ and $\Exp\bigg|\sigma(\frac{a\tran x_i - b}{c}) - \sigma(\frac{a\tran y_{\tau^*(i)} - b}{c})\bigg|$ in the following analysis. For the proof that follows, we consider $\supp(\sigma) \subseteq [0,1]$. For support in $[\gamma_1, \gamma_2]$, the proof follows similarly by sifting the distributions of $b, c$ by $\gamma_1$ and scaling by $\gamma_2 - \gamma_1$.

\paragraph{Computing $\mathbf{\Exp \abs{\sigma\round{\frac{a\tran y_j - b}{c}}}}$:}

Note that, $\sigma(\frac{a\tran y_j - b}{c}) = 0, \forall b \geq a\tran y_j$. Also, for a given $b \in (-1, a\tran y_j)$ we have that:
\begingroup
\allowdisplaybreaks
\begin{align*}
    \forall c \in (0, a\tran y_j- b), \frac{a\tran y_j - b}{c} \geq 1 \implies \sigma\bigg(\frac{a\tran y_j - b}{c}\bigg) = 0
\end{align*}
\endgroup
Thus, conditioned on a given $a \in \S^{d-1}$, we can write the conditional expectation over $b, c$ as follows:
\begingroup
\allowdisplaybreaks
\begin{align*}
    &\Exp_{b,c} \curly{\sigma\big(\frac{a\tran y_j - b}{c}\big)\bigg| a}  \\
    = & \frac{1}{4}\int_{b=-1}^{a\tran y_j} \int_{c=a\tran y_j - b}^2 \bigg|\sigma\bigg(\frac{a\tran y_j - b}{c}\bigg) - \sigma(0)\bigg|\dif c \dif b,  \text{ using } \sigma(0) = 0\\
    \leq & \frac{\kappa_2}{4}\int_{b=-1}^{a\tran y_j} \int_{c=a\tran y_j - b}^2 \bigg(\frac{a\tran y_j - b}{c}\bigg) \dif c \dif b, \text{as }\sigma\text{ is a \hhat activation, it's Lipschitz}\\
    = & \frac{\kappa_2}{4}\int_{b=-1}^{a\tran y_j} (a\tran y_j - b)\ln\round{\frac{2}{a\tran y_j - b}} \dif b = \frac{\kappa_2}{4}\int_{a\tran y_j + 1}^0 z\ln\round{\frac 2z} (-\dif z), z:= a\tran y_j - b\\
    = &\frac{\kappa_2}{4}\int_0^{a\tran y_j + 1} z\ln\round{\frac 2z} \dif z = \frac{\kappa_2}{4}\cdot\frac{a\tran y_j + 1}2\round{\frac{a\tran y_j + 1}2 + \round{a\tran y_j + 1}\ln\round{\frac{2}{a\tran y_j + 1}}}\\
    \leq & \frac{\kappa_2}4 \cdot (a\tran y_j + a\tran a) = \frac{\kappa_2}4 \abs{a\tran(y_j + a)} \leq \frac{\kappa_2}4 \norm{y_j + a}\leq \frac{\kappa_2}4\norm{y_j - z}
\end{align*}
\endgroup
Now, taking an expectation over $a \sim \Unif(\S^{d-1})$, the inequality is preserved and we get that, for "isolated " $y_j$'s, $\Exp\sigma\round{\frac{a\tran - b}c} \leq \frac{\kappa_2}4\norm{y_j - z}$

Thus for "isolated" elements from the larger set that get coupled with the padded element $z$ while computing augmented-$\W^{(k)}$, we have shown that they are upper bounded by their distance from the padding $z$. Hence, it remains to show a similar Lipschitz upper bound for the elements $x_1, \cdots x_m \in S_1$ which have a corresponding $y_{i(1)}, \cdots, y_{i(m)} \in S_2$ in the optimal coupling\\

\paragraph{Computing $\mathbf{\Exp\bigg|\sigma(\frac{a\tran x_i - b}{c}) - \sigma(\frac{a\tran y_{\tau^*(i)} - b}{c})\bigg|}$: \\}

    Now, we have, for any $i \in [M]$, and a given $a \in \S^{d-1}$ the conditional expectation $\Exp_{b,c}\bigg|\sigma(\frac{a\tran x_i - b}{c}) - \sigma(\frac{a\tran y_{\tau^*(i)} - b}{c})\bigg|$ is symmetric in $x_i$ and $y_{\tau^*(i)}$. Thus, WLOG we can assume that, $a\tran x_i \geq a\tran y_{\tau^*(i)}$. Note that we then have:
    \begingroup
    \allowdisplaybreaks
    \begin{align*}
        &\bigg|\sigma(\frac{a\tran x_i - b}{c}) - \sigma(\frac{a\tran y_{\tau^*(i)} - b}{c})\bigg| \\
        =&\begin{cases}
            &\bigg|\sigma(\frac{a\tran x_i - b}{c})\bigg|, \forall b \in (a\tran y_{\tau^*(i)}, a\tran x_i), \forall c \in (a\tran x_i - b, 2)\\
            & \bigg|\sigma(\frac{a\tran x_i - b}{c}) - \sigma(\frac{a\tran y_{\tau^*(i)} - b}{c})\bigg|,  \forall b \in (-1, a\tran y_{\tau^*(i)}), \forall c \in (a\tran y_{\tau^*(i)} - b, 2)\\
            & 0, \text{ otherwise}
        \end{cases}
    \end{align*}
    \endgroup
Hence, conditioned on a specific $a \in \S^{d-1}$, we can write the conditional expectation over $b, c$ as:
\begingroup
\allowdisplaybreaks
\begin{align*}
    & \Exp_{b,c}\bigg|\sigma(\frac{a\tran x_i - b}{c}) - \sigma(\frac{a\tran y_{\tau^*(i)} - b}{c})\bigg| = \int_{b=a\tran y_{\tau^*(i)}}^{a\tran x_i} \int_{c = a\tran x_i - b}^2\sigma\round{\frac{a\tran x_i - b}c} \dif c \dif b\\
    & + \int_{b=-1}^{a\tran y_{\tau^*(i)}} \int_{c = a\tran y_{\tau^*(i)} - b}^2 \bigg|\sigma(\frac{a\tran x_i - b}{c}) - \sigma(\frac{a\tran y_{\tau^*(i)} - b}{c})\bigg| \dif c \dif b  \\
    \leq & \frac{\kappa_2}4\int_{b=a\tran y_{\tau^*(i)}}^{a\tran x_i} (a\tran x_i - b)\ln\round{\frac{2}{a\tran x_i - b}} \dif b \\
    &+\frac{\kappa_2}4 \abs{a\tran(x_i - y_{\tau^*(i)})}\int_{b=-1}^{a\tran y_{\tau^*(i)}} \ln\round{\frac{2}{a\tran y_{\tau^*(i)} - b}} \dif b  \\
    \leq & \frac{\kappa_2}4\round{\int_0^{a\tran(x_i - y_{\tau^*(i)})}z\ln\round{\frac 2z}\dif z + \abs{a\tran(x_i - y_{\tau^*(i)})}\int_0^{a\tran y_{\tau^*(i)}+1}\ln\round{\frac 2z}\dif z}\\
    \leq & \frac{\kappa_2}4\round{\abs{a\tran(x_i - y_{\tau^*(i)})}+2\abs{a\tran(x_i - y_{\tau^*(i)})}} < \kappa_2 \abs{a\tran(x_i - y_{\tau^*(i)})}\\
    \leq & \kappa_2 \norm{x_i - y_{\tau^*(i)}}, \text{ by Cauchy Schwarz}
\end{align*}
\endgroup
Thus taking one more expectation over $a \sim \S^{d-1}$, and summing over all $i \in [M]$, we get:
\begin{align*}
    \Exp_{a, b, c}\sum_{i=1}^M\bigg|\sigma(\frac{a\tran x_i-b}{c}) - \sigma(\frac{a\tran y_{\tau^*(i)}-b}{c})\bigg| \leq \kappa_2\sum_{i=1}^M \norm{x_i - y_{\tau^*(i)}}
\end{align*}
Thus, combining our computations for "isolated" $y_j$'s and "coupled" $(x_i, y_{\tau^*(i)})$'s we get that:
\begin{align*}
    \Exp_{a, b,c} |F(S) - F(T)| \leq \frac{\kappa_2}4 \sum_{j \neq \tau^*(i)}\norm{y_j - z}  + \kappa_2\sum_{i=1}^M \norm{x_i - y_{\tau^*(i)}} \leq \kappa_2 \W^{(k)}(S, T)\\
\end{align*}
And our proof of $F$ being upper Lipschitz is complete. 
\end{proof}

\subsection{Universal approximators with \mas functions}
\universal*
\begin{proof}
\label{universality}
A \mas function is in particular invertible, and therefore we can write $f(S)=f\circ F^{-1} \circ F(S) $. On the image of $F$ we can define the function $M=f\circ F^{-1}$ which is a vector-to-vector functions, and it is monotone on the image of $F$, because if $v=F(S), u=F(T)$ for some sets $S,T$, and if $v\leq u$, then by separability $S \subseteq T$, and therefore  
$$M(v)=f\circ F^{-1}(v)= f(S)\leq f(T)=f\circ F^{-1}(u)=M(u) $$

Finally, we need to show that $M$ can be extended to a monotone function on all of $\RR^m$. We accomplish this by defining 
$$M(v)=\max\{M(u)| \quad u \leq v,  u \in \mathrm{Image}(F) \} .$$
\end{proof}

\section{Details about our model}
\label{app:model}
\subsection{Architecture details:}
In all 4 of \our models, we use an elementwise neural network $NN_\theta: \R^d \to \R^m$, where the ground set $V \subseteq \R^d$ and the output is in $\R^m$. We follow that with a sum aggregation, followed by a monotone vector-to-vector neural network $M_{2, \phi}: \R^m \to \R^m$. For the embedding Neural Network, we use a 1 hidden layer NN with $\relu$ activation in the hidden layers and $\relu$ or \hhat activation in the output, depending on whether we're using \relumas or \hhat activation based \our. To make $M_{2, \phi}$ monotone, we use non-negative weights by taking the absolute value of parameters before applying the linear transformation, and use all monotonic activations like $\relu$ in the intermediate layers. Now, as discussed before, we give details of our model \integralmas.
\subsection{\integralmas}
In this appendix we design universal approximators for the class of Hat functions to learn $\sigma$. We parametrize For that, we derive equivalent conditions on the derivatives of $\sigma$ (per embedding dimension) and use the techniques of learning functions by modelling the derivatives using neural networks and then using a numerical integration as in \citet{wehenkel2021unconstrainedmonotonicneuralnetworks}. 
\begin{lemma}
\label{lemma:hat_derivative}
    If $\sigma: \R \to \R_{\geq 0}$ belongs to the hat activation class , then $\sigma\dash$ satisfies the following conditions for some positive constants $c, C > 0$ and $\alpha \in \R, \beta > 0, \gamma \in (0,1)$:
    \begin{enumerate}
    \item $\supp\round{\sigma\dash} \subseteq [\alpha, \alpha+\beta]$
    \item $\int_\alpha^{\alpha+\beta}\sigma\dash(x) \dif x = 0$
    \item $\sigma\dash(x) \leq C, \forall x \in \R$ 
    \item $\sigma\dash(x) \geq c, \forall x \in (\alpha, \alpha+\gamma \cdot \beta)$ 
    \end{enumerate}
The above 4 conditions, along with $\sigma(\alpha) = 0$, are the necessary and sufficient conditions that characterize the hat activation class.
\end{lemma}
\begin{proof}
    We first prove that, if $\sigma$ belongs to the hat activation class as defined in \ref{def:hhat}, then conditions 1-4 are satisfied. Note that, if $\sigma$ is a \hhat activation, then $\sigma$ has compact support in some $[\alpha, \alpha + \beta]$ by definition, and it's piecewise continuously differentiable. Thus, outside $[\alpha, \alpha + \beta]$, we have $\sigma \equiv 0 \implies \sigma\dash \equiv 0$. Hence, $\supp(\sigma\dash)\subseteq [\alpha, \alpha + \beta]$. We also have that, $0 = \sigma(\alpha) = \sigma(\alpha + \beta) \implies \sigma(\alpha+\beta) - \sigma(\alpha) = 0 \implies \int_\alpha^{\alpha+\beta}\sigma\dash(t) \dif t = 0$. Finally, $\sigma$ being Lipschitz is equivalent to $\abs{\sigma\dash(t)}\leq C$ for some $C > 0$ and by continuity of $\sigma\dash$ in $(\alpha, \alpha + \beta)$ we must have that $2c := \lim_{t\to\alpha^+}\sigma\dash(t) > 0 \implies \sigma\dash(t) \geq c, \forall t \in (\alpha , \alpha + \gamma \cdot \beta)$ for some $\gamma \in (0,1)$.  Thus, conditions 1-4 are implied by $\sigma$ belonging to the \hhat activation class. 
    
    Moreover, if conditions 1-4 are satisfied, and $\sigma(\alpha) = 0$, then condition-2 implies $\sigma(\alpha+\beta)=0$. This, along with condition-1 implies that $\sigma(t) = 0, \forall t \not \in [\alpha, \alpha + \beta]$. Condition-4 immediately implies that $\sigma(\alpha + \gamma \cdot \beta) \geq c \cdot \gamma\beta > 0$, thus $\sigma \not \equiv 0$. On the other hand, we get that $\sigma' \leq C$ implies by LMVT that, $\abs{\sigma(x) - \sigma(y)} \leq C \cdot \abs{x-y}$. Hence, $\sigma$ belongs to the class of \hhat activations following the definition from \ref{def:hhat}.
\end{proof}
\paragraph{Neural Parametrization of \integralmas :}Consider $h_\theta^1(.), h_\phi^2(.)$ to be non-negative + bounded, and bounded fully connected Neural Networks, respectively, of one hidden layer each. Also consider the trainable support parameters to be $\bm{\alpha} \in \R^m, \bm{\beta} \in \R^m_+, \bm{\gamma} \in (0,1)^m$ where $m$ is the output dimension and $\bm{\alpha}, \bm{\beta}, \bm{\gamma}$ are the support parameters, aggregated for all output dimensions.
Now, we define an integral based model of parametric hat functions.
Let $\Theta = (\theta, \phi, \alpha, \beta, m)$ be the parameter space.
\begin{equation}
\label{eq:neural_hatfn}
\begin{split}
    \sigma_\Theta(x) &= \int_\alpha^x h_\theta^1(z)\indic{\alpha \leq z\leq\alpha + \gamma\beta} \dif z - \sbrac{\frac{\int_\alpha^{\alpha+\gamma\beta}h_\theta^1(z)dz}{\int_{\alpha+\gamma\beta}^{\alpha + \beta}h_\phi^2(z)dz}}\round{\int_{\alpha + \gamma\beta}^{x}h_\phi^2(z)\indic{\alpha + \gamma\beta\leq z\leq\alpha + \beta}dz}   
\end{split}
\end{equation}
In the above formulation, in addition to the support parameters we also learn the function itself rather than performing a linear interpolation, this makes the above class a universal approximator of hat functions.
\begin{restatable}[Universal Hat approximator]{lemma}{lemma:universal_hat_approximator}
    The family $\sigma_\Theta$ defined in~\cref{eq:neural_hatfn} is an universal approximator of the class of \hhat functions having support in $[\alpha, \alpha + \beta]$ with $\sigma\dash(t) > 0, \forall t \in (\alpha, \alpha + \gamma \beta)$
\end{restatable}
\begin{proof}
    We show that the given parametric model is an universal approximator of \hhat functions using their equivalent formulation in Lemma~\ref{lemma:hat_derivative}. Let $\sigma_1$ be $\sigma$ restricted to the interval $[\alpha, \alpha + \gamma\beta]$ and $\sigma_2$ be $\sigma$ restricted to the interval $[\alpha + \gamma\beta, \alpha + \beta]$. Using Theorem-1 from~\citet{lu2017expressivepowerneuralnetworks}, we can get $\relu$ neural network $h_\theta^1: \R \to \R$ and $h_\phi^2: \R \to \R$ of width 5 each such that they approximate the derivatives of the restricted functions $\sigma_1\dash$ and $\sigma_2\dash$. Thus, the following hold for any $\epsilon > 0$ and any constant $c \in \R$:
    \begin{equation}
    \label{eq:univ_approx}
        \int_\alpha^{\alpha + \gamma\beta}\abs{\sigma_1\dash(z) - h_\theta^1(z)}\dif z \leq \epsilon \text{ and } \int_{\alpha+ \gamma\beta}^{\alpha + \beta}\abs{\sigma_2\dash(t) + c \cdot h_\phi^2(t)}\dif t \leq \epsilon
    \end{equation}
    We now observe that, if $x \in (-\infty, \alpha)$, both the indicator functions: $\indic{\alpha \leq z\leq\alpha + \gamma\beta}$ and $\indic{\alpha + \gamma\beta\leq z\leq\alpha + \beta}$ as in the integrands of ~\cref{eq:neural_hatfn} evaluate to 0, thus $\sigma_\Theta$ exactly coincides with $\sigma$. On the other hand, if $x \in (\alpha + \beta, \infty)$ then we would have the first integral of \ref{eq:neural_hatfn} evaluate to $\int_\alpha^{\alpha +\gamma\beta} h_\theta^1(z) \dif z$ (due to presence of the indicator $\indic{\alpha \leq z \leq \alpha +\gamma\beta}$ and the second expression evaluates to $- \sbrac{\frac{\int_\alpha^{\alpha+\gamma\beta}h_\theta^1(z)dz}{\int_{\alpha+\gamma\beta}^{\alpha + \beta}h_\phi^2(z)dz}}\round{\int_{\alpha + \gamma\beta}^{\alpha + \beta}h_\phi^2(z)dz} = -\int_\alpha^{\alpha+\gamma\beta}h_\theta^1(z)dz$. Thus, both the expressions sum to $0$ and $\sigma_\Theta$ coincides with $\sigma$ for $x \in (\alpha +\beta, \infty)$ as well. 
    
    Now we shall show that, for any $x \in (\alpha , \alpha + \beta)$ we must also have $|\sigma_\Theta(x) - \sigma(x)| \leq \epsilon$, implying convergence in $\sup$ norm. We consider the following cases:
    \begin{itemize}
        \item \textbf{Case-I:} When $x \in (\alpha, \alpha +\gamma\beta)$, we have that:
        \begingroup
        \allowdisplaybreaks
        \begin{align*}
            &\abs{\sigma(x) - \sigma_\Theta(x)} \\
            = & \abs{\int_\alpha^x \sigma_1\dash(z) \dif z - \int_\alpha^x h_\theta^1(z) \dif z}\\
            \leq & \int_\alpha^x \abs{\sigma_1\dash(z) -  h_\theta^1(z)} \dif z, \text{by triangle ineq.}\\
            \leq & \int_\alpha^{\alpha + \beta\gamma} \abs{\sigma_1\dash(z) -  h_\theta^1(z)} \dif z, \text{by non-negativity of integrand}\\
            \leq & \epsilon, \text{by ~\cref{eq:univ_approx}}
        \end{align*}
        \endgroup
        \item \textbf{Case-II:} When $x \in (\alpha + \gamma\beta, \alpha +\beta)$ we have:
        \begingroup
        \allowdisplaybreaks
        \begin{align*}
            &\abs{\sigma(x) - \sigma_\Theta(x)} \\
            = & \abs{\int_\alpha^x \sigma\dash(z) \dif z - \int_\alpha^{\alpha +\gamma\beta} h_\theta^1(z) \dif z - \int_{\alpha+\gamma\beta}^x h_\phi^2(z) \dif z}\\
            = & \abs{\int_\alpha^{\alpha+\gamma\beta}\round{\sigma_1\dash(z) - h_\theta^1(z)}\dif z +\int_{\alpha +\gamma\beta}^x\round{\sigma_2\dash(z) + \sbrac{\frac{\int_\alpha^{\alpha+\gamma\beta}h_\theta^1(z)dz}{\int_{\alpha+\gamma\beta}^{\alpha + \beta}h_\phi^2(z)dz}} h_\phi^2(z)}\dif z}\\
            \leq & \int_\alpha^{\alpha+\gamma\beta}\abs{\sigma_1\dash(z) - h_\theta^1(z)} \dif z + \int_{\alpha+\gamma\beta}^x \abs{\sigma_2\dash(z) + \sbrac{\frac{\int_\alpha^{\alpha+\gamma\beta}h_\theta^1(z)dz}{\int_{\alpha+\gamma\beta}^{\alpha + \beta}h_\phi^2(z)\dif z}} h_\phi^2(z)}\dif z, \text{tri. ineq.}\\
            \leq & \epsilon + \int_{\alpha+\gamma\beta}^{\alpha + \beta} \abs{\sigma_2\dash(z) + \sbrac{\frac{\int_\alpha^{\alpha+\gamma\beta}h_\theta^1(z)dz}{\int_{\alpha+\gamma\beta}^{\alpha + \beta}h_\phi^2(z)\dif z}} h_\phi^2(z)}\dif z, \text{by~\ref{eq:univ_approx} on 1st term \& tri. ineq. on 2nd}\\
            \leq & 2\epsilon
        \end{align*}
        \endgroup
        where last inequality is obtained  $c= \frac{\int_\alpha^{\alpha+\gamma\beta}h_\theta^1(z)dz}{\int_{\alpha+\gamma\beta}^{\alpha + \beta}h_\phi^2(z)\dif z}$ in the universal approximator for $\sigma_2\dash$ in ~\cref{eq:univ_approx}. Thus, just we see that, for any $\epsilon > 0$ and $\alpha \in \R, \beta > 0, \gamma \in (0, 1)$ we can find a universal approximator for the \hhat function class
    \end{itemize}
\end{proof}
\paragraph{Parametrizing $\alpha, \beta, \gamma$} We have provided an universal approximator for the \hhat function class given $\alpha, \beta, \gamma$ but we'd also want to have $\alpha, \beta, \gamma$ as learnable parameters. This is valid for both \hatmas and \integralmas, as in both cases we seek to learn the support parameters. Now, there is no constraint on $\alpha$, so we can directly initialize $\bm{\alpha} \in \R^{3m}$ and learn it through gradient descent. However, we have a constraint on $\beta$ that $\beta > 0$. For this, we first initialize $\bm{\beta_0}\in \R^m$ and we obtain $\bm{\beta}\in \R^m$ by applying co-ordinatewise positive transformation on $\bm{\beta_0}$. This we do in two ways, depending on which works better on a given task: \textbf{(I)} by simply taking $\bm{\beta_0}\mapsto \abs{\bm{\beta_0}}$, where the absolute value is taken pointwise. \textbf{(II)}by taking the pointwise transformation $\bm{\beta_0}\mapsto \text{ELU}\round{{\bm{\beta_0}}; \upsilon} +\upsilon$, where $\text{ELU}\round{\cdot; \upsilon}$ is the Exponential Linear Unit with hyperparameter $\upsilon > 0$. Note that, the ELU function is given by: $\text{ELU}\round{x; \upsilon} = \begin{cases}
    x , \text{if } x > 0\\
    \upsilon\round{e^x - 1}, \text{if }x \leq 0
\end{cases}$. For parametrizing $\gamma \in (0,1)$, we similarly initialize $\bm{\gamma_0}\in\R^m$ and then apply a pointwise transfprmation that takes each co-ordinate to $(0,1)$. We use tempered sigmoid function: $\bm{\gamma_0} \mapsto \text{Sigmoid}\round{\tau \cdot \bm{\gamma_0}}$ where $\tau > 0$ is a hyperparameter

\paragraph{Designing soft indicator functions:} In the integrands for \integralmas, we have some indicator functions of the form $\indic{a \leq \cdot \leq b}(x) = \begin{cases}
    1, \text{if } x \in [a, b]\\
    0, \text{o.w}
\end{cases}$. We want these indicator functions to have non zero gradients as we intend to learn the "support parameters" $a, b$ of such indicator function. If we chose binary indicators, then that is not differentiable everywhere and have gradient $0$ in most places, making it difficult to learn $a, b$. Thus, we design a "soft" indicator function with the help of tempered sigmoid with hyperparameter $\tau_S$ as follows: 
\begin{align*}
    \indic{a \leq \cdot \leq b}(x) = \text{Sigmoid}\round{\tau_S \cdot (x - a)}\cdot \text{Sigmoid}\round{\tau_S \cdot (b-x)}
\end{align*}

\subsection{Making \our Mini Batch Consistent}
There has been a chain of works to make set-based neural models to be mini batch consistent in~\cite{MBC1_slotset, MBC2} which have applications in large scale machine learning. The goal of these Mini Batch Consistent (MBC) works is to design a set function $F$, which, if applied on any subsets from partitions of a set $S$ and pooled through an activation $g$, should return $F(S)$. Thus, at a high level, MBC functions analyse and design functions for the following batch consistency condition: $g(F(S_1), \cdots, F(S_n)) = F(S)$. MBC methods achieve this by imposing certain restrictions on attention-based set architectures. Similar to this, our work also puts restrictions on the inner embedding transformation and the outer vector-to-vector transformations. It is thus a natural question to ask, if \our can be made minibatch consistent. Since set containment is the major theme of our work, with potential applications in retrieval and recommendation systems, processing huge sets at once can be a key bottleneck, and it would be nice to make \our process over mini-batches and then design some aggregation mechanism to make the aggregated embedding equal to the embedding of the entire set.

If we consider a partition of a given set $S = \bigcup_{i=1}^k S_i$, where $S_i$'s are all disjoint, then $\our (S_i) \leq \our(S)$, by monotonicity. But, we can try to find an appropriate pooling mechanism over the subset embeddings to make the aggregated embedding same as that of the entire set. For example, if the outer transformation $M_{\theta_2}$, as in~\cref{eq:masnet} was invertible, and then we may define a particular permutation-invariant pooling function $g$ such that, $g(X_1, \cdots, X_k) = M_{\theta_2}\round{\sum_{i=1}^kM_{\theta_2}^{-1}\round{X_i}}$, then we'd have: $g\round{\our(S_1), \cdots, \our(S_k)} = M_{\theta_2}\round{M_{\theta_2}^{-1}\round{\our(S_i)}} = M_{\theta_2}\round{\sum_{i=1}^k\sum_{x \in S_i}\sigma \circ M_{\theta_1}(x)} = M_{\theta_2}\round{\sum_{x \in S} \sigma \circ M_{\theta_1}(x)} = \our (S)$. his is irrespective of how we partition $S$ into $S_1, \cdots S_k$. And this would enable us to process the subsets independently and aggregate them accordingly. 

For our purposes, we have needed the outer transformation $M_{\theta_2}$ to be a vector-to-vector monotone function in~\cref{eq:masnet}. But that doesn't guarantee injectivity or invertibility. Thus, we need to restrict the choice of $M_{\theta_2}$ to be monotone and invertible for \our to be MBC consistent. One possible choice is to use ideas from relevant works like~\cite{InvertibleNN, wehenkel2021unconstrainedmonotonicneuralnetworks} to design monotone and invertible neural networks, and analyse the expressibility properties of the resulting set model. This forms an interesting direction for future work. 

\section{Additional details about experiments and more experiments}
\label{app:addl-details}
\subsection{Hardware details:} All experiments were performed in a compute server, running the OS of GNU Linux Version 12, equipped with a 16 core Intel(R) Xeon(R) Gold 6130 CPU @ 2.10GHz CPU architecture and equipped with a cluster of 6 NVIDIA RTX A6000 GPUs with a memory of 49GB each.

\subsection{Dataset details and more experiments}
We now give an account of preparation details of each dataset and accounts on more experiments on each.

\paragraph{Synthetic datasets and related experiments: }

We begin with a controlled synthetic setting where we generate pairs $(S, T)$ such that $k_S < k_T$. 

Each pair is labeled either positive or negative. For positive examples, we first sample $k_T$ vectors in $\R^d$ from $\Normal(0, \Id)$, which form the set $T$. A subset of size $k_S$ is then uniformly sampled from $T$ to obtain $S$. For negative examples, we again sample $T$ as above, but then independently draw $k_S$ vectors from $\R^d$ to form $S$ .

To simulate a real world scenario of information loss, we inject gaussian noise into the vectors of $S$ to get $S'$ but keep the boolean label unchanged.
\begin{figure}[h]
    \centering
    \begin{minipage}{0.48\textwidth}
        \centering
        \includegraphics[width=\textwidth]{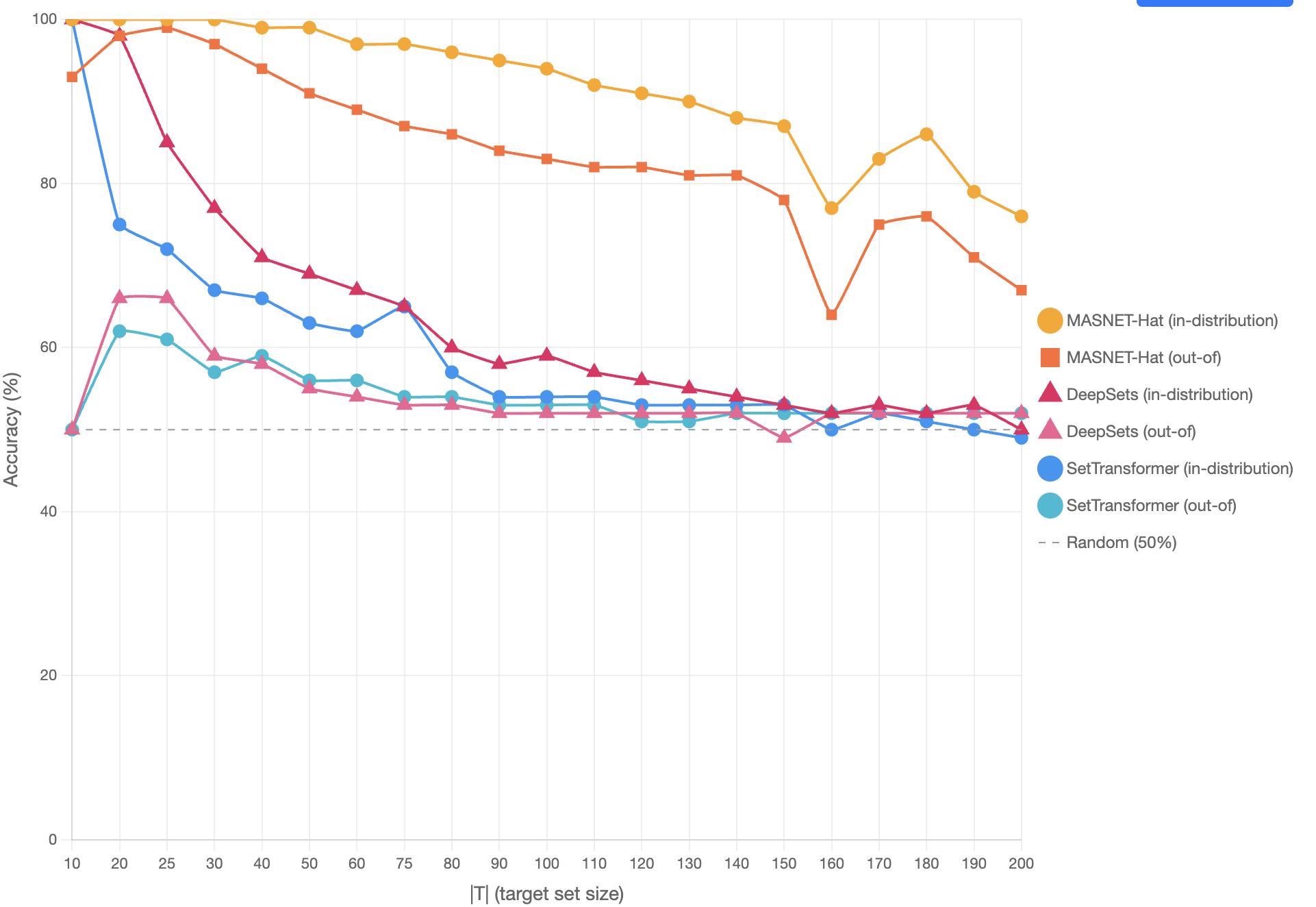}
        \caption{Acc vs $|T|$ for all models $\geq 2$ layer MLP}
        \label{fig:deep1}
    \end{minipage}
    \hfill
    \begin{minipage}{0.48\textwidth}
        \centering
        \includegraphics[width=\textwidth]{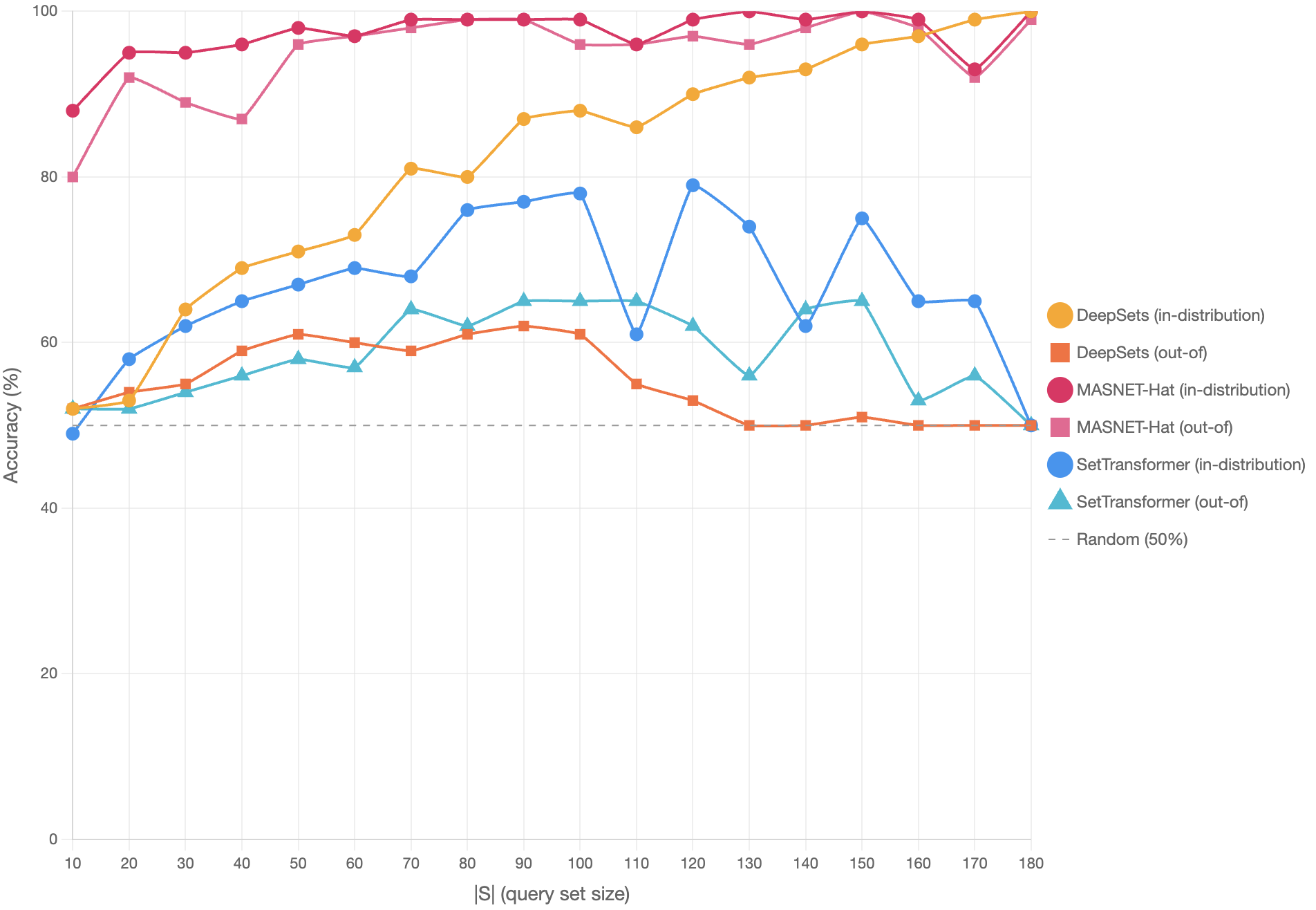}
        \caption{Acc vs $|S|$ for all models $\geq 2$ layer MLP}
        \label{fig:deep2}
    \end{minipage}
\end{figure}

On this, we provide two more plots: Figs. \ref{fig:deep1} and \ref{fig:deep2} that compares the accuracies of $\our$ vs baselines, by varying the gap between target set size $|T|$ and query set size $|S|$. Note that the more this gap is, the more it becomes challenging to separate two non-subsets. In all cases, the embedding MLP $M_{\theta_1}$ from~\cref{eq:masnet} is comprising of $\geq 2$ layers.

\paragraph{Text-based datasets and related experiments:}

We evaluate our models on three text based real-world datasets: MSWEB, MSNBC, and the Amazon Baby Registry datasets, all of which reflect naturally occurring set containment structures coming out of user behavior in websites and recommendation engines.

\paragraph{MSWEB and MSNBC:} These datasets consist of user activity logs from \textit{www.microsoft.com} and \textit{www.msnbc.com}, respectively. Each user session is treated as a bag of page identifiers, which are embedded into 768-dimensional vectors using a pre-trained BERT model. We construct query-target pairs \((S, T)\) by sampling \(S\) from a session \(V\), and letting \(T = V \setminus S\). A pair is labeled \(1\) if \(S \subseteq T\), and \(0\) otherwise. To simulate noisy real-world conditions, Gaussian noise is added to each element in \(S\) while preserving the label. The positive-to-negative ratio was taken to be 0.1, which is the number of labels 1 to labels 0 in the entire dataset.

\paragraph{Amazon Baby Registry:} This dataset contains subsets of products selected by customers, each tagged by category (e.g., “toys” or “feeding”). Product descriptions are embedded using BERT. We filter out sets of size \(<2\) or \(>30\), and for each valid subset \(S\), we generate \(T \supseteq S\) of size 30 by sampling additional items from the same category. For negative samples, \(T\) is sampled randomly and verified to satisfy \(S \not\subseteq T\).

\begin{wraptable}[7]{r}{0.6\textwidth}
\vspace{-15pt}
\centering
\small
\maxsizebox{\hsize}{!}{\tabcolsep 1pt 
\begin{tabular}{l|c|c|c|c}
\hline
\textbf{Model}  & \textbf{Bedding} & \textbf{Feeding} & \textbf{MSWEB} & \textbf{MSNBC} \\
\hline
DeepSets        & \underline{0.91}           & 0.90          & 0.93          & \textbf{0.97}           \\
SetTransformer  & \underline{0.91}           & 0.90          & 0.94          & \underline{0.96}         \\
FlexSubNet      & 0.90           & 0.91          & 0.94          & 0.94              \\
Neural SFE      & 0.91           & 0.91          & 0.91          & 0.93              \\
\hline
\relumas        & \textbf{0.95}  & \textbf{0.93} & \textbf{0.97} & \textbf{0.97}          \\
\hatmas         & \underline{0.91}  & \underline{0.92} & \underline{0.95} & \textbf{0.97}           \\
\hline
\end{tabular}}
\caption{Set containment in text datasets with 90:10 ratio}
\label{tab:noisy_accuracy_9010}
\end{wraptable}

Noise is added to \(S\) in both cases. This models scenarios such as predicting whether a set of displayed products \(T\) includes a customer’s interest \(S\). On these datasets, we provide three sets of additional experiments. In the first table, we perform the same set-containment task as done in the main text, but with a different negative-to-positive ratio of $90:10$. 

Note that, due to the inductive bias of monotonicity, all the positive examples are correctly classified by design is the case of \our. Thus, it only has to learn to identify the negative examples. Thus, if the test set has a higher proportion of negative examples, then it throws a model a tougher task to learn. This is shown in the following Table~\ref{tab:noisy_accuracy_9010}. Other than the modified class ratio, we provide two additional tables: in the first one, we compare shallow(1 layer) vs deep ($\geq 2$ layers) embedding MLP $M_{\theta_1}$ in \our , which is shown in Table~\ref{tab:shallow_deep}. White noise of std 0.1 added for \inexact set containment.

\begin{wraptable}{r}{0.6\textwidth}
\centering
\scriptsize
\renewcommand{\arraystretch}{1.1}
\setlength{\tabcolsep}{3pt}
\begin{tabular}{l|c|c|c|c|}
    \toprule
    \textbf{Model} & \textbf{bedding} & \textbf{feeding} & \textbf{MSWEB} & \textbf{MSNBC} \\
    \midrule
    Shallow-ReLU & 0.69 & 0.36 & 0.92 & 0.93 \\
    Shallow-\trimas & 0.83 & 0.76 & 0.94 & 0.95 \\
    Shallow-\hatmas & 0.89 & 0.77 & \emph{0.97} & 0.96 \\
    Shallow-\integralmas & 0.90 & 0.91 & 0.92 & 0.95 \\
    \hline
    (Deep)\relumas & \textbf{0.95} & \textbf{0.93} & \emph{0.97} & \textbf{0.97} \\
    Deep-\trimas & 0.92 & \emph{0.93} & \textbf{0.98} & 0.96 \\
    Deep-\hatmas & 0.91 & 0.92 & 0.95 & \emph{0.97} \\
    Deep-\integralmas & \emph{0.93} & 0.92 & 0.96 & \emph{0.97} \\
    \bottomrule
\end{tabular}
\caption{Comparison of shallow vs. deep variants of \our across datasets.}
\label{tab:shallow_deep}
\end{wraptable}

\paragraph{Ablation studies on text datasets:}
From the definition of \our in~\cref{eq:masnet}, we see that, the formulation is quite similar to Deepsets, which is of the form $F(S) = \mtwo\round{\sum_{x \in S}\relu(a\tran\mone(x) + b)}$. As shown earlier, using $\relu$ in the last layer of an elementwise function with a universally approximating $\mone$ is an instance of \our. 

\begin{wraptable}{r}{0.6\textwidth}
\centering
\scriptsize
\renewcommand{\arraystretch}{1.1}
\setlength{\tabcolsep}{3pt}
\begin{tabular}{l|c|c|c|c|}
    \toprule
    \textbf{Model} & \textbf{bedding} & \textbf{feeding} & \textbf{MSWEB} & \textbf{MSNBC} \\
    \midrule
    DeepSets & \textit{0.91} & 0.90 & 0.93 & \textit{0.97} \\
    DeepSets, monotone $\mtwo$ & 0.90 & 0.91 & 0.94 & 0.93 \\
    \relumas & \textbf{0.95} & \textbf{0.93} & \textbf{0.97} & \textbf{0.97} \\
    \hatmas & 0.91 & \textit{0.92} & \textit{0.95} & \textit{0.97} \\
    \hatmas (No division) & 0.87 & 0.92 & 0.92 & 0.94 \\
    \bottomrule
\end{tabular}
\caption{Ablation study }
\label{tab:ablation_study}
\end{wraptable}

But it is different from DeepSets in a key point, namely: $\bullet$ For set containment tasks, we don't have an outer $\mtwo$, which DeepSets have; for universal approximation tasks, we use a monotonically increasing $\mtwo$ that we enforce by taking positive weights and increasing activation functions. Also, for \hhat function based \our models, $\bullet$ we use a re-parametrization in which we perform a division based scaling. We now show the effect of the outer monotonic $\mtwo$ on DeepSets and division based re-parametrization on \hatmas in the  ablation study in table~\ref{tab:ablation_study}. White noise of std 0.1 added for \inexact containment.

\paragraph{Pointcloud datasets and related experiments}

ModelNet40 \cite{modelnetwu20153dshapenetsdeeprepresentation} is a benchmark dataset of 12,311 CAD models across 40 object categories, with each object represented as a 3D point cloud. We frame our task as checking if a given point cloud $S$ is a segment of a target point cloud $T$.

\begin{wraptable}[8]{r}{0.40\textwidth}
\vspace{-10pt}
\centering
\maxsizebox{\hsize}{!}{\tabcolsep 10pt
\begin{tabular}{l|c|c|c}
\hline
$|S|\rightarrow$ & $ 128$ & $ 256$ & $ 512$ \\
\hline
DeepSets        & \underline{0.90} & 0.90 & 0.91 \\
SetTransformer  & \underline{0.90} & \underline{0.91} & 0.91 \\
FlexSubNet      & 0.84 & 0.88 & 0.92 \\
Neural SFE      & 0.89 & 0.89 & 0.90 \\
\hline
\relumas        & \textbf{0.91}    & \textbf{0.93}    & \textbf{0.98} \\
\hatmas         & 0.87 & \underline{0.91} & \underline{0.94} \\
\hline
\end{tabular}}
\captionsetup{font=small}
\caption{Performance on Point cloud for different values of $|S|$, for 90:10 class ratio}
\label{tab:pointcloud_9010}
\end{wraptable}

Firstly, we choose an object from an object category $C_1$, and randomly sample $1024$ points from that object to get the target point cloud $T$. To obtain a positive sample (i.e true subset) from $T$, we first sample a random center point from $T$, and then extract $S$ using a hybrid approach: selecting the nearest point to the center, a few local neighbors via k-NN, and the rest via importance-weighted sampling (inverse-distance from center with noise). This makes sure that we're selecting a true local region from the point cloud, which is in-line with the actual task of detecting whether a given segment of an object is contained in a target object. For a negative sample(non-subset), we sample $S$ from an object of a different category $C_2$.

\begin{wraptable}[8]{r}{0.40\textwidth}
\vspace{-3pt}
\centering
\maxsizebox{\hsize}{!}{\tabcolsep 10pt
\begin{tabular}{l|c|c|c}
\hline
$|S|\rightarrow$ & $ 128$ & $ 256$ & $ 512$ \\
\hline
DeepSets        & \underline{0.89} & 0.90 & 0.90 \\
SetTransformer  & \underline{0.90} & \underline{0.91} & 0.91 \\
\hline
\relumas        & 0.87    & \textbf{0.91}    & \textbf{0.93} \\
\hatmas         & 0.85 & \underline{0.91} & 0.89 \\
\trimas        & 0.83    & 0.93    & 0.87 \\
\integralmas   & 0.87 & 0.90 & \underline{0.92} \\
\hline
\end{tabular}}
\captionsetup{font=small}
\caption{Point cloud with DGCNN for different values of $|S|$.}
\label{tab:pointcloud_dgcnn}
\end{wraptable}

We first give the accuracy table for PointNet encoder, but with the modified negative-to-positive class ratio of 90:10 to make the task harder for \our, which correctly classifies the actual subsets by the inductive bias of monotonicity. The numbers with the modified class-ratio are given in Tab~\cref{tab:pointcloud_9010}:

We also give the accuracy numbers for DGCNN encoder, in Table~\ref{tab:pointcloud_dgcnn}

\paragraph{Details on Linear assignment problem}
In this problem, we are given a positive matrix $M\in \R^{n \times m},n\leq m$, where $M_{i,j} $ 
represents the salary that a 'worker' will be paid to do a 'job' $j$. The goal of the task is to maximize the average salary obtained by all workers. This is done by finding the optimal assignment $\pi \in S_{n,m} $ which maps a worker $i\in [n]$ to a 'job' $\pi(i)\in [m]$, where by construction each worker can be mapped to at most one job, so $\pi(i)\in \{0,1\}$. This gives us the following maximization problem:
\begin{align*}
  \textstyle F(M)= \frac{1}{n}\max_{\pi \in S_{n,m}}\sum^{n}_{i=1} M_{i,\pi(i)}.
\end{align*}
Thinking of the matrix $M$ as a set of $m$ columns $M=[M_1,\ldots,M_m] $, we see that the function $F$ is permutation invariant and monotone. Accordingly, our goal will be to evaluate our \our model and baselines.

\clearpage
\addtocontents{toc}{\protect\setcounter{tocdepth}{1}}
\printbibliography[
    heading=subbibintoc,
    title={References}
]

\endgroup

\end{refsection}
\chapter{FSW-GNN: A Bi-Lipschitz WL-Equivalent Graph Neural Network}
\label{chap:paper4}

\noindent
\textbf{Yonatan Sverdlov, Yair Davidson, Nadav Dym, and Tal Amir.}\\
Learning on Graphs Conference (LoG), 2025.

\bigskip

\begin{refsection}[papers/paper4/main.bib]

\begingroup

\graphicspath{{papers/paper4/figures/}}

\let\citep\parencite
\let\citet\textcite

\newif\ifinappendix
\inappendixfalse
\newcommand{\onappendix}[1]{\ifinappendix#1\fi}
\newcommand{\onmain}[1]{\ifinappendix\else#1\fi}
\pgfplotsset{width=6cm,compat=newest,table/search path={papers/paper4/csv/}}
\newtheorem*{theorem*}{Theorem}
\newtheorem*{lemma*}{Lemma}
\newtheorem*{claim*}{Claim}
\newtheorem*{proposition*}{Proposition}
\newtheorem*{corollary*}{Corollary}
\newtheorem*{definition*}{Definition}
\newtheorem*{remark*}{Remark}
\newtheorem*{note*}{Note}
\newtheorem*{question*}{Question}
\newtheorem*{example*}{Example}

%
%
%

\makeatletter
\def\input@path{{papers/paper4/csv/}{papers/paper4/text/}{papers/paper4/tables/}{papers/paper4/figures/}}
\makeatother

\providetoggle{show_our_comments}
\settoggle{show_our_comments}{false} 

\newcommand{\nd}[1]{
\iftoggle{show_our_comments}{{\color{red}{\bf[Nadav:} \textup{#1}{\bf]}}}{}}

\newcommand{\ta}[1]{\iftoggle{show_our_comments}{{\color{blue}{\bf[Tal:} \textup{#1}{\bf]}}}{}}

\newcommand{\tared}[1]{\iftoggle{show_our_comments}{{\color{red}{\bf[Tal:} \textup{#1}{\bf]}}}{}}

\newcommand{\ys}[1]{\iftoggle{show_our_comments}{{\color{green}{\bf[Yonatan:} \textup{#1}{\bf]}}}{}}

\newcommand{\yd}[1]{
\iftoggle{show_our_comments}{{\color{orange}{\bf[Yair:} \textup{#1}{\bf]}}}{}}

\newcommand{\csvpath}{papers/paper4/csv/} 
\graphicspath{{papers/paper4/figures/}}


\newcommand{\myparagraph}[1]{\paragraph{{#1}.}}

\newcommand{\redline}[1]{ \iftoggle{show_our_comments} {
  \noindent
  \parbox{\linewidth}{
    \vspace{0.5em}
    \hrulefill
    \textcolor{red}{\rule{\linewidth}{0.8pt}}\\
    \centering #1\\
    \vspace{-0.45em}
    \textcolor{red}{\rule{\linewidth}{0.8pt}}
  }
} {} }

%
%
%
%


\renewcommand{\of}[1]{{\left({#1}\right)}} 
\newcommand{\multiset}[1]{{\{\!\!\{{#1}\}\!\!\}}} 

\newcommand{\setstspace}{{\ \,}}
\newcommand{\setst}[2]{ { \left\{ {#1} \setstspace\middle|\setstspace {#2} \right\} } }
\newcommand{\bigsetst}[2]{ { \bigbrc{ {#1} \setstspace\bigm|\setstspace {#2} } } }
\newcommand{\Bigsetst}[2]{ { \Bigbrc{ {#1} \setstspace\Bigm|\setstspace {#2} } } }
\newcommand{\biggsetst}[2]{ { \biggbrc{ {#1} \setstspace\biggm|\setstspace {#2} } } }
\newcommand{\Biggsetst}[2]{ { \Biggbrc{ {#1} \setstspace\Biggm|\setstspace {#2} } } }

\newcommand{\br}[1]{{\left({#1}\right)}} 
\newcommand{\brs}[1]{{\left[{#1}\right]}} 
\newcommand{\brc}[1]{{\left\{{#1}\right\}}} 
\newcommand{\bra}[1]{{\left\langle{#1}\right\rangle}} 
\newcommand{\norm}[1]{{\left\lVert{#1}\right\rVert}} 
\newcommand{\abs}[1]{{\left\lvert{#1}\right\rvert}} 
\newcommand{\floor}[1]{{\left\lfloor{#1}\right\rfloor}}  
\newcommand{\ceil}[1]{{\left\lfloor{#1}\right\rfloor}} 

\newcommand{\bigbr}[1]{{\bigl({#1}\bigr)}} 
\newcommand{\bigbrs}[1]{{\bigl[{#1}\bigr]}} 
\newcommand{\bigbrc}[1]{{\bigl\{{#1}\bigr\}}} 
\newcommand{\bigbra}[1]{{\bigl\langle{#1}\bigr\rangle}} 
\newcommand{\bignorm}[1]{{\bigl\lVert{#1}\bigr\rVert}} 
\newcommand{\bigabs}[1]{{\bigl\lvert{#1}\bigr\rvert}} 
\newcommand{\bigfloor}[1]{{\bigl\lfloor{#1}\bigr\rfloor}} 
\newcommand{\bigceil}[1]{{\bigl\lfloor{#1}\bigr\rfloor}} 

\newcommand{\Bigbr}[1]{{\Bigl({#1}\Bigr)}} 
\newcommand{\Bigbrs}[1]{{\Bigl[{#1}\Bigr]}} 
\newcommand{\Bigbrc}[1]{{\Bigl\{{#1}\Bigr\}}} 
\newcommand{\Bigbra}[1]{{\Bigl\langle{#1}\Bigr\rangle}} 
\newcommand{\Bignorm}[1]{{\Bigl\lVert{#1}\Bigr\rVert}} 
\newcommand{\Bigabs}[1]{{\Bigl\lvert{#1}\Bigr\rvert}} 
\newcommand{\Bigfloor}[1]{{\Bigl\lfloor{#1}\Bigr\rfloor}} 
\newcommand{\Bigceil}[1]{{\Bigl\lfloor{#1}\Bigr\rfloor}} 

\newcommand{\biggbr}[1]{{\biggl({#1}\biggr)}} 
\newcommand{\biggbrs}[1]{{\biggl[{#1}\biggr]}} 
\newcommand{\biggbrc}[1]{{\biggl\{{#1}\biggr\}}} 
\newcommand{\biggbra}[1]{{\biggl\langle{#1}\biggr\rangle}} 
\newcommand{\biggnorm}[1]{{\biggl\lVert{#1}\biggr\rVert}} 
\newcommand{\biggabs}[1]{{\biggl\lvert{#1}\biggr\rvert}} 
\newcommand{\biggfloor}[1]{{\biggl\lfloor{#1}\biggr\rfloor}} 
\newcommand{\biggceil}[1]{{\biggl\lfloor{#1}\biggr\rfloor}} 

\newcommand{\Biggbr}[1]{{\Biggl({#1}\Biggr)}} 
\newcommand{\Biggbrs}[1]{{\Biggl[{#1}\Biggr]}} 
\newcommand{\Biggbrc}[1]{{\Biggl\{{#1}\Biggr\}}} 
\newcommand{\Biggbra}[1]{{\Biggl\langle{#1}\Biggr\rangle}} 
\newcommand{\Biggnorm}[1]{{\Biggl\lVert{#1}\Biggr\rVert}} 
\newcommand{\Biggabs}[1]{{\Biggl\lvert{#1}\Biggr\rvert}} 
\newcommand{\Biggfloor}[1]{{\Biggl\lfloor{#1}\Biggr\rfloor}} 
\newcommand{\Biggceil}[1]{{\Biggl\lfloor{#1}\Biggr\rfloor}} 


\DeclareDocumentCommand \expect{ o m } {{ \IfNoValueTF{#1}{\mathbb{E}\brs{#2}}{\mathbb{E}_{#1}\brs{#2}} }}
\DeclareDocumentCommand \stdev{ o m } {{ \IfNoValueTF{#1}{\textup{STD}\brs{#2}}{\textup{STD}_{#1}\brs{#2}} }}
\DeclareDocumentCommand \variance{ o m } {{ \IfNoValueTF{#1}{\textup{Var}\brs{#2}}{\textup{Var}_{#1}\brs{#2}} }}

\DeclareDocumentCommand \rootp{ o m } {{ \IfNoValueTF{#1} { {\sqrt[\leftroot{-2}\uproot{2}{p}]{#2}} } { {\sqrt[\leftroot{-2}\uproot{2}{#1}]{#2}} } }}

\newcommand{\eqdef}{\coloneqq} 
\newcommand{\inteq}[1]{{\xlongequal[{#1}]{}}} 

\newcommand{\eqxspace}{{\hspace{0.0pt}}} 
\newcommand{\eqx}[1][]{{\eqxspace\overset{{\textup{#1}}}{=}\eqxspace}}
\newcommand{\leqx}[1][]{{\eqxspace\overset{{\textup{#1}}}{\leq}\eqxspace}}
\newcommand{\geqx}[1][]{{\eqxspace\overset{{\textup{#1}}}{\geq}\eqxspace}}
\newcommand{\ltx}[1][]{{\eqxspace\overset{{\textup{#1}}}{\lt}\eqxspace}}
\newcommand{\gtx}[1][]{{\eqxspace\overset{{\textup{#1}}}{\gt}\eqxspace}}
\newcommand{\neqx}[1][]{{\eqxspace\overset{{\textup{#1}}}{\neq}\eqxspace}}
\newcommand{\eqdefx}[1][]{{\eqxspace\overset{{\textup{#1}}}{\eqdef}\eqxspace}}
\newcommand{\subseteqx}[1][]{{\eqxspace\overset{{\textup{#1}}}{\subseteq}\eqxspace}}
\newcommand{\supseteqx}[1][]{{\eqxspace\overset{{\textup{#1}}}{\supseteq}\eqxspace}}
\newcommand{\subsetx}[1][]{{\eqxspace\overset{{\textup{#1}}}{\subset}\eqxspace}}
\newcommand{\supsetx}[1][]{{\eqxspace\overset{{\textup{#1}}}{\supset}\eqxspace}}

\newcommand{\argdot}{{\hspace{0.18em}{{\bm{\cdot}}}\hspace{0.18em}}} 

\newcommand{\dist}{{\textup{dist}}}


\newcommand{\figleft}{{\em (Left)}}
\newcommand{\figcenter}{{\em (Center)}}
\newcommand{\figright}{{\em (Right)}}
\newcommand{\figtop}{{\em (Top)}}
\newcommand{\figbottom}{{\em (Bottom)}}
\newcommand{\captiona}{{\em (a)}}
\newcommand{\captionb}{{\em (b)}}
\newcommand{\captionc}{{\em (c)}}
\newcommand{\captiond}{{\em (d)}}

\newcommand{\newterm}[1]{{\bf #1}}

\def\figref#1{figure~\ref{#1}}
\def\Figref#1{Figure~\ref{#1}}
\def\twofigref#1#2{figures \ref{#1} and \ref{#2}}
\def\quadfigref#1#2#3#4{figures \ref{#1}, \ref{#2}, \ref{#3} and \ref{#4}}
\def\secref#1{section~\ref{#1}}
\def\Secref#1{Section~\ref{#1}}
\def\twosecrefs#1#2{sections \ref{#1} and \ref{#2}}
\def\secrefs#1#2#3{sections \ref{#1}, \ref{#2} and \ref{#3}}
\def\eqref#1{equation~\ref{#1}}
\def\Eqref#1{Equation~\ref{#1}}
\def\plaineqref#1{\ref{#1}}
\def\chapref#1{chapter~\ref{#1}}
\def\Chapref#1{Chapter~\ref{#1}}
\def\rangechapref#1#2{chapters\ref{#1}--\ref{#2}}
\def\algref#1{algorithm~\ref{#1}}
\def\Algref#1{Algorithm~\ref{#1}}
\def\twoalgref#1#2{algorithms \ref{#1} and \ref{#2}}
\def\Twoalgref#1#2{Algorithms \ref{#1} and \ref{#2}}
\def\partref#1{part~\ref{#1}}
\def\Partref#1{Part~\ref{#1}}
\def\twopartref#1#2{parts \ref{#1} and \ref{#2}}

\def\ceil#1{\lceil #1 \rceil}
\def\floor#1{\lfloor #1 \rfloor}
\def\1{\bm{1}}
\newcommand{\train}{\mathcal{D}}
\newcommand{\valid}{\mathcal{D_{\mathrm{valid}}}}
\newcommand{\test}{\mathcal{D_{\mathrm{test}}}}

\def\eps{{\epsilon}}

\def\reta{{\textnormal{$\eta$}}}
\def\ra{{\textnormal{a}}}
\def\rb{{\textnormal{b}}}
\def\rc{{\textnormal{c}}}
\def\rd{{\textnormal{d}}}
\def\re{{\textnormal{e}}}
\def\rf{{\textnormal{f}}}
\def\rg{{\textnormal{g}}}
\def\rh{{\textnormal{h}}}
\def\ri{{\textnormal{i}}}
\def\rj{{\textnormal{j}}}
\def\rk{{\textnormal{k}}}
\def\rl{{\textnormal{l}}}
\def\rn{{\textnormal{n}}}
\def\ro{{\textnormal{o}}}
\def\rp{{\textnormal{p}}}
\def\rq{{\textnormal{q}}}
\def\rr{{\textnormal{r}}}
\def\rs{{\textnormal{s}}}
\def\rt{{\textnormal{t}}}
\def\ru{{\textnormal{u}}}
\def\rv{{\textnormal{v}}}
\def\rw{{\textnormal{w}}}
\def\rx{{\textnormal{x}}}
\def\ry{{\textnormal{y}}}
\def\rz{{\textnormal{z}}}

\def\rvepsilon{{\mathbf{\epsilon}}}
\def\rvtheta{{\mathbf{\theta}}}
\def\rva{{\mathbf{a}}}
\def\rvb{{\mathbf{b}}}
\def\rvc{{\mathbf{c}}}
\def\rvd{{\mathbf{d}}}
\def\rve{{\mathbf{e}}}
\def\rvf{{\mathbf{f}}}
\def\rvg{{\mathbf{g}}}
\def\rvh{{\mathbf{h}}}
\def\rvu{{\mathbf{i}}}
\def\rvj{{\mathbf{j}}}
\def\rvk{{\mathbf{k}}}
\def\rvl{{\mathbf{l}}}
\def\rvm{{\mathbf{m}}}
\def\rvn{{\mathbf{n}}}
\def\rvo{{\mathbf{o}}}
\def\rvp{{\mathbf{p}}}
\def\rvq{{\mathbf{q}}}
\def\rvr{{\mathbf{r}}}
\def\rvs{{\mathbf{s}}}
\def\rvt{{\mathbf{t}}}
\def\rvu{{\mathbf{u}}}
\def\rvv{{\mathbf{v}}}
\def\rvw{{\mathbf{w}}}
\def\rvx{{\mathbf{x}}}
\def\rvy{{\mathbf{y}}}
\def\rvz{{\mathbf{z}}}

\def\erva{{\textnormal{a}}}
\def\ervb{{\textnormal{b}}}
\def\ervc{{\textnormal{c}}}
\def\ervd{{\textnormal{d}}}
\def\erve{{\textnormal{e}}}
\def\ervf{{\textnormal{f}}}
\def\ervg{{\textnormal{g}}}
\def\ervh{{\textnormal{h}}}
\def\ervi{{\textnormal{i}}}
\def\ervj{{\textnormal{j}}}
\def\ervk{{\textnormal{k}}}
\def\ervl{{\textnormal{l}}}
\def\ervm{{\textnormal{m}}}
\def\ervn{{\textnormal{n}}}
\def\ervo{{\textnormal{o}}}
\def\ervp{{\textnormal{p}}}
\def\ervq{{\textnormal{q}}}
\def\ervr{{\textnormal{r}}}
\def\ervs{{\textnormal{s}}}
\def\ervt{{\textnormal{t}}}
\def\ervu{{\textnormal{u}}}
\def\ervv{{\textnormal{v}}}
\def\ervw{{\textnormal{w}}}
\def\ervx{{\textnormal{x}}}
\def\ervy{{\textnormal{y}}}
\def\ervz{{\textnormal{z}}}

\def\rmA{{\mathbf{A}}}
\def\rmB{{\mathbf{B}}}
\def\rmC{{\mathbf{C}}}
\def\rmD{{\mathbf{D}}}
\def\rmE{{\mathbf{E}}}
\def\rmF{{\mathbf{F}}}
\def\rmG{{\mathbf{G}}}
\def\rmH{{\mathbf{H}}}
\def\rmI{{\mathbf{I}}}
\def\rmJ{{\mathbf{J}}}
\def\rmK{{\mathbf{K}}}
\def\rmL{{\mathbf{L}}}
\def\rmM{{\mathbf{M}}}
\def\rmN{{\mathbf{N}}}
\def\rmO{{\mathbf{O}}}
\def\rmP{{\mathbf{P}}}
\def\rmQ{{\mathbf{Q}}}
\def\rmR{{\mathbf{R}}}
\def\rmS{{\mathbf{S}}}
\def\rmT{{\mathbf{T}}}
\def\rmU{{\mathbf{U}}}
\def\rmV{{\mathbf{V}}}
\def\rmW{{\mathbf{W}}}
\def\rmX{{\mathbf{X}}}
\def\rmY{{\mathbf{Y}}}
\def\rmZ{{\mathbf{Z}}}

\def\ermA{{\textnormal{A}}}
\def\ermB{{\textnormal{B}}}
\def\ermC{{\textnormal{C}}}
\def\ermD{{\textnormal{D}}}
\def\ermE{{\textnormal{E}}}
\def\ermF{{\textnormal{F}}}
\def\ermG{{\textnormal{G}}}
\def\ermH{{\textnormal{H}}}
\def\ermI{{\textnormal{I}}}
\def\ermJ{{\textnormal{J}}}
\def\ermK{{\textnormal{K}}}
\def\ermL{{\textnormal{L}}}
\def\ermM{{\textnormal{M}}}
\def\ermN{{\textnormal{N}}}
\def\ermO{{\textnormal{O}}}
\def\ermP{{\textnormal{P}}}
\def\ermQ{{\textnormal{Q}}}
\def\ermR{{\textnormal{R}}}
\def\ermS{{\textnormal{S}}}
\def\ermT{{\textnormal{T}}}
\def\ermU{{\textnormal{U}}}
\def\ermV{{\textnormal{V}}}
\def\ermW{{\textnormal{W}}}
\def\ermX{{\textnormal{X}}}
\def\ermY{{\textnormal{Y}}}
\def\ermZ{{\textnormal{Z}}}

\def\vzero{{\bm{0}}}
\def\vone{{\bm{1}}}
\def\vmu{{\bm{\mu}}}
\def\vtheta{{\bm{\theta}}}
\def\va{{\bm{a}}}
\def\vb{{\bm{b}}}
\def\vc{{\bm{c}}}
\def\vd{{\bm{d}}}
\def\ve{{\bm{e}}}
\def\vf{{\bm{f}}}
\def\vg{{\bm{g}}}
\def\vh{{\bm{h}}}
\def\vi{{\bm{i}}}
\def\vj{{\bm{j}}}
\def\vk{{\bm{k}}}
\def\vl{{\bm{l}}}
\def\vm{{\bm{m}}}
\def\vn{{\bm{n}}}
\def\vo{{\bm{o}}}
\def\vp{{\bm{p}}}
\def\vq{{\bm{q}}}
\def\vr{{\bm{r}}}
\def\vs{{\bm{s}}}
\def\vt{{\bm{t}}}
\def\vu{{\bm{u}}}
\def\vv{{\bm{v}}}
\def\vw{{\bm{w}}}
\def\vx{{\bm{x}}}
\def\vy{{\bm{y}}}
\def\vz{{\bm{z}}}

\def\evalpha{{\alpha}}
\def\evbeta{{\beta}}
\def\evepsilon{{\epsilon}}
\def\evlambda{{\lambda}}
\def\evomega{{\omega}}
\def\evmu{{\mu}}
\def\evpsi{{\psi}}
\def\evsigma{{\sigma}}
\def\evtheta{{\theta}}
\def\eva{{a}}
\def\evb{{b}}
\def\evc{{c}}
\def\evd{{d}}
\def\eve{{e}}
\def\evf{{f}}
\def\evg{{g}}
\def\evh{{h}}
\def\evi{{i}}
\def\evj{{j}}
\def\evk{{k}}
\def\evl{{l}}
\def\evm{{m}}
\def\evn{{n}}
\def\evo{{o}}
\def\evp{{p}}
\def\evq{{q}}
\def\evr{{r}}
\def\evs{{s}}
\def\evt{{t}}
\def\evu{{u}}
\def\evv{{v}}
\def\evw{{w}}
\def\evx{{x}}
\def\evy{{y}}
\def\evz{{z}}

\def\mA{{\bm{A}}}
\def\mB{{\bm{B}}}
\def\mC{{\bm{C}}}
\def\mD{{\bm{D}}}
\def\mE{{\bm{E}}}
\def\mF{{\bm{F}}}
\def\mG{{\bm{G}}}
\def\mH{{\bm{H}}}
\def\mI{{\bm{I}}}
\def\mJ{{\bm{J}}}
\def\mK{{\bm{K}}}
\def\mL{{\bm{L}}}
\def\mM{{\bm{M}}}
\def\mN{{\bm{N}}}
\def\mO{{\bm{O}}}
\def\mP{{\bm{P}}}
\def\mQ{{\bm{Q}}}
\def\mR{{\bm{R}}}
\def\mS{{\bm{S}}}
\def\mT{{\bm{T}}}
\def\mU{{\bm{U}}}
\def\mV{{\bm{V}}}
\def\mW{{\bm{W}}}
\def\mX{{\bm{X}}}
\def\mY{{\bm{Y}}}
\def\mZ{{\bm{Z}}}
\def\mBeta{{\bm{\beta}}}
\def\mPhi{{\bm{\Phi}}}
\def\mLambda{{\bm{\Lambda}}}
\def\mSigma{{\bm{\Sigma}}}

\def\gA{{\mathcal{A}}}
\def\gB{{\mathcal{B}}}
\def\gC{{\mathcal{C}}}
\def\gD{{\mathcal{D}}}
\def\gE{{\mathcal{E}}}
\def\gF{{\mathcal{F}}}
\def\gG{{\mathcal{G}}}
\def\gH{{\mathcal{H}}}
\def\gI{{\mathcal{I}}}
\def\gJ{{\mathcal{J}}}
\def\gK{{\mathcal{K}}}
\def\gL{{\mathcal{L}}}
\def\gM{{\mathcal{M}}}
\def\gN{{\mathcal{N}}}
\def\gO{{\mathcal{O}}}
\def\gP{{\mathcal{P}}}
\def\gQ{{\mathcal{Q}}}
\def\gR{{\mathcal{R}}}
\def\gS{{\mathcal{S}}}
\def\gT{{\mathcal{T}}}
\def\gU{{\mathcal{U}}}
\def\gV{{\mathcal{V}}}
\def\gW{{\mathcal{W}}}
\def\gX{{\mathcal{X}}}
\def\gY{{\mathcal{Y}}}
\def\gZ{{\mathcal{Z}}}

\def\sA{{\mathbb{A}}}
\def\sB{{\mathbb{B}}}
\def\sC{{\mathbb{C}}}
\def\sD{{\mathbb{D}}}
\def\sF{{\mathbb{F}}}
\def\sG{{\mathbb{G}}}
\def\sH{{\mathbb{H}}}
\def\sI{{\mathbb{I}}}
\def\sJ{{\mathbb{J}}}
\def\sK{{\mathbb{K}}}
\def\sL{{\mathbb{L}}}
\def\sM{{\mathbb{M}}}
\def\sN{{\mathbb{N}}}
\def\sO{{\mathbb{O}}}
\def\sP{{\mathbb{P}}}
\def\sQ{{\mathbb{Q}}}
\def\sR{{\mathbb{R}}}
\def\sS{{\mathbb{S}}}
\def\sT{{\mathbb{T}}}
\def\sU{{\mathbb{U}}}
\def\sV{{\mathbb{V}}}
\def\sW{{\mathbb{W}}}
\def\sX{{\mathbb{X}}}
\def\sY{{\mathbb{Y}}}
\def\sZ{{\mathbb{Z}}}

\def\emLambda{{\Lambda}}
\def\emA{{A}}
\def\emB{{B}}
\def\emC{{C}}
\def\emD{{D}}
\def\emE{{E}}
\def\emF{{F}}
\def\emG{{G}}
\def\emH{{H}}
\def\emI{{I}}
\def\emJ{{J}}
\def\emK{{K}}
\def\emL{{L}}
\def\emM{{M}}
\def\emN{{N}}
\def\emO{{O}}
\def\emP{{P}}
\def\emQ{{Q}}
\def\emR{{R}}
\def\emS{{S}}
\def\emT{{T}}
\def\emU{{U}}
\def\emV{{V}}
\def\emW{{W}}
\def\emX{{X}}
\def\emY{{Y}}
\def\emZ{{Z}}
\def\emSigma{{\Sigma}}

\newcommand{\pdata}{p_{\rm{data}}}
\newcommand{\ptrain}{\hat{p}_{\rm{data}}}
\newcommand{\Ptrain}{\hat{P}_{\rm{data}}}
\newcommand{\pmodel}{p_{\rm{model}}}
\newcommand{\Pmodel}{P_{\rm{model}}}
\newcommand{\ptildemodel}{\tilde{p}_{\rm{model}}}
\newcommand{\pencode}{p_{\rm{encoder}}}
\newcommand{\pdecode}{p_{\rm{decoder}}}
\newcommand{\precons}{p_{\rm{reconstruct}}}

\newcommand{\laplace}{\mathrm{Laplace}} 

\newcommand{\Ls}{\mathcal{L}}
\newcommand{\R}{\mathbb{R}}
\newcommand{\emp}{\tilde{p}}
\newcommand{\lr}{\alpha}
\newcommand{\reg}{\lambda}
\newcommand{\rect}{\mathrm{rectifier}}
\newcommand{\softmax}{\mathrm{softmax}}
\newcommand{\sigmoid}{\sigma}
\newcommand{\softplus}{\zeta}
\newcommand{\KL}{D_{\mathrm{KL}}}
\newcommand{\Var}{\mathrm{Var}}
\newcommand{\standarderror}{\mathrm{SE}}
\newcommand{\Cov}{\mathrm{Cov}}
\newcommand{\normlzero}{L^0}
\newcommand{\normlone}{L^1}
\newcommand{\normltwo}{L^2}
\newcommand{\normlp}{L^p}
\newcommand{\normmax}{L^\infty}

\newcommand{\parents}{Pa} 

\providecommand{\argmax}{\operatorname*{arg\,max}}
\providecommand{\argmin}{\operatorname*{arg\,min}}

\providecommand{\sign}{\operatorname{sign}}
\providecommand{\Tr}{\operatorname{Tr}}
\let\ab\allowbreak

%
%
%

\newcommand{\wleq}{{\mathrel{\overset{\tiny\textup{WL}}{\sim}}}}
\newcommand{\wlneq}{{\mathrel{\overset{\tiny\textup{WL}}{\nsim}}}}

\newcommand{\rhods}{{\rho_{\textup{DS}}}}
\newcommand{\rhotmd}{{\rho_{\textup{TMD}}}}

\DeclareDocumentCommand \TMD{ o } 
{ {\textup{TMD}\IfNoValueF{#1}{^{{#1}}}{}} } 
\newcommand{\GG}{\mathbb{G}}
\newcommand{\Gd}{\mathbb{G}(d)}
\newcommand{\GdN}{\mathcal{G}_{N}(\Omega)}
\newcommand{\GdNdistinct}{\mathbb{G}_{N}^{\mathrm{distinct}}(d)}
\newcommand{\GdNgeneric}{\mathbb{G}_{\mathrm{generic}}(d,N)}
\newcommand{\vvv}{{\vv}}
\newcommand{\vvx}{{\vx}}
\newcommand{\finv}{f_{\mathrm{inv}}}
\newcommand{\pair}{\mathbb{G}^{\text{pair}}} 

\newcommand{\X}{\textbf{X}}
\newcommand{\Y}{\textbf{Y}}
\newcommand{\Z}{\textbf{Z}}
\renewcommand{\C}{\textbf{C}}
\newcommand{\Gdomain}{G_{\leq n}(\Omega)}

\newcommand{\relu}{\mathrm{ReLU}}
\renewcommand{\upsilon}{\eta} 
\newcommand{\SnOmega}{\mathcal{S}_{\le n}(\Omega)}
\newcommand{\SnOmegaEqual}{\mathcal{S}_{=n}(\Omega)}
\newcommand{\Holder}{H\"{o}lder}
\newcommand{\lms}{ \{\!\!\{ }
\newcommand{\rms}{ \}\!\!\} }
\newcommand{\bigO}{\mathcal{O}}
\newcommand{\mbf}{\mathbf}

\newcommand{\dX}{D_{\X}}
\newcommand{\Dint}{\dX}
\newcommand{\enc}{\mathrm{enc}}

\newcommand{\GNof}[1]{{\G_{\leq N}\of{{#1}}}}
\newcommand{\GNofe}[1]{{\G_{= N}\of{{#1}}}}
\newcommand{\GNofWL}[1]{{\G^{\textup{WL}}_{\leq N}\of{{#1}}}}
\newcommand{\GNd}{{\GNof{\RR^d}}}
\newcommand{\GNdWL}{{\GNofWL{\RR^d}}}

\renewcommand{\k}{\kappa}
\renewcommand{\F}{\mathcal{F}}
\newcommand{\M}{\mathbb{M}}
\renewcommand{\W}{\mathbb{W}}
\newcommand{\G}{\mathcal{G}}
\newcommand{\GnS}{\G_{\leq n}(\Sigma)}
\newcommand{\Gnd}{\G_{\leq N}(\RR^d)}
\renewcommand{\V}{\mathcal{V}}
\newcommand{\E}{\textbf{E}}
\renewcommand{\K}{\mathcal{K}}
\newcommand{\D}{\mathcal{D}}
\renewcommand{\N}{\mathcal{N}}
\renewcommand{\P}{\mathcal{P}}
\renewcommand{\H}{\mathcal{H}}
\newcommand{\HH}{\mathcal{H}}
\renewcommand{\L}{\mathcal{L}}
\renewcommand{\A}{\textbf{A}}
\newcommand{\B}{\textbf{B}}
\renewcommand{\S}{\mathcal{S}}
\newcommand{\NN}{\mathbb{N}}
\newcommand{\CC}{\mathbb{C}}
\newcommand{\RR}{\mathbb{R}}
\newcommand{\ZZ}{\mathbb{Z}}
\newcommand{\EE}{\mathbb{E}}

\newcommand{\OO}{\mathcal{O}}

\newcommand{\Eus}{E_{\mathrm{quant}}}
\newcommand{\SnR}{{\S_{\leq n}\of{\RR}}}
\newcommand{\SnRd}{{\S_{\leq n}\of{\RR^d}}}
\newcommand{\SnRdequal}{{\S_{= n}\of{\RR^d}}}
\newcommand{\SnSig}{{\S_{\leq n}\of{\Sigma}}}
\newcommand{\MnSig}{{\mathcal{M}_{\leq n}\of{\Sigma}}}
\newcommand{\Mntheta}{\mathcal{M}_{\leq n}\of{\Omega\of{\theta_1}}}
\newcommand{\PnK}{{\P_{\leq n}\of{K}}}
\newcommand{\PnR}{{\P_{\leq n}\of{\RR}}}
\newcommand{\PnRd}{{\P_{\leq n}\of{\RR^d}}}
\newcommand{\Pnof}[1]{{\P_{\leq n}\br{{#1}}}}
\newcommand{\Snof}[1]{{\S_{\leq n}\br{{#1}}}}
\newcommand{\Pnnof}[2]{{\P_{{#1}}\br{{#2}}}}
\newcommand{\Snnof}[2]{{\S_{{#1}}\br{{#2}}}}
\newcommand{\toas}[2]{\,{ \xrightarrow[{{#2}}]{}}\,{{#1}}}
 
\newcommand{\Sneq}{\S_{\leq n}(\Omega)}
\newcommand{\Sndeq}{\S_{\leq n}(\RR^d)}
\newcommand{\Sn}{\S_{\leq n}(\Omega)}
\newcommand{\GnO}{\mathcal{G}_{\leq n}(\Omega)}
\newcommand{\GnSig}{\mathcal{G}_{\leq n}(\Sigma)}
\newcommand{\Snd}{\S_{\leq n}(\RR^d)}
\newcommand{\MnR}{\mathcal{M}_{\leq n}(\RR)}
\newcommand{\Mnd}{\mathcal{M}_{\leq n}(\RR^d)}
\newcommand{\Mn}{\mathcal{M}_{\leq n}(\Omega)}
\newcommand{\sort}{\textup{sort}}
\newcommand{\rowsort}{\mathbf{rowsort}}
\newcommand{\columnsort}{\mathbf{columnsort}}
\newcommand{\poly}{\mathrm{poly}}
\newcommand{\Dparam}{D_{param}}

\renewcommand{\T}{\mathcal{T}}
\newcommand{\bT}{\boldsymbol{\T}}
\newcommand{\bpsi}{\boldsymbol{\psi}}
\newcommand{\ba}{\boldsymbol{a}}
\newcommand{\bbeta}{\boldsymbol{\beta}}
\newcommand{\bb}{\boldsymbol{b}}
\newcommand{\bd}{\boldsymbol{d}}
\newcommand{\bA}{\boldsymbol{A}}

\DeclareDocumentCommand \bx{ o } {{ {\boldsymbol{x}} \IfNoValueF {#1} {^{\br{#1}}} }}

\DeclareDocumentCommand \btx{ o } {{ {\tilde{\boldsymbol{x}}} \IfNoValueF {#1} {^{\br{#1}}} }}

\DeclareDocumentCommand \by{ o } {{ {\boldsymbol{y}} \IfNoValueF {#1} {^{\br{#1}}} }}

\DeclareDocumentCommand \bv{ o } {{ {\boldsymbol{v}} \IfNoValueF {#1} {^{\br{#1}}} }}
\DeclareDocumentCommand \xik{ o } {{ {\xi} \IfNoValueF {#1} {^{\br{#1}}} }}

\DeclareDocumentCommand \bu{ o } {{ {\boldsymbol{u}} \IfNoValueF {#1} {^{\br{#1}}} }}

\DeclareDocumentCommand \Wass{ o o } {{ {\mathcal{W}} \IfNoValueF {#1} {_{{#1}}} \IfNoValueF {#2} {^{{#2}}} }}
\newcommand{\Wassp}{{\Wass[p]}}
\newcommand{\Wasspp}{{\Wass[p][p]}}

\newcommand{\Wassoo}{{{\widetilde{\mathcal{W}}}_1}}

\newcommand{\WassSqr}{{\mathcal{W}^2}}
\newcommand{\WassQuad}{{\mathcal{W}^4}}
\newcommand{\SWass}{{\mathcal{SW}}}
\newcommand{\SWassSqr}{{\mathcal{SW}^2}}
\newcommand{\SWassQuad}{{\mathcal{SW}^4}}

\DeclareDocumentCommand \bp{ o } {{ {\boldsymbol{p}} \IfNoValueF {#1} {^{\br{#1}}} }}
\DeclareDocumentCommand \bq{ o } {{ {\boldsymbol{q}} \IfNoValueF {#1} {^{\br{#1}}} }}
\DeclareDocumentCommand \bw{ o } {{ {\boldsymbol{w}} \IfNoValueF {#1} {^{\br{#1}}} }}
\DeclareDocumentCommand \btw{ o } {{ {\tilde{\boldsymbol{w}}} \IfNoValueF {#1} {^{\br{#1}}} }}

\newcommand{\bbm}{\boldsymbol{m}} 
\newcommand{\bmsg}{\boldsymbol{m}}
\newcommand{\bh}{\boldsymbol{h}}
\newcommand{\bht}{\boldsymbol{\tilde{h}}}
\newcommand{\bg}{\boldsymbol{g}}
\newcommand{\bM}{\boldsymbol{M}}
\newcommand{\bX}{\boldsymbol{X}}
\newcommand{\btX}{\tilde{\boldsymbol{X}}}
\newcommand{\bV}{\boldsymbol{V}}
\newcommand{\bY}{\boldsymbol{Y}}
\newcommand{\bz}{\boldsymbol{z}}
\newcommand{\bC}{\boldsymbol{C}}
\newcommand{\bzero}{\boldsymbol{0}}
\newcommand{\btheta}{\boldsymbol{\theta}}
\newcommand{\bth}{\btheta}

\newcommand{\Quant}{{\textup{Q}}}
\newcommand{\Qtrans}{\costrans{\Quant}}
\newcommand{\Quanthat}{{\hat{\mathup{Q}}}}

\newcommand{\emb}{{{E}}}
\newcommand{\embfsw}{{{E}_{\textup{FSW}}}}
\newcommand{\embfswglob}{{{E}_{\textup{FSW}}^{\textup{Glob}}}}
\DeclareDocumentCommand \embfswt{ o } {{ {{E}}_{\textup{FSW}} \IfNoValueTF {#1} {^\br{t}} {^\br{{#1}}} }}
\DeclareDocumentCommand \embfswm{ o } {{ {{E}}^{\textup{FSW}} \IfNoValueTF {#1} {_m} {_{{#1}}} }}
\DeclareDocumentCommand \embt{ o } {{ {\textup{FSW}}\IfNoValueTF {#1} {^\br{t}} {^\br{{#1}}} }}

\newcommand{\Dnosqr}{{\Delta}}
\newcommand{\Dsqr}{{\Delta^2}}
\newcommand{\Dquad}{{\Delta^4}}

\newcommand{\Symn}{{\textup{S}_n}}

\newcommand{\kmin}{k_{\min}}

\newcommand{\Sbb}{\mathbb{S}}

\newcommand{\ord}[1]{_{\br{{#1}}}}

\newcommand{\essmin}{\textup{ess}\min}
\newcommand{\essmax}{\textup{ess}\max}

\newcommand{\essinf}{\textup{ess}\inf}
\newcommand{\esssup}{\textup{ess}\sup}

\DeclareDocumentCommand \Lp{ o } {{ \IfNoValueTF{#1} {{L}^p}{{L}^{#1}} }}
\DeclareDocumentCommand \LpR{ o } {{ \IfNoValueTF{#1} {{L}^p\of{\RR}}{{L}^{#1}\of{\RR}} }}

\newcommand{\costrans}[1]{{\invbreve{#1}}}

\newcommand{\distxi}{{\mathcal{D}_{\xi}}}

\newcommand{\cone}[1]{{\mathcal{C}_{{#1}}}}


\newcommand{\tmd}{\mathrm{TMD}}
\newcommand{\TD}{\mathrm{TD}}
\newcommand{\tree}{\mathbf{T}}
\newcommand{\blankT}{\Bar{\tree}_0}


%
%

\begin{abstract} 
%
%

Famously, the ability of Message Passing Neural Networks (MPNN) to distinguish between graphs is limited to graphs separable by the Weisfeiler-Leman (WL) graph isomorphism test, and the strongest MPNNs, in terms of separation power, are WL-equivalent. 
However, it was demonstrated that the quality of separation provided by standard WL-equivalent MPNNs can be very low, resulting in WL-separable graphs being mapped to very similar, hardly distinguishable outputs. This phenomenon can be explained by the recent observation that standard MPNNs are not lower-Lipschitz. 
This paper addresses this issue by introducing FSW-GNN, the first MPNN that is fully bi-Lipschitz with respect to standard WL-equivalent graph metrics.   Empirically, we show that our MPNN is competitive with standard MPNNs for several graph learning tasks and is far more accurate in long-range tasks, due to its ability to avoid oversmoothing and oversquashing. Our code is available at \url{https://github.com/yonatansverdlov/Over-squashing}.
\end{abstract}
%

\ta{Hide the TOC by setting \texttt{show\_our\_comments} to false in \texttt{base/settings.tex}.}



\section{Introduction}
\label{sec_introduction}
Graph neural networks are a central research topic in contemporary machine learning. As shown by \citet{Gilmer2017NeuralMP}, many of the most popular models can be seen as instantiations of Message Passing Neural Networks (MPNNs). 

A well-known limitation of MPNNs is that they cannot differentiate between all distinct pairs of graphs. In fact, a pair of distinct graphs that cannot be separated by the Weisfeiler-Leman (WL) graph isomorphism test will not be separated by \emph{any} MPNN \citep{xu2018how}. Accordingly, the most expressive MPNNs are those that are  \emph{WL-equivalent}, which means they can separate all pairs of graphs that are separable by WL. WL-equivalent MPNNs were proposed in the seminal works of \citet{xu2018how,morris2019WL}, and the complexity of these constructions was later improved in \citep{aamand2022exponentially,amir2023neural}. 

While separation is theoretically guaranteed with WL-equivalent MPNNs, in some cases, their separation in practice is so weak that it cannot be observed with 32-bit floating-point numbers; see \cite{pmlr-v235-bravo24a}. Moreover, using many MPNN iterations often leads to almost-identical node features (oversmoothing), or features that are barely affected by changes in far-off nodes (oversquashing). These observations motivate the development of \emph{quantitative} estimates of MPNN separation by means of bi-Lipschitz stability guarantees. These guarantees would ensure that  Euclidean distances in the MPNN feature space are neither much larger nor much smaller than distances in the original graph space, which are defined by a suitable metric on graphs. Consequently, as we shall see, detrimental phenomena like oversmoothing and oversquashing can be fundamentally mitigated.

\begin{figure}[t]
\centering
\begin{minipage}{0.7\textwidth}

\definecolor{darkgreen}{rgb}{0.0, 0.5, 0.0} 
\definecolor{brown}{rgb}{0.6, 0.4, 0.2} 
\begin{tikzpicture}
\begin{axis}[
    width=\linewidth, 
    height=0.65\linewidth, 
    xlabel={Problem radius},
    ylabel={Accuracy},
    xlabel style={font=\small},
    ylabel style={font=\small},
    legend style={
        at={(0.04,0.04)}, 
        anchor=south west, 
        font=\small,
        nodes={scale=1, transform shape}, 
        legend columns=1 
    },
    grid=both,
    grid style={opacity=0.5},
    tick label style={
        /pgf/number format/fixed, 
        /pgf/number format/precision=2, 
        /pgf/number format/zerofill=false, 
    },
    yticklabel={\pgfmathprintnumber{\tick}\kern0.1em\%},
    scaled ticks=false 
        ]
\addplot[
    color=orange, 
    mark=square*, 
    mark size=2pt,
    line width=1pt,
    solid]
    coordinates {
        (2,100)
        (3,100)
        (4,100)
        (5,100)
        (6,100)
        (7,100)
        (8,100)
    };
\addlegendentry{FSW-GNN}
\addplot[
    color=brown, 
    mark=square*, 
    mark size=2pt,
    line width=1pt,
    dash dot]
    coordinates {
        (2,100.0)
        (3,100.0)
        (4,100.0)
        (5,100.0)
        (6,100.0)
        (7,100.0)
        (8,3.2343748956918716)
    };
\addlegendentry{Sort-MPNN}
\addplot[
    color=purple, 
    mark=diamond*, 
    mark size=3pt,
    line width=1pt,
    dotted]
    coordinates {
        (2,100.0)
        (3,100.0)
        (4,100.0)
        (5,100.0)
        (6,60.8)
        (7,57.2)
        (8,54.0)
    };
\addlegendentry{GAT}
\addplot[
    color=red, 
    mark=triangle*, 
    mark size=3pt,
    line width=1pt,
    dashed]
    coordinates {
        (2,100)
        (3,100)
        (4,100)
        (5,80)
        (6,18)
        (7,11)
        (8,5)
    };
\addlegendentry{GIN}
\addplot[
    color=blue, 
    mark=*, 
    mark size=2pt,
    line width=1pt,
    solid] 
    coordinates {
        (2,100)
        (3,100)
        (4,100)
        (5,58)
        (6,4)
        (7,3)
        (8,3)
    };
\addlegendentry{GCN}
\addplot[
    color=cyan, 
    mark=star, 
    mark size=3pt,
    line width=1pt,
    dashed]
    coordinates {
        (2,100)
        (3,100)
        (4,100)
        (5,20)
        (6,4)
        (7,3)
        (8,3)
    };
\addlegendentry{GGNN}
\end{axis}
\end{tikzpicture}
    \end{minipage}
    \caption{FSW-GNN handles tasks with large radius better than standard MPNNs, which are more prone to oversmoothing and oversquashing. Task taken from \cite{alon}}
    \label{fig:Trees}

\end{figure}
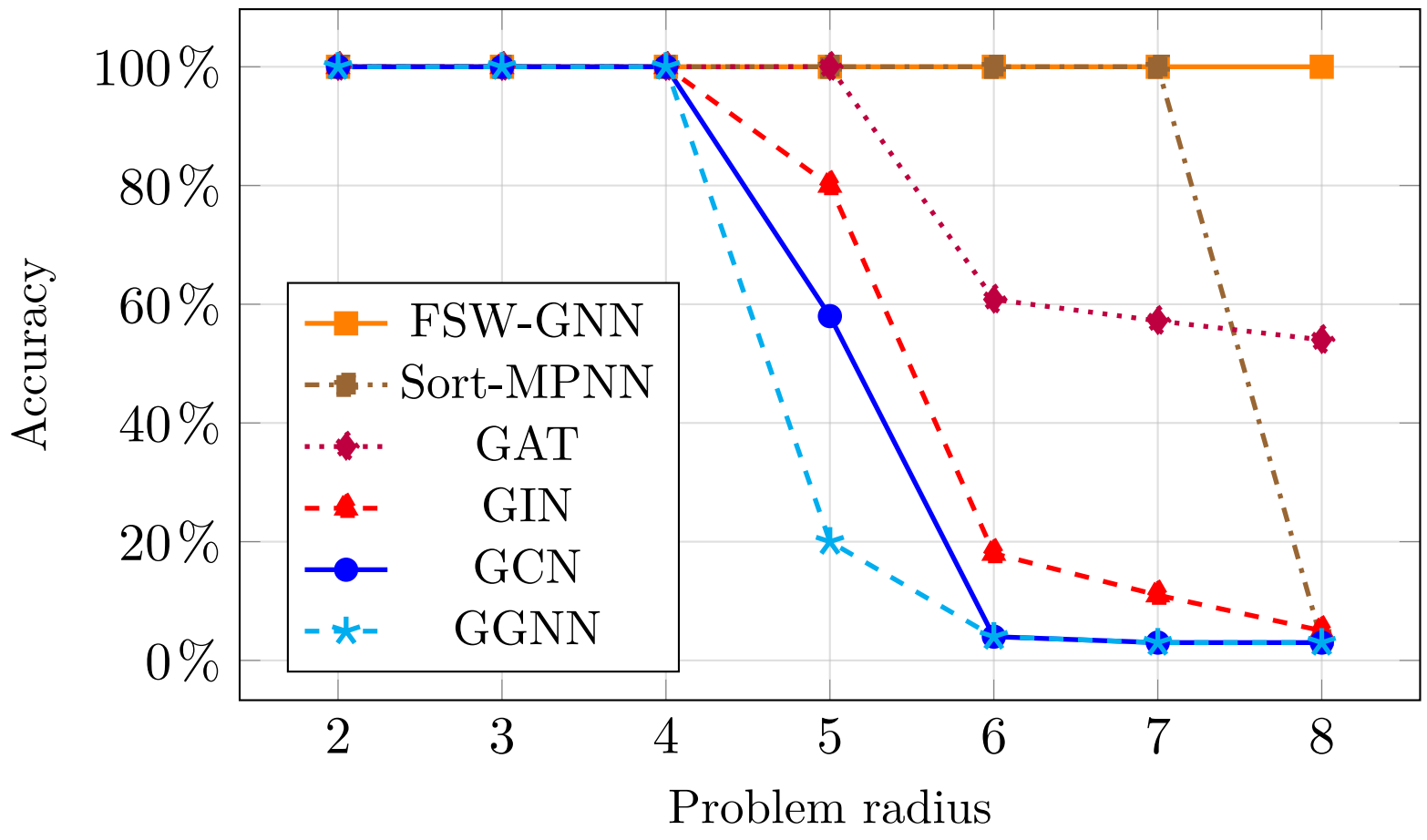


This paper introduces a novel MPNN, called \emph{FSW-GNN} (\emph{Fourier Sliced-Wasserstein GNN}), which is  bi-Lipschitz  with respect to two WL-equivalent graph metrics: (a) the Doubly Stochastic (DS) metric of \cite{grohe}, and (b) the Tree Mover's Distance
(TMD) metric of \citet{TMD}. 
Empirically, we show that FSW-GNN performs comparably or better than prevalent MPNNs on standard learning tasks, and  achieves significantly superior performance in long-range tasks, i.e., tasks that require a large number of message-passing iterations (for example see Figure \ref{fig:Trees}). This can be attributed to the bi-Lipschitzness of FSW-GNN, in contrast to standard MPNNs, which are not bi-Lipschitz \cite{davidson2024h}.
\subsection{Related Works}
\label{subsec_related_works}
%

%


\paragraph{Bi-Lipschitzness}\label{related_work:bi_lip}
Bi-Lipschitzness arises naturally in many domains, including frames \citep{balan1997stability}, phase retrieval \citep{bandeira2014saving, cheng2021stable}, group-invariant learning \citep{cahill2020complete, cahill2024bilipschitz} and  multisets \citep{amir2023neural, amir2024injective}. 
In the context of MPNNs, upper Lipschitzness was discussed in 
\citep{TMD, levie2023a}.
A recent survey by \citet{morris2024future} identifies \emph{bi-Lipschitzness} as a significant future goal for theoretical GNN research. 

\paragraph{SortMPNN}
First steps towards a Bi-Lipschitz MPNN have recently  been made by \citet{davidson2024h}. Their work analyzes a weaker notion of Lipschitz and H{\"o}lder guarantees---\emph{in expectation} over the model parameters. They show that essentially all popular MPNN models fail to be lower-Lipschitz, but are lower-H{\"o}lder in expectation, with an exponent that grows worse as the MPNN depth increases. In contrast, they propose a novel MPNN, called SortMPNN, and prove that it satisfies this weaker notion of bi-Lipschitzness in expectation. 

As we discuss below, the FSW-GNN satisfies the standard, stronger, notion of bi-Lipschitzness. Interestingly, the 
proof technique we present here for FSW-GNN also applies to SortMPNN, allowing us to establish that SortMPNN is bi-Lipschitz as well. Nonetheless, SortMPNN has a key limitation in its message aggregation mechanism: it handles neighborhoods of different sizes by padding them to a predetermined maximal size. This requires \emph{a priori} knowledge of the maximal neighborhood size in all future input graphs---an inherent constraint that significantly limits its applicability. 
In contrast, FSW-GNN does not share this constraint, since it treats vertex neighborhoods as distributions.
Moreover, due to this padding-based approach, SortMPNN only accommodates neighborhoods that are multisets, thus making it unsuitable for graphs with non-integer edge weights. In contrast, FSW-GNN naturally supports edge weights.

\paragraph{MPNNs with advanced pooling mechanisms}
In addition to SortMPNN, our approach is conceptually related to other MPNNs that replace the basic max- or sum-pooling, used for message aggregation, with more advanced pooling mechanisms, such as sorting \citep{balan2022permutation, DGCNN}, standard deviation \citep{pna}, or Sliced-Wasserstein embeddings via template distributions \citep{kolouriwasserstein}. However, these methods lack the bi-Lipschitzness guarantees that our model provides.

\paragraph{WL-equivalent metrics}
Bi-Lipschitz analysis of MPNNs requires a WL-equivalent graph metric. Several such metrics have been proposed, with notable examples being the \emph{Doubly-Stochastic (DS) metric} (also known as the \emph{tree metric}) \citep{grohe}; the \emph{Tree Mover's Distance (TMD)} \citep{TMD}; and the \emph{WL metric} \citep{wlmetric}.
In this paper, we prove that the graph embeddings computed by our FSW-GNN model are bi-Lipschitz with respect to both the DS and TMD metrics.
This analysis applies to graphs of bounded size with continuous, bounded node-features.
Weaker notions of equivalence between these metrics, in the context of graphs with unbounded cardinality and without node features, are discussed in \citep{boker2024fine,boker2021graph}.


\begin{table*}[t]
    \centering
    \caption{
        Node classification accuracy (\%) across homophilic and
        heterophilic graph benchmarks.
        The best and second-best results among the compared models are
        highlighted in bold and underlined, respectively.
        The SOTA row reports the best previously published result.
    }
    \label{tab:performance_comparison}

    \setlength{\tabcolsep}{5.5pt}
    \renewcommand{\arraystretch}{1.08}
    \begin{adjustbox}{max width=\textwidth}
    \begin{tabular}{lccccccccc}
        \toprule
        \textbf{Model}
        & \textbf{Cora}
        & \textbf{Cite.}
        & \textbf{Pubm.}
        & \textbf{Cham.}
        & \textbf{Squi.}
        & \textbf{Actor}
        & \textbf{Corn.}
        & \textbf{Texa.}
        & \textbf{Wisc.} \\
        \midrule

        GCN
        & 85.77
        & 73.68
        & \underline{88.13}
        & 28.18
        & 23.96
        & 26.86
        & 52.70
        & 52.16
        & 45.88 \\

        GAT
        & \textbf{86.37}
        & \underline{74.32}
        & 87.62
        & 42.93
        & 30.03
        & \underline{28.45}
        & \underline{54.32}
        & \underline{58.38}
        & \underline{49.41} \\

        FSW-GNN
        & \underline{86.35}
        & \textbf{75.44}
        & \textbf{88.17}
        & \underline{51.18}
        & \underline{36.38}
        & \textbf{34.66}
        & \textbf{72.43}
        & \textbf{75.68}
        & \textbf{81.56} \\

        Sort-MPNN
        & 83.46
        & 72.69
        & 85.15
        & \textbf{78.11}
        & \textbf{74.69}
        & 31.32
        & 67.03
        & 70.54
        & 73.92 \\

        \midrule
        \textit{Previously reported SOTA}
        & 90.16
        & 82.07
        & 91.31
        & 79.71
        & 76.71
        & 51.81
        & 92.72
        & 88.38
        & 94.99 \\
        \bottomrule
    \end{tabular}
    \end{adjustbox}
\end{table*}

\section{Problem Setting}
\label{sec_problem_setting}
In this section, we outline the problem setting, first providing the theoretical background of the problem and then stating our objectives.

\paragraph{Vertex-featured graphs}
Our main objects of study are graphs with vertex features, represented as triplets $G=\br{V,E,X}$, where $V=\brc{v_i}_{i=1}^n$ is the set of vertices, $E\subseteq \setst{ \brc{v_i,v_j}}{i,j\in\brs{n}}$ are the undirected edges in $G$, and $X=\brs{\bx_1,\ldots,\bx_n}$ is a matrix containing the vertex feature vectors $\bx_i\in\Omega$, with  $\Omega \subseteq \RR^d$ being the \emph{feature domain}. We denote by $\GNof\Omega$ the set of all vertex-featured graphs with at most $N$ vertices and corresponding features in $\Omega$. Throughout the paper, we use $\brc{}$ to denote multisets. We note that our results readily extend to graphs with edge features.\ta{; see XX. (to do)}
%

%

\paragraph{Weisfeiler-Leman Graph Isomorphism test}
Two graphs are \emph{isomorphic} if they are identical up to relabeling of their nodes. Perhaps surprisingly, the problem of determining whether two given graphs are isomorphic is rather challenging. To date, no known algorithm can solve it in polynomial time \citep{babai2016graph}. However, there exist various heuristics that provide an incomplete but often adequate method to test whether a given pair of graphs is isomorphic. The most notable example is the Weisfeiler-Leman (WL) graph isomorphism test.

The WL test can be described as assigning to each graph $G=\br{V,E,X}$ a feature vector $c_G^T$ according to the formula
\begin{equation}
\begin{split}\label{eqdef_wl}
    c^{0}_{v} \eqdefx & \X_{v}, \quad v \in V; 
    \quad \textup{for}\ \ 1 \leq t \leq T:
    \\ c^{t}_{v} \eqdefx & \textup{Combine}\br{c^{t-1}_{v},\textup{Aggregate}\br{\setst{c^{t-1}_{u}}{u \in \N_{v}}}};
    \\ c^{T}_G \eqdefx & \textup{Readout}\br{\brc{c^{T}_{v_{1}},\ldots,c^{T}_{v_{n}}}},
\end{split}
\end{equation}
where Aggregate and Readout are functions that injectively map multisets of vectors in Euclidean space into another Euclidean space, Combine is an injective function from one Euclidean space to another, and $\N_{v}$ denotes the neighborhood of the vertex $v$ in $G$. 

\begin{definition*}[WL graph equivalence]
Two vertex-featured graphs $G$ and $\tilde{G}$ are said to be \emph{WL-equivalent}, denoted by $G \wleq \tilde G$, if $c^T_G = c^T_{\tilde G}$ for all $T \geq 0$. Otherwise, they are said to be \emph{WL-separable}.
\end{definition*}

It is a known fact \citep{grohe2021logic,morris2023weisfeiler} that for $G,\tilde G \in \GNd$, if the equality $c^T_G = c^T_{\tilde G}$ is satisfied for $T=N$, then it is satisfied for all $T \geq 0$, and thus $G \wleq \tilde G$.

While the WL test can separate most pairs of non-isomorphic graphs, there exist examples of non-isomorphic graph pairs that WL cannot separate; see \citep{fail_example}.
%

\paragraph{Message passing neural networks}
Message Passing Neural Networks (MPNNs) 
operate on a similar principle to the WL test, but with the purpose of performing predictions on single graphs rather than determining if pairs of them are isomorphic. 
Their core mechanism is the message-passing procedure, which maintains a hidden feature for each vertex and iteratively updates it as a function of the neighbors' features.
This process is outlined as follows:

\begin{enumerate}
    \item \textbf{Initialization: } The hidden feature $\bh^{(0)}_v$ of each node is initialized by its input feature $\bx_v$.
    \item \textbf{Message aggregation:} Each node $v\in V$ aggregates messages from its neighbors by
    \begin{equation}
        \bmsg_v^{\br{t}} \eqdef \textup{Aggregate}\of{ \setst{ \bh^{\br{t-1}}_u } {u \in \N_v} } 
    \end{equation}
  Where Aggregate is a multiset-to-vector function.   %
    \item \textbf{Update step:} Each node updates its own hidden feature according to its aggregated messages and its previous hidden feature, using a vector-to-vector \emph{update function}:
    \begin{equation}
        \bh^{\br{t}}_v \eqdef \textup{Update}\of{ \bmsg_v^{\br{t}}, \bh^{\br{t-1}}_v },
    \end{equation}
    \item \textbf{Readout:} After $T$ iterations of steps 2-3, a graph-level feature $\bh_G$ is computed by a multiset-to-vector \emph{readout function}:
    \begin{equation*}
        \bh_G \eqdef \textup{Readout}\of{\setst{ \bh^{\br{T}}_v}{v\in V}}.
    \end{equation*}
\end{enumerate}
Numerous MPNNs were proposed in recent years, including GIN \citep{xu2018how}, GraphSAGE \citep{hamilton17graphsage}, GAT \citep{Velickovic18gat}, and GCN \citep{kipf2017semisupervised},  the main differences between them being the specific choices of the aggregation, update, and readout functions. 

An MPNN computes an \emph{embedding} of a graph $G$ to a vector $F\of{G}= \bh_G$. The obtained embedding is often further processed by standard machine-learning tools for vectors, such as multi-layer perceptrons (MLPs), to obtain a final graph prediction.  
The ability of such a model to approximate functions on graphs is closely related to the separation properties of $F$: if $F$ can differentiate between any pair of non-isomorphic graphs, then a model of the form $\mathrm{MLP}\circ F$ would be able to approximate any function on graphs \cite{chen2019equivalence}.

Unfortunately, an MPNN cannot separate any pair of WL-equivalent graphs, even if they are not truly isomorphic \citep{xu2018how,morris2019WL}. Accordingly, the best we can hope for from an MPNN, in terms of separation, is \emph{WL equivalence}: for every pair of graphs $G, G'\in \GNof\Omega$, $F(G)=F(G')$ \emph{ if and only if } $G \wleq G'$. While MPNNs based on max- or mean-pooling cannot be WL-equivalent \citep{xu2018how}, it is possible to construct WL-equivalent MPNNs based on sum-pooling, as discussed in \citep{xu2018how,morris2019WL,aamand2022exponentially,amir2023neural,bravo2024dimensionality}. 
Theoretically, a properly tuned graph model based on a WL-equivalent MPNN should be capable of perfectly solving any binary classification task, provided that no two WL-equivalent graphs have different ground-truth labels. However, this separation does not always manifest in practice. One reason is that WL-equivalent functions may map two input graphs that are far apart in the input space to outputs that are numerically indistinguishable in the output Euclidean space. In fact, \citet{davidson2024h} provide an example of graph pairs that are not WL-equivalent, yet are mapped to near-identical outputs by standard sum-based MPNNs. Consequently, these MPNNs fail on binary classification tasks for such graphs.

This paper aims to address this limitation by devising an MPNN whose embeddings preserve distances in the bi-Lipschitz sense. To state our goal formally, we first need to define a notion of distance on the input space of graphs.

\paragraph{WL metric for graphs}
WL metrics quantify the \emph{extent} to which two graphs are not WL-equivalent:
\begin{definition}[WL metric]
    \label{WL_metric}
    A \emph{WL metric on $\GNof\Omega$} is a function $\rho:\GNof\Omega \times \GNof\Omega \to \RR_{\geq 0}$ that satisfies the standard requirements for a metric, with the exception that 
    $\rho\of{G_1, G_2} =0 $ if and only if $G_1 \wleq G_2$.
\end{definition}

For convenience, we use the term WL metric, despite the fact that strictly speaking, WL metrics are \emph{pseudometrics} on $\GNd$.

%


\paragraph{Tree Mover's Distance}
The first WL metric we consider is the Tree Mover's Distance (TMD), defined in \citep{TMD}. This metric is based on building computation trees $\tree_v^{(t)}$, which simulate the WL procedure used to create the node features $h_v^{(t)}$, and calculating distances 
\begin{equation}\label{eq:TDshort}
\TD(\tree_v^{(t)},\tree_u^{(t)}) 
\end{equation}
recursively between sub-trees using optimal transport.
These node-level distances are zero if, and only if, the features $c_v^{(t)}$ and $c_u^{(t)}$, constructed by the WL test, are equal.  A graph-level distance $\tmd(G,G')$ is obtained by aggregating all node-level distances.  For the full definition, see \cref{app:TMD} and \cite{TMD}. Under the assumption that the feature domain $\Omega$ does not contain the zero vector,   \citet{TMD} proved that $\tmd(G,G') $ is a WL-metric. 

The second  WL metric we consider is the DS metric \citep{grohe2021logic}. Originally, this metric was defined only for featureless graphs of fixed cardinality. In the next section, we extend this metric to the more general case of $\GNd$.

\paragraph{Bi-Lipschitzness}
Once a WL-metric is defined to measure distances between graphs, one can bound the distortion incurred by a graph embedding with respect to that metric, using the notion of bi-Lipschitzness:
\begin{definition*}[Bi-Lipschitz embedding]
    Let $\rho$ be a WL-metric on $\GNof\Omega$. An embedding $\emb:\GNof\Omega\to\RR^m$ is said to be \emph{bi-Lipschitz with respect to $\rho$ on $\GNof\Omega$} if there exist constants $0<c\leq C < \infty$ such that $\forall G_1, G_2 \in \GNof\Omega$,
    \begin{equation}\label{eqdef_bilip}
    \begin{split}
    \scalebox{0.95}{$
        c \cdot \rho\of{G_1,G_2}
        \leq \norm{\emb\of{G_1}-\emb\of{G_2}}_2
        \leq C \cdot \rho\of{G_1,G_2}.$}
    \end{split}
    \end{equation}
    If $\emb$ only satisfies the left- or right-hand side of \cref{eqdef_bilip}, it is said to be \emph{lower-} or \emph{upper-Lipschitz} respectively.
\end{definition*}
%

Bi-Lipschitzness ensures that the embedding maps the input space $\GNof\Omega$ into the output Euclidean space with bounded distortion, with the ratio $\tfrac{C}{c}$ serving as an upper bound on the distortion, akin to the condition number of a matrix. 

\section{Main Contributions}
\label{sec_main_contributions}
In this section, we discuss our main contributions. We begin by defining our generalized DS metric for vertex-featured graphs. We then discuss our proposed MPNN and show that it is bi-Lipschitz with respect to the DS and TMD metrics.

\paragraph{The DS metric}
The DS metric originates from a relaxation of the \emph{graph isomorphism problem}. Two graphs $G$ and $\tilde G$, each with $n$ vertices, and corresponding adjacency matrices  $A$, $\tilde A$ are isomorphic if and only if there exists a \emph{permutation matrix} $P$ such that $AP=P\tilde A $. Since checking whether graphs are isomorphic is intractable, an approximate solution can be sought by considering the equation $AS=S\tilde A$ with $S \in \D_n$, the collection of $n \times n$ \emph{doubly stochastic} matrices, which is the convex hull of all permutation matrices.
Remarkably, this equation admits a solution $S\in\D_n$ if and only if the graphs $G$ and $\tilde G$ are WL-equivalent \citep{book}. Accordingly, a WL-metric between featureless graphs with the same number of vertices $n$ can be defined by the minimization problem
\begin{equation}
    \label{eqdef_metric_ds_nofeat}
    \rhods\of{G,\tilde G}
    =
    \min_{S\in\D_n} \ 
    \norm{AS-S\tilde A},
\end{equation}
where $\norm{\argdot}$ could denote any $p$ norm. The fact that this is indeed a pseudometric was established in \cite{bento}. The optimization problem \cref{eqdef_metric_ds_nofeat} can be solved by off-the-shelf convex optimization solvers and was considered as a method for finding the correspondence between two graphs in several papers, including \cite{aflalo,risk,dym2018exact,dym2017ds++,Bernard_2018_CVPR}. 

The idea of using the DS metric for MPNN stability analysis was introduced in \citep{grohe} and further discussed by \citet{boker2021graph}. To apply this idea to our setting, we need to adapt this metric to vertex-featured graphs with varying numbers of vertices. We do this by augmenting it as follows:
\begin{equation}\label{eqdef_metric_ds}
    \rhods\of{G,\tilde G}
    = 
    \abs{n-\tilde n}
    +
    \min_{S\in\Pi\of{n,\tilde n}}\left( \ 
    \norm{AS-S\tilde A}_1
     + 
    \sum_{i\in\brs{n}, j\in\brs{\tilde n}} S_{ij}
    \norm{\bx_i-\btx_j}_1 \right)
\end{equation}
where $n$ and $\tilde{n}$ denote the number of vertices in $G$ and $\tilde{G}$, $\bx_i$ and $\btx_j$ denote the vertex features of $G$ and $\tilde G$, and   $\Pi\of{n,\tilde n}$ is the set of $n \times \tilde n$ matrices $S$ with non-negative entries, whose rows and columns sum to $\frac{1}{n}$ and $\frac{1}{\tilde n}$, respectively. 
%

%
\begin{restatable}{theorem}{thmRhoDSIsWLMetric}
	\label{thm_rhods_is_wl_metric}
	\onmain{\textup{(Proof in \cref{subapp_proofs_ds_metric})\ }}
$\rhods$ is a WL-equivalent metric on $\GNd$.
\end{restatable}



%

%
\paragraph{Bi-Lipschitz MPNN} 
We now present our main contribution: a novel MPNN that is not only WL-equivalent, but also bi-Lipschitz, with respect to both $\rhods$ and TMD.

\paragraph{FSW Embedding} The core innovation in our MPNN lies in its message aggregation method. Specifically, we employ the \emph{Fourier Sliced-Wasserstein (FSW) Embedding}---a method for embedding multisets of vectors into Euclidean space, proposed by \citet{amir2024injective}, where it was also shown to be bi-Lipschitz. This property makes it plausible, a priori, that an MPNN based on FSW aggregation will be bi-Lipschitz for \emph{graphs}. In this work, we formally establish that this is indeed the case. We begin by describing the FSW embedding and then introduce our FSW-GNN architecture.

The FSW embedding maps an input multiset $ \bX = \brc{\bx_1,\ldots,\bx_n}$, where $\bx_1,\ldots,\bx_n \in \RR^d$, to an output vector $\bz = \br{z_1,\ldots,z_m} \in \RR^m$. 
It is denoted by
\begin{equation*}
\bz 
=
\embfsw\of{\bX; \br{\bv_k,\xi_k}_{k=1}^{m-1}},
\quad 
\bX = \brc{\bx_1,\ldots,\bx_n}.
\end{equation*}
In addition to the input multiset $\bX$, it depends on parameters $\br{\bv_k, \xi_k}_{k=1}^{m-1}$, where each $\bv_k \in \S^{d-1}$ represents a direction vector and $\xi_k \in \RR$ represents a frequency. The embedding is computed in three steps: First, the direction $\bv_k$ is used to project the original multiset of vectors to  a multiset of scalars 
$\brc{\bra{\bv_k,\bx_i}}_{i=1}^n$,
which is then sorted: $\by_k = \sort\of{\brc{\bra{\bv_k,\bx_i}}_{i=1}^n}$. This step is similar to the sort-type embedding used in SortMPNN, and was shown to be bi-Lipschitz on multisets of fixed size \citep{balan2022permutation}. However, taking $\by_k$ directly as the embedding leads to an output dimension dependent on the input multiset's size, thus making the embedding unsuitable for varying-size multisets. The next steps address and resolve this limitation.

In the second step, the vector $\by_k$ is identified with a step function $Q_{\by_k}: \brs{0,1}\to\RR$, namely the \emph{quantile function} of the multiset $\brc{\bra{\bv_k,\bx_i}}_{i=1}^n$; see illustration in \cref{fig_quantile_function}. Then, in the third step, the \emph{cosine transform}, a variant of the Fourier transform, is applied to $Q_{\by_k}$, and sampled at the given frequency $\xi_k$, to obtain the final output coordinates $z_k$, $k=1,\ldots,m-1$. This is summarized by:
%
\begin{wrapfigure}[12]{r}{0.3\linewidth}
    \centering
    \includegraphics[width=\linewidth]{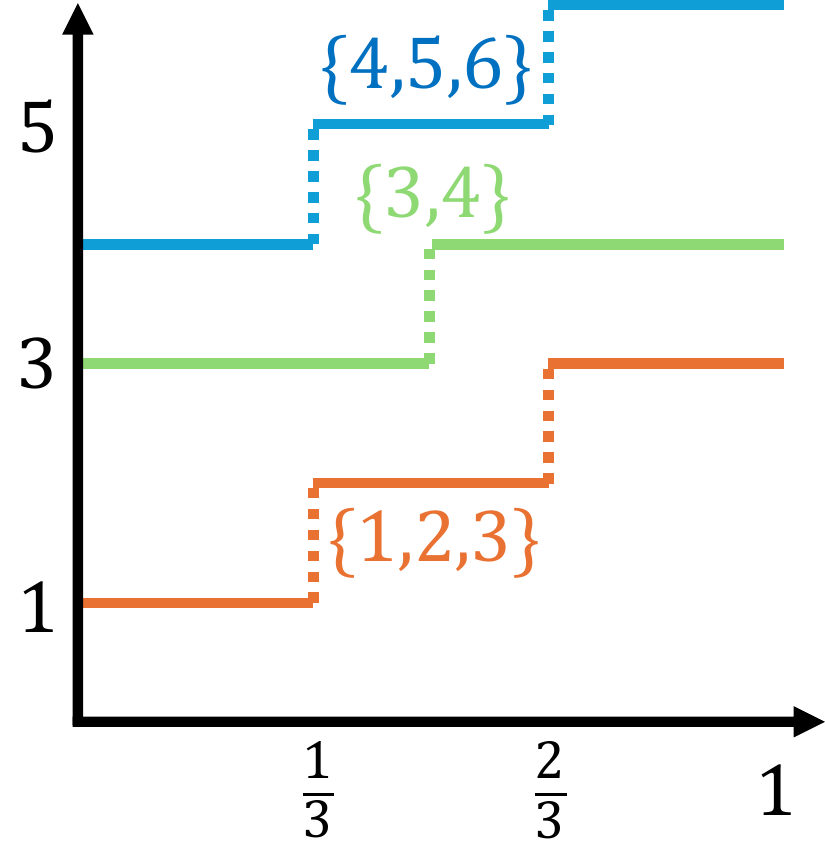}     \caption{\label{fig_quantile_function}The quantile function for three different multisets}      
\end{wrapfigure}
%
\begin{align}
\by_k
&=
\resizebox{!}{0.9\height}{$\br{y_{k1},\ldots,y_{kn}}$}
=
\mathrm{sort}\br{\resizebox{!}{0.85\height}{$\bra{\bv_k, \bx_1},\ldots, \bra{\bv_k, \bx_n}$} } \label{eq:sort}\\
Q_{\by_k}(t)
&=
%
%
y_{ki} \ \textup{with $i$ such that}\ t \in [\tfrac{i-1}{n},\tfrac{i}{n}) \label{eq:quantile}
\\
z_k&=2(1+\xi_k)\int_0^1 Q_{\by_k}(t) \cos(2\pi \xi_k t)dt \label{eq:cosine}
\end{align}

Lastly, note that \cref{eq:sort,eq:cosine,eq:quantile} treat input multisets as uniform distributions over their elements, and thus are agnostic to the multiset size by design. To address this and ensure that multisets of different sizes but identical element-proportions are mapped to distinct outputs, the last output coordinate $z_m$ is set to the cardinality $\abs{\bX}$ of the input multiset $\bX$. Further details appear in \citep[Appendix A.1]{amir2024injective}.

Note that the integral in \cref{eq:cosine} has a closed-form solution, and the whole embedding can be computed with a complexity of $\mathcal{O}\of{m n d + m n \log n}$, similarly to simple aggregation functions such as sum pooling. Furthermore, the embedding parameters and output dimension do not depend on $n$, making this method suitable for multisets of different sizes.
\paragraph{FSW-GNN} The FSW-GNN model processes input graphs $G=\br{V,E,X}$ by $T$ message-passing iterations: 
\begin{equation}\label{eqdef_fswgnn}
\begin{split}
    \bh^{\br{0}}_{v} \eqdefx & \bx_{v},
    %
    \\ \bq_v^{\br{t}} 
    \eqdefx &
    \embfswt[t]\of{ \setst{ \bh^{\br{t-1}}_u } {u \in \N_v} }, 
    \ \ \ 1 \leq t \leq T,
    \\ \bh^{\br{t}}_v
    \eqdefx &
    \Phi^{\br{t}} \of{ \brs{\bh^{\br{t-1}}_v;\bq^{\br{t}}_v}},
\end{split}
\end{equation}
where the functions $\embfswt[t]$ are all instances of the FSW embedding,
%
$\Phi^{\br{t}}$ are MLPs,
and $\brs{\bx; \by}$ denotes column-wise concatenation of column vectors $\bx$ and $\by$. Finally, a graph-level output is computed by:
\begin{equation}\label{eqdef_fswgnn_readout}
    \bh_G \eqdef 
    \Psi \circ
    \embfswglob\of{\setst{ \bh^{\br{T}}_v}{v\in V}},
\end{equation}
where, again, $\embfswglob$ is an FSW embedding, and $\Psi$ an MLP.

The following \lcnamecref{thm_fswgnn_is_wl_equivalent} shows that, with the appropriate choice of MLP sizes and number of iterations $T$, our proposed architecture is WL equivalent:

\begin{restatable}{theorem}{thmFSWGNNIsWLEquivalent}\label{thm_fswgnn_is_wl_equivalent}\onmain{\textup{(Proof in \cref{subapp_proofs_fswgnn_wl_equivalence})}}
    Consider the FSW-GNN architecture for input graphs in $\GNd$, with $T=N$ iterations, where $\Phi^{\br{t}}, \Psi$  are just linear functions, and all features (except for input features) are of dimension $m \geq 2Nd+2$. Then for almost every choice of model parameters, the graph embedding defined by the architecture is WL equivalent. 
\end{restatable}


%

The proof of \cref{thm_fswgnn_is_wl_equivalent} is based on the theory of $\sigma$-subanalytic functions and the \emph{Finite Witness Theorem}, introduced in \citep{amir2023neural}.

It is worth noting that the output dimension $m$ required in practice is typically considerably lower than the one required in \cref{thm_fswgnn_is_wl_equivalent}. This can be explained intuitively by the following fact: if all input graphs originate from a subset of $\GNd$ with intrinsic dimension $D$ that is lower than the ambient dimension $n\cdot d$, then it can be shown that $m=2D+2$ suffices for WL-equivalence.

\paragraph{From separation to bi-Lipschitzness}
In general, WL-equivalence does not imply bi-Lipschitzness. As mentioned above, sum-based MPNN can be injective but are never bi-Lipschitz. In contrast, we shall now prove that for FSW-GNN,
 WL-equivalence does imply bi-Lipschitzness, under the assumption that the feature domain $\Omega$ is compact:

\begin{restatable}{theorem}{thmFSWGNNIsBilip}\label{thm_fswgnn_is_bilip}
\onmain{\textup{(Proof in \cref{subapp_proofs_fswgnn_bilipschitz})}}
Let $\Omega \subset \RR^d$ be compact. 
Under the assumptions of
\cref{thm_fswgnn_is_wl_equivalent}, the FSW-GNN is bi-Lipschitz with respect to $\rhods$ on $\GNof\Omega$. If, additionally, $\Omega$ does not contain $\bzero$, then the FSW-GNN is bi-Lipschitz with respect to TMD on $\GNof\Omega$. 
\end{restatable}

We now give a high-level explanation of the proof idea. We rely on the following facts: (1) the output of FSW-GNN for an input graph $G=\br{V,E,X}$ is piecewise-linear with respect to the vertex-feature matrix $X$. This follows from properties of the FSW embedding functions $\embfswt[t]$ and $\embfswglob$ used in \cref{eqdef_fswgnn,eqdef_fswgnn_readout}. (2) both metrics $\rhods$ and TMD can be transformed, with bounded distortion, into metrics that are piecewise-linear, by choosing all the vector norms they employ to be the $\ell_1$ norm. The claim then follows from these observations and the following \lcnamecref{thm_pwl_ratio_bound}, which shows that two functions that are piecewise linear and have the same zero set, are bi-Lipschitz with respect to one another:
\begin{restatable}{lemma}{thmPWLRatioBound}
\label{thm_pwl_ratio_bound}
    \onmain{\textup{(Proof in \cref{subapp_proofs_fswgnn_bilipschitz})}}
    Let $f,g: M \to \RR_{\geq 0}$ be nonnegative piecewise-linear functions defined on a compact polygon $M \subset \RR^d$. Suppose that for all $x\in M$, $f(x) = 0$ if and only if $g(x)=0$. Then there exist real constants $c,C > 0$ such that
    \begin{equation}\label{eq_pwl_functions_ratio_bound}
        c \cdot g\of{x} \leq f\of{x} \leq C \cdot g\of{x}, \quad \forall x \in M.
    \end{equation}
\end{restatable}
We note that the assumption of a compact domain is essential; see Remark~6 in \cite{balan2024stability}. While \cref{thm_pwl_ratio_bound} is rather intuitive, we are not aware of it appearing previously in the literature. This lemma easily implies\footnotemark\ previous bi-Lipschitzness results in the literature, such as Theorem~1.4 in \citep{balan2023ginvariant}, Theorem~1 in \citep{balan2024stability}, and Theorem~3.10 in \citep{balan2022permutation}. We believe the lemma has the potential to serve as a valuable tool for proving bi-Lipschitzness results in additional domains in the future. To illustrate this, in Appendix \ref{app:subgraph} we show how the lemma can be used to construct a bi-Lipschitz \emph{subgraph aggregation MPNN}.

\footnotetext{In all these scenarios, the proposed embedding and metric are \emph{homogeneous}, so that bi-Lipschitzness on a compact polygon implies global bi-Lipschitzness.}

To complement \cref{thm_fswgnn_is_bilip}, we now present an analogous result for node-level tasks: we show that the node features computed by FSW-GNN are bi-Lipschitz with respect to the \emph{Tree Distance} discussed in \cref{eq:TDshort}. 
\begin{restatable}{theorem}{thmNodeLevelBiLip}\label{thm_node_level_bilip}
\onmain{\textup{(Proof in \cref{subapp_proofs_fswgnn_bilipschitz})}}
Suppose that $\Omega \subset \RR^d$ is a compact set that does not contain zero.
Under the assumptions of
\cref{thm_fswgnn_is_wl_equivalent}, the node features computed by FSW-GNN are bi-Lipschitz with respect to the Tree Distance metric on $\GNof\Omega$. 
\end{restatable}
%
\section[Oversmoothing, Oversquashing, and Bi-Lipschitz MPNN]
{Oversmoothing, Oversquashing, and Bi-Lipschitz\\ MPNN}
\label{sec:over}
Training \emph{deep} MPNNs is one of the core challenges in graph neural networks \citep{morris2024future}. The difficulty is often attributed to \emph{oversmoothing} \citep{Rusch2023ASO} or \emph{oversquashing} \citep{alon}. Both phenomena can be regarded as a form of metric distortion induced by the MPNN as the number of iterations  grows. Oversmoothing is the phenomenon where, for large $t$, 
\begin{equation}\label{eq:oversmoothing}
\norm{\bx_v^{(t)}- \bx_u^{(t)}} \approx 0, \quad \forall v,u\in V.
\end{equation}
Oversquashing, as defined in \cite{bron}, refers the phenomenon where a non-negligible change in an input node feature $\bx_v^{(0)}$ to a new value $\hat{\bx}_v^{(0)}$ results in only a negligible change 
in features of far-away nodes, namely, 
\begin{equation}\label{eq:oversquashing}
	\norm{ \bx_u^{(t)}- \hat{\bx}_u^{(t)} }  \ll \norm{ \bx_v^{(0)}- \hat{\bx}_v^{(0)} }, \text{ for } u \text{\ far from\ } v.
\end{equation}
In other words, information fails to propagate adequately between distant nodes through the message-passing process.

Accordingly, an MPNN that perfectly preserves distances between node features would provably not encounter oversmoothing or oversquashing. We note that FSW-GNN is not in this position: while it is bi-Lipschitz and thus has bounded distortion, we cannot rule out the possibility that its distortion grows with $t$. Nonetheless, we conjecture that FSW-GNN should perform much better than standard MPNNs on long range tasks. This is because, as shown in \cite{davidson2024h}, standard MPNNs are only lower H\"{o}lder in expectation, namely, the inequality
$$\|x_v^{(t)}- x_u^{(t)}\| \geq C_t \left[\TD(\tree_v^{(t)},\tree_u^{(t)}) \right]^{\alpha_t} $$ 
holds only in expectation, 
with constants $\frac{1}{C_t}$ and exponents $\alpha_t$ that grow with $t$. In contrast, for FSW-GNN we have $\alpha_t=1$ for all $t$, and only $\frac{1}{C_t}$ may grow with $t$. 
We provide empirical evidence for our conjecture in \cref{sec_numerical_experiments}, where we show that our bi-Lipschitz MPNN outperforms standard MPNNs on long-range tasks.
%
%

\section{Numerical Experiments}
\label{sec_numerical_experiments}
We compare the performance of FSW-GNN with standard MPNNs and Sort-MPNN on both real-world benchmarks and synthetic long-range tasks. 


\begin{figure*}
    \centering
    \begin{minipage}[b]{0.32\textwidth}
        \centering
\definecolor{darkgreen}{rgb}{0.0, 0.5, 0.0} 
\definecolor{brown}{rgb}{0.6, 0.4, 0.2} 
\definecolor{teal}{rgb}{0.0, 0.5, 0.5} 

\begin{tikzpicture}
\begin{axis}[
    legend to name=LegendOversquashing, 
    width=\linewidth, 
    height=0.85\linewidth, 
    xlabel={Number of MPNN Layers},
    xlabel style={font=\scriptsize},
    ylabel style={font=\scriptsize},
    xmin=1, xmax=16,
    ymin=10, ymax=110,
    legend style={
        at={(0.04,0.04)}, 
        anchor=south west, 
        font=\small,
        nodes={scale=1, transform shape}, 
        legend columns=6 
    },
    grid=major,
    grid style={opacity=0.5},
    tick label style={
        /pgf/number format/fixed, 
        /pgf/number format/precision=2, 
        /pgf/number format/zerofill=false, 
        font=\scriptsize,
    },
    xtick={2,4,6,8,10,12,14},
    ytick={0, 20, 40, 60, 80, 100},
    minor x tick num=1,
    tick style={
        major tick length=5pt, 
        minor tick length=2pt, 
    },
    extra x ticks={15},        
    extra x tick style={
        tick style={thin},
        major tick length=2pt, 
        grid=none, 
        tick label style={opacity=0} 
    },  
    yticklabel={\pgfmathprintnumber{\tick}\kern0.1em\%},
    scaled ticks=false, 
    enlargelimits=false, 
    axis line style={}, 
    clip=false 
]

\addplot[
    color=orange, 
    mark=square*, 
    mark size=2pt,
    mark options={scale=0.7},
    line width=1pt,
    solid]
    coordinates {
        (2,100)
        (3,100)
        (4,100)
        (5,100)
        (6,100)
        (7,100)
        (8,100)
        (9,100)
        (10,100)
        (11,100)
        (12,100)
        (13,100)
        (14,100)
        (15,100)
    };
\addlegendentry{FSW-GNN}

\addplot[
    color=red, 
    mark=triangle*, 
    mark size=3pt,
    mark options={scale=0.7},
    line width=1pt,
    dashed]
    coordinates {
        (2,100)
        (3,100)
        (4,100)
        (5,60)
        (6,30)
        (7,20)
        (8,20)
        (9,20)
        (10,20)
        (11,20)
        (12,20)
        (13,20)
        (14,20)
        (15,20)
    };
\addlegendentry{GIN}

\addplot[
    color=purple, 
    mark=diamond*, 
    mark size=3pt,
    mark options={scale=0.7},
    line width=1pt,
    dotted]
    coordinates {
        (2,100)
        (3,100)
        (4,100)
        (5,100)
        (6,100)
        (7,100)
        (8,80)
        (9,20)
        (10,20)
        (11,20)
        (12,20)
        (13,20)
        (14,20)
        (15,20)
    };
\addlegendentry{GAT}

\addplot[
    color=blue, 
    mark=*, 
    mark size=2pt,
    mark options={scale=0.7},
    line width=1pt,
    solid]
    coordinates {
        (2,100)
        (3,100)
        (4,100)
        (5,100)
        (6,100)
        (7,100)
        (8,100)
        (9,85)
        (10,70)
        (11,30)
        (12,25)
        (13,22)
        (14,20)
        (15,20)
    };
\addlegendentry{GCN}

\addplot[
    color=brown, 
    mark=square*, 
    mark size=2pt,
    mark options={scale=0.7},
    line width=1pt,
    dash dot]
    coordinates {
        (2,100.0)
        (3,100.0)
        (4,100.0)
        (5,100.0)
        (6,100.0)
        (7,100.0)
        (8,80.00000029802322)
        (9,20.000000298023224)
        (10,20.000000298023224)
        (11,20.000000298023224)
        (12,20.000000298023224)
        (13,20.000000298023224)
        (14,20.000000298023224)
        (15,20.000000298023224)
    };
\addlegendentry{Sort-MPNN}

\addplot[
    color=teal, 
    mark=star, 
    mark size=3pt,
    mark options={scale=0.7},
    line width=1pt,
    dash dot]
    coordinates {
        (2,100)
        (3,100)
        (4,100)
        (5,100)
        (6,100)
        (7,100)
        (8,22)
        (9,20)
        (10,22)
        (11,20)
        (12,20)
        (13,20)
        (14,20)
        (15,20)
    };
\addlegendentry{Graph-SAGE}

\end{axis}
\end{tikzpicture}
        \vspace{-10pt}
        \small (a) CliquePath
        \label{fig:cliquepath}
    \end{minipage}
    \hfill
    \begin{minipage}[b]{0.32\textwidth}
        \centering
\definecolor{darkgreen}{rgb}{0.0, 0.5, 0.0} 
\definecolor{brown}{rgb}{0.6, 0.4, 0.2} 
\definecolor{teal}{rgb}{0.0, 0.5, 0.5} 

\begin{tikzpicture}
\begin{axis}[
    legend to name=LegendRing, 
    width=\linewidth, 
    height=0.85\linewidth, 
    xlabel={Number of MPNN Layers},
    xlabel style={font=\scriptsize},
    ylabel style={font=\scriptsize},
    xmin=1, xmax=16,
    ymin=10, ymax=110,
    legend style={
        at={(0.04,0.04)}, 
        anchor=south west, 
        font=\small,
        nodes={scale=0.05, transform shape}, 
        legend columns=1 
    },
    grid=major,
    grid style={opacity=0.5},
    tick label style={
        /pgf/number format/fixed, 
        /pgf/number format/precision=2, 
        /pgf/number format/zerofill=false, 
        font=\scriptsize,
    },
    xtick={2,4,6,8,10,12,14},
    ytick={0, 20, 40, 60, 80, 100},
    minor x tick num=1,
    tick style={
        major tick length=5pt, 
        minor tick length=2pt, 
    },
    extra x ticks={15},        
    extra x tick style={
        tick style={thin},
        major tick length=2pt, 
        grid=none, 
        tick label style={opacity=0} 
    },  
    yticklabel={\pgfmathprintnumber{\tick}\kern0.1em\%},
    scaled ticks=false, 
    enlargelimits=false, 
    axis line style={}, 
    clip=false 
]

\addplot[
    color=orange, 
    mark=square*, 
    mark size=2pt,
    mark options={scale=0.7},
    line width=1pt,
    solid]
    coordinates {
        (15,100)
        (14,100)
        (13,100)
        (12,100)
        (11,100)
        (10,100)
        (9,100)
        (8,100)
        (7,100)
        (6,100)
        (5,100)
        (4,100)
        (3,100)
        (2,100)
    };
\addlegendentry{FSW-GNN}

\addplot[
    color=red, 
    mark=triangle*, 
    mark size=3pt,
    mark options={scale=0.7},
    line width=1pt,
    dashed]
    coordinates {
        (15,20)
        (14,20)
        (13,25)
        (12,55)
        (11,60)
        (10,20)
        (9,80)
        (8,85)
        (7,60)
        (6,80)
        (5,100)
        (4,100)
        (3,100)
        (2,100)
    };
\addlegendentry{GIN}

\addplot[
    color=purple, 
    mark=diamond*, 
    mark size=3pt,
    mark options={scale=0.7},
    line width=1pt,
    dotted]
    coordinates {
        (15,20)
        (14,20)
        (13,40)
        (12,50)
        (11,100)
        (10,100)
        (9,100)
        (8,100)
        (7,100)
        (6,100)
        (5,100)
        (4,100)
        (3,100)
        (2,100)
    };
\addlegendentry{GAT}

\addplot[
    color=blue, 
    mark=*, 
    mark size=2pt,
    mark options={scale=0.7},
    line width=1pt,
    solid]
    coordinates {
        (15,20)
        (14,20)
        (13,20)
        (12,20)
        (11,30)
        (10,80)
        (9,100)
        (8,100)
        (7,100)
        (6,100)
        (5,100)
        (4,100)
        (3,100)
        (2,100)
    };
\addlegendentry{GCN}

\addplot[
    color=brown, 
    mark=square*, 
    mark size=2pt,
    mark options={scale=0.7},
    line width=1pt,
    dash dot]
    coordinates {
        (15,20.000000298023224)
        (14,20.000000298023224)
        (13,20.000000298023224)
        (12,20.000000298023224)
        (11,20.000000298023224)
        (10,40.00000059604645)
        (9,100.0)
        (8,100.0)
        (7,100.0)
        (6,100.0)
        (5,100.0)
        (4,100.0)
        (3,100.0)
        (2,100.0)
    };
\addlegendentry{Sort-MPNN}

\addplot[
    color=teal, 
    mark=star, 
    mark size=3pt,
    mark options={scale=0.7},
    line width=1pt,
    dash dot]
    coordinates {
        (15,20)
        (14,20)
        (13,20)
        (12,20)
        (11,20)
        (10,20)
        (9,80)
        (8,100)
        (7,100)
        (6,100)
        (5,100)
        (4,100)
        (3,100)
        (2,100)
    };
\addlegendentry{SAGE}

\end{axis}
\end{tikzpicture}
        \vspace{-10pt}
        \small (b) Ring
        \label{fig:ring}
    \end{minipage}
    \hfill
    \begin{minipage}[b]{0.32\textwidth}
        \centering   
\definecolor{darkgreen}{rgb}{0.0, 0.5, 0.0} 
\definecolor{brown}{rgb}{0.6, 0.4, 0.2} 
\definecolor{teal}{rgb}{0.0, 0.5, 0.5} 

\begin{tikzpicture}
\begin{axis}[
    legend to name=LegendCrossRing, 
    width=\linewidth, 
    height=0.85\linewidth, 
    xlabel={Number of MPNN Layers},
    xlabel style={font=\scriptsize},
    ylabel style={font=\scriptsize},
    xmin=1, xmax=16,
    ymin=10, ymax=110,
    legend style={
        at={(0.04,0.04)}, 
        anchor=south west, 
        font=\small,
        nodes={scale=0.05, transform shape}, 
        legend columns=1 
    },
    grid=major,
    grid style={opacity=0.5},
    tick label style={
        /pgf/number format/fixed, 
        /pgf/number format/precision=2, 
        /pgf/number format/zerofill=false, 
        font=\scriptsize,
    },
    xtick={2,4,6,8,10,12,14},
    ytick={0, 20, 40, 60, 80, 100},
    minor x tick num=1,
    tick style={
        major tick length=5pt, 
        minor tick length=2pt, 
    },
    extra x ticks={15},        
    extra x tick style={
        tick style={thin},
        major tick length=2pt, 
        grid=none, 
        tick label style={opacity=0} 
    },  
    yticklabel={\pgfmathprintnumber{\tick}\kern0.1em\%},
    scaled ticks=false, 
    enlargelimits=false, 
    axis line style={}, 
    clip=false 
]

\addplot[
    color=orange, 
    mark=square*, 
    mark size=2pt,
    mark options={scale=0.7},
    line width=1pt,
    solid]
    coordinates {
        (2,100)
        (3,100)
        (4,100)
        (5,100)
        (6,100)
        (7,100)
        (8,100)
        (9,100)
        (10,100)
        (11,100)
        (12,100)
        (13,100)
        (14,100)
        (15,100)
    };
\addlegendentry{FSW-GNN}

\addplot[
    color=red, 
    mark=triangle*, 
    mark size=3pt,
    mark options={scale=0.7},
    line width=1pt,
    dashed]
    coordinates {
        (2,100)
        (3,100)
        (4,100)
        (5,100)
        (6,100)
        (7,100)
        (8,100)
        (9,60)
        (10,100)
        (11,90)
        (12,50)
        (13,40)
        (14,20)
        (15,20)
    };
\addlegendentry{GIN}

\addplot[
    color=purple, 
    mark=diamond*, 
    mark size=3pt,
    mark options={scale=0.7},
    line width=1pt,
    dotted]
    coordinates {
        (2,100)
        (3,100)
        (4,100)
        (5,100)
        (6,100)
        (7,100)
        (8,100)
        (9,100)
        (10,100)
        (11,40)
        (12,20)
        (13,20)
        (14,20)
        (15,20)
    };
\addlegendentry{GAT}

\addplot[
    color=blue, 
    mark=*, 
    mark size=2pt,
    mark options={scale=0.7},
    line width=1pt,
    solid]
    coordinates {
        (2,100)
        (3,100)
        (4,100)
        (5,100)
        (6,100)
        (7,100)
        (8,100)
        (9,100)
        (10,100)
        (11,20)
        (12,20)
        (13,20)
        (14,20)
        (15,20)
    };
\addlegendentry{GCN}

\addplot[
    color=brown, 
    mark=square*, 
    mark size=2pt,
    mark options={scale=0.7},
    line width=1pt,
    dash dot]
    coordinates {
        (2,100.0)
        (3,100.0)
        (4,100.0)
        (5,100.0)
        (6,100.0)
        (7,100.0)
        (8,100.0)
        (9,40.0)
        (10,40.0)
        (11,20.0)
        (12,20.0)
        (13,20.0)
        (14,20.0)
        (15,20.0)
    };
\addlegendentry{Sort-MPNN}

\addplot[
    color=teal, 
    mark=star, 
    mark size=3pt,
    mark options={scale=0.7},
    line width=1pt,
    dash dot]
    coordinates {
        (2,100)
        (3,100)
        (4,100)
        (5,100)
        (6,100)
        (7,100)
        (8,100)
        (9,85)
        (10,20)
        (11,20)
        (12,20)
        (13,20)
        (14,20)
        (15,20)
    };
\addlegendentry{SAGE}

\end{axis}
\end{tikzpicture}
        \vspace{-10pt}
        \small (c) CrossRing
        \label{fig:crossring}
    \end{minipage}
    %
    
    \vspace{5pt}
    \pgfplotslegendfromname{LegendOversquashing}
    \caption{\small Performance comparison of MPNN models across the CliquePath, Ring, and CrossRing graph transfer tasks as presented in \cite{bron}.}
    \label{fig:comparison}
\end{figure*}
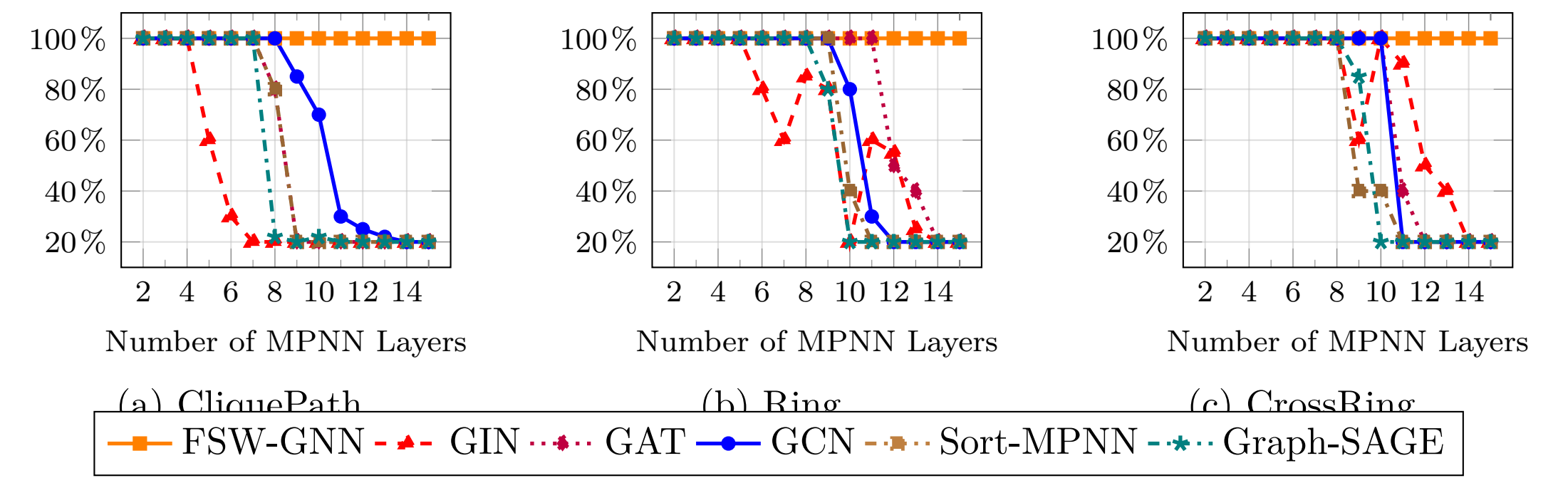

\paragraph{Empirical distortion evaluation}
%
\begin{figure*}[t]
    \centering
    \begin{minipage}{0.45\textwidth}
        \centering
\definecolor{darkgreen}{rgb}{0.0, 0.5, 0.0} 

\begin{tikzpicture}
\begin{axis}[
    width=\linewidth, 
    height=0.85\linewidth, 
    xlabel={Number of MPNN iterations},
    ylabel={Distortion},
    xlabel style={font=\small},
    ylabel style={font=\small},
    legend style={
        at={(0.75,0.125)}, 
        anchor=south, 
        font=\small,
        nodes={scale=0.75, transform shape}, 
        legend columns=1 
    },
    xmin=0.5, xmax=9,
    ymin=0.1, ymax=1000000000000000,
    grid=both,
    grid style={opacity=0.5},
    ymode=log, 
    log basis y=10,
    xtick=data,
    ytick={1, 100, 10000, 1000000, 100000000, 10000000000, 1000000000000, 100000000000000},
    yticklabels={$1$, $10^2$, $10^4$, $10^6$, $10^8$, $10^{10}$, $10^{12}$, $10^{14}$},
    tick label style={/pgf/number format/fixed}
]

\addplot[
    color=blue, 
    mark=*, 
    mark size=2pt, 
    line width=1pt, 
    solid 
] coordinates {
        (1,12638093.039936073)
        (2,36277547773546.97)
        (3,21264609154794.703)
        (4,22973356464791.957)
        (5,15057111489491.791)
        (6,15648307286362.354)
        (7,8043080153472.388)
        (8,17840817430754.05)
        (9,9681136511164.129)
    };
\addlegendentry{GCN}

\addplot[
    color=red, 
    mark=triangle*, 
    mark size=3pt, 
    line width=1pt, 
    dashed 
] coordinates {
        (1,39113311332096.984)
        (2,10076621645407.969)
        (3,2996697271242.111)
        (4,1585106348852.8674)
        (5,3351681634641.4043)
        (6,1113652984109.949)
        (7,686299729795.78)
        (8,477337721718.20074)
        (9,169431990829.66888)
    };
\addlegendentry{GIN}

\addplot[
    color=purple, 
    mark=diamond*, 
    mark size=3pt, 
    line width=1pt, 
    dotted 
] coordinates {
        (1,448641667.3591167)
        (2,219778059.42747307)
        (3,175375299.03847378)
        (4,98356652.1972862)
        (5,79738332.28032017)
        (6,158271843.73325607)
        (7,113370485.23310857)
        (8,268947205.2025125)
        (9,43110305.69958279)
    };
\addlegendentry{GAT}

\addplot[
    color=orange, 
    mark=square*, 
    mark size=2pt, 
    line width=1pt, 
    dash dot 
] coordinates {
        (1,882.5296361486475)
        (2,1683539.3638225684)
        (3,9097933.48438621)
        (4,11246181.00008503)
        (5,12575413.39619732)
        (6,12594059.93482434)
        (7,12538425.462116275)
        (8,14015738.771661198)
        (9,13848807.118490659)
    };
\addlegendentry{Sort-MPNN}

\addplot[
    color=darkgreen, 
    mark=square*, 
    mark size=2pt, 
    line width=1pt, 
    solid 
] coordinates {
        (1,1.6564223455271248)
        (2,1.7647409190382553)
        (3,1.9515738503254876)
        (4,2.230946474760214)
        (5,2.2427286882117063)
        (6,2.3206199577695625)
        (7,2.513434676666042)
        (8,2.5251651897621628)
        (9,2.622899074399135)
    };
\addlegendentry{FSW-GNN}

\end{axis}
\end{tikzpicture}
        \caption{\centering Empirical distortion evaluation with respect to the Doubly-Stochastic (DS) metric}
        \label{fig_empirical_distortion_ds}
    \end{minipage}
    \hfill
    \begin{minipage}{0.45\textwidth}
        \centering
        \vspace{-8pt}
\definecolor{darkgreen}{rgb}{0.0, 0.5, 0.0} 
\definecolor{brown}{rgb}{0.6, 0.4, 0.2} 
\definecolor{teal}{rgb}{0.0, 0.5, 0.5} 

\begin{tikzpicture}
\begin{axis}[
    width=\linewidth, 
    height=0.85\linewidth, 
    xlabel={Number of MPNN iterations},
    ylabel={MAD energy},
    xlabel style={font=\small},
    ylabel style={font=\small},
    ymin=-0.05, ymax=0.5,
    legend style={
        at={(0.98,0.98)}, 
        anchor=north east, 
        font=\small,
        nodes={scale=0.93, transform shape}, 
        legend columns=2 
    },
    grid=both,
    grid style={opacity=0.5},
    tick label style={
        /pgf/number format/fixed, 
        /pgf/number format/precision=2, 
        /pgf/number format/zerofill=false 
    },
    scaled ticks=false 
]

\addplot[
    color=orange, 
    mark=square*, 
    mark size=2pt,
    line width=1pt,
    solid]
    coordinates {
        (1,0.42338799238204955)
        (2,0.3808883249759674)
        (3,0.35066452622413635)
        (4,0.3483756065368652)
        (5,0.345891010761261)
        (6,0.32809064984321595)
        (7,0.32408376336097716)
        (8,0.2752107441425323)
        (9,0.29241743087768557)
        (10,0.30117576122283934)
        (11,0.26300403475761414)
        (12,0.25773012042045595)
        (13,0.292334121465683)
        (14,0.24901258945465088)
    };
\addlegendentry{FSW-GNN}

\addplot[
    color=red, 
    mark=triangle*, 
    mark size=3pt,
    line width=1pt,
    dashed]
    coordinates {
        (1,0.3294026255607605)
        (2,0.2805932998657227)
        (3,0.2677850604057312)
        (4,0.22886189818382263)
        (5,0.20339438915252686)
        (6,0.19521003365516662)
        (7,0.18596392273902893)
        (8,0.17703982889652253)
        (9,0.16854530274868013)
        (10,0.15637784004211425)
        (11,0.1473350077867508)
        (12,0.144667911529541)
        (13,0.1307902991771698)
        (14,0.1273142069578171)
    };
\addlegendentry{GIN}

\addplot[
    color=teal, 
    mark=star, 
    mark size=3pt,
    line width=1pt,
    dash dot]
    coordinates {
        (1,0.14689984023571015)
        (2,0.05464031249284744)
        (3,0.025625161826610565)
        (4,0.02811870090663433)
        (5,0.011014155112206936)
        (6,0.004303085897117853)
        (7,0.0032731398940086366)
        (8,0.001661011902615428)
        (9,0.0014552110340446233)
        (10,0.0007656027097254992)
        (11,0.0004141142126172781)
        (12,0.0002915089018642902)
        (13,0.00018575575668364764)
        (14,0.00016290307976305485)
    };
\addlegendentry{Graph-SAGE}

\addplot[
    color=purple, 
    mark=diamond*, 
    mark size=3pt,
    line width=1pt,
    dotted]
    coordinates {
        (1,0.0888837456703186)
        (2,0.03975565209984779)
        (3,0.025218838453292848)
        (4,0.011862923018634319)
        (5,0.0012912253616377712)
        (6,0.00046252908650785686)
        (7,0.0003909442573785782)
        (8,9.717643406474963e-05)
        (9,6.694108305964619e-05)
        (10,3.220222279196605e-05)
        (11,1.3201932597439736e-05)
        (12,1.4777596697967965e-05)
        (13,5.527479606826091e-06)
        (14,5.419056014943635e-06)
    };
\addlegendentry{GAT}

\addplot[
    color=blue, 
    mark=*, 
    mark size=2pt,
    line width=1pt,
    solid]
    coordinates {
        (1,0.07149463295936584)
        (2,0.044893496483564374)
        (3,0.005825106054544449)
        (4,0.003122424567118287)
        (5,0.000672326993662864)
        (6,0.0013081908924505114)
        (7,0.0003982309252023697)
        (8,0.0002689235785510391)
        (9,0.00015448511112481356)
        (10,0.00012930767406942322)
        (11,5.556717660510912e-05)
        (12,3.080964233959094e-05)
        (13,2.2826025815447793e-05)
        (14,9.54349889070727e-06)
    };
\addlegendentry{GCN}

\end{axis}
\end{tikzpicture}
        \vspace{8pt}
        \caption{MAD Energy vs. Number of MPNN layers for various models on the Ring long-range task}
        \label{fig:oversmoothing}
    \end{minipage}        
\end{figure*}

First, to assess the metric distortion induced by different MPNNs, we compared the distances induced by each MPNN vs. the TMD and DS metric on a particularly challenging set of graph pairs; see \cref{app_numerical_experiments} for details. Empirical estimates $\hat C, \hat c$ of the constants $C$,$c$ of \cref{eqdef_bilip} were computed, and the distortion estimate was taken as the ratio $\hat{C}/\hat{c}$. The results appear in \cref{fig_empirical_distortion_ds}. Similar results were obtained for the TMD; see \cref{fig_empirical_distortion_tmd} in the appendix. As seen in the figures, our method yields considerably lower distortion than all competitors, which aligns with our theoretical guarantees.



\paragraph{Transductive Learning}
Next, we compare FSW-GNN with GCN, GAT and Sort-MPNN for nine transductive learning tasks taken from \cite{geom-gcn}. As shown in  \cref{tab:performance_comparison}, FSW-GNN outperforms the competition in six out of nine tasks. SortMPNN is a clear winner in two of the other graphs, both of which have relatively large average degrees (see Table \ref{tab:common_graph_stats}). In addition, we include the state-of-the-art results known to us for each dataset: For Cora \cite{best_cora}; for Citeseer \cite{best_cite}; for PubMed \cite{best_pubm}; for Chameleon \cite{best_cham}; for Squirrel \cite{best_squirrel}; for Actor \cite{best_actor}; for Cornell \cite{best_corn}; for Texas \cite{best_texas}; for Wisconsin \cite{best_wisc}. These results are typically achieved by models that are strictly more powerful and computationally expensive than any MPNN.


\paragraph{Graph classification and regression}
\vspace{-0.5em} 
In the appendix, we include results on standard graph classification and regression tasks: the peptides-func and peptides-struct datasets of the LRGB benchmark \cite{lrgb2022}, and the MUTAG  and Protein datasets \cite{TUD}. These results appear in \cref{tab:lrgb_datasets}
and  \cref{tab:mutag}.




\paragraph{Long range tasks} Next, we consider several synthetic long-range tasks proposed in the oversquashing literature \cite{alon,bron}. In these problems, one can control the problem radius $r$, which is the number of MPNN iterations needed to solve the problem. 

We first consider the NeighborsMatch problem from
\cite{alon}.  In \cref{fig:Trees}, we compare the performance of  FSW-GNN with standard MPNN on the NeighborsMatch problem with $2 \leq r \leq 8$. Our FSW-GNN achieves perfect accuracy for all values of $r$, while Sort-MPNN fails at $r=8$ and the other competing methods falter for $r\geq 6$. 

Next, we consider the ‘graph transfer’ tasks \citep{bron}, in which a ‘source’ node feature is propagated to a ‘target’ node that is $r$ steps away. All other nodes have ‘blank’ features. Here we used three graph topologies proposed by \cite{bron}: clique, ring, and crossring, with problem radii $r$ varying from $2$ to $15$.

As shown in \cref{fig:comparison}, FSW-GNN is the only method to achieve $100\%$ accuracy across all three graphs and radii. Other models start failing at much smaller $r$, which depends on the graph topology. We do find that our model's performance  deteriorates when $r>20$.

Finally, we evaluate the relation between our empirical results and oversmoothing on the ring experiment, using the Mean Average Distance (MAD) \cite{Rusch2023ASO}. 
As seen in \cref{fig:oversmoothing}, all methods except for FSW-GNN and GIN exhibit oversmoothing starting from some $r$.

We note that long-range issues can be alleviated by graph rewiring methods, which effectively reduce problem radii \cite{gutteridge2023drew}, or by adding global information using spectral filters \cite{best_func_model} or graph transformers. A key distinction of FSW-GNN is that it performs well \emph{without modifying} the original graph topology. Thus, this work directly addresses the problem of relaying long-range messages, rather than circumventing it by architectural modifications.

\nd{Thinks to address about figs, if possible: (a) Table 2 text is too small. (b) Figure 3 I think tal was correct that we should send one half of it to an appendix, and discuss there why we get an essentially identical figure. (I think I understand why this could make sense)}

\section{Conclusion}
\label{sec_conclusion}
In this paper, we introduced FSW-GNN, the first bi-Lipschitz MPNN. Empirically, we found that FSW-GNN is highly effective, particularly for long-range problems, and appears to mitigate oversmoothing and oversquashing due to its inherent ability to preserve graph metrics.

A slight drawback of FSW-GNN is that its runtime is somewhat higher than that of standard MPNNs; see \cref{comparison_time_fswgnn}. An interesting direction for future work is to obtain quantitative estimates of its bi-Lipschitz constants.


\newpage

\clearpage

\newpage
\inappendixtrue
\section*{Appendix}
\phantomsection
\addcontentsline{toc}{section}{Appendix}
\addtocontents{toc}{\protect\setcounter{tocdepth}{0}}

\setcounter{section}{0}
\renewcommand{\thesection}{\Alph{section}}
\renewcommand{\theHsection}{appendix.\thechapter.\Alph{section}}

\begin{table}[H]
    \caption{Performance of different models on the LRGB results.}
    \label{tab:lrgb_datasets}
    \centering
    \begin{tabular}{lcc}
        \toprule
        Dataset & peptides-func (AP$\uparrow$) & peptides-struct (MAE$\downarrow$) \\ \hline
        GINE & 0.6621$\pm$0.0067 & \underline{0.2473$\pm$0.0017} \\
        GCN & 0.6860$\pm$0.0050 & \textbf{0.2460$\pm$0.0007} \\
        GatedGCN & 0.6765$\pm$0.0047 & 0.2477$\pm$0.0009 \\
        SortMPNN & \textbf{0.6914$\pm$0.0056} & 0.2494$\pm$0.0021 \\
        FSW-GNN  & \underline{0.6864$\pm$0.0048} & 0.2489$\pm$0.00155 \\ \hline
        SOTA & \textbf{0.73} & \textbf{0.242} \\ \hline
    \end{tabular}
\end{table}
\begin{table}[H]
\caption{Performance of different models on the MUTAG and Protein datasets.}
\label{tab:mutag}
\centering
\footnotesize
\setlength{\tabcolsep}{5pt} 
\renewcommand{\arraystretch}{0.95} 
\begin{tabular}{@{}lll@{}}
\toprule
\textbf{Model} & \textbf{MUTAG} & \textbf{Protein} \\
\midrule
GIN {\scriptsize\citep{gin}} & $89.4 \pm 5.6$ & $76.2 \pm 2.8$ \\
GCN & $85.6 \pm 5.8$ & $76.0 \pm 3.2$ \\
GraphSAGE & $85.1 \pm 7.6$ & $75.9 \pm 3.2$ \\
SortMPNN & $\mathbf{90.99 \pm 6.2}$ & \underline{$76.46 \pm 3.68$} \\
AdaptMPNN & $90.41 \pm 6.1$ & $75.12 \pm 3.64$ \\
FSW-GNN & \underline{$90.55 \pm 6.1$} & $\mathbf{76.93 \pm 7.64}$ \\
\bottomrule
SOTA & \underline{$96.66 \pm 1.23$} & $\mathbf{84.91 \pm 1.62}$ \\
\bottomrule
\end{tabular}
\end{table}
\section{Statistics on our benchmarks}
We present some statistics of the learning problems considered in the main text. In Table \ref{tab:common_graph_stats}, we present the statistics for each transductive dataset, and for the MUTAG dataset, including the number of nodes, edges, features, classes, average degree, and density, measuring the number of edges divided by the number of maximal edges. As we can see in the table, these datasets are very sparse.

Table \ref{tab:peptides_graph_stats} shows the same statistics for  the Peptides tasks from the  LRGB dataset.

\begin{table}[h!]
\caption{Graph statistics for transductive learning and MUTAG.}
\label{tab:common_graph_stats}
\centering
\resizebox{\textwidth}{!}{%
\begin{tabular}{@{}lcccccccccc@{}}
\toprule
\textbf{Dataset}     & \textbf{Cora} & \textbf{Cite.} & \textbf{Pubm.} & \textbf{Cham.} & \textbf{Squi.} & \textbf{Actor} & \textbf{Corn.} & \textbf{Texa.} & \textbf{Wisc.} & \textbf{MUTAG} \\ \midrule
\# Nodes             & 2708          & 3327           & 19717          & 2277           & 5201           & 7600           & 183            & 183            & 251            & 17.93            \\
\# Edges             & 5429          & 4732           & 44338          & 36101          & 217073         & 33544          & 295            & 309            & 499            & 19.79            \\
\# Features          & 1433          & 3703           & 500            & 2325           & 2089           & 931            & 1703           & 1703           & 1703           & 7              \\
\# Classes           & 7             & 6              & 3              & 5              & 5              & 5              & 5              & 5              & 5              & 2              \\
\textbf{Avg. Degree} & 4             & 2              & 4              & 31             & 83             & 8              & 3              & 3              & 4              & 8              \\
\textbf{Density}     & 0.0007        & 0.0009         & 0.0001         & 0.0159         & 0.0161         & 0.0012         & 0.0177         & 0.0184         & 0.0041         & 0.0422         \\ \bottomrule
\end{tabular}
}
\end{table}

\begin{table}[h!]
\caption{Graph statistics for Peptides datasets.}
\label{tab:peptides_graph_stats}
\centering
\begin{tabular}{@{}lcc@{}}
\toprule
\textbf{Dataset}       & \textbf{Peptides-Func} & \textbf{Peptides-Struct} \\ \midrule
\# Graphs              & 15,535                & 15,535                  \\
\# Nodes (Avg.)        & 150.94                & 150.94                  \\
\# Edges (Avg.)        & 2.04                  & 2.04                    \\
\textbf{Avg. Degree}   & 2.04                  & 2.04                    \\
\textbf{Density}       & \(1.74 \times 10^{-6}\) & \(1.74 \times 10^{-6}\) \\
\# Classes             & -                     & 10                      \\ \bottomrule
\end{tabular}
\end{table}
\section{Implementation details}
For the Transductive learning and the LRGB dataset, we looked for each learning rate in the set $1e^{-3},5e^{-3}$ and weight decay in the set $0.0,1e^{-5}$ and learning rate factor of 0.7. For each such configuration we stop according to the validation and take the best configuration of the validation set. For all Over-squashing tasks we took a learning rate of $0.001$ and weight decay $0$, and learning rate factor of 0.7. In all datasets we took the original dataset splits to train, test and validation sets.

\section{Relation to Tree Mover's Distance}\label{app:TMD}
%
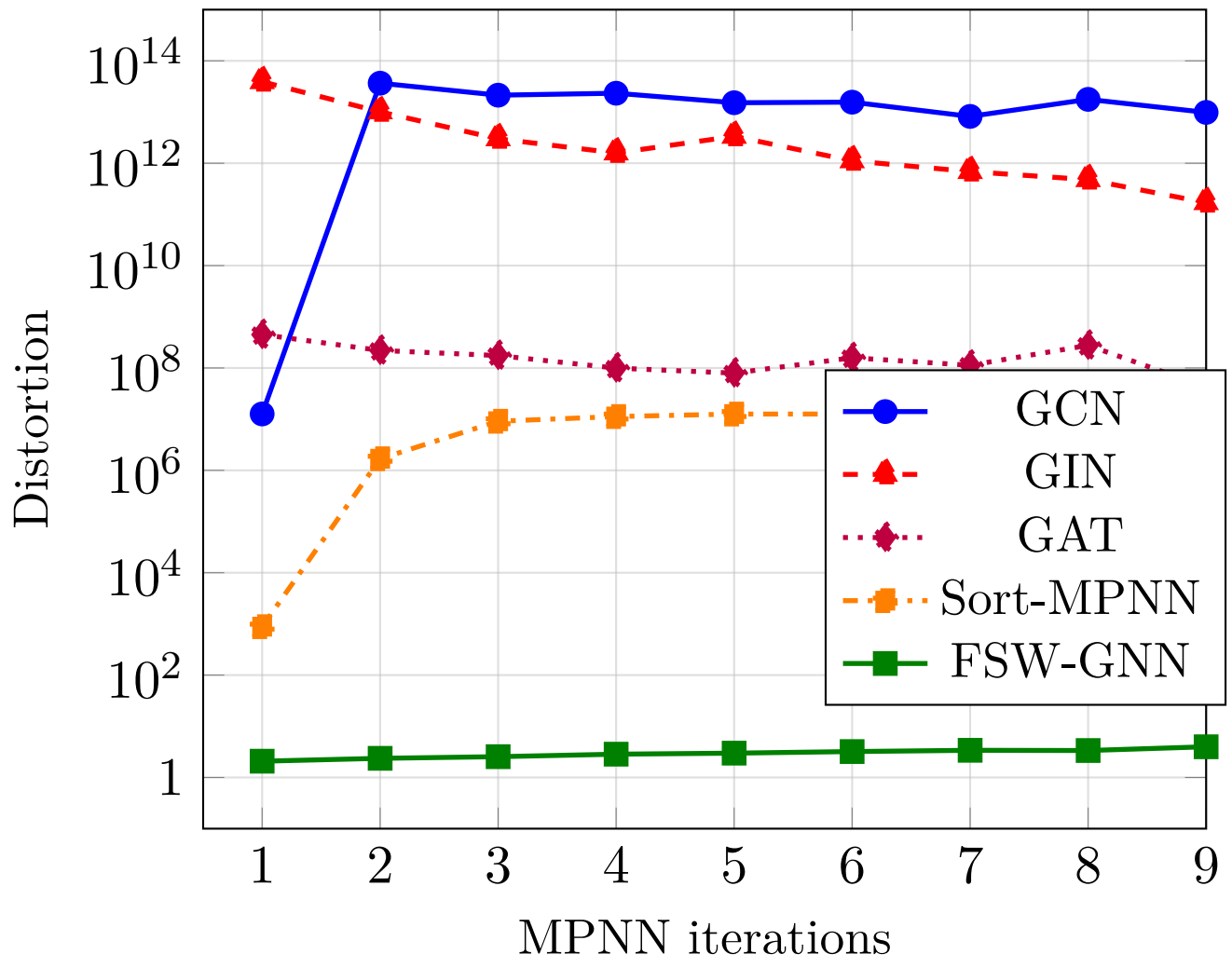
\begin{figure*}[ht]
    \centering
    \begin{minipage}{0.6\linewidth}
        \centering
\definecolor{darkgreen}{rgb}{0.0, 0.5, 0.0} 

\begin{tikzpicture}
\begin{axis}[
    width=\linewidth, 
    height=0.85\linewidth, 
    xlabel={MPNN iterations},
    ylabel={Distortion},
    xlabel style={font=\small},
    ylabel style={font=\small},
    legend style={
        at={(0.82,0.15)}, 
        anchor=south, 
        font=\small,
        nodes={scale=1, transform shape}, 
        legend columns=1 
    },
    xmin=0.5, xmax=9,
    ymin=0.1, ymax=1000000000000000,
    grid=both,
    grid style={opacity=0.5},
    ymode=log, 
    log basis y=10,
    xtick=data,
    ytick={1, 100, 10000, 1000000, 100000000, 10000000000, 1000000000000, 100000000000000},
    yticklabels={$1$, $10^2$, $10^4$, $10^6$, $10^8$, $10^{10}$, $10^{12}$, $10^{14}$},
    tick label style={/pgf/number format/fixed}
]

\addplot[
    color=blue, 
    mark=*, 
    mark size=2pt, 
    line width=1pt, 
    solid 
] coordinates {
        (1,12638093.044023262)
        (2,36596146788997.96)
        (3,21342411339985.98)
        (4,23295034986163.297)
        (5,15191815626010.543)
        (6,15582168275756.059)
        (7,8209843063968.8)
        (8,17679456368908.965)
        (9,9829893133962.688)
    };
\addlegendentry{GCN}

\addplot[
    color=red, 
    mark=triangle*, 
    mark size=3pt, 
    line width=1pt, 
    dashed 
] coordinates {
        (1,39113311332097.03)
        (2,10076621645407.967)
        (3,2996697271242.1055)
        (4,1585106348852.862)
        (5,3351681634641.4062)
        (6,1113652984109.948)
        (7,686299729795.7797)
        (8,477337721718.2003)
        (9,169431990829.6688)
    };
\addlegendentry{GIN}

\addplot[
    color=purple, 
    mark=diamond*, 
    mark size=3pt, 
    line width=1pt, 
    dotted 
] coordinates {
        (1,448641667.32040673)
        (2,219847796.7868665)
        (3,174747129.25470394)
        (4,99340914.06196953)
        (5,79364615.09907597)
        (6,158263693.97921783)
        (7,112927918.7426942)
        (8,276225665.5825091)
        (9,42500354.08787507)
    };
\addlegendentry{GAT}

\addplot[
    color=orange, 
    mark=square*, 
    mark size=2pt, 
    line width=1pt, 
    dash dot 
] coordinates {
        (1,882.5296363001007)
        (2,1683539.3644449934)
        (3,9097933.487546248)
        (4,11246181.004833749)
        (5,12575413.400833698)
        (6,12594059.93961346)
        (7,12538425.465092795)
        (8,14015738.776621256)
        (9,13848807.123872679)
    };
\addlegendentry{Sort-MPNN}

\addplot[
    color=darkgreen, 
    mark=square*, 
    mark size=2pt, 
    line width=1pt, 
    solid 
] coordinates {
        (1,2.078650265843728)
        (2,2.3656003540728796)
        (3,2.5477499395702012)
        (4,2.851521166123101)
        (5,2.9864821502300796)
        (6,3.2084230973056034)
        (7,3.3943916605178064)
        (8,3.3606438447809266)
        (9,3.979510453215072)
    };
\addlegendentry{FSW-GNN}

\end{axis}
\end{tikzpicture}
    \end{minipage}
    \caption{Empirical distortion evaluation with respect to the Tree Mover's Distance (TMD)}
    \label{fig_empirical_distortion_tmd}
\end{figure*}


In this appendix we give a full description of 
the Tree Mover's Distance
as defined in \citet{TMD}.

We first review Wasserstein distances. Recall that if $(X,d) $ is a metric space, $\Omega \subseteq X$ is a subset,  then the Wasserstein distance can be defined on the space  of multisets consisting of $n$ elements in $\Omega$ via
\[
{W}_{d}( x_1,\ldots,x_n, y_1,\ldots,y_n) =\min_{\tau \in S_n} \sum_{j=1}^n d(x_j,y_{\tau(j)})
\]
For multisets of different size, the authors of \citep{TMD} used an augmentation map, which, for a fixed parameter $n$, augments multisets of size $r \leq n$ by padding with $n-r$ instances of an element $z$ in $X\setminus \Omega $, namely
$$\Gamma^z\left( x_1,\ldots,x_r \right)= (x_1,\ldots,x_r,x_{r+1}=z,\ldots.x_n=z)$$
and the augmented distance on multi-sets of size up to $n$ is defined by
$$W_{d}(X,\hat X)=W_{d}^0\br{\Gamma^z\of X, \Gamma^z\of {\hat X } }.$$ 

We now return to define the TMD. We consider the space of graphs $\GNof\Omega$, consisting of graphs with $\leq N$ nodes, with node features coming from a compact domain $\Omega \subseteq \RR^d$ such that $0\notin\Omega$. The TMD is defined using the notion of computation trees:
\begin{definition}
    (Computation Trees). Given a graph $\G=(V,E,X)$ with node features $\brc{ x_v }_{v\in V} $, let $\tree_v^1$ be the rooted tree with a single node $v$, which is also the root of the tree, and node features $x_v$. For $t \in \NN$  let $\tree_v^t$ be the depth-$t$ computation tree of node $v$ constructed by connecting the neighbors of the leaf nodes of $\tree_v^{t-1}$ to the tree. Each node is assigned the same node feature it had in the original graph $\G$. The multiset of depth-$t$ computation trees defined by $\G$ is denoted by $\mathcal{T}_G^K:=\brc{ \tree_v^t }_{v\in V}$.
    Additionally, for a tree $\tree$ with root $r$, we denote by $\mathcal{T}_r$ the multiset of subtrees that root at the descendants of $r$. 
\end{definition}

\begin{definition}
    (Blank Tree). A blank tree $\blankT$ is a tree (graph) that contains a single node and no edge, where the node feature is the zero vector $0$. 
\end{definition}

Recall that by assumption, all node features will come from the compact set $\Omega$, and $0 \not \in \Omega$. Therefore, the blank tree is not included in the space of trees with features in $\Omega$, and can be used for augmentation.

We can now define the tree distance:
\begin{definition}
    (Tree Distance).\footnote{This definition slightly varies from the original definition in \cite{TMD}, due to our choice to set the depth weight to 1 and using the 1-Wasserstein which is equivalent to optimal transport.} 
    The distance between two trees $\tree_a, \tree_b$ with features from $\Omega$ and $0\not \in \Omega$, is defined recursively as
    \[
    \TD(\tree_a, \tree_b) := \begin{cases}
      \|x_{r_a}-x_{r_b}\|_1 + W_{\TD}^{\blankT} (\mathcal{T}_{r_a}, \mathcal{T}_{r_b}) & \text{if $t>1$} \\
      \|x_{r_a}-x_{r_b}\|_1 & \text{otherwise}  
    \end{cases}
    \]
    where $t$ denotes the maximal depth of the trees $\tree_a$ and $\tree_b$. Here $W_{\TD}^{\blankT} $ denotes the Wasserstein metric obtained from the metric $\TD$ on the space of trees of smaller depth, with augmentation by blank trees $\blankT$.
\end{definition}

\begin{definition}
    (Tree Mover's Distance). Given two graphs, $G_a, G_b$ and $w,t \ge 0$, the tree mover's distance is defined by
    \[
    \tmd^t(G_a, G_b) = W^{\blankT}_{\TD}(\mathcal{T}_{G_a}^t, \mathcal{T}_{G_b}^t),
    \]
\end{definition}
where $\mathcal{T}_{G_a}^t$ and $\mathcal{T}_{G_b}^t$ denote the multiset of all depth $t$ computational trees arising from the graphs $G_a$ and $G_b$, respectively. 
\citet{TMD} proved that $\tmd^t(G_a, G_b)$ is a pseudo-metric that fails to distinguish only graphs that cannot be separated by $t+1$ iterations of the WL test. Thus, assuming that $0\notin\Omega$, $\tmd^t(G_a, G_b)$ is WL equivalent on $\GNof\Omega$ whenever $t+1\geq N$.

In addition, it is easy to see from the definition of TMD that it satisfies the following properties:
\begin{enumerate}
    \item  $\tmd^t((\A, \alpha \cdot \X),(\B, \alpha \cdot \Y)) = \alpha \cdot \tmd^t(\X,\Y)$ for any $\alpha \geq 0$.
    \item For fixed $\A,\B$, the TMD metric is piecewise linear in $\br{\X,\Y}$.
\end{enumerate}

These properties will be used to show that under the above assumptions, the embedding computed by FSW-GNN is bi-Lipschitz with respect to TMD.

\section{Proofs}
\label{app_proofs}
%
\subsection{DS metric}
\label{subapp_proofs_ds_metric}\label{metric_property}
Here we prove \cref{thm_rhods_is_wl_metric} 

\thmRhoDSIsWLMetric*

\begin{proof}
$\rhods$ is  symmetric,  since if a pair's minimum is obtained at $S$ then the exact value is obtained for the opposite pair with $S^{T}$ and vice-versa.

To prove the triangle inequality, let $(V_1, \A,\X),(V_2, \B,\Y),(V_3, \C,\Z)$ be three arbitrary graphs with $\abs{V_i}=n_i$, $i=1,2,3$. Then the following holds:
\begin{align*}
   \alpha:= \rhods(\A,\B) = \abs{n_1-n_2} + min_{S\in \Pi\of{n_1,n_2}}||\A S-S\B||_1 +\sum_{i,j}S_{i,j}||\X_{i}-\Y_{j}||_1 \\
  \beta:= \rhods(\B,\C) = \abs{n_2-n_3} + min_{S\in \Pi\of{n_2,n_3}}||\B S-S\C||_1 +\sum_{i,j}S_{i,j}||\Y_{i}-\Z_{j}||_1\\
  \gamma:=\rhods(\A,\C) = \abs{n_1-n_3} + min_{S\in \Pi\of{n_1,n_3}}||\A S-S\C||_1 +\sum_{i,j}S_{i,j}||\X_{i}-\Z_{j}||_1\\
\end{align*}
We want to prove $\gamma \leq \beta+\alpha$. Let $S^{1}, S^{2}$ be the minimizers of the first two equations. 
Define $S^{3} = n_2\cdot S^{1}\cdot S^{2}$ and note that $S^3 \in \Pi\of{n_1,n_3}$ because, using the notation $1_d$ for the all one $d$-dimensional vector, we have
\begin{align*}
S^31_{n_3}&=n_2\cdot S^{1}\cdot S^{2}1_{n_3}=S^{1}1_{n_2}=\frac{1}{n_1}1_{n_1} \\
1_{n_1}^TS^3&=n_2\cdot 1_{n_1}^TS_1\cdot S_2=1_{n_2}^TS_2=\frac{1}{n_3}1_{n_3}.
\end{align*}
Next, we prove the following lemma (all matrix norms in this lemma are entrywise $\ell_1$ norms).
\begin{lemma}
Let $S \in \Pi\of{n_1,n_2}$, then for all $B\in \RR^{n_2\times b}$ and $A\in \RR^{a\times n_1}$

$$\|SB\|_1\leq \frac{1}{n_2}\|B\|_1, \quad \|AS\|_1\leq \frac{1}{n_1}\|A\|_1 $$
\end{lemma}
\begin{proof}

\begin{align*}
 \|SB\|_1&=\sum_{i=1}^{n_1}\sum_{j=1}^{b} |(SB)_{ij}|\\
 &\leq  \sum_{i=1}^{n_1}\sum_{j=1}^{b} \sum_{k=1}^{n_2} S_{ik}|B_{kj}|\\
 &=\sum_{j=1}^{b} \sum_{k=1}^{n_2} |B_{kj}| \cdot    \sum_{i=1}^{n_1}S_{ik}\\
 &=\frac{1}{n_2}\|B\|_1
\end{align*}
A symmetric argument can be used to show that $\|AS\|_1\leq \frac{1}{n_1} \|A\|_1$
\end{proof}

Given the lemma, we now have
\begin{align*}
    \frac{1}{n_2}||\A S^{3}-S^{3}\C||_1&=||\A S^{1}S^{2}-S^{1}S^{2}\C||_1=||\A S^{1}S^{2}-S^{1}\B S^{2}+S^{1}\B S^{2}-S^{1}S^{2}\C||_1 \\
    &\leq 
    ||\A S^{1}S^{2}-S^{1}\B S^{2}||_1 + ||S^{1}\B S^{2}-S^{1}S^{2}\C||_1 = 
    ||(\A S^{1}-S^{1}\B)\cdot S^{2}||_1
    \\
    &+
    ||S^{1}\cdot(\B S^{2}-S^{2}\C)||_1
    \leq 
    \frac{1}{n_2} ||\A S^{1}-S^{1}\B||_1 +
    \frac{1}{n_2}\cdot||\B S^{2}-S^{2}\C||_1.
\end{align*}
Next, we show the second part is also smaller:
\begin{align*}
    \frac{1}{n_2}\sum_{i,j}S^{3}_{i,j}\cdot||\X_{i}-\Z_{j}||_1
    =& \sum_{i,j}\sum_{k} S^{1}_{i,k}\cdot S^{2}_{k,j} \cdot ||\X_{i}-\Z_{j}||_1
    \\ \leq& \sum_{i,j}\sum_{k} S^{1}_{i,k}\cdot S^{2}_{k,j} \cdot
    (||\X_{i}-\Y_{k}||_1+||\Y_{k}-\Z_{j}||_1)\\
    =& 
    \sum_{i,j}\sum_{k} S^{1}_{i,k} \cdot S^{2}_{k,j} \cdot ||\X_{i}-\Y_{k}||_1
    +
    \sum_{i,j}\sum_{k} S^{1}_{i,k} \cdot S^{2}_{k,j} \cdot ||\Y_{k}-\Z_{j}||_1.
\end{align*}
We now open both sums:
\begin{align*}
    \sum_{i,j}\sum_{k} S^{1}_{i,k} \cdot S^{2}_{k,j} \cdot ||\X_{i}-\Y_{k}||_1
    =&\sum_{i}\sum_{k} S^{1}_{i,k}  \cdot ||\X_{i}-\Y_{k}||_1 \cdot \sum_{j}  S^{2}_{k,j}
    \\=&\frac{1}{n_2} \sum_{i}\sum_{k} S^{1}_{i,k}  \cdot ||\X_{i}-\Y_{k}||_1.
\end{align*}
With the same argument, we obtain that 
\begin{align*}
    \sum_{i,j}\sum_{k}S^{1}_{i,k}\cdot S^{2}_{k,j}\cdot ||\Y_{k} - \Z_{j}||_1
    = \frac{1}{n_2} \sum_{k,j} S^{2}_{k,j}\cdot ||\Y_{k} - \Z_{j}||_1.
\end{align*}
So overall we obtain
\begin{align*}
\gamma \leq& \abs{n_1-n_3} +||\A S^{3}-S^{3}\C||_1 +\sum_{i,j}S^{3}_{i,j}||\X_{i}-\Z_{j}||_1 \\
\leq &\abs{n_1-n_2}+ ||\A S^{1}-S^{1}\B||_1 +\sum_{i,j}S^{1}_{i,j}||\X_{i}-\Y_{j}||_1+\abs{n_2-n_3}
\\+&||\B S^{2}-S^{2}\C||_1 +\sum_{i,j}S^{2}_{i,j}||\Y_{i}-\Z_{j}||_1 
 = \alpha + \beta.
\end{align*}

Now, we show that our metric $\rhods$ is equivalent to WL. Clearly, any pair of graphs with different numbers of vertices are distinguished both by WL and by $\rhods$. Thus, in the following, we assume that the two graphs have the same number of vertices.

We begin first with some needed definitions.
\paragraph{Partitions}
Most of our techniques are inspired by \cite{book}.
Given $\G = (\A, \X)$, we define a stable partition $\V = \P_{1}\cup \P_{2}...\cup \P_{k}$ if 
\begin{align}
    \forall u,v \in \P_{i}, \X_{u} =& \X_{v} \\
    \forall l \in [k],i \in [k], u, v \in  \P_{i}, |\N_{u}\cap \P_{l}| =& |\N_{v} \cap \P_{l}| 
\end{align}
In simple words, two nodes in the same partition must have the same feature and the same number of neighbors with the same feature.
Note that the union of all singletons is a valid stable partition.
We can characterize a partition as a $k$ tuple, and in the $i'th$ place, we put a tuple of the common feature and a vector telling the number of neighbors in other partitions.
We say that graphs have the same stable partition if, up to permuting the $k$-tuple indices, we have the same $k$-tuple. 
\begin{lemma}
    \label{stable}
    The number of colors in 1-WL can't increase at level $n$. In addition, all nodes with the same color at step $n$ make a stable partition.
\end{lemma}
\begin{proof}
The number of colors of $1$-WL can only increase as the number of iterations increases. It is also known that, if at some step $t$ the number of colors does not increase, then the process is 'stuck'. Accordingly,  after at most $n$ iterations, the number of different colors will stay the same.  Denote by $\P = (\P_{1},..,\P_{s})$ a partition of the nodes with the same coloring at the $n$-th iteration, and let $c\in C$ be some color, we claim that nodes in the same partition, have the same number of neighbors with color $c$. As otherwise, those nodes that now have the same coloring will have different coloring in the next $n+1$ iteration. But as the number of colors doesn't decrease (because of the concatenation of the current color), we have at least one more color, a contradiction. So, we found a stable partition.
\end{proof}
\begin{lemma}
    \label{equivalence}
    The following conditions are equivalent:
    \begin{itemize}
        \item $\G \cong_{1-WL} \HH$
        \item Both graphs have a common stable partition. That is, there is a relabeling of $G$ such that the same partition $\P_{1},..,\P_{s} $ of $V$ is a stable partition for both $\G$ and $\HH$, and the features inside each partition are the same, in both $\G$ and $\HH$.   
    \end{itemize}
\end{lemma}
\begin{proof}
    Assume we have the same common partition.
    Up to renaming the names of the vertices of the nodes in the first graphs, we assume we have the same stable partition with the same parameters.
    Assume that a common stable partition exists $\P = \P_{1}\cup..\cup P_{k}$.  
    Then, by simple induction on the number of iterations, we will prove that all nodes in the same partition have the same color in both graphs.
    \newline
    \textbf{Basis} 
    Nodes in the same partition have the same initial feature and, thus, have the same color (in both graphs).
    \newline
    \textbf{Step}
    By the induction hypothesis, all nodes in the same partition iteration $T$ have the same color; 
    and they both have the same number of neighbors with the same color, so the aggregation yields the same output.
    Thus, in iteration $T+1$, those nodes also have the same color.
    Note that this argument is symmetric to both graphs; thus, the 1-WL test will not distinguish them.
    \ys{Nadav take a look}
    On the other hand, if $\G_1,\G_2$ have the same 1-WL embedding, then we partition the nodes to those classes with the same color at iteration $n$, denoted by $\P = \P_{1}\cup..\cup P_{k}$. 
    We have to show this partition is valid.
    First, by definition, those nodes $u,v \in \P_{i}$ have the same node feature (by iteration 1); next, if $u,v$ don't have the same number of neighbors in $\P_{l}$ for some $l\in [k]$, then their color won't be the same at iteration $n+1$. But, as shown in lemma \ref{stable}, the number of distinct colors can't increase between $n$ to $n+1$. Thus, we found a common stable partition.            
\end{proof}
Before proving the following lemma, we revise a definition from \cite{book}.
We define direct the direct sum of matrices \( S_1 \) and \( S_2 \) by:
\[
S_1 \oplus S_2 =
\begin{bmatrix}
S_1 & 0 \\
0 & S_2
\end{bmatrix}
\]
Given $S \in \R^{n \times n}$, we say $S$ is composable if there exists $P,Q,S_1,S_2$ such that 
\begin{align*}
    S = P \cdot (S_{1} \oplus S_{2}) \cdot Q
\end{align*}
Such that $P,Q$ are permutation matrices.
By simple induction, we can write $M$ as a direct sum of an indecomposable
\begin{align*}
     S = P \cdot (S_{1} \oplus S_{2}\oplus ... \oplus S_{t}) \cdot Q
\end{align*}
\begin{lemma}
    \label{features}
    Let $S \in \Pi\of{n,n}$ and assume it's in the form of $S = S_{1} \oplus ... \oplus S_{k}$ such that all blocks are indecomposable and $\sum_{i,j}S_{i,j}\cdot||\X_{i}-\Y_{j}||=0$. Denote by $i_{1},...,i_{k+1}$ the indices that define the start and the end of $S_{1},...,S_{t}$ and by $I_t:=\{i_t,\ldots,i_{t+1}-1\}$.
    Then $\X_{i}=\Y_{j},\forall t \in [k], \forall i,j \in I_{t}$.
\end{lemma}
\begin{proof}
    Given $t\in [k]$, we build a bipartite graph from $I_{t}$ to itself, such that two indexes $i,j$ are connected if $S_{i,j}>0$. Note that this graph is connected, as otherwise, $S_{t}$ could be composed into its connected components, and thus $S_{t}$ would be composable. 
    Given a path $P=(i_1,\ldots,i_l)$ in the bipartite support graph, note that for every edge in the path connecting a row vertex $i$ to a column vertex $j$, we have $S_{i,j}>0$. Since the metric vanishes, this implies that $\X_i=\Y_j$. Propagating these equalities along the path, we obtain $\X_i=\Y_j$ for all $i,j\in I_k$, and we are done.
\end{proof}
\begin{theorem}
    Let $\G,\HH$ be two featured graphs. Then 
    \begin{align*}
        d(\G,\HH) = 0 \iff \G \wleq \HH
    \end{align*}
\end{theorem}
We first prove that if the two graphs are 1-WL equivalent, then this metric vanishes. We may assume they have the same stable partition as we proved above. Take $\P=\P_{1}\cup..\cup \P_{t}$, rename the nodes such that they come in consecutive order and denote by $n_{1},n_{2},..,n_{k}$ the sizes of the partitions. Note that by definition, $\forall u,v\in \P_{i}$, both have the same feature and number of neighbors in all $\P_{l}$.
As in the book \cite{book}, define
\[
S := \frac{1}{n}\left(
\frac{1}{n_1}J_{n_1}\oplus\cdots\oplus\frac{1}{n_k}J_{n_k}
\right).
\] and from the book \cite{book}, we know that $\A S=S\B$. As all nodes in the same partition have the same feature, $\X_{i}=\Y_{j},\forall i,j \in I_{k} = [n_{t},n_{t+1}],\forall t \in [k]$, so as $S$ is non-zero only on indexes in the same partition then also this metric vanishes, so also the sum vanishes on $S$. 

For the second direction, let $\G = (\A,\X),\HH = (\B,\Y)$ be two graphs and assume there exists a matrix of the form
$S=P(S^{1}\oplus\cdots\oplus S_{t})Q$ such that the metric vanishes. Denote by
$D=S^{1}\oplus\cdots\oplus S_{t}$. We show that, up to relabeling the nodes of the two graphs, we can choose $S$ to be a block diagonal matrix.

Since the metric vanishes, we have $\A S=S\B$. Therefore,
\begin{align*}
    0
    &=\|\A PDQ-PDQ\B\|_1\\
    &=\|P\cdot(P^{-1}\A P\cdot D-D\cdot Q\B Q^{-1})\cdot Q\|_1\\
    \stackrel{(*)}{=}
    &\|P^{-1}\A P\cdot D-D\cdot Q\B Q^{-1}\|_1,
\end{align*}
where $(*)$ follows from the fact that permutation matrices preserve the norm.

Similarly, for the feature term, by reindexing the sum according to the permutations defined by $P$ and $Q$, we obtain
\begin{align*}
    0
    &=\sum_{i,j}(PDQ)_{i,j}\cdot\|\X_i-\Y_j\|_1\\
    &=\sum_{i,j}D_{i,j}\cdot
    \|(P^{-1}\X)_i-(Q\Y)_j\|_1.
\end{align*}

We define two graphs
\[
\hat{\G}=(P^{-1}\A P,P^{-1}\X)\cong\G,
\qquad
\hat{\HH}=(Q\B Q^{-1},Q\Y)\cong\HH.
\]
Thus, for $\hat{\G}$ and $\hat{\HH}$, the metric vanishes using the block diagonal matrix $D$. Hence, up to isomorphism, we can assume that $S=D$ is a block diagonal matrix.

Note that $D$ defines a partition $\P_1,\ldots,\P_s$ of the nodes, and we will prove that it is a stable partition of both graphs.
From
\[
(P^{-1}\A P)D=D(Q\B Q^{-1}),
\]
we obtain, as in \cite{book}, that every two nodes $u,v\in\P_k$ have the same number of neighbors in $\P_l$, for every $l\in[s]$.

Next, by Lemma \ref{features}, we know that
$\hat{\X}_i=\hat{\Y}_j$ for all $i,j\in\P_k$, so nodes in the same partition have the same feature. Thus, this partition is stable.

Therefore, $\hat{\G}$ and $\hat{\HH}$ have exactly the same stable partition. Consequently, $\G$ and $\HH$ have the same stable partition up to isomorphism, and by Lemma \ref{equivalence}, the two graphs are 1-WL equivalent.
\end{proof}

\subsection{FSW-GNN: Equivalence to WL}
\label{subapp_proofs_fswgnn_wl_equivalence}
We now prove \cref{thm_fswgnn_is_wl_equivalent} that says the FSW-GNN is equivalent to WL.

\thmFSWGNNIsWLEquivalent*
\begin{proof}
This proof employs the finite witness theorem \cite{amir2023neural}, and the basic methodology to use it for graph neural network injectivity, first presented by \citet{hordan2024complete}. 

It is sufficient to show that the functions $\Phi^{\br{t}}$ and $\Psi$ are injective vector-to-vector functions, and that the FSW embedding maps used in the FSW-GNN are injective multiset to vector functions. The injectivity of all these functions ensures that the MPNN is WL-equivalent \cite{gin}.

By Theorem 4.1 in \citep{amir2024injective}, if we take  the FSW embedding with an output dimension of $2Nd+1$, defined on the space of multisets with at most $N$ features in $\RR^d$, then the embedding will be injective for almost every choice of its parameters. This injectivity is in the sense that two multisets will attain the same value if and only if they define the same distribution. This can happen when the multisets are not equal, but have different cardinalities. However, the last coordinate of the FSW embedding, which encodes the multiset cardinality, ensures that this will not be an issue. Accordingly the FSW embedding is injective on the multiset space. 

We note that the output of the FSW embedding will now be node features of a higher dimension of $2Nd+2$, rather than just $d$. However, the set of all possible features obtained from the previous iteration will be a semialgebraic set of dimension $\leq Nd$. As long as  this is the case, reapplying the FSW embedding with output dimension of $2Nd+2$ will still be injective, for almost every choice of parameters.

The same argument can also be applied to the linear functions $\Phi^{(t)}$ and $\Psi$ used in FSW-GNN. Indeed, it is an easy consequence of the finite witness theorem \cite{amir2023neural} that, for a semialgebraic set of dimension at most $Nd$, almost every linear map into $\mathbb{R}^m$ is injective whenever $m>2Nd$. Since $m\geq 2Nd+2$, the functions $\Phi^{(t)}$ and $\Psi$ are injective on their respective sets of attainable inputs for almost every choice of parameters. This concludes the proof.
\end{proof}
\subsection{FSW-GNN: Bi-Lipschitzness}
\label{subapp_proofs_fswgnn_bilipschitz}
We first prove \cref{thm_pwl_ratio_bound}.
\thmPWLRatioBound*
\nd{olde proof commented out}
\begin{proof}
  
   By definition, there is a partition of $M$ into a finite union of compact polygons, so that the restriction of $f$ to each polygon is an affine function. $g$ also has such a partition, and by taking the mutual refinement of the two partitions, we obtain a finite partition of $M$ into compact polygons, such that the restriction of both $f$ and $g$ to each polygon is affine. 

   Accordingly, it is sufficient to prove the claim on a single compact polygon $P\subseteq M$ on which the restriction of both $f$ and $g$ is affine. A compact polygon has a finite number of extreme points $v_1,\ldots,v_k$, and every point in the polygon can be written as a convex combination of these points. Now let us denote 
   $$I=\{i\in [k]| \quad f(v_i)>0 \}=\{i\in [k]| \quad g(v_i)>0 \} .$$
 \nd{Yonatan for your question:} 
  We note that if $I$ is empty, then $f$ and $g$ are identically zero on $P$, and there is nothing to prove. Otherwise, the following quantities are well defined and strictly positive:
  
   \begin{align*}
   m_f&=\min_{i\in I} f(v_i), \quad M_f=\max_{i\in I} f(v_i)\\ 
   m_g&=\min_{i\in I} g(v_i), \quad M_g=\max_{i\in I} g(v_i)
\end{align*}
Then for every $x\in P$, there exist non-negative $\theta_1,\ldots,\theta_k $ which sum to $1$ such that $x=\sum_{i=1}^k \theta_iv_i$. 

It follows that 
\begin{align*}
f(x)&=f(\sum_{i=1}^k \theta_iv_i)
=\sum_{i=1}^k \theta_i f(v_i)
=\sum_{i\in I} \theta_i f(v_i)\\
&\geq m_f \sum_{i\in I} \theta_i
=\frac{m_f}{M_g}\cdot M_g \sum_{i\in I} \theta_i
\geq \frac{m_f}{M_g}\cdot \sum_{i\in I} \theta_i g(v_i)\\
&= \frac{m_f}{M_g}\cdot g(\sum_{i=1}^k \theta_i v_i)
=\frac{m_f}{M_g}\cdot g(x).
\end{align*}    
By reversing the roles of $f$ and $g$, we obtain that for every $x\in P$,
$$g(x) \geq \frac{m_g}{M_f}f(x) .$$
This concludes the proof.
   \end{proof}

We now turn to the full proof of \cref{thm_fswgnn_is_bilip}.

\thmFSWGNNIsBilip*

\begin{proof}
Following the reasoning in the main text, the rest of the proof is as follows: Let $G,\tilde G \in \GNof\Omega$, $G=\br{V,E,X}$ and $\tilde G=\br{\tilde V, \tilde E, \tilde X}$, with $n$ and $\tilde n$ vertices respectively. Since there is a finite number of choices of possible $n,\tilde n \leq N$ and edges $E$, $\tilde E$. It is enough to show that once these parameters are chosen, the bi-Lipschitz ratio is bounded on all choices of $X \in \Omega^n$, $\tilde X \in \Omega^{\tilde n}$. Define the function $f\of{X,\tilde X} = \norm{\bh_G - \bh_{\tilde G}}_1$, $g\of{X,\tilde X} = \rho\of{G,\tilde G}$, where $\rho$ is either $\rhods$ or TMD. By the comment above, both $f$ and $g$ are piecewise linear. Since both FSW-GNN and the metric $\rho$ are WL-equivalent, $f$ and $g$ have the same zero set. Lastly, if $\rho=\rho_{\mathrm{DS}}$, since $\Omega$ is compact, it is contained in a compact polygon. If $\rho=\mathrm{TMD}$, then additionally $0\notin\Omega$, and therefore the distance from $\Omega$ to $0$ is strictly positive. Hence, $\Omega$ is contained in a finite union of compact polygons that don't contain $0$ (e.g., by taking the set difference between a compact box containing $\Omega$ and a sufficiently small $\ell_\infty$ ball around $0$). Thus, Lemma 3.4 guarantees the existence of Lipschitz constants $c,C$. Thus, \cref{thm_pwl_ratio_bound} guarantees the existence of Lipschitz constants $c,C$. Finally, since there are only finitely many choices of graph sizes, adjacency
matrices, vertices $v,\tilde v$, and iterations $t\leq N$, we can take the
minimum of the corresponding lower Lipschitz constants and the maximum of the
upper Lipschitz constants. Thus, the constants $c,C$ can be chosen uniformly
over $\mathcal{G}_{\leq N}(\Omega)$.

Finally, note that by equivalence of all norms on a finite dimensional space, the same bi-Lipschitz equivalence holds if we replace the one-norm in the definition of $f$ with any other norm.
\end{proof}

\thmNodeLevelBiLip
\begin{proof}
Let $G = \br{V,A,\bX}$, $\tilde G = \br{\tilde V, \tilde A, \btX}$ be two graphs in $\GNof\Omega$. Let $t\in\NN$ denote the number of MPNN/WL iterations applied. Let $v\in V$, $\tilde v \in \tilde V$. Let $\bh^{\br{t}}_v$ and $\bh^{\br{t}}_{\tilde v}$ be the output features of vertices $v$ and $\tilde v$ computed by FSW-GNN. Then $\textup{TD}\of{\tree_v^{(t)},\tree_{\tilde v}^{(t)}} = 0$ if and only if $v$ and $\tilde v$ are assigned the same features by the WL test, which in turn takes place if and only if $\bh_v^{(t)} = \bh_{\tilde v}^{(t)}$. Note that $\norm{\bh_v - \bh_{\tilde v}}_1$ is a piecewise-linear function of the input vertex features $\brc{\bx_v}_{v \in V}$ and $\brc{\btx_{\tilde v}}_{\tilde v \in \tilde V}$. Moreover, $\mathrm{TD}(T_v^{(t)},T_{\tilde v}^{(t)})$ is equivalent, up to bounded distortion,
to a piecewise-linear function of the input vertex features
$\{x_u\}_{u\in V}$ and $\{\tilde{x}_{\tilde u}\}_{\tilde u\in\tilde V}$. Lastly, similarly to the proof of \cref{thm_fswgnn_is_bilip}, $\Omega$ can be extended to a compact polygon that does not contain $\bzero$. Thus, by \cref{thm_pwl_ratio_bound}, there exist constants $0<c,C < \infty$ such that
\begin{equation*}
c \cdot\textup{TD}\of{\tree_v^{(t)},\tree_{\tilde v}^{(t)}}
\leq
\norm{\bh_v^{(t)} - \bh_{\tilde v}^{(t)}}_1
\leq
C \cdot \textup{TD}\of{\tree_v^{(t)},\tree_{\tilde v}^{(t)}}.
\end{equation*}
\end{proof}

\section{Experiment Details}
\label{app_numerical_experiments}
We used the Adam optimizer for all experiments.

For the empirical distortion evaluation, we used pairs of graphs $G$, $\tilde G$, each consisting of four vertices and the edges $1-2-3-4-1$. Two random vectors $v_0, \Delta v \in \RR^d$ were drawn i.i.d. Gaussian and normalized to unit length. In $G$ all vertex features were set to $v_0$, whereas in $\tilde G$ they were set to $v_0 + \varepsilon \sigma \Delta v$, with $\sigma = 1$ for $v_2, v_4$ and $-1$ for $v_1, v_3$. We used $\varepsilon = \brc{1,10^{-1},10^{-2},\ldots,10^{-6}}$, and generated 100 pairs for each value of $\varepsilon$, and evaluated the constants $C$, $c$ for the resulting pairs. This experiment was repeated 10 times and the average distortion was taken. To ensure accurate results, we used 64-bit floating-point arithmetic in this experiment.

For the NeighborsMatch problem from \citep{alon}, we used the protocol developed in their paper: we used  their implementation for the MPNNs we compared to, with a hidden dimension of $64$ for all models, searched for the best hyperparameters for each model, and reported the training accuracy. For fair comparison with rival models, we repeated each sample 100 times, as was done in \cite{alon}.

For the Ring dataset, we used the results from \cite{bron} and trained our models with a hidden dimension of 64.

For the LRGB dataset, we trained all models under the constraint of 500K parameters. In contrast, for the MolHIV dataset, there was no restriction, and we trained the models with 40K parameters.

For the transductive learning tasks, we used a hidden dimension of 128 across all models.

\paragraph{Timing} Here, we show the testing time for FSW-GNN, GIN, GCN, GGNN, GAT, and GraphSAGE on the MUTAG graph classification task.
\begin{table}[h!]
    \caption{Comparison of Average testing time in seconds in evaluation mode for different models on MUTAG.}
    \label{comparison_time_fswgnn}
    \centering
    \begin{tabular}{|c|c|}
        \hline
        Model          & Avg Time per Testing  \\ \hline
        GIN            & 0.09                    \\ \hline
        GCN            & 0.08                    \\ \hline
        GAT            & 0.09                    \\ \hline
        GGNN           & 0.08                    \\ \hline
        GraphSAGE      & 0.07                    \\ \hline
        FSW-GNN        & 0.11                 \\ \hline
    \end{tabular}
\end{table}

\section{Ordered subgraph aggregation}\label{app:subgraph}
In this section, we recall sub-graph aggregation networks and show that FSW-GNN could be used to create a bi-Lipschitz embedding with respect to the metric equivalent to sub-graph WL.
\subsection{Ordered subgraph 1-WL}. 
The OSWL (Ordered sub-graph WL) is taken from \cite{OSWL,OSWL2}.
Given a 1-WL procedure using $T$ iterations denoted by $WL^{T}$, and deterministic graph sampling $\Pi$(could be node/edge deletion), we define the following three steps: 
\begin{itemize}
    \item Step 1: Deterministic graph sampling (e.g. node/edge deletion) $\Pi(\G) = \{\G_1,\ldots,\G_m\}$ .
    \item $y_i = WL^{T}(\G_i),\G_i \in \Pi(\G)$
    \item $y_G = \phi(\{y_1,\ldots,y_m\})$
\end{itemize}
Where $WL^T $ denotes the application of $T$ iterations of the WL test, and $\phi$ is an injective multiset function.
\begin{definition}
    We say $\G_1, \G_2$ are equivalent up to OSWL, if $OSWL^{T}(\G_1) = OSWL^{T}(\G_2),\forall T \in \NN$, and we denote this by $\G_1 \cong \G_2$. As for WL, checking equality for $T =|\G_1|$ is sufficient to ensure equality for all $T$.
\end{definition}
\paragraph{Ordered subgraph MPNN}
Ordered subgraph MPNNs are very similar to OSWL. Given an MPNN and a multi-set to vector embedding $\phi$, we define its sub-graph version: 
\begin{itemize}
    \item Step 1: Deterministic graph sampling  $\Pi(\G) = \{\G_1,\ldots,\G_m\}$ 
    \item $y_i = MPNN(\G_i),\G_i\in \Pi(\G)$
    \item $y_G = \phi(\{y_1,\ldots,y_m\})$
\end{itemize}
\paragraph{OSWL equivalent metric}
We define a metric OSWL equivalent similarly to \ref{WL_metric}:
\begin{definition*}[WL metric]
    An \emph{OSWL metric on $\GNofe\Omega$} is a function $\rho:\GNofe\Omega \times \GNofe\Omega \to \RR_{\geq 0}$ that satisfies the following conditions for all $G_1, G_2, G_3 \in \GNofe\Omega$:
    \begin{subequations}
    \begin{align}        
        \rho\of{G_1, G_2} &\eqx \rho\of{G_2, G_1}
        & \quad &\textup{\emph{Symmetry} }
        \\ \rho\of{G_1, G_3} &\leqx \rho\of{G_1, G_2} + \rho\of{G_2, G_3}
        & \quad &\textup{\emph{Triangle inequality} }
        \\ \rho\of{G_1, G_2} &= 0 \Longleftrightarrow G_1 \cong G_2.
        & \quad &\textup{\emph{OSWL equivalence} }
    \end{align}
    \end{subequations}
\end{definition*}
\begin{theorem}
    Let a 1-WL equivalent metric $d_{WL}$, and sampling procedure, the following metric is OSWL equivalent:  
    \begin{align*}
    d_{sub}(\G,\HH) = min_{\pi \in S_m}\sum^{m}_{i=1} d_{WL}(\G_i,\HH_{\pi(i)})
\end{align*}
\end{theorem}
\begin{proof}
    We prove only the last part, which is not trivial.
    Assume first that the number of nodes is the same, then:
    \begin{align*}
        d_{sub}(\G,\HH) = 0 \iff \exists \pi \in S_m:\forall i\in [m]: d_{WL}(\G_i,\HH_{\pi(i)}) = 0 \iff \\ 
        \iff \exists \pi \in S_m: \forall i \in [m],y^{\G}_i = y^{\HH}_{\pi(i)}  \\
        \phi(\{y^{\G}_1,\ldots,y^{\G}_m\}) = \phi(\{y^{\HH}_1,\ldots,y^{\HH}_m\}) \iff OSWL(\G) = OSWL(\HH)
    \end{align*}
\end{proof}
\begin{theorem} Assume that the node features belong to a compact domain $\Omega$. Assume we use FSW-GNN as the MPNN, FSW aggregation as $\phi$, and $d_{DS}$ as the 1-WL equivalent metric. Then the overall OSMPNN is OSWL bi-Lipschitz. 
\end{theorem} 
\begin{proof} Note that the OSMPNN is piecewise linear as a composition of piecewise linear functions. Also, $d_{sub}$ is piecewise linear as a sum of piecewise linear functions. In addition, both functions vanish on the same set. Since the input domain is compact, by Theorem~\ref{thm_pwl_ratio_bound}, the OSMPNN is bi-Lipschitz with respect to $d_{sub}$.
\end{proof}



\addtocontents{toc}{\protect\setcounter{tocdepth}{1}}

\printbibliography[
    heading=subbibintoc,
    title={References}
]

\endgroup

\end{refsection}

\chapter{Discussion}
\label{chap:discussion}

\section{Summary of the Main Findings}

This thesis studies the expressive power of neural networks for structured data, with an emphasis on two related questions: how the symmetries of the input constrain the architecture, and what information a symmetry-respecting representation preserves beyond the minimal requirement of invariance or equivariance. The results span geometric graphs, multi-permutation symmetries, set functions, and message-passing graph neural networks. Although the four works address different settings, they share a common viewpoint: the structure of the input domain should be reflected explicitly in the representation, and doing so can lead both to sharper theoretical characterizations and to improved practical models.

The first work, presented in Chapter~\ref{chap:paper1}, studies the expressive power of message-passing networks on sparse geometric graphs. A central distinction is between networks whose intermediate features are rotation invariant and networks that are allowed to propagate rotation-equivariant features. For equivariant geometric GNNs, we prove that generic completeness is possible precisely when the underlying graph is connected. In contrast, for invariant geometric GNNs, connectivity is not sufficient: generic completeness holds precisely when the graph is generically globally rigid. This gives a sharp characterization of how the choice of intermediate representation changes the amount of geometric information that can be recovered from sparse neighborhoods. We further introduce EGenNet, a simple equivariant message-passing architecture that attains the corresponding generic separation guarantees. Empirically, the architecture performs competitively on synthetic separation tasks and molecular regression benchmarks, suggesting that generic rather than worst-case completeness may be sufficient in many practical regimes.

The second work, presented in Chapter~\ref{chap:paper2}, approaches equivariant neural networks from a representation-theoretic perspective. Instead of deriving equivariant linear layers by enumerating parameter-sharing patterns, we decompose the relevant representation into irreducible components and use Schur's lemma to characterize all admissible linear maps. This recovers classical results such as DeepSets and second-order invariant graph networks, substantially simplifies the derivation of equivariant layers for deep weight spaces, and leads to a new characterization for wreath-product symmetries. In particular, for the action of $\mathcal G\wr S_k$ on $\mathcal V^k$, if the fixed subspace of $\mathcal V$ has dimension $s$, then the full space of equivariant maps contains $s^2$ families of non-Siamese interactions in addition to the equivariant maps applied independently to each component. The experiments show that these additional interactions can be necessary for tasks in which information must be exchanged across otherwise symmetrically interchangeable inputs.

The third work, presented in Chapter~\ref{chap:paper3}, considers a different structural relation: set containment. Rather than requiring only permutation invariance or injectivity, the goal is to construct set representations whose coordinatewise order reflects containment. The work characterizes when monotone and separating set functions can exist, gives dimension bounds and impossibility results, and develops a relaxed stable construction for settings where exact separation is impossible. While this direction is somewhat distinct from the graph-focused results of the thesis, it reinforces the broader principle that a useful representation can be required to preserve more structure than equality alone.

The fourth work, presented in Chapter~\ref{chap:paper4}, develops this principle quantitatively for message-passing GNNs. Standard expressivity theory asks whether two graphs that are distinguished by the Weisfeiler--Leman procedure are also assigned different representations by the network. This is an essential requirement, but it is purely qualitative: two distinguishable graphs may still be mapped arbitrarily close to one another. We introduce FSW-GNN, which uses Fourier Sliced-Wasserstein aggregation, and prove that it is both WL-equivalent and bi-Lipschitz with respect to WL-compatible graph metrics, including the doubly-stochastic metric and Tree Mover's Distance. Thus, graph distances are preserved up to multiplicative constants rather than merely remaining nonzero. Empirically, FSW-GNN exhibits substantially lower metric distortion than standard MPNNs and performs particularly well on long-range tasks. In the graph-transfer benchmarks, it is the only evaluated message-passing method to achieve perfect accuracy for all tested radii up to $15$ across the considered graph topologies, while standard MPNNs deteriorate at substantially shorter distances. These observations suggest that quantitative preservation of information is relevant not only as a mathematical strengthening of injectivity, but also as a useful design principle for deep message passing.

\section{Symmetry as a Constraint on Expressive Architectures}

The first two works highlight two complementary roles played by symmetry. First, symmetry determines which distinctions a model should deliberately ignore. A representation of a geometric graph should not change under relabeling of the vertices or under a global rigid motion, and a representation of an unordered collection should not depend on the order in which its elements are presented. Second, once these required identifications are fixed, symmetry also constrains the possible mechanisms by which information can be propagated and combined. Expressivity is therefore not independent of the symmetry model: the chosen invariant or equivariant feature space can determine which information remains available to later layers.

This is particularly clear in the geometric setting of Chapter~\ref{chap:paper1}. An invariant geometric message-passing layer communicates quantities that are unchanged by rotations. In the considered setting, this means that the network can reconstruct geometric structure only through invariant local measurements, most importantly distances along edges. If these distances do not uniquely determine the geometric realization, two non-isomorphic realizations can remain indistinguishable. Global rigidity is therefore not merely a graph-theoretic condition that happens to arise in the proof; it describes exactly when the information visible to an invariant architecture is sufficient to determine a generic realization. Rotation-equivariant intermediate features alter this situation because they can transport directional information. As a result, connectedness alone becomes sufficient for generic separation. The comparison provides a concrete example in which increasing the type of representation carried by the hidden features changes the graph-theoretic condition required for expressivity.

This distinction also gives a useful interpretation of generic completeness. Fully complete geometric models must account for exceptional configurations, including highly symmetric or algebraically degenerate inputs. Doing so may require substantially more elaborate higher-order representations. The results of Chapter~\ref{chap:paper1} show that, away from these exceptional configurations, a much simpler architecture can already preserve all information required to distinguish geometric graphs up to the relevant symmetries. The experimental results support the practical relevance of this regime: EGenNet is competitive with more complicated geometric architectures on the considered molecular tasks. This does not imply that generic completeness is sufficient for every application, but it suggests that worst-case expressive power and useful practical expressive power need not coincide.

Chapter~\ref{chap:paper2} addresses a different aspect of the same question. Once a symmetry group and its action on the feature space are specified, one still has to characterize the full family of linear maps that respect this action. Parameter-sharing arguments solve this problem by working directly in coordinates and identifying parameters that must be equal. This is effective for familiar actions but can become cumbersome when the action has several interacting permutation symmetries. The irreducible-representation viewpoint separates the coordinate description from the underlying algebraic structure. After decomposing the representation, Schur's lemma identifies exactly which components may interact. The resulting characterization is therefore both a derivation technique and a way of exposing interactions that may be obscured by a coordinate-level parameter-sharing argument.

The wreath-product result illustrates this point sharply. A Siamese architecture applies the same $\mathcal G$-equivariant map independently to each of the $k$ components. Such an architecture respects the symmetry, but symmetry does not force the components to be processed independently. The full characterization shows that information may also pass through the $\mathcal G$-fixed subspace. When this fixed subspace has dimension $s$, there are $s^2$ types of such cross-component interactions. For graph-valued components, for example, $s=2$, producing four non-Siamese interaction families. The anomaly-detection experiments demonstrate why this matters: detecting which member of a symmetric collection is anomalous is inherently relational, and a purely Siamese architecture cannot solve the task because it never compares information across members. Thus, satisfying an equivariance constraint is weaker than exploiting the entire space of functions permitted by that constraint.

Together, Chapters~\ref{chap:paper1} and~\ref{chap:paper2} suggest a general lesson. Symmetry should not be viewed only as a regularizer or as a way to reduce the number of parameters. It determines the quotient structure of the learning problem and, through the representation chosen for hidden features, determines which communication channels are available to the network. A complete analysis of an equivariant architecture should therefore ask both whether the output obeys the required transformation law and whether the architecture uses the full set of symmetry-compatible interactions needed for the task.

\section{Beyond Qualitative Expressivity in Message Passing}

The results of Chapter~\ref{chap:paper4} address a limitation of the standard notion of GNN expressivity. In the classical WL-based viewpoint, the main question is whether a network produces different outputs whenever the WL procedure distinguishes two graphs. This viewpoint has been extremely useful for classifying architectures, but it treats all nonzero separations as equivalent. From the perspective of numerical representation, however, an embedding distance of $10^{-1}$ and one of $10^{-20}$ are qualitatively identical but computationally very different. The latter leaves the graphs formally distinguishable while making the distinction fragile and difficult for subsequent layers to exploit.

FSW-GNN was designed to control this quantitative aspect of representation. Its bi-Lipschitz property gives both an upper and a lower comparison between distances in graph space and distances in representation space. The upper bound prevents small perturbations of the graph from causing arbitrarily large changes in the representation, while the lower bound prevents distinct graph structures from collapsing arbitrarily close together relative to the chosen graph metric. The important conceptual point is not the definitions themselves, which were developed in Chapter~\ref{chap:methods}, but what the resulting theorem adds to existing expressivity guarantees: it replaces the statement that information is not completely lost by the stronger statement that information is retained at a controlled scale.

The choice of WL-compatible metrics is central to the interpretation of this result. Since an MPNN cannot separate graphs that are WL-equivalent, no meaningful lower-Lipschitz guarantee can hold with respect to a metric that assigns positive distance to such pairs. The doubly-stochastic metric and Tree Mover's Distance instead measure graph differences at the resolution that message passing can in principle observe. Bi-Lipschitzness with respect to these metrics should therefore be understood as preservation of essentially all quantitatively meaningful information available to a WL-equivalent MPNN. In this sense, the FSW result is a metric refinement of the usual WL expressivity theory rather than an alternative to it.

The empirical distortion experiments support this interpretation. On challenging graph pairs, FSW-GNN produces substantially smaller estimated distortion than the competing message-passing architectures, for both doubly-stochastic and tree-based graph metrics. This observation is important because the theoretical theorem by itself guarantees only the existence of finite constants; it does not give useful numerical estimates of their magnitude. The experiments therefore provide complementary evidence that the advantage is not merely asymptotic or pathological: the representations produced by FSW aggregation appear to retain graph distances more faithfully in finite-dimensional networks.

The most striking empirical effect appears in long-range message passing. Standard MPNNs often degrade as the number of required propagation steps grows. Oversmoothing describes the collapse of node representations toward one another, while oversquashing describes the weak influence of distant information after it has passed through repeated aggregations. Although these phenomena are different, both involve a loss of quantitative separation. Bi-Lipschitzness does not by itself prove that FSW-GNN is immune to either phenomenon, because the bi-Lipschitz constants may worsen with depth. Nevertheless, it rules out the more severe form of collapse in which a finite-depth architecture loses its lower metric control altogether.

The long-range experiments are consistent with this picture. On NeighborsMatch, FSW-GNN retains perfect accuracy throughout the tested range up to radius $8$, whereas the competing methods fail at smaller radii. On the graph-transfer tasks, FSW-GNN achieves perfect accuracy for all tested radii up to $15$ on the clique-path, ring, and cross-ring topologies, without rewiring the graph or introducing global communication. Its performance eventually deteriorates for still larger radii, which is an important qualification: the theoretical bi-Lipschitz property does not imply depth-independent distortion. The experiments should therefore not be interpreted as proving that bi-Lipschitzness completely solves long-range message passing. Rather, they provide evidence that preserving aggregation geometry can substantially delay the degradation that appears in standard MPNNs.

This distinction is also useful when comparing FSW-GNN with methods that address long-range dependencies by changing graph topology. Rewiring, virtual nodes, spectral methods, and global attention can reduce the effective path length between distant vertices. FSW-GNN takes a complementary approach: it leaves the original topology unchanged and attempts to improve how faithfully information is transmitted through the existing sequence of local aggregations. Consequently, its strong performance on the synthetic transfer tasks isolates the aggregation mechanism itself as a significant contributor to long-range behavior.

\section{Broader Implications}

A recurring theme across the thesis is that binary notions of correctness are often too coarse for understanding neural representations. Invariance specifies which inputs must be identified, but does not specify which non-equivalent inputs remain distinguishable. Injectivity or WL-equivalence adds separation, but does not quantify the robustness of this separation. The FSW results add metric control, while the set-function work shows that in other domains the relevant structure may instead be an order relation. These are different mathematical requirements, but they can be organized as increasingly informative questions about a learned representation: which distinctions should disappear, which distinctions must remain, and which additional relations between inputs should be reflected in the representation space.

This viewpoint also changes how architectural choices can be interpreted. An aggregation rule, a hidden representation type, or a parameter-sharing scheme is not merely an implementation detail. Each choice determines what aspects of the underlying structure are retained and what interactions can be expressed. The invariant-versus-equivariant distinction in sparse geometric graphs changes the required graph rigidity; the complete wreath-product characterization reveals interactions absent from Siamese architectures; and replacing sum aggregation with a bi-Lipschitz multiset embedding changes the quantitative geometry of message passing. In all three cases, a local architectural decision has a global consequence for expressive behavior.

Another implication is that expressive-power results are most informative when matched to the resolution of the task. Full geometric completeness is a strong worst-case requirement, but generic completeness may be sufficient when non-generic configurations are rare in the data distribution. WL-equivalence is the natural qualitative resolution for standard MPNNs, but long-range tasks may require additional quantitative control within the WL quotient. Similarly, a set representation used for containment queries should preserve containment rather than merely distinguish sets. There is therefore unlikely to be a single notion of expressivity that is optimal for every structured learning problem. A more useful goal is to identify the structural relation that matters for a task and design representations that preserve that relation with the strongest feasible guarantees.

The thesis also suggests a productive interaction between characterization results and constructive architectures. The theoretical analyses are not only impossibility or classification statements. The geometric characterization motivates EGenNet; the representation-theoretic characterization motivates SchurNet; and the metric analysis motivates FSW-GNN. In each case, understanding the mathematical source of an expressive limitation leads directly to an architectural modification. This connection is particularly valuable because expressive-power theory is sometimes criticized for distinguishing models in ways that do not translate to empirical improvements. The results here provide several examples where the theoretical distinction is observable in controlled experiments and, in the geometric and FSW settings, also translates to competitive performance on broader benchmarks.

\section{Limitations and Open Questions}

The results leave several important limitations and open directions. First, the geometric results of Chapter~\ref{chap:paper1} are statements about generic configurations. They deliberately exclude a measure-zero family of non-generic inputs, including configurations with exact symmetries or degeneracies. Such configurations can be important in scientific applications, where symmetry may be physically meaningful rather than accidental. A natural direction is to characterize efficient sparse geometric architectures that obtain stronger worst-case separation guarantees without resorting to prohibitively expensive high-order representations. Higher-order geometric WL procedures and subgraph-based geometric methods are possible candidates, but their expressive power and computational cost on sparse graphs remain to be understood more completely.

Second, the representation-theoretic framework of Chapter~\ref{chap:paper2} gives a complete description of linear equivariant maps once an irreducible decomposition is available, but computing useful decompositions can itself become difficult for more complicated representations. Extending this approach to higher-order invariant graph networks is especially interesting. Such networks operate on tensor spaces whose coordinate dimension grows quickly with the order, while their irreducible components may admit a substantially more economical description. This raises both theoretical and practical questions: whether irreducible feature spaces can lead to lower-dimensional architectures with the same symmetry guarantees, how nonlinearities should be chosen between such components, and whether the resulting parameterizations improve optimization or generalization in addition to reducing algebraic bookkeeping.

Third, the bi-Lipschitz guarantees for FSW-GNN are qualitative with respect to the constants. They establish finite upper and lower Lipschitz bounds on the relevant bounded graph domains, but do not provide sharp estimates of the distortion or its dependence on depth, width, graph size, and the number of FSW slices. This is arguably the main theoretical gap between the current guarantee and a complete explanation of the long-range experiments. If the distortion were to grow rapidly with depth, bi-Lipschitzness at every fixed depth would offer only limited protection for very long propagation distances. Obtaining explicit depth-dependent bounds, and identifying architectural choices that keep these bounds small, would directly connect the metric theory to oversmoothing and oversquashing.

Relatedly, the empirical correlation between lower distortion and improved long-range performance should not yet be interpreted as a causal theorem. FSW aggregation differs from sum-based aggregation in several ways, and the present results do not isolate which part of the empirical gain is explained specifically by the bi-Lipschitz property. Controlled comparisons between different injective aggregators with matched capacity and different distortions, together with sharper theoretical lower bounds for long-range tasks, could help establish this connection more directly.

There are also computational trade-offs. FSW aggregation is more expensive than the simplest sum or mean aggregators, and the current implementation has a higher runtime than standard MPNNs. Improving the computational efficiency of the embedding while retaining its injective and bi-Lipschitz properties would make the approach more attractive for large-scale graphs. Approximate or structured choices of projection directions and frequencies may provide useful trade-offs between runtime, output dimension, and distortion.

Finally, a broader open question raised by the thesis is how to develop a unified theory of structure-preserving neural representations. Equivalence relations induced by symmetry, partial orders such as set containment, and metrics on WL equivalence classes are mathematically different objects, yet in each case the goal is to transfer a relation from the input domain into a simpler feature space. Understanding which relations admit finite-dimensional stable embeddings, what dimensions are required, and which neural architectures can realize them could provide a more general framework for expressive power than separation alone.

\section{Concluding Remarks}

The main conclusion of this thesis is that expressive power for structured data should be studied relative to the structure one aims to preserve. Symmetry determines the transformations that should not affect the representation, but this is only the starting point. The choice between invariant and equivariant hidden features can change what geometric information is recoverable; the full space of symmetry-compatible linear maps can contain interactions omitted by common architectural templates; and qualitative separation of graphs can be strengthened to quantitative preservation of graph geometry.

From this perspective, the four works form a progression from identifying admissible symmetries to understanding what remains after quotienting by them. The geometric and representation-theoretic results characterize how symmetry shapes expressive architectures. The set-function work illustrates that relations richer than equality can be built into a representation. The FSW-GNN work then shows, in the graph setting, that preserving not only distinguishability but also distance can have concrete consequences for long-range message passing. Together, these results support a view of expressivity in which the central question is not simply whether two structured objects can be distinguished, but how faithfully the relations that matter in the original domain survive in the learned representation.

The research presented in this thesis is part of a broader effort to understand expressive neural architectures for structured data. Several related works extend the ideas developed here in complementary directions. Our work on short-range oversquashing investigates additional limitations of message-passing architectures \cite{mishayev2025shortRange}, while our work on bi-Lipschitz geometric models extends the study of metric preservation to point clouds \cite{sverdlov2025towardBiLipschitz}. Our subsequent study of canonization shifts the focus from expressivity to generalization in invariant learning \cite{sverdlov2026canonize}. Finally, our work on finite probing extends the investigation of expressivity in neural network weight spaces by studying identifiability and universality through probing-based representations \cite{sarangi2026finiteProbes}. Together, these works broaden the scope of the thesis beyond its four constituent papers.

\makebackmatter

\end{document}